\documentclass[twoside,journal]{IEEEtran}

\usepackage{cite}
\usepackage{graphicx}
\usepackage{url}
\usepackage{amsmath}
\usepackage{amsfonts}
\usepackage{amssymb}
\usepackage{algorithmic}
\usepackage{array}
\usepackage{enumitem}
\usepackage[caption=false,font=footnotesize]{subfig}
\usepackage{hyperref}                       % hyperlinks
\usepackage{booktabs}                       % professional-quality tables
\usepackage{multirow}                       % for multirow macro in tables
\usepackage{makecell}                       % for nice line breaks in tables
\usepackage{nicefrac}                       % compact symbols for 1/2, etc.
\usepackage{bm}                             % bold vector notation
\usepackage{dsfont}                         % to denote the indicator function
\usepackage{pifont}                         % for checkmark and xmark symbols
\usepackage{tikz}                           % for drawing plots
\usepackage{xcolor}                         % for adding color
\usepackage[framemethod=TikZ]{mdframed}     % for nice framing boxes
\usepackage[font=itshape]{quoting}          % for nice in-text quoting

\newcommand{\bc}{\bm{c}}
\newcommand{\bw}{\bm{w}}
\newcommand{\bx}{\bm{x}}
\newcommand{\bz}{\bm{z}}
\newcommand{\bmu}{\bm{\mu}}
\newcommand{\bsigma}{\bm{\sigma}}

\newcommand{\bbE}{\mathbb{E}}
\newcommand{\bbN}{\mathbb{N}}
\newcommand{\bbP}{\mathbb{P}}
\newcommand{\bbQ}{\mathbb{Q}}
\newcommand{\bbR}{\mathbb{R}}

\newcommand{\calA}{\mathcal{A}}
\newcommand{\calB}{\mathcal{B}}
\newcommand{\calD}{\mathcal{D}}
\newcommand{\calF}{\mathcal{F}}
\newcommand{\calM}{\mathcal{M}}
\newcommand{\calN}{\mathcal{N}}
\newcommand{\calR}{\mathcal{R}}
\newcommand{\calT}{\mathcal{T}}
\newcommand{\calU}{\mathcal{U}}
\newcommand{\calX}{\mathcal{X}}
\newcommand{\calY}{\mathcal{Y}}
\newcommand{\calZ}{\mathcal{Z}}

\newcommand{\Pgen}{\mathbb{P}}
\newcommand{\Pnorm}{\mathbb{P}^\texttt{\textbf{+}}}
\newcommand{\Pout}{\mathbb{P}^\texttt{\textbf{-}}}

\newcommand{\pgen}{p}
\newcommand{\pnorm}{p^\texttt{\textbf{+}}}
\newcommand{\pout}{p^\texttt{\textbf{-}}}
\newcommand{\phat}{\hat{p}}

\newcommand{\bbInd}{\mathds{1}}
\newcommand{\diff}{\mathop{}\!\mathrm{d}}
\newcommand{\compactcdots}{\cdotp\mkern-2mu\cdotp\mkern-2mu\cdotp}

\DeclareMathOperator*{\argmin}{argmin}
\DeclareMathOperator*{\argmax}{argmax}
\DeclareMathOperator*{\arginf}{arginf}
\DeclareMathOperator*{\supp}{supp}
\DeclareMathOperator*{\diag}{diag}
\DeclareMathOperator*{\KL}{KL}  % KL divergence
\DeclareMathOperator*{\Fr}{Fr}  % Frobenius norm

\newcommand{\enc}{\phi_{e}}
\newcommand{\dec}{\phi_{d}}

\newcommand{\xmark}{\ding{55}}

\definecolor{darkblue}{rgb}{0.0, 0.0, 0.55}
\definecolor{aliceblue}{rgb}{0.94, 0.97, 1.0}
\definecolor{neural_green}{RGB}{220,230,220}
\definecolor{neural_blue}{RGB}{220,230,240}
\definecolor{neural_red}{RGB}{240,220,220}

\begin{document}

\title{A Unifying Review\\ of Deep and Shallow Anomaly Detection}

\author{Lukas~Ruff,
        Jacob~R.~Kauffmann,
        Robert~A.~Vandermeulen,
        Gr{\'e}goire~Montavon,
        Wojciech~Samek,~\IEEEmembership{Member,~IEEE},\\
        Marius~Kloft\textsuperscript{*},~\IEEEmembership{Senior Member,~IEEE},
        Thomas~G.~Dietterich\textsuperscript{*},~\IEEEmembership{Member,~IEEE},
        Klaus-Robert~M{\"u}ller\textsuperscript{*},~\IEEEmembership{Member,~IEEE}.%
\thanks{\textsuperscript{*}~Corresponding authors: M.~Kloft, T.~G.~Dietterich, and K.-R.~M{\"u}ller.}%
\thanks{L.~Ruff, J.~R.~Kauffmann, R.~A.~Vandermeulen, and G.~Montavon are with the ML Group, Technische Universit{\"a}t Berlin, 10587 Berlin, Germany.}%
\thanks{W.~Samek is with the Dept.\ of Artificial Intelligence, Fraunhofer Heinrich Hertz Institute, 10587 Berlin, Germany.}%
\thanks{M.~Kloft is with the Dept.\ of Computer Science, Technische Universit{\"a}t Kaiserslautern, 67653 Kaiserslautern, Germany (e-mail: \href{mailto:kloft@cs.uni-kl.de}{kloft@cs.uni-kl.de}).}%
\thanks{T.~G.~Dietterich is with the School of Electrical Engineering and Computer Science, Oregon State University, Corvallis, OR 97331, USA (e-mail: \href{mailto:tgd@oregonstate.edu}{tgd@cs.orst.edu}).}%
\thanks{K.-R.~M{\"u}ller is with the Brain Team, Google Research, 10117 Berlin, Germany and the ML Group, Technische Universit{\"a}t Berlin, 10587 Berlin, Germany, and also with the Dept.\ of Artificial Intelligence, Korea University, Seoul 136-713, South Korea and Max Planck Institute for Informatics, 66123 Saarbr{\"u}cken, Germany (e-mail: \href{mailto:klaus-robert.mueller@tu-berlin.de}{klaus-robert.mueller@tu-berlin.de}).}}

% The paper headers
%\markboth{Proceedings of the IEEE}%
\markboth{}%
{Ruff \MakeLowercase{\textit{et al.}}: Unifying Review of Deep and Shallow Anomaly Detection}

\maketitle

\begin{abstract}
    Deep learning approaches to anomaly detection have recently improved the state of the art in detection performance on complex datasets such as large collections of images or text. 
    These results have sparked a renewed interest in the anomaly detection problem and led to the introduction of a great variety of new methods. 
    With the emergence of numerous such methods, including approaches based on generative models, one-class classification, and reconstruction, there is a growing need to bring methods of this field into a systematic and unified perspective. 
    In this review we aim to identify the common underlying principles as well as the assumptions that are often made implicitly by various methods. 
    In particular, we draw connections between classic `shallow' and novel deep approaches and show how this relation might cross-fertilize or extend both directions. 
    We further provide an empirical assessment of major existing methods that is enriched by the use of recent explainability techniques, and present specific worked-through examples together with practical advice.
    Finally, we outline critical open challenges and identify specific paths for future research in anomaly detection.
\end{abstract}


\section{Introduction}
\label{sec:introduction}

% Brief introduction to anomaly detection (AD) in general
An \emph{anomaly} is an observation that deviates considerably from some concept of normality.
Also known as \emph{outlier} or \emph{novelty}, such an observation may be termed unusual, irregular, atypical, inconsistent, unexpected, rare, erroneous, faulty, fraudulent, malicious, unnatural, or simply strange\,---\,depending on the situation.
\emph{Anomaly detection} (or \emph{outlier detection} or \emph{novelty detection}) is the research area that studies the detection of such anomalous observations through methods, models, and algorithms based on data.
Classic approaches to anomaly detection include Principal Component Analysis (PCA) \cite{pearson1901,hotelling1933,scholkopf1998,hoffmann2007,huber2009}, the One-Class Support Vector Machine (OC-SVM) \cite{scholkopf2001}, Support Vector Data Description (SVDD) \cite{tax2004}, nearest neighbor algorithms \cite{knorr2000,ramaswamy2000,breunig2000}, and Kernel Density Estimation (KDE) \cite{rosenblatt1956,parzen1962}.

What the above methods have in common is that they are all \emph{unsupervised}, which constitutes the predominant approach to anomaly detection. 
This is because in standard anomaly detection settings labeled anomalous data is often non-existent. When available, it is usually insufficient to fully characterize all notions of anomalousness. 
This typically makes a supervised approach ineffective.
Instead, a central idea in anomaly detection is to learn a model of normality from normal data in an unsupervised manner, so that anomalies become detectable through deviations from the model.

The study of anomaly detection has a long history and spans multiple disciplines including engineering, machine learning, data mining, and statistics.
While the first formal definitions of so-called `discordant observations' date back to the 19th century \cite{edgeworth1887}, the problem of anomaly detection has likely been studied informally even earlier, since anomalies are phenomena that naturally occur in diverse academic disciplines such as medicine and the natural sciences.
Anomalous data may be useless, for example when caused by measurement errors, or may be extremely informative and hold the key to new insights, such as very long surviving cancer patients. 
Kuhn \cite{kuhn-1970} claims that persistent anomalies drive scientific revolutions (see section VI `Anomaly and the Emergence of Scientific Discoveries' in \cite{kuhn-1970}).

Anomaly detection today has numerous applications across a variety of domains.
Examples include intrusion detection in cybersecurity \cite{patcha2007,liao2013,ahmed2016,kwon2017,xin2018,malaiya2018}, fraud detection in finance, insurance, healthcare, and telecommunication \cite{bolton2002,bhattacharyya2011,joudaki2015,ahmed2016b,abdallah2016,vanCapelleveen2016,zheng2018}, industrial fault and damage detection \cite{rabatel2011,marzat2012,marti2015,yan2015,lopez2017,hundman2018,atha2018,ramotsoela2018,zhao2019}, the monitoring of infrastructure \cite{borghesi2019,Sipple2020} and stock markets \cite{golmohammadi2015,golmohammadi2017}, acoustic novelty detection \cite{rabaoui2008,marchi2015,lim2017,principi2017,koizumi2018}, medical diagnosis \cite{tarassenko1995,chauhan2015,leibig2017,litjens2017,schlegl2017,chen2018,iakovidis2018,latif2018,pawlowski2018,baur2019,schlegl2019,seebock2019,guo2020,naud2020,tuluptceva2020} and disease outbreak detection \cite{wong2003,wong2005}, event detection in the earth sciences \cite{blender1997,verbesselt2012,fisher2017,flach2017,wu2018,jiang2020}, and scientific discovery in chemistry \cite{oprea2002,gromski2019}, bioinformatics \cite{min2017}, genetics \cite{tomlins2005,tibshirani2007}, physics \cite{cerri2019,kharkov2020}, and astronomy \cite{protopapas2006,dutta2007,henrion2013,reyes2020}.
The data available in these domains is continually growing in size. 
It is also expanding to include complex data types such as images, video, audio, text, graphs, multivariate time series, and biological sequences, among others.
For applications to be successful on such complex and high-dimensional data, a meaningful representation of the data is crucial \cite{bengio2013}.

% Brief introduction to deep learning
\emph{Deep learning} \cite{lecun2015,schmidhuber2015,goodfellow2016} follows the idea of \emph{learning} effective representations from the data itself by training flexible, multi-layered (`deep') neural networks and has greatly improved the state of the art in many applications that involve complex data types.
Deep neural networks provide the most successful solutions for many tasks in domains such as computer vision \cite{krizhevsky2012,simonyan2015,szegedy2015,long2015,ren2015,gatys2016,he2016,redmon2016,karras2019,xie2020}, speech recognition \cite{lee2009b,dahl2011,mohamed2011,hinton2012,graves2013,hannun2014,amodei2016b,chan2016,chorowski2019,schneider2019}, or natural language processing \cite{bengio2003,mikolov2013,pennington2014,cho2014,bojanowski2017,joulin2017,peters2018,devlin2019,wu2019,brown2020}, and have contributed to the sciences \cite{lengauer2007bioinformatics,baldi2014searching,schutt2017quantum,carleo2017solving,schnorb, jurmeister2019machine,klauschen2018scoring,arcadu2019deep,ardila2019end,esteva2019guide}. 
Methods based on deep neural networks are able to exploit the hierarchical or latent structure that is often inherent to data through their multi-layered, distributed feature representations. Advances in parallel computation, stochastic gradient descent optimization, and automated differentiation make it possible to apply deep learning at scale using large datasets.

% Brief deep anomaly detection overview
Recently, there has been a surge of interest in developing deep learning approaches for anomaly detection. 
This is motivated by a lack of effective methods for anomaly detection tasks which involve complex data, for instance cancer detection from multi-gigapixel whole-slide images in histopathology \cite{faust2018}.
As in other adoptions of deep learning, the goal of \emph{deep anomaly detection} is to mitigate the burden of manual feature engineering and to enable effective, scalable solutions.
However, unlike supervised deep learning, it is less clear what useful representation learning objectives for deep anomaly detection are, due to the mostly unsupervised nature of the problem.

The major approaches to deep anomaly detection include deep autoencoder variants \cite{chalapathy2017,chen2017b,principi2017,zhou2017,zong2018,aytekin2018,chen2018,pawlowski2018,abati2019,huang2019,gong2019,oza2019b,nguyen2019,kim2020}, deep one-class classification \cite{erfani2016,ruff2018,sabokrou2018,oza2019,ruff2019,perera2019a,perera2019c,wang2019d,ruff2020,ghafoori2020}, methods based on deep generative models such as Generative Adversarial Networks (GANs) \cite{schlegl2017,chalapathy2018,deecke2018,akcay2018,choi2018,pidhorskyi2018,zenati2018a,schlegl2019}, and recent self-supervised methods \cite{golan2018,hendrycks2019d,wang2019c,bergman2020b,tack2020}.
In comparison to traditional anomaly detection methods, where a feature representation is fixed a priori (e.g., via a kernel feature map), these approaches aim to \emph{learn} a feature map of the data $\phi_\omega: \bx \mapsto \phi_\omega(\bx)$, a deep neural network parameterized with weights $\omega$, as part of their learning objective.

% Motivation for a unifying AD review
Due to the long history and diversity of anomaly detection research, there exists a wealth of review and survey literature \cite{markou2003a,markou2003b,hodge2004,walfish2006,chandola2009,hadi2009,gogoi2011,singh2012,zimek2012,aguinis2013,zhang2013,pimentel2014,gupta2014,agrawal2015,akoglu2015,ranshous2015,tamboli2016,goldstein2016,xu2019,wang2019} as well as books \cite{barnett1994,rousseeuw2005,aggarwal2017} on the topic. 
Some very recent surveys focus specifically on deep anomaly detection \cite{chalapathy2019,diMattia2019,pang2020}.
However, an integrated treatment of deep learning methods in the overall context of anomaly detection research\,---\,in particular its kernel-based learning part \cite{scholkopf2001,muller2001introduction,tax2004}\,---\,is still missing.

\begin{figure}[!t]
\centering
\includegraphics[width=\linewidth]{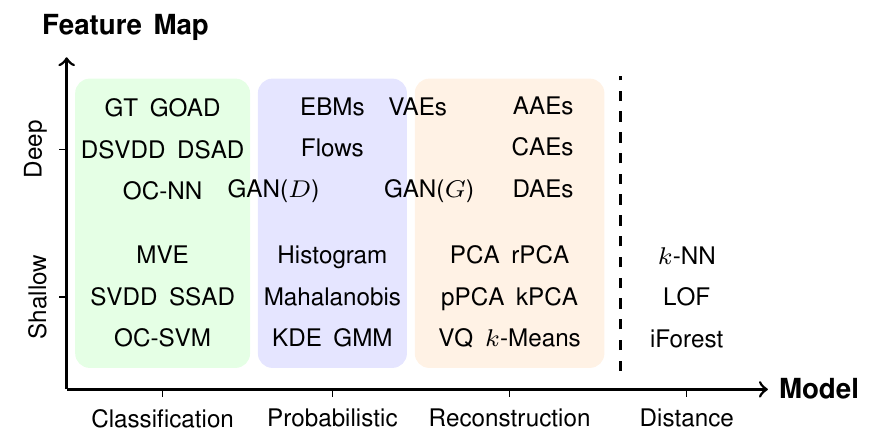}
\caption{Anomaly detection approaches arranged in the plane spanned by two major components (\emph{Model} and \emph{Feature Map}) of our unifying view. Based on shared principles, we distinguish One-Class \emph{Classification}, \emph{Probabilistic} models, and \emph{Reconstruction} models as the three main groups of approaches which all formulate \emph{Shallow} and \emph{Deep} models (see Table \ref{tab:abbrev} for a list of abbreviations). 
These three groups are complemented by purely \emph{Distance}-based methods.
Besides \emph{Model} and \emph{Feature Map}, we identify \emph{Loss}, \emph{Regularization}, and \emph{Inference Mode} as other important modeling components of the anomaly detection problem.}
\label{fig:taxonomy_intro}
\end{figure}

% Scope of this review
In this review article, we aim to fill this gap by presenting a unifying view that connects traditional shallow and novel deep learning approaches.
We will summarize recent exciting developments, present different classes of anomaly detection methods, provide theoretical insights, and highlight the current best practices when applying anomaly detection.
Fig.~\ref{fig:taxonomy_intro} gives an overview of the categorization of anomaly detection methods within our unifying view.
Note finally, that we do not attempt an encyclopedic treatment of all available anomaly detection literature; rather, we present a slightly biased point of view (drawing from our own work on the subject) illustrating the main topics and provide ample reference to related work for further reading.

%%%%%%%%%%%%%%%%%%%%%%%%%%%%%%%%%%%%%%%%%%%%%%%%%%%%%%%%%%%%%%%%%%%%%%%%%%%%%%%%
%%%%%%%%%%%%%%%%%%%%%%%%%%%%%%%%%%%%%%%%%%%%%%%%%%%%%%%%%%%%%%%%%%%%%%%%%%%%%%%%

\section{An Introduction to Anomaly Detection}
\label{sec:introduction_to_AD}

\subsection{Why Should We Care About Anomaly Detection?}
\label{ssec:AD_motivation}
Though we may not realize it, anomaly detection is part of our daily life. Operating mostly unnoticed, anomaly detection algorithms are continuously monitoring our credit card payments, our login behaviors, and companies' communication networks. 
If these algorithms detect an abnormally expensive purchase made on our credit card, several unsuccessful login attempts made from an alien device in a distant country, or unusual ftp requests made to our computer, they will issue an alarm. While warnings such as ``someone is trying to login to your account'' can be annoying when you are on a business trip abroad and just want to check your e-mails from the hotel computer, the ability to detect such anomalous patterns is vital for a large number of today's applications and services and even small improvements in anomaly detection can lead to immense monetary savings\footnote{In 2019, UK's online banking fraud has been estimated to be 111.8 million GBP (source: \href{https://www.statista.com/statistics/326169/united-kingdom-uk-online-banking-losses/}{https://www.statista.com/}).}.

In addition, the ability to detect anomalies is also an important ingredient in ensuring fail-safe and robust design of deep learning-based systems, for instance in medical applications or autonomous driving. Various international standardization initiatives have been launched towards this goal (e.g., ITU/WHO FG-AI4H, ISO/IEC CD TR 24029-1, or IEEE P7009).

Despite its importance, discovering a reliable distinction between `normal' and `anomalous' events is a challenging task. First, the variability within normal data can be very large, resulting in misclassifying normal samples as being anomalous (type I error) or not identifying the anomalous ones (type II error). Especially in biological or biomedical datasets, the variability between the normal data (e.g., person-to-person variability) is often as large or even larger than the distance to anomalous samples (e.g., patients). Preprocessing, normalization, and feature selection are potential means to reduce this variability and improve detectability \cite{guyon2003,garcia2015,aggarwal2017}. If such steps are neglected, the features with wide value ranges, noise, or irrelevant features can dominate distance computations and `mask' anomalies \cite{zimek2012} (see example \ref{ssec:example_uci}).
Second, anomalous events are often very rare, which results in highly imbalanced training datasets. Even worse, in most cases the dataset is unlabeled, so that it remains unclear which data points are anomalies and why. Hence, anomaly detection reduces to an unsupervised learning task with the goal to learn a valid model of the majority of data points. 
Finally, anomalies themselves can be very diverse, so that it becomes difficult to learn a complete model for them. Likewise the solution is again to learn a model for the normal samples and treat deviations from it as anomalies. However, this approach can be problematic if the distribution of the normal data changes (non-stationarity), either intrinsically or due to environmental changes (e.g., lighting conditions, recording devices from different manufacturers, etc.). 

As exemplified and discussed above, we note that anomaly detection has a broad practical relevance and impact. Moreover, (accidentally) detecting the \emph{unknown unknowns} \cite{rumsfeld2011} is a strong driving force in the sciences. If applied in the sciences, anomaly detection can help us to identify new, previously unknown patterns in data, which can lead to novel scientific insights and hypotheses.

\subsection{A Formal Definition of Anomaly Detection}
\label{ssec:problem_definition}

In the following, we formally introduce the anomaly detection problem.
We first define in probabilistic terms what an anomaly is, explain what types of anomalies there are, and delineate the subtle differences between an anomaly, an outlier, and a novelty.
Finally we present a fundamental principle in anomaly detection\,---\,the so-called \emph{concentration assumption}\,---\,and give a theoretical problem formulation that corresponds to density level set estimation.

\subsubsection{What is an Anomaly?}

We opened this review with the following definition:
\begin{quoting}
    An anomaly is an observation that deviates considerably from some concept of normality.
\end{quoting}
To formalize this definition, we here specify two aspects more precisely: a `concept of normality' and what `deviates considerably' signifies.
Following many previous authors \cite{edgeworth1887,anscombe1960,grubbs1969,hawkins1980,barnett1994}, we rely on probability theory.

% Probabilistic definition of an anomaly
Let $\calX \subseteq \bbR^D$ be the data space given by some task or application.
We define a concept of normality as the distribution $\Pnorm$ on $\calX$ that is the \emph{ground-truth law of normal behavior} in a given task or application.
An observation that deviates considerably from such a law of normality\,---\emph{an anomaly}---\,is then a data point $\bx \in \calX$ (or set of points) that lies in a low probability region under $\Pnorm$.
Assuming that $\Pnorm$ has a corresponding probability density function (pdf) $\pnorm(\bx)$, we can define a \emph{set of anomalies} as
\begin{equation}
\label{eqn:anomaly_set}
\calA = \{ \bx \in \calX \; | \; \pnorm(\bx) \leq \tau \}, \quad \tau \geq 0,
\end{equation}
where $\tau$ is some threshold such that the probability of $\calA$ under $\Pnorm$ is `sufficiently small' which we will specify further below.

\subsubsection{Types of Anomalies}
\label{sssec:anomaly_types}

%%%%%%%%%%%%%%%%%%%%%%%%%%%%%%%%%%%%%%%%%%%%%%%%%%%%%%%%%%%%%%%%%%%%%%%%%%%%%%%%
\begin{figure}[!t]
\centering
\setlength{\fboxsep}{0em}

\fbox{\includegraphics[width=0.33\linewidth]{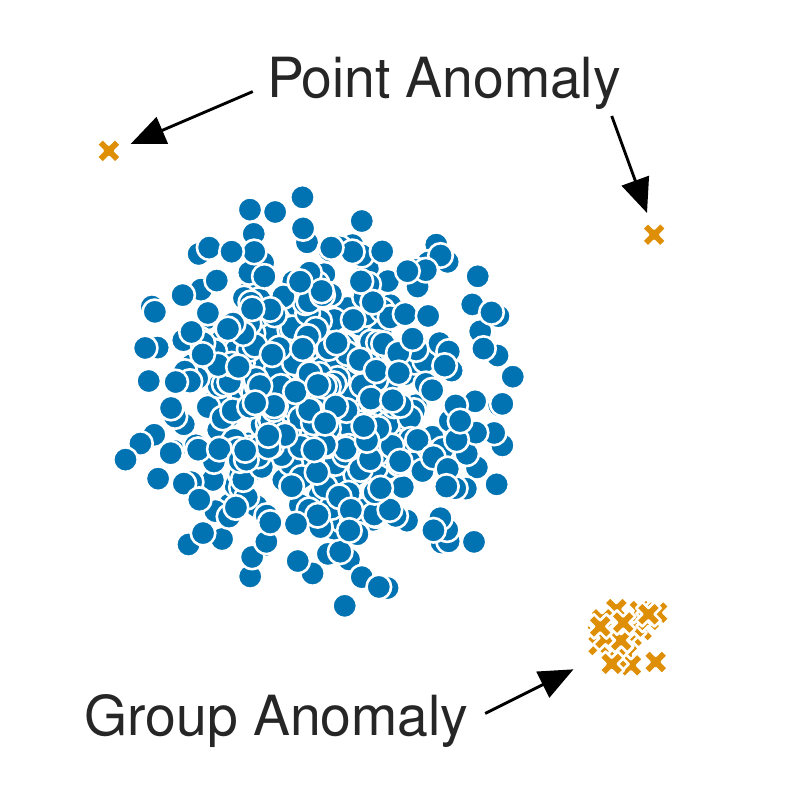}}
\hspace{4mm}
\includegraphics[width=0.33\linewidth]{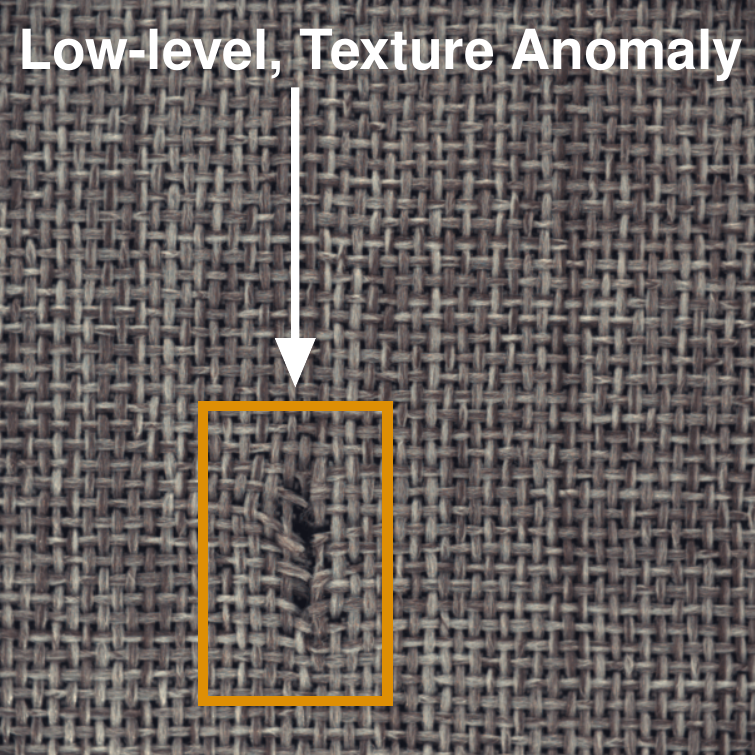}

\vspace{1mm}

\fbox{\includegraphics[width=0.33\linewidth]{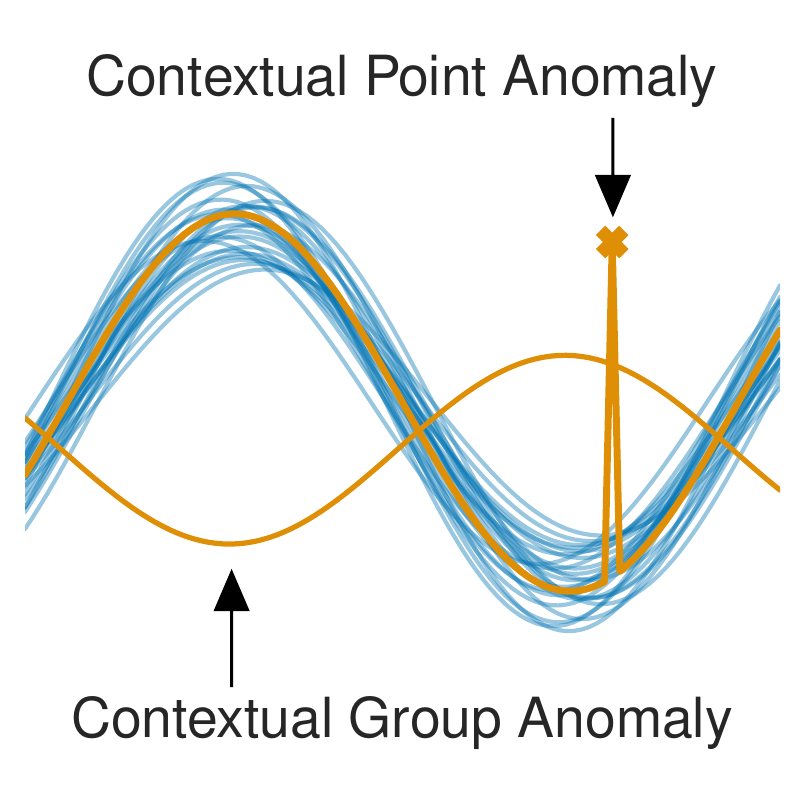}}
\hspace{4mm}
\includegraphics[width=0.33\linewidth]{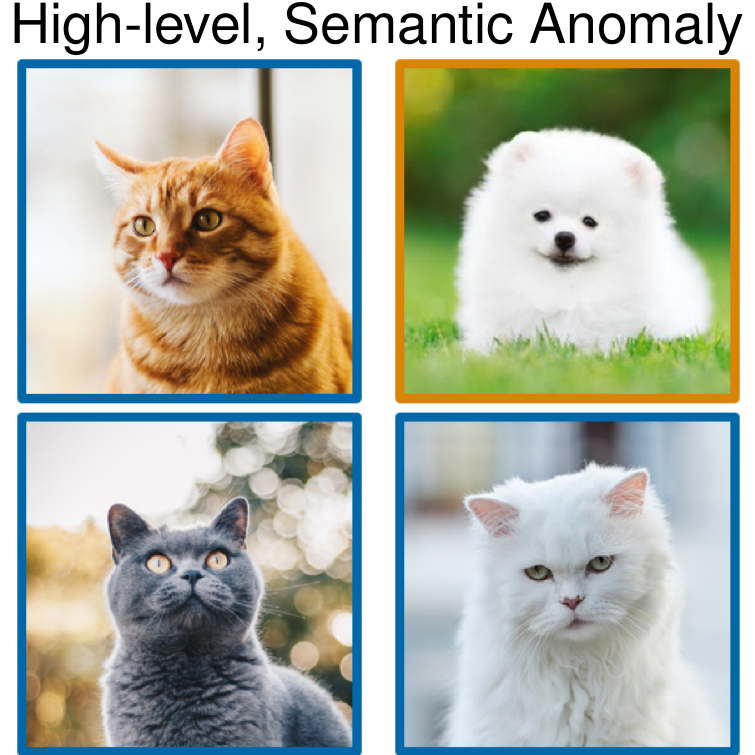}

\caption{An illustration of the types of anomalies: A \emph{point anomaly} is a single anomalous point.
A \emph{contextual point anomaly} occurs if a point deviates in its local context, here a spike in an otherwise normal time series.
A \emph{group anomaly} can be a cluster of anomalies or some series of related points that is anomalous under the joint series distribution (\emph{contextual group anomaly}). 
Note that both contextual anomalies have values that fall into the global (time-integrated) range of normal values.
A low-level sensory anomaly deviates in the low-level features, here a cut in the fabric texture of a carpet \cite{bergmann2019}. 
A semantic anomaly deviates in high-level factors of variation or semantic concepts, here a dog among the normal class of cats. 
Note that the white cat is more similar to the dog than to the other cats in low-level pixel space.}
\label{fig:anomaly_types}
\end{figure}
%%%%%%%%%%%%%%%%%%%%%%%%%%%%%%%%%%%%%%%%%%%%%%%%%%%%%%%%%%%%%%%%%%%%%%%%%%%%%%%%

% Point vs. contextual vs. collective
Various types of anomalies have been identified in the literature \cite{chandola2009,aggarwal2017}. These include point anomalies, conditional or contextual anomalies \cite{song2007,smets2009,chandola2010,gupta2014,akoglu2015,lu2017,samek2017robust}, and group or collective anomalies \cite{chandola2010,xiong2011,muandet2013,yu2015,bontemps2016,chalapathy2018}.
We extend these three established types by further adding low-level sensory anomalies and high-level semantic anomalies \cite{ahmed2020}, a distinction that is particularly relevant for choosing between deep and shallow feature maps.

A \emph{point anomaly} is an individual anomalous data point $\bx \in \calA$, for example an illegal transaction in fraud detection or an image of a damaged product in manufacturing. 
This is arguably the most commonly studied type in anomaly detection research. 

A \emph{conditional} or \emph{contextual anomaly} is a data instance that is anomalous in a specific context such as time, space, or the connections in a graph.
A price of \$1 per Apple Inc.\ stock might have been normal before 1997, but as of today (2021) would be an anomaly.
A mean daily temperature below freezing point would be an anomaly in the Amazon rainforest, but not in the Antarctic desert.
For this anomaly type, the normal law $\Pnorm$ is more precisely a conditional distribution $\Pnorm \equiv \Pnorm_{X|T}$ with conditional pdf $\pnorm(\bx \, | \, t)$ that depends on some contextual variable $T$.
Time-series anomalies \cite{fox1972,tsay1988,tsay2000,gupta2014,lavin2015,samek2017robust} are the most prominent example of contextual anomalies.
Other examples include spatial \cite{chawla2006,schubert2014}, spatio-temporal \cite{smets2009}, or graph-based \cite{noble2003,akoglu2015,honer2017minimizing} anomalies.

A \emph{group} or \emph{collective anomaly} is a \emph{set} of related or dependent points $\{\bx_j \in \calX \; | \; j \in J \}$ that is anomalous, where $J \subseteq \bbN$ is an index set that captures some relation or dependency.
A cluster of anomalies such as similar or related network attacks in cybersecurity form a collective anomaly for instance \cite{kwon2017,honer2017minimizing,ahmed2018}.
Often, collective anomalies are also contextual such as anomalous time (sub-)series or biological (sub-)sequences, for example, some series or sequence $\{\bx_t, \ldots, \bx_{t+s-1} \}$ of length $s \in \bbN$.
It is important to note that although each individual point $\bx_j$ in such a series or sequence might be normal under the time-integrated marginal $\pnorm(\bx) = \int \pnorm(\bx, t) \diff t$ or under the sequence-integrated, time-conditional marginal $\pnorm(\bx \, | \, t)$ given by
\begin{equation*}
\int{\compactcdots}\int \pnorm(\bx_t, {...}\, ,\bx_{t+s-1} \, | \, t) \diff \bx_t \compactcdots \diff \bx_{j-1} \diff \bx_{j+1} \compactcdots \diff \bx_{t+s-1}
\end{equation*}
the full series or sequence $\{\bx_t, \ldots, \bx_{t+s-1} \}$ can be anomalous under the \emph{joint} conditional density $\pnorm(\bx_t, \ldots, \bx_{t+s-1} \, | \, t)$, which properly describes the distribution of the collective series or sequences.

In the wake of deep learning, a distinction between \emph{low-level sensory anomalies} and \emph{high-level semantic anomalies} \cite{ahmed2020} has become important.
Low and high here refer to the level in the feature hierarchy of some hierarchical distribution, for instance, the hierarchy from pixel-level features such as edges and textures to high-level objects and scenes in images or the hierarchy from individual characters and words to semantic concepts and topics in texts.
It is commonly assumed that data with such a hierarchical structure is generated from some semantic latent variables $Z$ and $Y$ that describe higher-level factors of variation $Z$ (e.g., the shape, size or orientation of an object) and concepts $Y$ (e.g., the object class identity) \cite{bengio2013,locatello2019}.
We can express this via a law of normality with conditional pdf $\pnorm(\bx \, | \, \bz, y)$, where we usually assume $Z$ to be continuous and $Y$ to be discrete.
Low-level anomalies could be texture defects or artifacts in images, or character typos in words.
In comparison, semantic anomalies could be images of objects from non-normal classes \cite{ahmed2020}, for instance, or misposted reviews and news articles \cite{ruff2019}.
Note that semantic anomalies can be very close to normal instances in the raw feature space $\calX$.
For example a dog with a fur texture and color similar to that of some cat can be more similar in raw pixel space than various cat breeds among themselves (see Fig.~\ref{fig:anomaly_types}).
Similarly, low-level background statistics can also result in a high similarity in raw pixel space even when objects in the foreground are completely different \cite{ahmed2020}.
Detecting semantic anomalies is thus innately tied to finding a semantic feature representation (e.g., extracting the semantic features of cats such as whiskers, slit pupils, triangular snout, etc.), which is an inherently difficult task in an unsupervised setting \cite{locatello2019}.

\subsubsection{Anomaly, Outlier, or Novelty?}
% Anomalies vs. outliers vs. novelties
Some works make a concrete (albeit subtle) distinction between what is an anomaly, an outlier, or a novelty. 
While all three refer to instances from low probability regions under $\Pnorm$ (i.e., are elements of $\calA$), an anomaly is often characterized as being an instance from a distinct distribution other than $\Pnorm$ (e.g., when anomalies are generated by a different process than the normal points), an outlier as being a rare or low-probability instance from $\Pnorm$, and a novelty as being an instance from some new region or mode of an evolving, non-stationary $\Pnorm$.
Under the distribution $\Pnorm$ of cats, for instance, a dog would be an anomaly, a rare breed of cats such as the LaPerm would be an outlier, and a new breed of cats would be a novelty.
Such a distinction between anomaly, outlier, and novelty may reflect slightly different objectives in an application: whereas anomalies are often the data points of interest (e.g., a long-term survivor of a disease), outliers are frequently regarded as `noise' or `measurement error' that should be removed in a data preprocessing step (`outlier removal'), and novelties are new observations that require models to be updated to the `new normal'.
The methods for detecting points from low probability regions, whether termed `anomaly', `outlier', or `novelty', are essentially the same, however. 
For this reason, we make no distinction between these terms and call any instance $\bx \in \calA$ an `anomaly.'

\subsubsection{The Concentration Assumption}
\label{sssec:concentration_assumption}

While in most situations the data space $\calX \subseteq \bbR^D$ is unbounded, a fundamental assumption in anomaly detection is that the region where the normal data lives can be bounded.
That is, that there exists some threshold $\tau \geq 0$ such that
\begin{equation}
\label{eqn:concentration_assumption}
\calX \setminus \calA = \{ \bx \in \calX \; | \; \pnorm(\bx) > \tau \}
\end{equation}
is non-empty and small (typically in the Lebesgue-measure sense, which is the ordinary notion of volume in $D$-dimensional space).
This is known as the so-called \emph{concentration} or \emph{cluster assumption} \cite{scholkopf2002,steinwart2005,chapelle2006}.
Note that the concentration assumption does not imply that the full support $\supp(\pnorm) = \{ \bx \in \calX \, | \, \pnorm(\bx) > 0 \}$ of the normal law $\Pnorm$ must be bounded; only that some high-density subset of the support is bounded.
A standard univariate Gaussian's support is the full real axis, for example, but approximately 95\% of its probability mass is contained in the interval $[-1.96, 1.96]$.
In contrast, the set of anomalies $\calA$ need not be concentrated and can be unbounded.

\subsubsection{Density Level Set Estimation}
\label{sssec:density_level_set}

% Transition from ground-truth law to estimation from data
A law of normality $\Pnorm$ is only known in a few application settings, such as for certain laws of physics.
Sometimes a concept of normality might also be user-specified (as in juridical laws).
In most cases, however, the ground-truth law of normality $\Pnorm$ is unknown because the underlying process is too complex.
For this reason, we must estimate $\Pnorm$ from data.

% Data-generating distribution vs. normal data distribution
Let $\Pgen$ be the \emph{ground-truth data-generating distribution} on data space $\calX \subseteq \bbR^D$ with corresponding density $\pgen(\bx)$, that is, the distribution that generates the observed data.
For now, we assume that this data-generating distribution exactly matches the normal data distribution, i.e.\ $\Pgen \equiv \Pnorm$ and $\pgen \equiv \pnorm$.
This assumption is often invalid in practice, of course, as the data-generating process might be subject to noise or contamination as we will discuss in section \ref{ssec:data_settings}.

\begin{figure}[!t]
\centering
\includegraphics[width=0.495\linewidth]{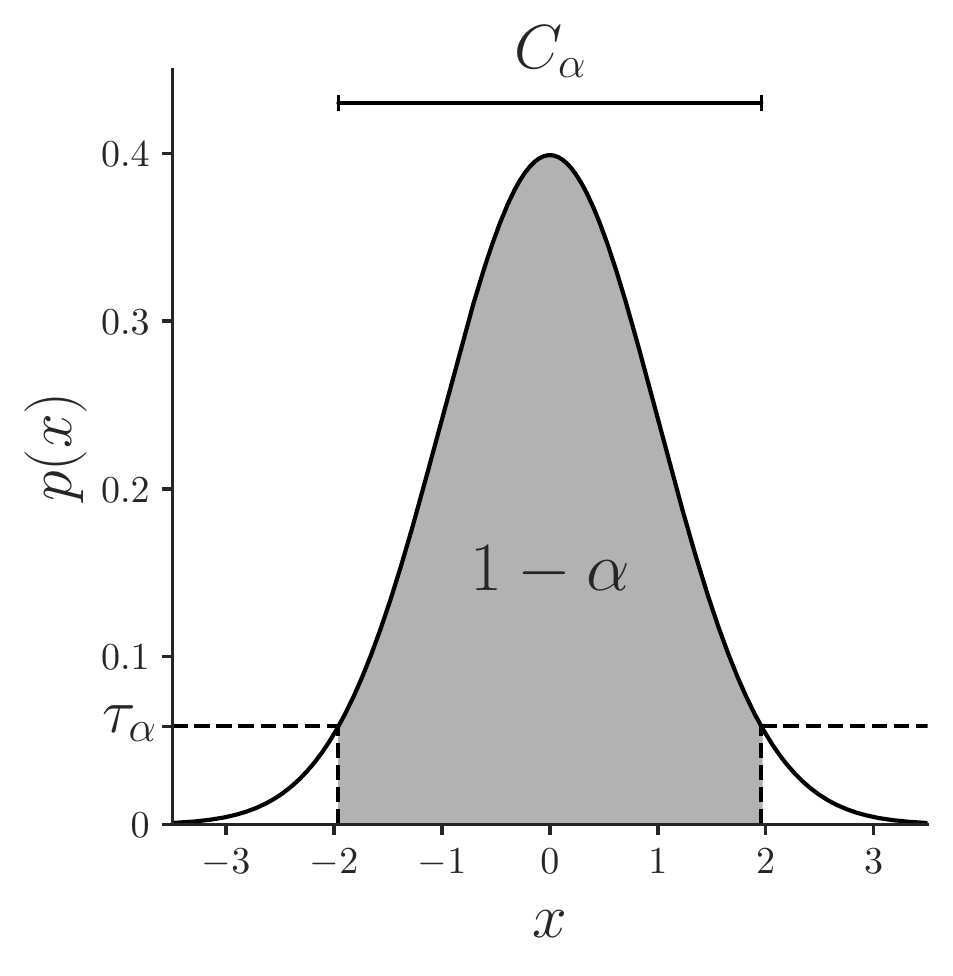}~
\includegraphics[width=0.495\linewidth]{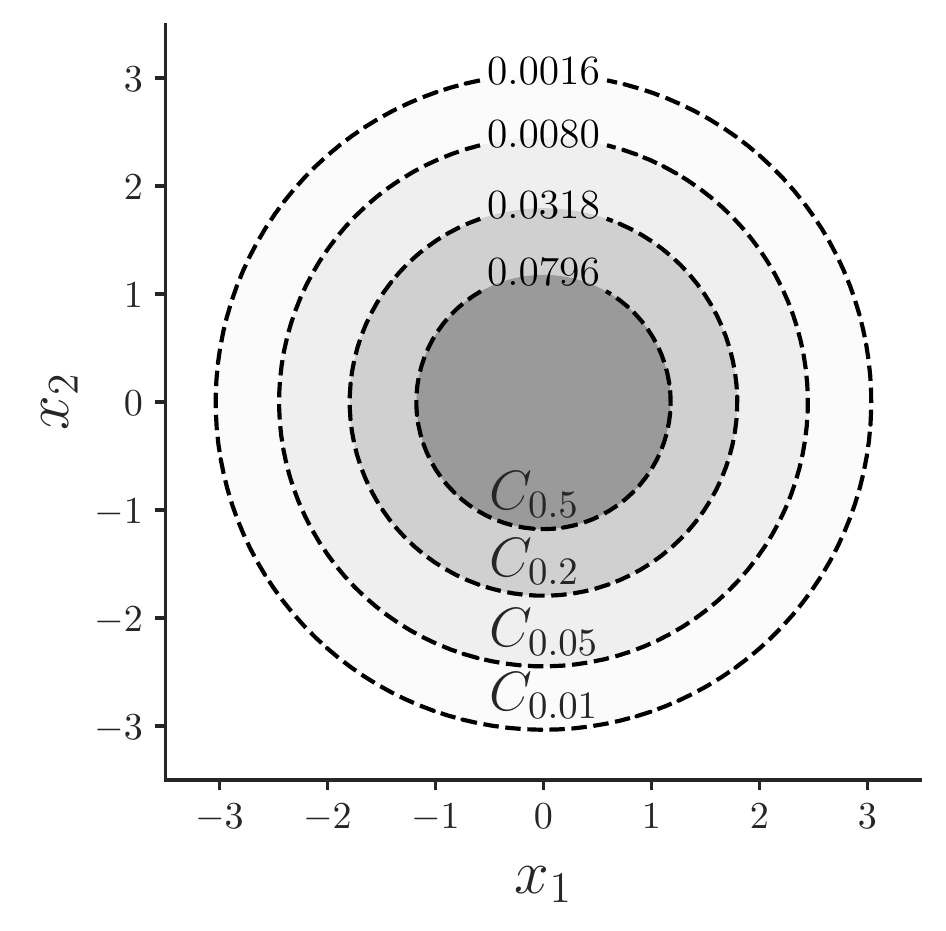}
\caption{An illustration of the $\alpha$-density level sets $C_\alpha$ with threshold $\tau_\alpha$ for a univariate (left) and bivariate (right) standard Gaussian distribution.}
\label{fig:Gauss_level_set}
\end{figure}

% Density level sets and optimal threshold detector
Given data points $\bx_1, \ldots, \bx_n \in \calX$ generated by $\Pgen$ (usually assumed to be drawn from i.i.d.\ random variables following $\Pgen$), the goal of anomaly detection is to learn a model that allows us to predict whether a new test instance $\tilde{\bx} \in \calX$ is an anomaly or not, i.e.\ whether $\tilde{\bx} \in \calA$.
Thus, the anomaly detection objective is to (explicitly or implicitly) estimate the low-density regions (or equivalently high-density regions) in data space $\calX$ under the normal law $\Pnorm$.
We can formally express this objective as the problem of \emph{density level set estimation} \cite{polonik1995,tsybakov1997,ben1997,rigollet2009} which is equivalent to \emph{minimum volume set estimation} \cite{polonik1997,garcia2003,scott2006} for the special case of density-based sets.
The density level set of $\Pgen$ for some threshold $\tau \geq 0$ is given by $C = \{ \bx \in \calX \, | \, \pgen(\bx) > \tau \}$.
For some fixed level $\alpha \in [0,1]$, the \emph{$\alpha$-density level set} $C_\alpha$ of distribution $\Pgen$ is then defined as the smallest density level set $C$ that has a probability of at least $1 - \alpha$ under $\Pgen$, that is,
\begin{equation}
\label{eqn:density_level_set}
\begin{split}
    C_\alpha &= \arginf_C \, \{ \lambda(C) \; | \; \Pgen(C) \geq 1 - \alpha \}\\
    &= \{ \bx \in \calX \, | \, \pgen(\bx) > \tau_\alpha \}
\end{split}
\end{equation}
where $\tau_\alpha \geq 0$ denotes the corresponding threshold and $\lambda$ is typically the Lebesgue measure.
The extreme cases of $\alpha = 0$ and $\alpha \to 1$ result in the full support $C_0 = \{ \bx \in \calX \, | \, \pgen(\bx) > 0 \} = \supp(\pgen)$ and the most likely modes $\argmax_{\bx} \pgen(\bx)$ of $\Pgen$ respectively.
If the aforementioned concentration assumption holds, there always exists some level $\alpha$ such that a corresponding level set $C_\alpha$ exists and can be bounded.
Fig.~\ref{fig:Gauss_level_set} illustrates some density level sets for the case that $\Pgen$ is the familiar standard Gaussian distribution.
Given a level set $C_\alpha$, we can define a corresponding threshold anomaly detector $c_\alpha : \calX \to \{\pm 1\}$ as
\begin{equation}
\label{eqn:detector}
\begin{split}
    c_\alpha(\bx) = \; \left\{ \begin{array}{lr} +1 & \text{if } \bx \in C_\alpha,\\ -1 & \text{if } \bx \not\in C_\alpha. \end{array} \right.
\end{split}
\end{equation}

%%%%%%%%%%%%%%%%%%%%%%%%%%%%%%%%%%%%%%%%%%%%%%%%%%%%%%%%%%%%%%%%%%%%%%%%%%%%%%%%
%%%%%%%%%%%%%%%%%%%%%%%%%%%%%%%%%%%%%%%%%%%%%%%%%%%%%%%%%%%%%%%%%%%%%%%%%%%%%%%%

\subsubsection{Density Estimation for Level Set Estimation}
% The plug-in density level set estimator 
An obvious approach to density level set estimation is through density estimation. 
Given some estimated density model $\phat(\bx) = \phat(\bx; \, \bx_1, \ldots, \bx_n) \approx \pgen(\bx)$ and some target level $\alpha \in [0,1]$, one can estimate a corresponding threshold $\hat{\tau}_\alpha$ via the empirical $p$-value function:
\begin{equation}
\label{eqn:p_value}
    \hat{\tau}_\alpha = \inf_\tau \, \left\{ \tau \geq 0 \; \Big| \; \frac{1}{n} \sum_{i=1}^n \bbInd_{[0, \phat(\bx_i))}(\tau) \geq 1 - \alpha \right\},
\end{equation}
where $\bbInd_{A}(\cdot)$ denotes the indicator function for some set $A$.
Using $\hat{\tau}_\alpha$ and $\phat(\bx)$ in \eqref{eqn:density_level_set} yields the plug-in density level set estimator $\hat{C}_\alpha$ which can be used in \eqref{eqn:detector} to obtain the plug-in threshold detector $\hat{c}_\alpha(\bx)$.
Note however that density estimation is generally the most costly approach to density level set estimation (in terms of samples required), since estimating the full density is equivalent to first estimating the \emph{entire family} of level sets $\{C_\alpha \, | \, \alpha \in [0,1]\}$ from which the desired level set for some fixed $\alpha \in [0,1]$ is then selected \cite{ghaoui2003,menon2018}.
If there are insufficient samples, this density estimate can be biased.
This has also motivated the development of one-class classification methods that aim to estimate a collection \cite{menon2018} or single level sets \cite{tax1999,tax2001,scholkopf2001,tax2004} directly, which we will explain in section \ref{sec:one-class} in more detail.

\subsubsection{Threshold vs.\ Score}
\label{sssec:threshold_vs_score}
% Threshold detectors vs. score functions
The previous approach to level set estimation through density estimation is relatively costly, yet results in a more informative model that can rank inliers and anomalies according to their estimated density. 
In comparison, a pure threshold detector as in \eqref{eqn:detector} only yields a binary prediction.
Menon and Williamson \cite{menon2018} propose a compromise by learning a density outside the level set boundary. 
Many anomaly detection methods also target some strictly increasing transformation $T : [0, \infty) \to \bbR$ of the density for estimating a model (e.g., log-likelihood instead of likelihood). 
The resulting target $T(p(\bx))$ is usually no longer a proper density but still preserves the density ranking \cite{clemencon2013,goix2015}.
An \emph{anomaly score} $s : \calX \to \bbR$ can then be defined by using an additional order-reversing transformation, for example $s(\bx) = -T(p(\bx))$ (e.g., negative log-likelihood), so that high scores reflect low density values and vice versa.
Having such a score that indicates the `degree of anomalousness' is important in many anomaly detection applications.
As for the density in \eqref{eqn:p_value}, of course, we can always derive a threshold from the empirical distribution of anomaly scores if needed.

\subsubsection{Selecting a Level \texorpdfstring{$\alpha$}{TEXT}}
\label{sssec:alpha}
% Alpha corresponds to the (expected) false alarm rate 
As we will show, there are many degrees of freedom when attacking the anomaly detection problem which inevitably requires making various modeling assumptions and choices.
Setting the level $\alpha$ is one of these choices and depends on the specific application.
When the value of $\alpha$ increases, the anomaly detector focuses only on the most likely regions of $\Pgen$.
Such a detector can be desirable in applications where missed anomalies are costly (e.g., in medical diagnosis or fraud detection).
On the other hand, a large $\alpha$ will result in high false alarm rates, which can be undesirable in online settings where lots of data is generated (e.g., in monitoring tasks).
We provide a practical example for selecting $\alpha$ in section \ref{sec:guidelines_and_examples}.
Choosing $\alpha$ also involves further assumptions about the data-generating process $\Pgen$, which we have assumed here to match the normal data distribution $\Pnorm$.
In the following section \ref{ssec:data_settings}, we discuss the data settings that can occur in anomaly detection that may alter this assumption.

\subsection{Dataset Settings and Data Properties}
\label{ssec:data_settings}

% Brief introduction to the various data settings
The dataset settings (e.g., unsupervised or semi-supervised) and data properties (e.g., type or dimensionality) that occur in real-world anomaly detection problems can be diverse.
We here characterize these settings which may range from the standard unsupervised to a semi-supervised as well as a supervised setting and list further data properties that are relevant for modeling an anomaly detection problem.
But before we elaborate on these, we first observe that the assumptions made about the distribution of anomalies (often implicitly) are also crucial to the problem.

\subsubsection{A Distribution of Anomalies?}
\label{sssec:anomaly_distribution}

% An explanation of why the distribution of anomalies matters
Let $\Pout$ denote the \emph{ground-truth anomaly distribution} and assume that it exists on $\calX \subseteq \bbR^D$.
As mentioned above, the common concentration assumption implies that some high-density regions of the normal data distribution are concentrated whereas anomalies are assumed to be \emph{not} concentrated \cite{scholkopf2002,steinwart2005}.
This assumption may be modeled by an anomaly distribution $\Pout$ that is a uniform distribution over the (bounded\footnote{Strictly speaking, we are assuming that there always exists some data-enclosing hypercube of numerically meaningful values such that the data space $\calX$ is bounded and the uniform distribution is well-defined.}) data space $\calX$ \cite{tax2001}.
Some well-known unsupervised methods such as KDE \cite{parzen1962} or the OC-SVM \cite{scholkopf2001}, for example, implicitly make this assumption that $\Pout$ follows a uniform distribution which can be interpreted as a default uninformative prior on the anomalous distribution \cite{steinwart2005}.
This prior assumes that there are no anomalous modes and that anomalies are equally likely to occur over the valid data space $\calX$.
Semi-supervised or supervised anomaly detection approaches often depart from this uninformed prior and try to make a more informed a-priori assumption about the anomalous distribution $\Pout$ \cite{steinwart2005}. 
If faithful to $\Pout$, such a model based on a more informed anomaly prior can achieve better detection performance.
Modeling anomalous modes also can be beneficial in certain applications, for example, for typical failure modes in industrial machines or known disorders in medical diagnosis.
We remark that these prior assumptions about the anomaly distribution $\Pout$ are often expressed only implicitly in the literature, though such assumptions are critical to an anomaly detection model.

\subsubsection{The Unsupervised Setting}
\label{sssec:unsupervised}
The unsupervised anomaly detection setting is the case in which \emph{only unlabeled data}
\begin{equation}
\label{eqn:unsupervised_setting}
\bx_1, \ldots, \bx_n \in \calX
\end{equation}
is available for training a model.
This setting is arguably the most common setting in anomaly detection \cite{hodge2004,chandola2009,zimek2012,pimentel2014}.
We will usually assume that the data points have been drawn in an i.i.d.\ fashion from the data-generating distribution $\Pgen$.
For simplicity, we have so far assumed that the data-generating distribution is the same as the normal data distribution $\Pgen \equiv \Pnorm$. 
This is often expressed by the statement that the training data is `clean'. 
In practice, however, the data-generating distribution $\Pgen$ may contain noise or contamination.

\emph{Noise}, in the classical sense, is some inherent source of randomness $\varepsilon$ that is added to the signal in the data-generating process, that is, samples from $\Pgen$ have the form $\bx + \varepsilon$ where $\bx \sim \Pnorm$.
Noise might be present due to irreducible measurement uncertainties in an application, for example.
The greater the noise, the harder it is to accurately estimate the ground-truth level sets of $\Pnorm$, since informative normal features get obfuscated \cite{zimek2012}.
This is because added noise expands the regions covered by the observed data in input space $\calX$.
A standard assumption about noise is that it is unbiased ($\bbE[\varepsilon] = 0$) and spherically symmetric.

In addition to noise, the \emph{contamination} (or \emph{pollution}) of the unlabeled data with undetected anomalies is another important source of disturbance. 
For instance, some unnoticed anomalous degradation in an industrial machine might have already occurred during the data collection process.
In this case the data-generating distribution $\Pgen$ is a mixture of the normal data and the anomaly distribution, i.e., $\Pgen \equiv (1-\eta) \, \Pnorm \, + \, \eta \, \Pout$ with contamination (or pollution) rate $\eta \in (0,1)$.
The greater the contamination, the more the normal data decision boundary will be distorted by including the anomalous points.

In summary, a more general and realistic assumption is that samples from the data-generating distribution $\Pgen$ have the form of $\bx+\varepsilon$ where $\bx \sim (1-\eta) \, \Pnorm + \eta \, \Pout$ and $\varepsilon$ is random noise.
Assumptions on both, the noise distribution $\varepsilon$ and contamination rate $\eta$, are crucial for modeling a specific anomaly detection problem.
Robust methods \cite{hampel2005,huber2009,zhou2017} specifically aim to account for these sources of disturbance.
Note also that by increasing the level $\alpha$ in the density level set definition above, a corresponding model generally becomes more robust (often at the cost of a higher false alarm rate), since the target decision boundary becomes tighter and excludes the contamination.

\subsubsection{The Semi-Supervised Setting}
\label{sssec:semi_supervised}
The semi-supervised anomaly detection setting is the case in which both \emph{unlabeled and labeled data}
\begin{equation}
\label{eqn:semi-supervised_setting}
\bx_1, \ldots, \bx_n \in \calX \;\;\; \text{and} \;\;\; (\tilde{\bm{x}}_{1}, \tilde{y}_{1}), \ldots, (\tilde{\bm{x}}_{m}, \tilde{y}_{m}) \in \calX \times \calY
\end{equation}
are available for training a model with $\calY = \{\pm 1\}$, where we denote $\tilde{y}={+}1$ for normal and $\tilde{y}={-}1$ for anomalous points respectively.
Usually, we have $m \ll n$ in the semi-supervised setting, that is, most of the data is unlabeled and only a few labeled instances are available, since labels are often costly to obtain in terms of resources (time, money, etc.).
Labeling might for instance require domain experts such as medical professionals (e.g., pathologists) or technical experts (e.g., aerospace engineers).
Anomalous instances in particular are also infrequent by nature (e.g., rare medical conditions) or very costly (e.g., the failure of some industrial machine).
The deliberate generation of anomalies is mostly not an option. 
However, including known anomalous examples, if available, can significantly improve the detection performance of a model \cite{tax2001,liu2006,gornitz2013,min2018,kiran2018,ruff2020}. Labels are also sometimes available in the online setting where alarms raised by the anomaly detector have been investigated to determine whether they were correct. Some unsupervised anomaly detection methods can be incrementally updated when such labels become available \cite{Siddiqui2018}.
A recent approach called \emph{Outlier Exposure} \cite{hendrycks2019a} follows the idea of using large quantities of unlabeled data that is available in some domains as auxiliary anomalies (e.g., online stock photos for computer vision or the English Wikipedia for NLP), thereby effectively labeling this data with $\tilde{y}={-}1$. 
In this setting, we frequently have that $m \gg n$, but this labeled data has an increased uncertainty in the labels as the auxiliary data may not only contain anomalies and may not be representative of test time anomalies.
We will discuss this specific setting in sections \ref{ssec:negative_samples} and \ref{ssec:weak_supervision} in more detail.
Verifying unlabeled samples as indeed being normal can often be easier due to the more frequent nature of normal data.
This is one of the reasons why the special semi-supervised case of \emph{Learning from Positive and Unlabeled Examples} (LPUE) \cite{denis1998,zhang2008,duPlessis2014}, i.e., labeled normal and unlabeled examples, is also studied specifically in the anomaly detection literature \cite{chandola2009,munoz2010,blanchard2010,song2017,akcay2018}.

Previous work \cite{chandola2009} has also referred to the special case of learning exclusively from positive examples as the `semi-supervised anomaly detection' setting, which is confusing terminology. 
Although meticulously curated normal data can sometimes be available (e.g., in open category detection \cite{liu2018a}), such a setting in which entirely (and confidently) labeled normal examples are available is rather rare in practice.
The analysis of this setting is rather again justified by the \emph{assumption} that most of the given (unlabeled) training data is normal, but not the absolute certainty thereof.
This makes this setting effectively equivalent to the unsupervised setting from a modeling perspective, apart from maybe weakened assumptions on the level of noise or contamination, which previous works also point out \cite{chandola2009}.
We therefore refer to the more general setting as presented in \eqref{eqn:semi-supervised_setting} as the semi-supervised anomaly detection setting, which incorporates both labeled normal as well as anomalous examples in addition to unlabeled instances, since this setting is reasonably common in practice.
If some labeled anomalies are available, the modeling assumptions about the anomalous distribution $\Pout$, as mentioned in section \ref{sssec:anomaly_distribution}, become critical for effectively incorporating anomalies into training.
These include for instance whether modes or clusters are expected among the anomalies (e.g., group anomalies).

%%%%%%%%%%%%%%%%%%%%%%%%%%%%%%%%%%%%%%%%%%%%%%%%%%%%%%%%%%%%%%%%%%%%%%%%%%%%%%%%
\begin{table}[!t]
    \caption{Data properties relevant in anomaly detection.}
    \label{tab:data_properties}
    \centering
    \begin{tabular}{ll}
    \toprule
    Data Property               & Description \\
    \midrule
    Size $n+m$                  & \parbox[t][][t]{6.2cm}{Is algorithm scalability in dataset size critical? Are there labeled samples ($m>0$) for (semi-)supervision?} \\[12pt]
    Dimension $D$               & \parbox[t][][t]{6.2cm}{Low- or high-dimensional? Truly high-dimensional or embedded in some higher dimensional ambient space?} \\[12pt]
    Type                        & \parbox[t][][t]{6.2cm}{Continuous, discrete, or categorical?} \\[2pt]
    Scales                      & \parbox[t][][t]{6.2cm}{Are features uni- or multi-scale?} \\[2pt]
    Modality                    & \parbox[t][][t]{6.2cm}{Uni- or multimodal (classes and clusters)? Is there a hierarchy of sub- and superclasses (or -clusters)?} \\[12pt]
    Convexity                   & \parbox[t][][t]{6.2cm}{Is the data support convex or non-convex?} \\[2pt]
    Correlation                 & \parbox[t][][t]{6.2cm}{Are features (linearly or non-linearly) correlated?} \\[2pt]
    Manifold                    & \parbox[t][][t]{6.2cm}{Has the data a (linear, locally linear, or non-linear) subspace or manifold structure? Are there invariances (translation, rotation, etc.)?} \\[21pt]
    Hierarchy                   & \parbox[t][][t]{6.2cm}{Is there a natural feature hierarchy (e.g., images, video, text, speech, etc.)? Are low-level or high-level (semantic) anomalies relevant?} \\[21pt]
    Context                     & \parbox[t][][t]{6.2cm}{Are there contextual features (e.g., time, space, sequence, graph, etc.)? Can anomalies be contextual?} \\[12pt]
    Stationarity                & \parbox[t][][t]{6.2cm}{Is the distribution stationary or non-stationary? Is a domain or covariate shift expected?}\\[12pt]
    Noise                       & \parbox[t][][t]{6.2cm}{Is the noise level $\varepsilon$ large or small? Is the noise type Gaussian or more complex?}\\[12pt]
    Contamination               & \parbox[t][][t]{6.2cm}{Is the data contaminated with anomalies? What is the contamination rate $\eta$?}\\
    \bottomrule
\end{tabular}

\end{table}
%%%%%%%%%%%%%%%%%%%%%%%%%%%%%%%%%%%%%%%%%%%%%%%%%%%%%%%%%%%%%%%%%%%%%%%%%%%%%%%%

\subsubsection{The Supervised Setting}
\label{sssec:supervised}
The supervised anomaly detection setting is the case in which \emph{completely labeled data} 
\begin{equation}
\label{eqn:supervised_setting}
(\tilde{\bm{x}}_{1}, \tilde{y}_{1}), \ldots, (\tilde{\bm{x}}_{m}, \tilde{y}_{m}) \in \calX \times \calY
\end{equation}
is available for training a model, where again $\calY = \{\pm 1\}$ with $\tilde{y}={+}1$ denoting normal instances and $\tilde{y}={-}1$ denoting anomalies respectively.
If both the normal and anomalous data points are assumed to be representative for the normal data distribution $\Pnorm$ and anomaly distribution $\Pout$ respectively, this learning problem is equivalent to supervised binary classification.
Such a setting would thus not be an anomaly detection problem in the strict sense, but rather a classification task.
Although anomalous modes or clusters might exist, that is, some anomalies might be more likely to occur than others, \emph{anything} not normal is by definition an anomaly. 
Labeled anomalies are therefore rarely fully representative of some `anomaly class'.
This distinction is also reflected in modeling: in classification the objective is to learn a (well-generalizing) decision boundary that best separates the data according to some (closed set of) class labels, but the objective in anomaly detection remains the estimation of the normal density level set boundaries.
Hence, we should interpret supervised anomaly detection problems as label-informed density level set estimation in which confident normal (in-distribution) and anomalous (out-of-distribution) training examples are available.
Due to the above and also the high costs often involved with labeling, the supervised anomaly detection setting is the most uncommon setting in practice.

Finally, we note that labels may also carry more granular information beyond simply indicating whether some point $\tilde{\bx}$ is normal ($\tilde{y}={+}1$) or anomalous ($\tilde{y}={-}1$).
In out-of-distribution detection \cite{hendrycks2017} or open category detection \cite{liu2018a} problems, for example, the goal is to train a classifier while also detecting examples that are not from any of the known training set classes.
In these problems, the labeled data $(\tilde{\bx}_1, \tilde{y}_1), \ldots, (\tilde{\bx}_m, \tilde{y}_m)$ with $\tilde{y} \in \{1, \ldots, k\}$ also holds information about the $k$ (sub-)classes of the in-distribution $\Pnorm$.
Such information about the structure of the normal data distribution has been shown to be beneficial for semantic detection tasks \cite{che2019,schirrmeister2020}.
We will discuss such specific and related detection problems later in section \ref{ssec:related_lines}.

\subsubsection{Further Data Properties}
\begin{figure}[!t]
\centering
\setlength{\fboxsep}{0em}

\parbox{.44\linewidth}{\centering \footnotesize \sffamily Ground-truth normal law $\Pnorm$}
\hspace{.02\linewidth}
\parbox{.44\linewidth}{\centering \footnotesize \sffamily Observed data from $\Pgen = \Pnorm + \varepsilon$}

\vspace{0.7mm}

\fbox{\includegraphics[width=0.44\linewidth]{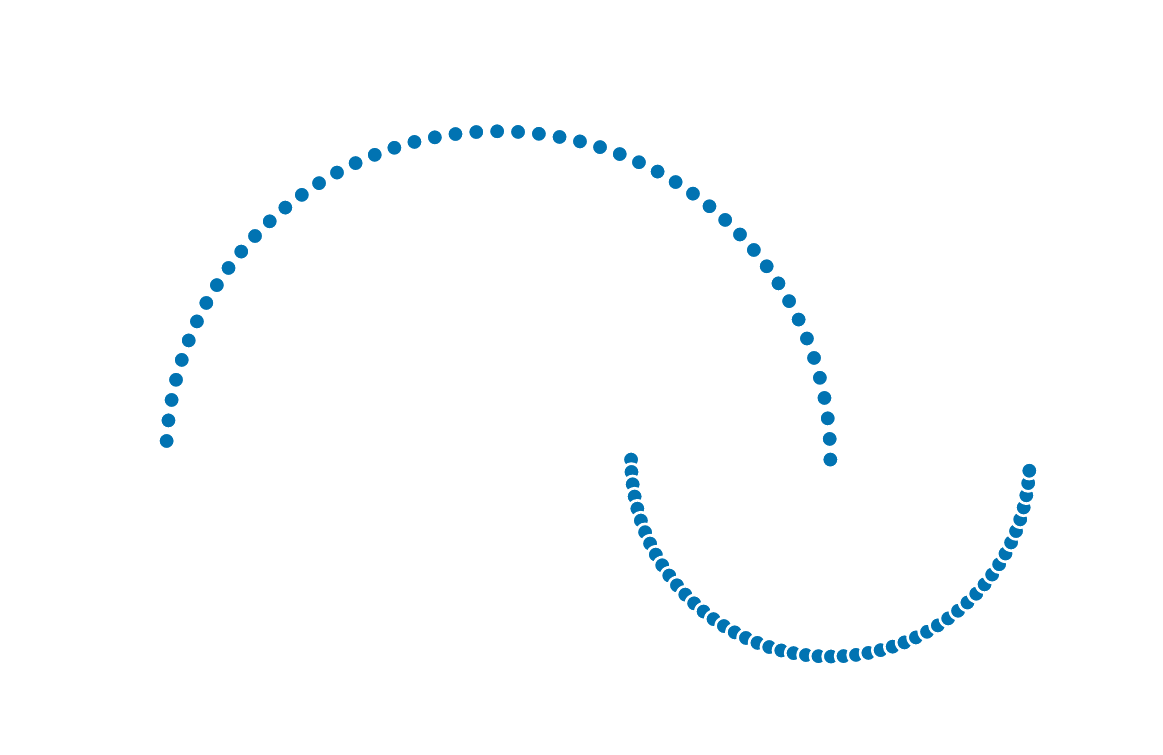}}
\hspace{.02\linewidth}
\fbox{\includegraphics[width=0.44\linewidth]{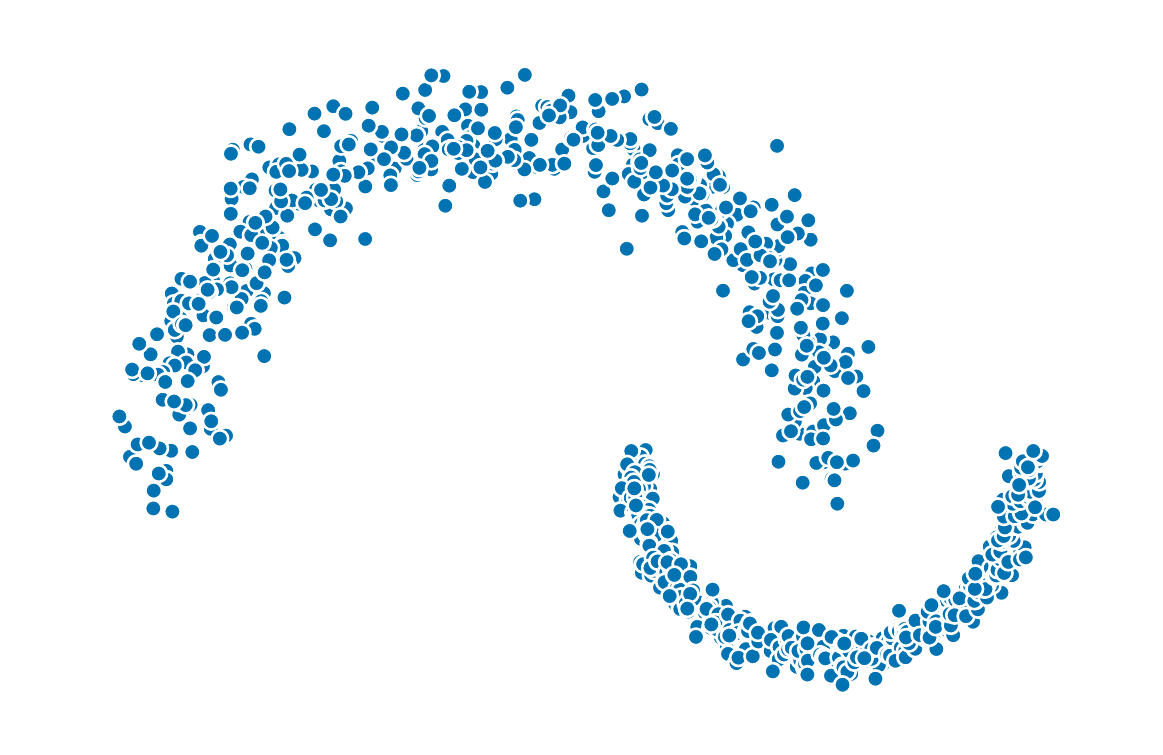}}

\caption{A two-dimensional \emph{Big Moon, Small Moon} toy example with real-valued ground-truth normal law $\Pnorm$ that is composed of two one-dimensional manifolds (bimodal, two-scale, non-convex). The unlabeled training data ($n$ = 1,000, $m$ = 0) is generated from $\Pgen = \Pnorm + \varepsilon$ which is subject to Gaussian noise $\varepsilon$. This toy data is non-hierarchical, context-free, and stationary. Anomalies are off-manifold points that may occur uniformly over the displayed range.}
\label{fig:2D_toy_example}
\end{figure}

Besides the settings described above, the intrinsic properties of the data itself are also crucial for modeling a specific anomaly detection problem.
We give a list of relevant data properties in Table \ref{tab:data_properties} and present a toy dataset with a specific realization of these properties in Fig.~\ref{fig:2D_toy_example} which will serve us as a running example.
The assumptions about these properties should be reflected in the modeling choices such as adding context or deciding among suitable deep or shallow feature maps which can be challenging.
We outline these and further challenges in anomaly detection next.

%%%%%%%%%%%%%%%%%%%%%%%%%%%%%%%%%%%%%%%%%%%%%%%%%%%%%%%%%%%%%%%%%%%%%%%%%%%%%%%%
%%%%%%%%%%%%%%%%%%%%%%%%%%%%%%%%%%%%%%%%%%%%%%%%%%%%%%%%%%%%%%%%%%%%%%%%%%%%%%%%

\subsection{Challenges in Anomaly Detection}
\label{ssec:challenges}

\begin{figure*}[t]
\centering
\includegraphics[width=\textwidth]{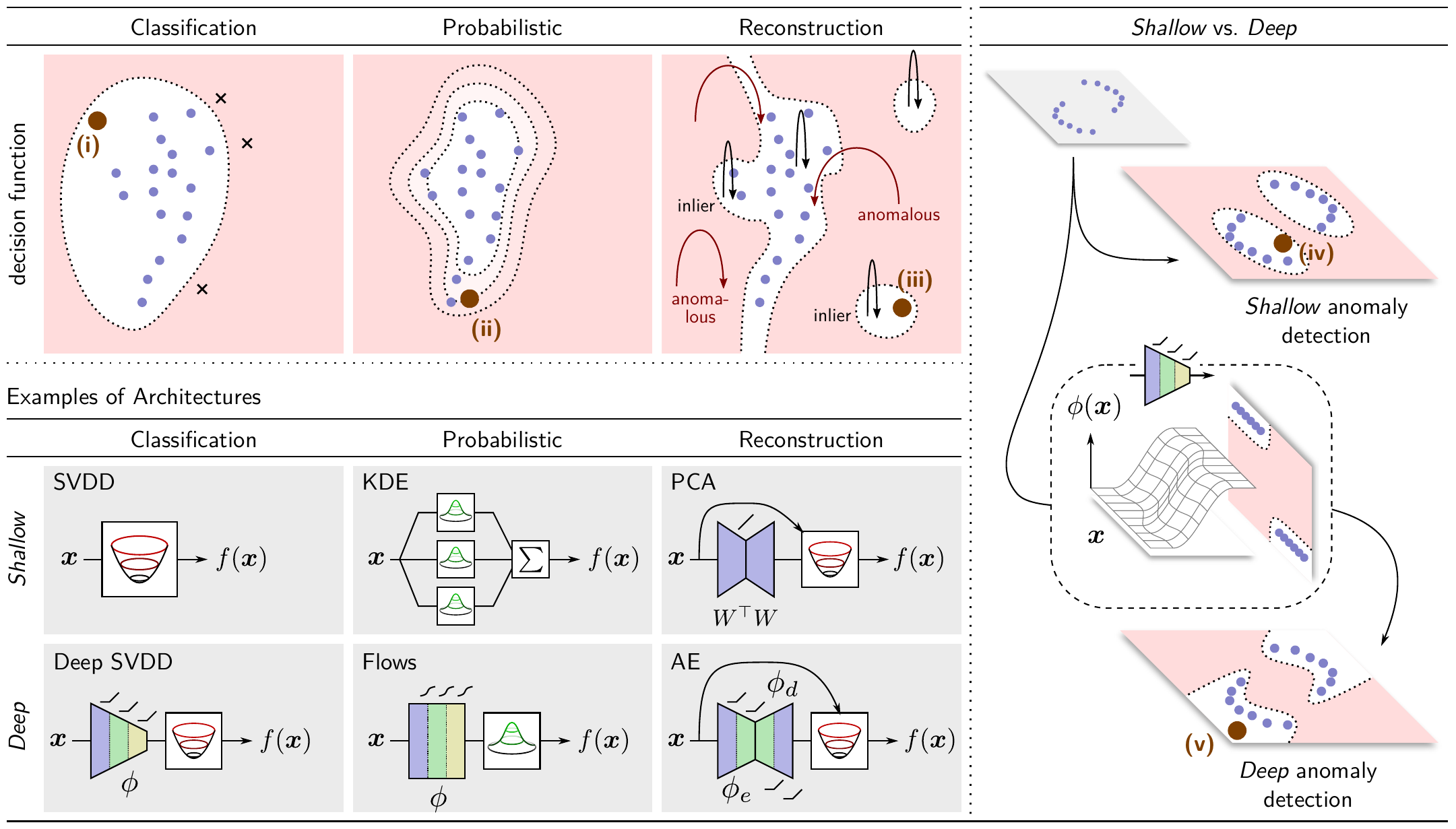}
\caption{An overview of the different approaches to anomaly detection. {\em Top}: Typical decision functions learned by the different anomaly detection approaches, where white corresponds to normal and red to anomalous decision regions. One-class \emph{Classification} models typically learn a discriminative decision boundary, \emph{Probabilistic} models a density, and \emph{Reconstruction} models some underlying geometric structure of the data (e.g., manifold or prototypes).
{\em Right}: Deep feature maps enable to learn more flexible, non-linear decision functions suitable for more complex data.
{\em Bottom:} Diagrams of architectures for a selection of different methods with deep and shallow feature maps.
{\em Points (i)--(v)}: Locations in input space, where we highlight some model-specific phenomena: (i) A too loose, biased one-class boundary may leave anomalies undetected; (ii) Probabilistic models may underfit (or overfit) the tails of a distribution; (iii) Manifold or prototype structure artifacts may result in good reconstruction of anomalies; (iv) Simple shallow models may fail to fit complex, non-linear distributions; (v) Compression artifacts of deep feature maps may create `blind spots' in input space.}
\label{fig:overview}
\end{figure*}

We conclude our introduction by briefly highlighting some notable challenges in anomaly detection, some of which directly arise from the definition and data characteristics detailed above.
Certainly, the fundamental challenge in anomaly detection is the mostly unsupervised nature of the problem, which necessarily requires assumptions to be made about the specific application, the domain, and the given data.
These include assumptions about the relevant types of anomalies (cf., \ref{sssec:anomaly_types}), possible prior assumptions about the anomaly distribution (cf., \ref{sssec:anomaly_distribution}) and, if available, the challenge of how to incorporate labeled data instances in a generalizing way (cf., \ref{sssec:semi_supervised} and \ref{sssec:supervised}).
Further questions include how to derive an anomaly score or threshold in a specific task (cf., \ref{sssec:threshold_vs_score})? 
What level $\alpha$ (cf., \ref{sssec:alpha}) strikes a balance between false alarms and missed anomalies that is reasonable for the task? 
Is the data-generating process subject to noise or contamination (cf., \ref{sssec:unsupervised}), that is, is robustness a critical aspect? 
Moreover, identifying and including the data properties given in Table \ref{tab:data_properties} into a method and model can pose challenges as well. 
The computational complexity in both the dataset size $n+m$ and dimensionality $D$ as well as the memory cost of a model at training time, but also at test time can be a limiting factor (e.g., for data streams or in real-time monitoring \cite{boracchi2018}). 
Is the data-generating process assumed to be non-stationary \cite{sugiyama2007covariate,quionero2009dataset,sugiyama2012machine} and are there distributional shifts expected at test time?
For (truly) high-dimensional data, the curse of dimensionality and resulting concentration of distances can be a major issue \cite{zimek2012}.
Here, finding a representation that captures the features that are relevant for the task and meaningful for the data and domain becomes vital.
Deep anomaly detection methods further entail new challenges such as an increased number of hyperparameters and the selection of a suitable network architecture and optimization parameters (learning rate, batch sizes, etc.).
In addition, the more complex the data or a model is, the greater the challenges of model interpretability (e.g., \cite{baehrens2010explain,montavon2018methods,lapuschkin-natcom19,DBLP:series/lncs/11700}) and decision transparency become. 
We illustrate some of these practical challenges and provide guidelines with worked-through examples in section \ref{sec:guidelines_and_examples}.

Considering the various facets of the anomaly detection problem we have covered in this introduction, it is not surprising that there is a wealth of literature and approaches on the topic.
We outline these approaches in the following sections, where we first examine density estimation and probabilistic models (section \ref{sec:probabilistic}), followed by one-class classification methods (section \ref{sec:one-class}), and finally reconstruction models (section \ref{sec:reconstruction}). 
In these sections, we will point out the connections between deep and shallow methods.
Fig.~\ref{fig:overview} gives an overview and intuition of the approaches.
Afterwards, in section \ref{sec:taxonomy}, we present our unifying view which will enable us to systematically identify open challenges and paths for future research.

\section{Density Estimation and Probabilistic Models}
\label{sec:probabilistic}
The first category of methods we introduce, predict anomalies through estimation of the normal data probability distribution. 
The wealth of existing probability models is therefore a clear candidate for the task of anomaly detection. 
This includes classic density estimation methods \cite{hardle1990applied} as well as deep statistical models. 
In the following, we describe the adaptation of these techniques to anomaly detection.

\subsection{Classic Density Estimation}
\label{ssec:density_classic}
One of the most basic approaches to multivariate anomaly detection is to compute the Mahalanobis distance from a test point to the training data mean \cite{laurikkala00}. This is equivalent to fitting a multivariate Gaussian distribution to the training data and evaluating the log-likelihood of a test point according to that model \cite{jain88}. Compared to modeling each dimension of the data independently, fitting a multivariate Gaussian captures linear interactions between pairs of dimensions. To model more complex distributions, nonparametric density estimators have been introduced, such as kernel density estimators (KDE) \cite{parzen1962,hardle1990applied}, histogram estimators, and Gaussian mixture models (GMMs) \cite{roberts1994b,bishop1994}. The kernel density estimator is arguably the most widely used nonparametric density estimator due to theoretical advantages over histograms \cite{devroye85} and the practical issues with fitting and parameter selection for GMMs \cite{fruhwirth06}. The standard kernel density estimator, along with a more recent adaptation that can deal with modest levels of outliers in the training data \cite{kim2012,vandermeulen13}, is therefore a popular approach to anomaly detection.
A GMM with a finite number of $K$ mixtures can also be viewed as a soft (probabilistic) clustering method that assumes $K$ prototypical modes (cf., section \ref{sssec:prototype}).
This has been used, for example, to represent typical states of a machine in predictive maintenance \cite{amruthnath2018}.

While classic nonparametric density estimators perform fairly well for low dimensional problems, they suffer notoriously from the curse of dimensionality: the sample size required to attain a fixed level of accuracy grows exponentially in the dimension of the feature space. One goal of deep statistical models is to overcome this challenge.

\subsection{Energy-Based Models}
Some of the earliest deep statistical models are energy based models (EBMs) \cite{fahlman83,hopfield1982,lecun2006}. An EBM is a model whose density is characterized by an energy function $E_\theta(\bx)$ with
\begin{equation}
p_\theta(\bx) = \frac{1}{Z(\theta)}\exp \left(-E_\theta(\bx)\right),
\end{equation}
where $Z(\theta) = \int \exp \left(-E_\theta(\bx)\right) \diff\bx$ is the so-called \emph{partition function} that ensures that $p_\theta$ integrates to $1$. These models are typically trained via gradient descent, and approximating the log-likelihood gradient $\nabla_\theta \log p_\theta(\bx)$ via Markov chain Monte Carlo (MCMC) \cite{hinton2002} or Stochastic Gradient Langevin Dynamics (SGLD) \cite{welling2011,grathwohl20}. While one typically cannot evaluate the density $p_\theta$ directly due to the intractability of the partition function $Z(\theta)$, the function $E_\theta$ can be used as an anomaly score since it is monotonically decreasing as the density $p_\theta$ increases.

Early deep EBMs such as Deep Belief Networks \cite{hinton06dbn} and Deep Boltzmann Machines \cite{salakhutdinov09} are graphical models consisting of layers of latent states followed by an observed output layer that models the training data. Here, the energy function depends not only on the input $\bx$, but also on a latent state $\bz$ so the energy function has the form $E_\theta(\bx, \bz)$. While including latent states allows these approaches to richly model latent probabilistic dependencies in data distributions, these approaches are not particularly amenable to anomaly detection since one must marginalize out the latent variables to recover some value related to the likelihood. Later works replaced the probabilistic latent layers with deterministic ones \cite{ngiam11} allowing for the practical evaluation of $E_\theta(\bx)$ for use as an anomaly score. This sort of model has been successfully used for deep anomaly detection \cite{zhai2016}. Recently, EBMs have also been suggested as a framework to reinterpret deep classifiers where the energy-based training has shown to improve robustness and out-of-distribution detection performance \cite{grathwohl20}.

\subsection{Neural Generative Models (VAEs and GANs)}
Neural generative models aim to learn a neural network that maps vectors sampled from a simple predefined source distribution $\bbQ$, usually a Gaussian or uniform distribution, to the actual input distribution $\Pnorm$. More formally, the objective is to train the network so that $\phi_\omega\left(\bbQ \right) \approx \Pnorm$ where $\phi_\omega\left(\bbQ \right)$ is the distribution that results from pushing the source distribution $\bbQ$ through neural network $\phi_\omega$.
The two most established neural generative models are Variational Autoencoders (VAEs) \cite{kingma2014,rezende2014,kingma2019} and Generative Adversarial Networks (GANs) \cite{goodfellow2014}. 

\subsubsection{VAEs}
\label{sssec:vae}
Variational Autoencoders learn deep latent-variable models where the inputs $\bx$ are parameterized on latent samples $\bz \sim \bbQ$ via some neural network so as to learn a distribution $p_\theta(\bx \, | \, \bz)$ such that $p_\theta(\bx) \approx \pnorm(\bx)$. A common instantiation of this is to let $\bbQ$ be an isotropic multivariate Gaussian distribution and let the neural network $\phi_{d,\omega} = (\bmu_\omega,\bsigma_\omega)$ (the \emph{decoder}) with weights $\omega$, parameterize the mean and variance of an isotropic Gaussian distribution, so $p_\theta(\bx \, | \, \bz) \sim \mathcal{N}(\bx; \bmu_\omega(\bz),\bsigma^2_\omega(\bz)I)$. Performing maximum likelihood estimation on $\theta$ is typically intractable. To remedy this an additional network $\phi_{e,\omega'}$ (the \emph{encoder}) is introduced to parameterize a variational distribution $q_{\theta'}(\bz \, | \, \bx)$, with $\theta'$ encapsulated by the output of $\phi_{e,\omega'}$, to approximate the latent posterior $p(\bz \, | \, \bx)$. The full model is then optimized via the evidence lower bound (ELBO) in a variational Bayes manner:
\begin{equation}\label{eqn:elbo}
    \max_{\theta,\theta'} \; -D_{\KL}\left(q_{\theta'}(\bz | \bx) \Vert p(\bz)\right) + \bbE_{q_{\theta'}(\bz | \bx)}\left[\log p_\theta(\bx | \bz) \right].
\end{equation}
Optimization proceeds using Stochastic Gradient Variational Bayes \cite{kingma2014}.
Given a trained VAE, one can estimate $p_\theta(\bx)$ via Monte Carlo sampling from the prior $p(\bz)$ and computing $\bbE_{\bz \sim p(\bz)}\left[p_\theta(\bx \, | \, \bz) \right]$. Using this score directly for anomaly detection has a nice theoretical interpretation, but experiments have shown that it tends to perform worse \cite{xu2018,nalisnick2019} than alternatively using the \emph{reconstruction probability} \cite{an2015} which conditions on $\bx$ to estimate $\bbE_{q_{\theta'}(\bz | \bx)} \left[\log p_\theta\left(\bx|\bz \right) \right]$. 
The latter can also be seen as a probabilistic reconstruction model using a stochastic encoding and decoding process (cf., section \ref{ssec:autoencoders}).

\subsubsection{GANs}
Generative Adversarial Networks pose the problem of learning the target distribution as a zero-sum-game: a generative model is trained in competition with an adversary that challenges it to generate samples whose distribution is similar to the training distribution. A GAN consists of two neural networks, a \emph{generator} network $\phi_\omega: \calZ \to \calX$ and a \emph{discriminator} network $\psi_{\omega'}: \calX \to (0,1)$ which are pitted against each other so that the discriminator is trained to discriminate between $\phi_\omega(\bz)$ and $\bx\sim \Pnorm$ where $\bz \sim \bbQ$. The generator is trained to fool the discriminator network thereby encouraging the generator to produce samples more similar to the target distribution. This is done using the following adversarial objective: 
\begin{equation}
\begin{split}
    \min_\omega \max_{\omega'} \quad &\bbE_{\bx \sim \Pnorm}\left[\log \psi_{\omega'}(\bx) \right] \\ 
    &+ \bbE_{\bz \sim \bbQ}\left[\log(1 - \psi_{\omega'}(\phi_\omega (\bz)))\right].
\end{split}
\end{equation}
Training is typically carried out via an alternating optimization scheme which is notoriously finicky \cite{salimans16}. 
There exist many GAN variants, for example the Wasserstein GAN \cite{arjovsky17,gulrajani17}, which is frequently used for anomaly detection methods using GANs, and StyleGAN, which has produced impressive high-resolution photorealistic images \cite{karras2019}.

Due to their construction, GAN models offer no way to assign a likelihood to points in the input space. Using the discriminator directly has been suggested as one approach to use GANs for anomaly detection \cite{sabokrou2018}, which is conceptually close to one-class classification (cf., section \ref{sec:one-class}).
Other approaches apply optimization to find a point $\tilde{\bz}$ in latent space $\calZ$ such that $\tilde{\bx} \approx \phi_\omega(\tilde{\bz})$ for the test point $\tilde{\bx}$. The authors of AnoGAN \cite{schlegl2017} recommend using an intermediate layer of the discriminator, $f_{\omega'}$, and setting the anomaly score to be a convex combination of the reconstruction loss $\Vert \tilde{\bx} - \phi_\omega(\tilde{\bz}) \Vert$ and the discrimination loss $\Vert f_{\omega'}(\tilde{\bx}) - f_{\omega'}(\phi_\omega(\tilde{\bz})) \Vert$. In AD-GAN \cite{deecke2018}, the authors recommend initializing the search for latent points multiple times to find a collection of $m$ latent points $\tilde{\bz}_1, \ldots, \tilde{\bz}_m$ while simultaneously adapting the network parameters $\omega_i$ individually for each $\tilde{\bz}_i$ to improve the reconstruction and using the mean reconstruction loss as an anomaly score:
\begin{equation}
\frac{1}{m}\sum_{i=1}^m \Vert \tilde{\bx} - \phi_{\omega_i}(\tilde{\bz}_i) \Vert.
\end{equation}
Viewing the generator as a stochastic decoder and the search for an optimal latent point $\tilde{\bz}$ as an (implicit) encoding of a test point $\tilde{\bx}$, utilizing a GAN this way with the reconstruction error for anomaly detection is similar to reconstruction methods, particularly autoencoders (cf., section \ref{ssec:autoencoders}).
Later GAN adaptations have added explicit encoding networks that are trained to find the latent point $\tilde{\bz}$. 
This has been used in a variety of ways, usually again incorporating the reconstruction error \cite{zenati2018a,akcay2018,schlegl2019}.

\subsection{Normalizing Flows}
Like neural generative models, normalizing flows \cite{dinh2014nice,papamakarios19,kobyzev2020} attempt to map data points from a source distribution $\bz\sim\bbQ$ (usually called \emph{base distribution} for normalizing flows) so that $\bx \approx \phi_\omega(\bz)$ is distributed according to $\pnorm$. The crucial distinguishing characteristic of normalizing flows is that the latent samples are $D$-dimensional, so they have the same dimensionality as the input space, and the network consists of $L$ layers $\phi_{i,\omega_i} : \bbR^D \to \bbR^D$ so $\phi_\omega = \phi_{L,\omega_L} \circ \cdots \circ \phi_{1,\omega_1}$ where each $\phi_{i,\omega_i}$ is designed to be invertible for all $\omega_i$, thereby making the entire network invertible. The benefit of this formulation is that the probability density of $\bx$ can be calculated exactly via a change of variables
\begin{equation}
p_{\bx}(\bx) = p_{\bz}(\phi_\omega^{-1}(\bx)) \prod_{i=1}^L \left|\det J\phi^{-1}_{i,\omega_i}\left(\bx_i\right)\right|
\end{equation}
where $\bx_L = \bx$ and $\bx_i = \phi_{i+1}^{-1} \circ \cdots \circ \phi^{-1}_L \left(\bx\right)$ otherwise. Normalizing flow models are typically optimized to maximize the likelihood of the training data. Evaluating each layer's Jacobian and its determinant can be very expensive. Consequently, the layers of flow models are usually designed so that the Jacobian is guaranteed to be upper (or lower) triangular, or have some other nice structure, such that one does not need to compute the full Jacobian to evaluate its determinant \cite{dinh2014nice,dinh17,huang18}. See \cite{noe2019boltzmann} for an application in physics.

An advantage of these models over other methods is that one can calculate the likelihood of a point directly without any approximation while also being able to sample from it reasonably efficiently. Because the density $p_{\bx}(\bx)$ can be computed exactly, normalizing flow models can be applied directly for anomaly detection \cite{nachman20,wellhausen20}.

A drawback of these models is that they do not perform any dimensionality reduction, which argues against applying them to images where the true (effective) dimensionality is much smaller than the image dimensionality. Furthermore, it has been observed that these models often assign high likelihood to anomalous instances \cite{nalisnick2019}. Recent work suggests that one reason for this seems to be that the likelihood in current flow models is dominated by low-level features due to specific network architecture inductive biases \cite{schirrmeister2020, kirichenko2020}. Despite present limitations, we have included normalizing flows here because we believe that they may provide an elegant and promising direction for future anomaly detection methods.
We will come back to this in our outlook in section \ref{sec:future_research}.

%%%%%%%%%%%%%%%%%%%%%%%%%%%%%%%%%%%%%%%%%%%%%%%%%%%%%%%%%%%%%%%%%%%%%%%%%%%%%%%%
\begin{figure}[!t]
\centering
\setlength{\fboxsep}{0em}

\parbox{.32\linewidth}{\centering \footnotesize \sffamily Gaussian (AUC=74.3)}
\parbox{.32\linewidth}{\centering \footnotesize \sffamily KDE (AUC=81.8)}
\parbox{.32\linewidth}{\centering \footnotesize \sffamily RealNVP (AUC=96.3)}

\vspace{0.7mm}

\fbox{\includegraphics[width=0.32\linewidth]{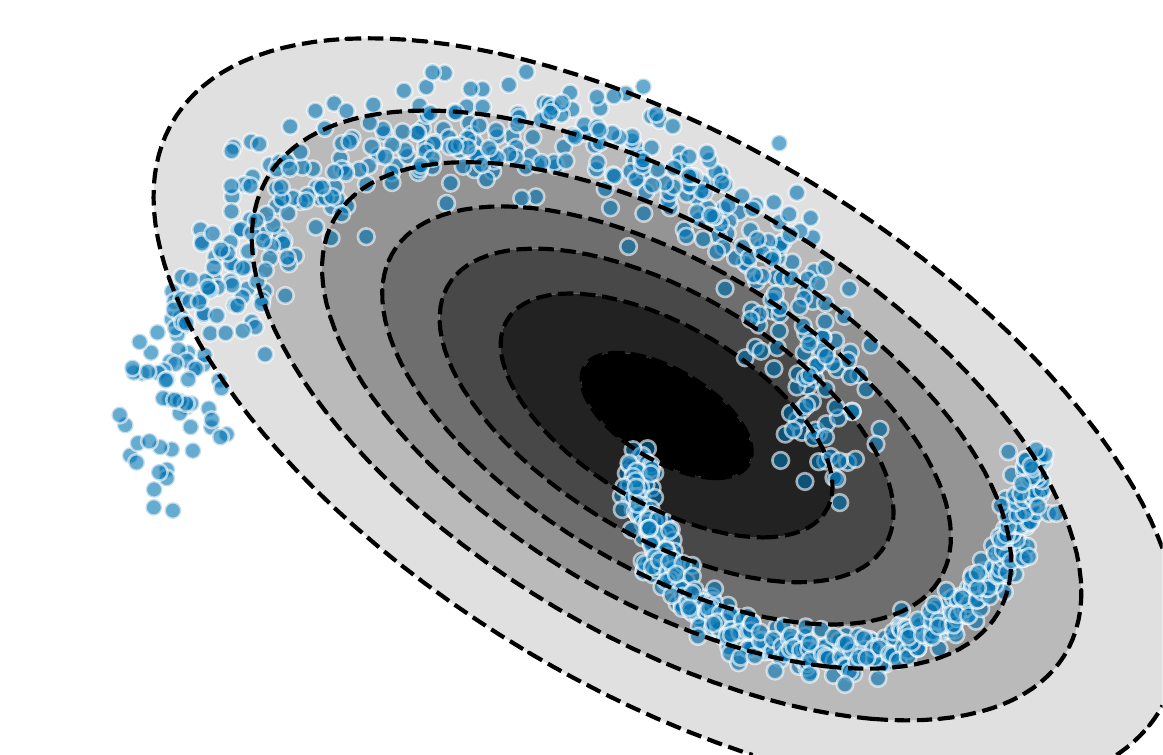}}
\fbox{\includegraphics[width=0.32\linewidth]{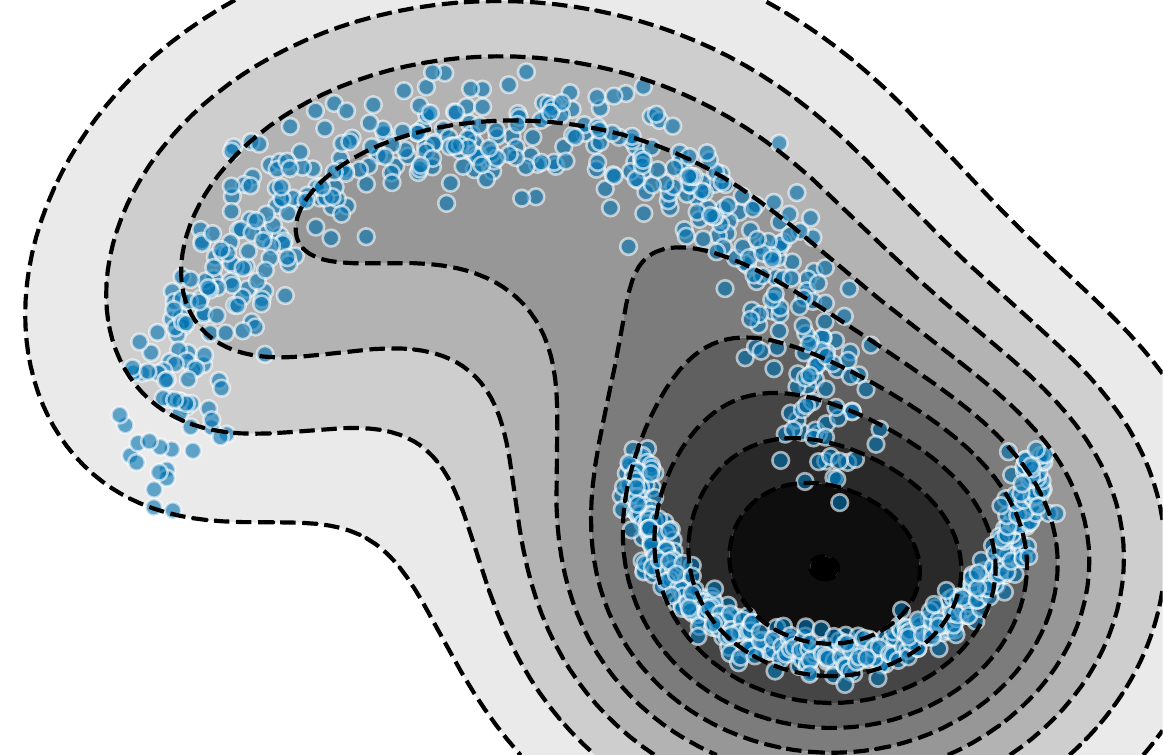}}
\fbox{\includegraphics[width=0.32\linewidth]{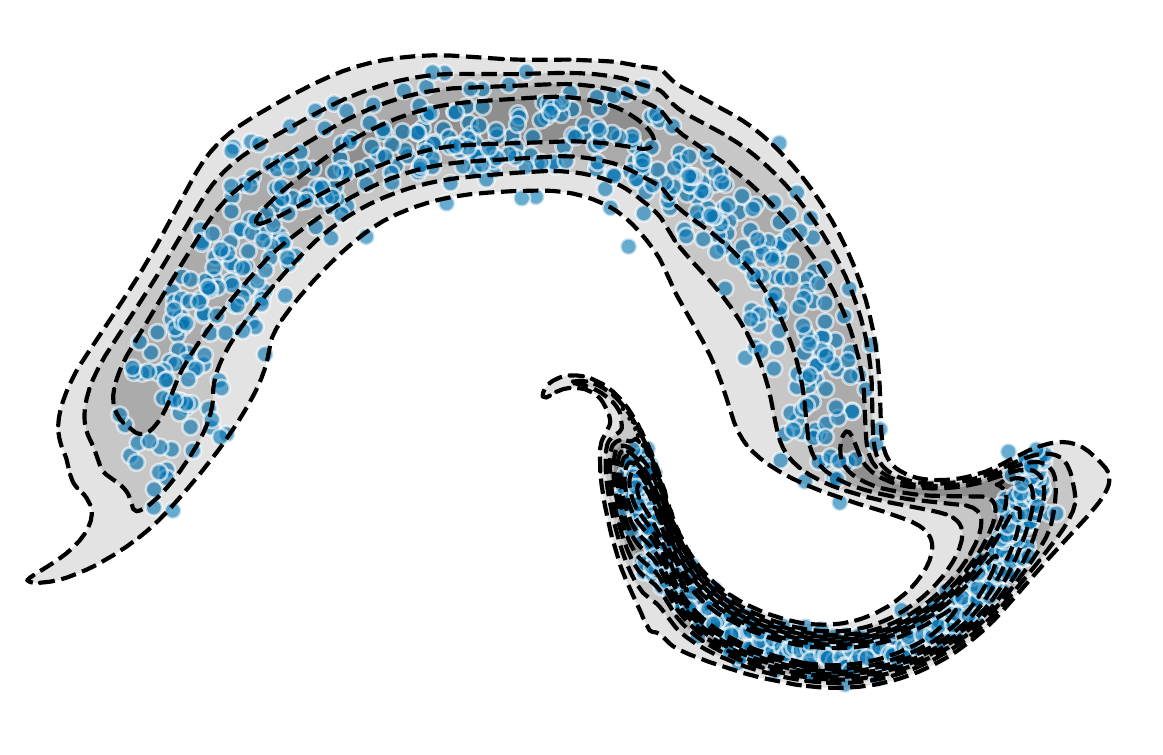}}

\caption{Density estimation models on the \emph{Big Moon, Small Moon} toy example (see Fig.~\ref{fig:2D_toy_example}). The parametric Gaussian model is limited to an ellipsoidal (convex, unimodal) density. KDE with an RBF kernel is more flexible, yet tends to underfit the (multi-scale) distribution due a uniform kernel scale. RealNVP is the most flexible model, yet flow architectures induce biases as well, here a connected support caused by affine coupling layers in RealNVP.}
\label{fig:2D_toy_probabilistic}
\end{figure}
%%%%%%%%%%%%%%%%%%%%%%%%%%%%%%%%%%%%%%%%%%%%%%%%%%%%%%%%%%%%%%%%%%%%%%%%%%%%%%%%

\subsection{Discussion}
Above, we have focused on the case of density estimation on i.i.d.\ samples of low-dimensional data and images. For comparison, we show in Fig.~\ref{fig:2D_toy_probabilistic} three canonical density estimation models (Gaussian, KDE, and RealNVP) trained on the \emph{Big Moon, Small Moon} toy data set, each of which makes use of a different feature representation (raw input, kernel, and neural network). 
It is worth nothing that there exist many deep statistical models for other settings. 
When performing conditional anomaly detection, for example, one can use GAN \cite{mirza14}, VAE \cite{suh16}, and normalizing flow \cite{abdelhamed19} variants which perform conditional density estimation. Likewise there exist many deep generative models for virtually all data types including time series data \cite{suh16,li2019}, text \cite{bowman2016,chen18}, and graphs \cite{jin18,bojchevski18,liao19}, all of which may potentially be used for anomaly detection.

It has been argued that full density estimation is not needed for solving the anomaly detection problem, since one learns all density level sets simultaneously when one really only needs a single density level set \cite{ben1997,scholkopf2001,tax2004}. This violates Vapnik's Principle: ``[W]hen limited amount of data is available, one should avoid solving a more general problem as an intermediate step to solve the original problem'' \cite{vapnik1998}. The methods in the next section seek to compute only a single density level set, that is, they perform one-class classification.

\section{One-Class Classification}
\label{sec:one-class}

\emph{One-class classification} \cite{moya1993,moya1996,tax1999,tax2001,khan2014}, occasionally also called \emph{single-class classification} \cite{minter1975,elYaniv2007}, adopts a discriminative approach to anomaly detection.
Methods based on one-class classification try to avoid a full estimation of the density as an intermediate step to anomaly detection. Instead, these methods aim to directly learn a decision boundary that corresponds to a desired density level set of the normal data distribution $\Pnorm$, or more generally, to produce a decision boundary that yields a low error when applied to unseen data.

\subsection{One-Class Classification Objective}
\label{ssec:one-class_objective}
We can see one-class classification as a particularly tricky classification problem, namely as binary classification where we only have (or almost only have) access to data from one class\,---\,the normal class.
Given this imbalanced setting, the one-class classification objective is to learn a one-class decision boundary that minimizes (i) falsely raised alarms for true normal instances (i.e., the false alarm rate or type I error), and (ii) undetected or missed true anomalies (i.e., the miss rate or type II error).
Achieving a low (or zero) false alarm rate, is conceptually simple: given enough normal data points, one could just draw some boundary that encloses all the points, for example a sufficiently large ball that contains all data instances. 
The crux here is, of course, to simultaneously keep the miss rate low, that is, to not draw this boundary too loosely.
For this reason, one usually \emph{a priori} specifies some target false alarm rate $\alpha \in [0,1]$ for which the miss rate is then sought to be minimized.
Note that this precisely corresponds to the idea of estimating an $\alpha$-density level set for some a priori fixed level $\alpha \in [0,1]$.
The key question in one-class classification thus is how to minimize the miss rate for some given target false alarm rate with access to no (or only few) anomalies.

We can express the rationale above in terms of the binary classification risk \cite{steinwart2005,menon2018}.
Let $Y \in \{\pm 1\}$ be the class random variable, where again $Y=+1$ denotes normal and $Y=-1$ denotes anomalous points, so we can then identify the normal data distribution as $\Pnorm \equiv \Pgen_{X|Y{=}+1}$ and the anomaly distribution as $\Pout \equiv \Pgen_{X|Y{=}-1}$ respectively.
Furthermore, let $\ell : \bbR \times \{\pm 1\} \to \bbR$ be a binary classification loss and $f: \calX \to \bbR$ be some real-valued score function.
The classification risk of $f$ under loss $\ell$ is then given by:
\begin{equation}
\label{eqn:one-class_risk}
    R(f) = \bbE_{X \sim \Pnorm}[\ell(f(X),+1)] + \bbE_{X \sim \Pout}[\ell(f(X),-1)].
\end{equation}
Minimizing the second term\,---\,the expected loss of classifying true anomalies as normal\,---\,corresponds to minimizing the (expected) miss rate.
Given some unlabeled data $\bx_1, \ldots, \bx_n \in \calX$, and potentially some additional labeled data $(\tilde{\bm{x}}_1, \tilde{y}_1), \ldots, (\tilde{\bm{x}}_m, \tilde{y}_m)$, we can apply the principle of empirical risk minimization to obtain
\begin{equation}
\label{eqn:one-class_objective}
    \min_f \quad \frac{1}{n} \sum_{i=1}^n \ell(f(\bx_i),+1) + \frac{1}{m} \sum_{j=1}^m \ell(f(\tilde{\bm{x}}_j), \tilde{y}_j) + \calR.
\end{equation}
This solidifies the empirical one-class classification objective.
Note that the second term is an empty sum in the unsupervised setting.
Without any additional constraints or regularization, the empirical objective \eqref{eqn:one-class_objective} would then be trivial. 
We add $\calR$ as an additional term to denote and capture regularization which may take various forms depending on the assumptions about $f$, but critically also about $\Pout$.
Generally, the regularization $\calR = \calR(f)$ aims to minimize the miss rate (e.g., via volume minimization and assumptions about $\Pout$) and improve generalization (e.g., via smoothing of $f$).
Further note, that the pseudo-labeling of $y = +1$ in the first term incorporates the assumption that the $n$ unlabeled training data points are normal.
This assumption can be adjusted, however, through specific choices of the loss (e.g., hinge) and regularization.
For example, requiring some fraction of the unlabeled data to get misclassified to include an assumption about the contamination rate $\eta$ or achieve some target false alarm rate $\alpha$.

\subsection{One-Class Classification in Input Space}
As an illustrative example that conveys useful intuition, consider the simple idea from above of fitting a data-enclosing ball as a one-class model. Given $\bx_1, \ldots, \bx_n \in \calX$, we can define the following objective:
\begin{equation}
\label{eqn:svdd_input_1}
\begin{split}
    &\min_{R, \bc, \bm{\xi}} \quad R^2 + \frac{1}{\nu n} \sum_{i=1}^n \xi_i \\
	\text{s.t.} \quad &\Vert \bm{x}_i - \bc \Vert^2 \leq R^2 + \xi_i, \quad \xi_i \geq 0, \quad \forall i.
\end{split}
\end{equation}
In words, we aim to find a hypersphere with radius $R>0$ and center $\bc \in \calX$ that encloses the data (${\Vert \bm{x}_i - \bc \Vert^2} \leq R^2$).
To control the miss rate, we minimize the volume of this hypersphere by minimizing $R^2$ to achieve a tight spherical boundary.
Slack variables $\xi_i \geq 0$ allow some points to fall outside the sphere, thus making the boundary soft, where hyperparameter $\nu \in (0,1]$ balances this trade-off.

Objective \eqref{eqn:svdd_input_1} exactly corresponds to Support Vector Data Description (SVDD) applied in the input space $\calX$, motivated above as in \cite{tax1999,tax2001,tax2004}.
Equivalently, we can derive \eqref{eqn:svdd_input_1} from the binary classification risk. Consider the (shifted, cost-weighted) hinge loss $\ell(s,y)$ defined by $\ell(s,+1) = \tfrac{1}{1+\nu} \max(0, s)$ and $\ell(s,-1) = \tfrac{\nu}{1+\nu} \max(0, -s)$ \cite{menon2018}.
Then, for a hypersphere model $f_\theta(\bx) = \Vert \bx - \bc \Vert^2 - R^2$ with parameters $\theta = (R, \bc)$, the corresponding classification risk \eqref{eqn:one-class_risk} is given by
\begin{equation}
\label{eqn:svdd_input_risk}
\begin{split}
    \min_\theta \quad &\bbE_{X \sim \Pnorm}[\max(0, \Vert X - \bc \Vert^2 - R^2)]\\
    &+ \nu \; \bbE_{X \sim \Pout}[\max(0, R^2 - \Vert X - \bc \Vert^2)].
\end{split}
\end{equation}
We can estimate the first term in \eqref{eqn:svdd_input_risk} empirically from $\bx_1, \ldots, \bx_n$, again assuming (most of) these points have been drawn from $\Pnorm$.
If labeled anomalies are absent, we can still make \emph{an assumption} about their distribution $\Pout$.
Following the basic, uninformed prior assumption that anomalies may occur uniformly on $\calX$ (i.e., $\Pout \equiv \calU(\calX)$), we can examine the expected value in the second term analytically:
\begin{equation}
\begin{split}
    &\bbE_{X \sim \calU(\calX)}[\max(0, R^2 - \Vert X - \bc \Vert^2)]\\
    = \quad & \frac{1}{\lambda(\calX)} \int_\calX \max(0, R^2 - \Vert \bx - \bc \Vert^2) \diff \lambda(\bx)\\
    \leq \quad & R^2 \; \frac{\lambda(\calB_{R}(\bc))}{\lambda(\calX)} \leq R^2,
\end{split}
\end{equation}
where $\calB_{R}(\bc) \subseteq \calX$ denotes the ball centered at $\bc$ with radius $R$ and $\lambda$ is again the standard (Lebesgue) measure of volume.\footnote{Again note that we assume $\lambda(\calX) < \infty$ here, i.e., that the data space $\calX$ can be bounded to numerically meaningful values.}
This shows that the minimum volume principle \cite{polonik1997,scott2006} naturally arises in one-class classification through seeking to minimize the risk of missing anomalies, here illustrated for an assumption that the anomaly distribution $\Pout$ follows a uniform distribution.
Overall, from \eqref{eqn:svdd_input_risk} we thus can derive the empirical objective
\begin{equation}
\label{eqn:svdd_input_2}
    \min_{R, \bc} \quad R^2 + \frac{1}{\nu n} \sum_{i=1}^n \max(0, \Vert \bm{x}_i - \bc \Vert^2 - R^2),
\end{equation}
which corresponds to \eqref{eqn:svdd_input_1} with the constraints directly incorporated into the objective function.
We remark that the cost-weighting hyperparameter $\nu \in (0,1]$ is purposefully chosen here, since it is an upper bound on the ratio of points outside and a lower bound on the ratio of points inside or on the boundary of the sphere \cite{scholkopf2001,ruff2018}.
We can therefore see $\nu$ as an approximation of the false alarm rate, that is, $\nu \approx \alpha$.

A sphere in the input space $\calX$ is of course a very limited model and only matches a limited class of distributions $\Pnorm$ (e.g., an isotropic Gaussian distribution).
Minimum Volume Ellipsoids (MVE) \cite{rousseeuw1985,rousseeuw2005} and the Minimum Covariance Determinant (MCD) estimator \cite{rousseeuw1999} are a generalization to non-isotropic distributions with elliptical support.
Nonparametric methods such as One-Class Neighbor Machines \cite{munoz2006} provide additional freedom to model multimodal distributions having non-convex support.
Extending the objective and principles above to general feature spaces (e.g., \cite{vapnik1998,scholkopf1999input,scholkopf2002}) further increases the flexibility of one-class models and enables decision boundaries for more complex distributions.

\subsection{Kernel-based One-Class Classification}
The kernel-based OC-SVM \cite{scholkopf2001,manevitz2001} and SVDD \cite{tax2001,tax2004} are perhaps the most well-known one-class classification methods.
Let $k : \calX \times \calX \to \bbR$ be some positive semi-definite (PSD) kernel with associated reproducing kernel Hilbert space (RKHS) $\calF_k$ and corresponding feature map $\phi_k : \calX \to \calF_k$, so $k(\bx, \tilde{\bx}) = \langle \phi_k(\bx), \phi_k(\tilde{\bx}) \rangle$ for all $\bx, \tilde{\bx} \in \calX$.
The objective of (kernel) SVDD is again to find a data-enclosing hypersphere of minimum volume.
The SVDD primal problem is the one given in \eqref{eqn:svdd_input_1}, but with the hypersphere model $f_\theta(\bx) = \Vert \phi_k(\bx) - \bc \Vert^2 - R^2$ defined in feature space $\calF_k$ instead.
In comparison, the OC-SVM objective is to find a hyperplane $\bw \in \calF_k$ that separates the data in feature space $\calF_k$ with maximum margin from the origin:
\begin{equation}
\label{eqn:ocsvm}
\begin{split}
    &\min_{\bw, \rho, \bm{\xi}} \quad \frac{1}{2} \Vert \bw \Vert^2 - \rho + \frac{1}{\nu n} \sum_{i=1}^n \xi_i \\
    \text{s.t.} \quad &\rho - \langle \phi_k(\bx_i), \bw \rangle \leq \xi_i, \quad \xi_i \geq 0, \quad \forall i.
\end{split}
\end{equation}
So the OC-SVM uses a linear model $f_\theta(\bx) = \rho - \langle \phi_k(\bx), \bm{w} \rangle$ in feature space $\calF_k$ with model parameters $\theta = (\bw, \rho)$.
The margin to the origin is given by $\tfrac{\rho}{\Vert \bw \Vert}$ which is maximized via maximizing $\rho$, where $\Vert \bw \Vert$ acts as a normalizer.

The OC-SVM and SVDD both can be solved in their respective dual formulations which are quadratic programs that only involve dot products (the feature map $\phi_k$ is implicit).
For the standard Gaussian kernel (or any kernel with constant norm $k(\bx,\bx) = c > 0$), the OC-SVM and SVDD are equivalent \cite{tax2001}.
In this case, the corresponding density level set estimator defined by
\begin{equation}
\label{eqn:level_set_estimator}
\hat{C}_\nu = \{ \bx \in \calX \, | \, f_\theta(\bx) < 0 \}
\end{equation}
is in fact an asymptotically consistent $\nu$-density level set estimator \cite{vert2006}.
The solution paths of hyperparameter $\nu$ have been analyzed for both the OC-SVM \cite{lee2007a} and SVDD \cite{sjostrand2006}.

Kernel-induced feature spaces considerably improve the expressive power of one-class methods and allow to learn well-performing models in multimodal, non-convex, and non-linear data settings.
Many variants of kernel one-class classification have been proposed and studied over the years such as hierarchical formulations for nested density level set estimation \cite{lee2009,glazer2013}, Multi-Sphere SVDD \cite{gornitz2017}, Multiple Kernel Learning for OC-SVM \cite{das2010,gautam2019}, OC-SVM for group anomaly detection \cite{muandet2013}, boosting via $L_1$-norm regularized OC-SVM \cite{ratsch2002}, One-Class Kernel Fisher Discriminants \cite{roth2005,roth2006,dufrenois2014}, Bayesian Data Description \cite{ghasemi2012}, and distributed \cite{stolpe2013}, incremental learning \cite{jiang2019}, or robust \cite{liu2014} variants.

\subsection{Deep One-Class Classification}
\label{ssec:deep-one-class}

Selecting kernels and hand-crafting relevant features can be challenging and quickly become impractical for complex data.
Deep one-class classification methods aim to overcome these challenges by learning useful neural network feature maps $\phi_\omega : \calX \to \cal Z$ from the data or transferring such networks from related tasks.
Deep SVDD \cite{ruff2018,ruff2020,wu2020,ghafoori2020} and deep OC-SVM variants \cite{erfani2016,chalapathy2018b} employ a hypersphere model $f_\theta(\bx) = \Vert \phi_\omega(\bx) - \bc \Vert^2 - R^2$ and linear model $f_\theta(\bx) = \rho - \langle \phi_\omega(\bx), \bm{w} \rangle$ with explicit neural feature maps $\phi_\omega(\cdot)$ in \eqref{eqn:svdd_input_1} and \eqref{eqn:ocsvm} respectively.
These methods are typically optimized with SGD variants \cite{lecun2012,kingma2015,goh2017}, which, together with GPU parallelization, makes them scale to large datasets.

The \emph{One-Class Deep SVDD} \cite{ruff2018,ruff2020b} has been introduced as a simpler variant compared to using a neural hypersphere model in \eqref{eqn:svdd_input_1}, which poses the following objective:
\begin{equation}
\label{eqn:one-class_dsvdd}
    \min_{\omega, \bc} \quad \frac{1}{n} \sum_{i=1}^n \Vert \phi_\omega(\bm{x}_i) - \bc \Vert^2 + \calR.
\end{equation}
Here, the neural network transformation $\phi_\omega(\cdot)$ is learned to minimize the mean squared distance over \emph{all} data points to center $\bc \in \calZ$.
Optimizing this simplified objective has been found to converge faster and be effective in many situations \cite{ruff2018,ruff2020,ruff2020b}.
In light of our unifying view, we will see that we may interpret One-Class Deep SVDD also as a single-prototype deep clustering method (cf., sections \ref{sssec:prototype} and \ref{ssec:clustering}).

A recurring question in deep one-class classification is how to meaningfully regularize against a feature map collapse $\phi_\omega \equiv \bc$. 
Without regularization, minimum volume or maximum margin objectives such as \eqref{eqn:svdd_input_1}, \eqref{eqn:ocsvm}, or \eqref{eqn:one-class_dsvdd} could be trivially solved with a constant mapping \cite{ruff2018,goyal2020}.
Possible solutions for this include adding a reconstruction term or architectural constraints \cite{ruff2018,wu2020}, freezing the embedding \cite{erfani2016,ruff2019,oza2019,perera2019c,kauffmann2020a}, inversely penalizing the embedding variance \cite{chong2020}, using true \cite{pang2019,ruff2020}, auxiliary \cite{hendrycks2019a,oza2019,ruff2020b,liznerski2020}, or artificial \cite{liznerski2020} negative examples in training, pseudo-labeling \cite{golan2018,hendrycks2019d,chong2020,bergman2020b}, or integrating some manifold assumption \cite{goyal2020}.
Further variants of deep one-class classification include multimodal \cite{ghafoori2020} or time-series extensions \cite{shen2020} and methods that employ adversarial learning \cite{sabokrou2018,perera2019a,sabokrou2020} or transfer learning \cite{oza2019,perera2019c}.

\begin{figure}[!t]
\centering
\setlength{\fboxsep}{0em}

\parbox{.32\linewidth}{\centering \footnotesize \sffamily MVE (AUC=74.7)}
\parbox{.32\linewidth}{\centering \footnotesize \sffamily SVDD (AUC=90.9)}
\parbox{.32\linewidth}{\centering \footnotesize \sffamily DSVDD (AUC=97.5)}

\vspace{0.7mm}

\fbox{\includegraphics[width=0.32\linewidth]{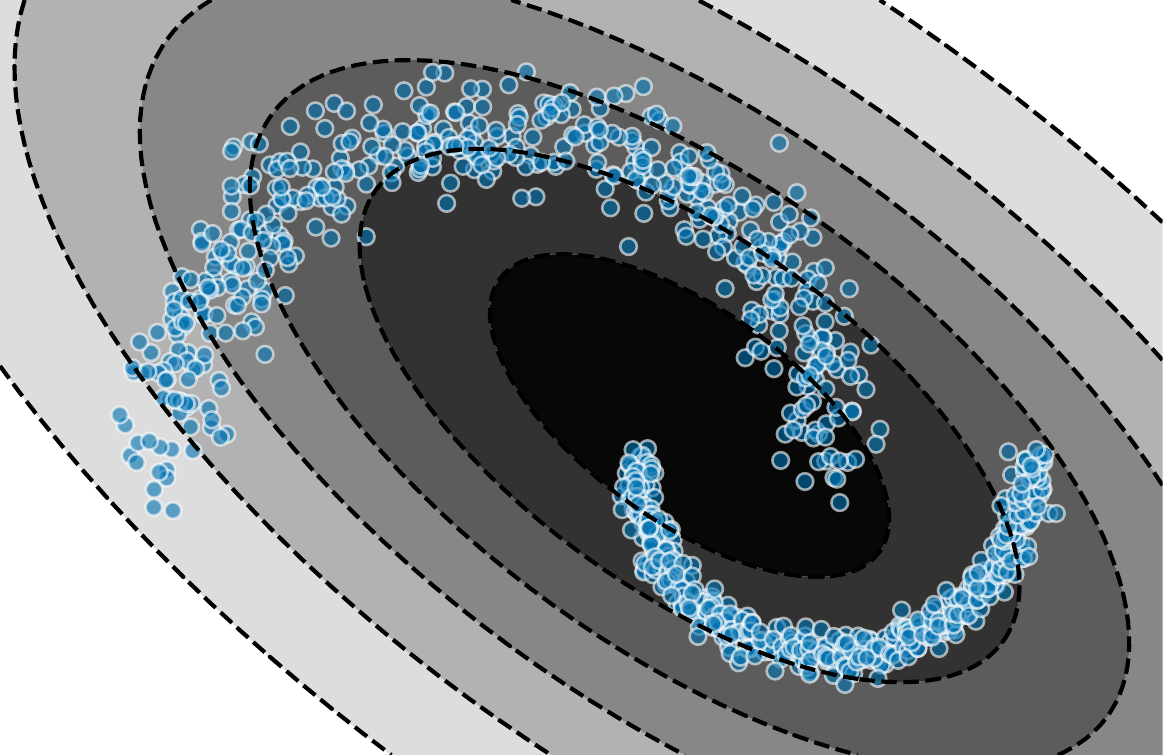}}
\fbox{\includegraphics[width=0.32\linewidth]{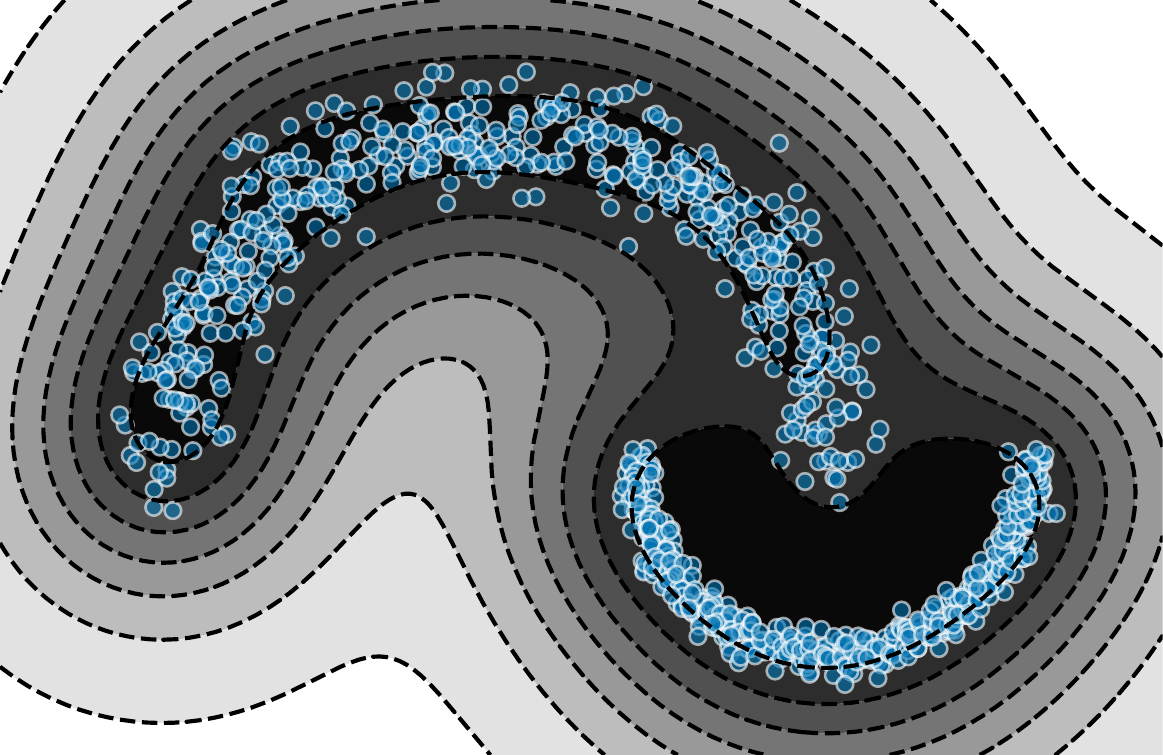}}
\fbox{\includegraphics[width=0.32\linewidth]{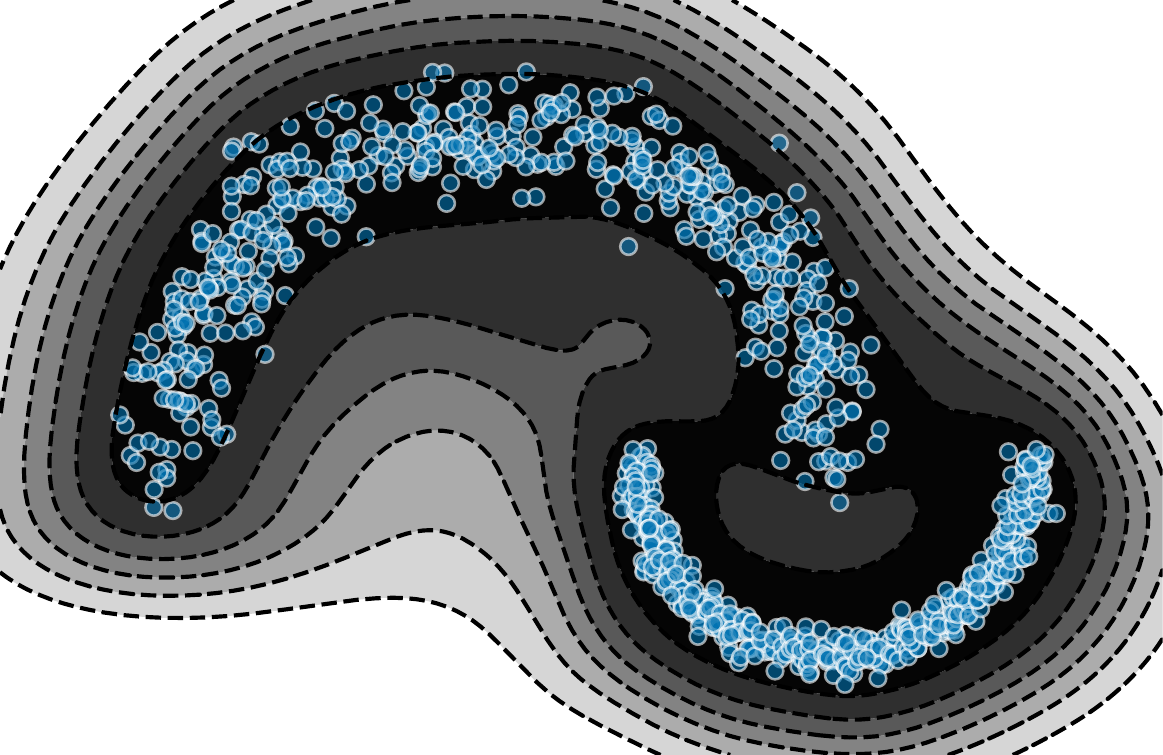}}

\caption{One-class classification models on the \emph{Big Moon, Small Moon} toy example (see Fig.~\ref{fig:2D_toy_example}). A Minimum Volume Ellipsoid (MVE) in input space is limited to enclose an ellipsoidal, convex region. By (implicitly) fitting a hypersphere in kernel feature space, SVDD enables non-convex support estimation. Deep SVDD learns an (explicit) neural feature map (here with smooth ELU activations) that extracts multiple data scales to fit a hypersphere model in feature space for support description.}
\label{fig:2D_toy_one-class}
\end{figure}

Deep one-class classification methods generally offer a greater modeling flexibility and enable the learning or transfer of task-relevant features for complex data.
They usually require more data to be effective though, or must rely on some informative domain prior (e.g., some pre-trained network).
However, the underlying principle of one-class classification methods\,---\,targeting a discriminative one-class boundary in learning\,---\,remains unaltered, regardless of whether a deep or shallow feature map is used.
We show three canonical one-class classification models (MVE, SVDD, and DSVDD) trained on the \emph{Big Moon, Small Moon} toy data set, each using a different feature representation (raw input, kernel, and neural network), in Fig.~\ref{fig:2D_toy_one-class} for comparison.

%%%%%%%%%%%%%%%%%%%%%%%%%%%%%%%%%%%%%%%%%%%%%%%%%%%%%%%%%%%%%%%%%%%%%%%%%%%%%%%%
%%%%%%%%%%%%%%%%%%%%%%%%%%%%%%%%%%%%%%%%%%%%%%%%%%%%%%%%%%%%%%%%%%%%%%%%%%%%%%%%

\subsection{Negative Examples}
\label{ssec:negative_samples}

One-class classifiers can usually incorporate labeled negative examples ($y=-1$) in a direct manner due to their close connection to binary classification as explained above.
Such negative examples can facilitate an empirical estimation of the miss rate (cf., \eqref{eqn:one-class_risk} and \eqref{eqn:one-class_objective}).
We here recognize three qualitative types of negative examples that have been studied in the literature, that we distinguish as \emph{artificial}, \emph{auxiliary}, and \emph{true} negative examples which increase in their informativeness in this order.

The idea to approach unsupervised learning problems through generating \emph{artificial} data points has been around for some time (see section 14.2.4 in \cite{hastie2009}).
If we assume that the anomaly distribution $\Pout$ has some form that we can generate data from, one idea would be to simply train a binary classifier to discern between the normal and the artificial negative examples. 
For the uniform prior $\Pout \equiv \calU(\calX)$, this approach yields an asymptotically consistent density level set estimator \cite{steinwart2005}. 
However, classification against uniformly drawn points from a hypercube quickly becomes ineffective in higher dimensions.
To improve over artificial uniform sampling, more informed sampling strategies have been proposed \cite{steinbuss2020} such as resampling schemes \cite{theiler2003}, manifold sampling \cite{davenport2006}, and sampling based on local density estimation \cite{fan2004,cheema2016} as well as active learning strategies \cite{abe2006,stokes2008,gornitz2009}.
Another recent idea is to treat the enormous quantities of data that are publicly available in some domains as \emph{auxiliary} negative examples \cite{hendrycks2019a}, for example images from photo sharing sites for computer vision tasks and the English Wikipedia for NLP tasks.
Such auxiliary examples provide more informative domain knowledge, for instance about the distribution of natural images or the English language in general, as opposed to sampling random pixels or words.
This approach, called \emph{Outlier Exposure} \cite{hendrycks2019a}, which trains on known anomalies can significantly improve deep anomaly detection performance in some domains \cite{hendrycks2019a,hendrycks2019d}.
Outlier exposure has also been used with density-based methods by employing a margin loss \cite{hendrycks2019a} or temperature annealing \cite{schirrmeister2020} on the log-likelihood ratio between positive and negative examples.
The most informative labeled negative examples are ultimately \emph{true} anomalies, for example verified by some domain expert.
Access to even a few labeled anomalies has been shown to improve detection performance significantly \cite{tax2001,gornitz2013,ruff2020}.
There also have been active learning algorithms proposed that include subjective user feedback (e.g., from an expert) to learn about the user-specific informativeness of particular anomalies in an application \cite{pelleg2005}.
Finally, we remark that negative examples have also been incorporated heuristically into reconstruction models via using a bounded reconstruction error \cite{du2019} since maximizing the unbounded error for negative examples can quickly become unstable. 
We will turn to reconstruction models next.

\section{Reconstruction Models}
\label{sec:reconstruction}

Models that are trained on a reconstruction objective are among the earliest \cite{japkowicz1995,hawkins2002} and most common \cite{chalapathy2019,pang2020} neural network approaches to anomaly detection. 
Reconstruction-based methods learn a model that is optimized to well-reconstruct normal data instances, thereby aiming to detect anomalies by \emph{failing} to accurately reconstruct them under the learned model.
Most of these methods have a purely geometric motivation (e.g., PCA or deterministic autoencoders), yet some probabilistic variants reveal a connection to density (level set) estimation.
In this section, we define the general reconstruction learning objective, highlight common underlying assumptions, as well as present standard reconstruction-based methods and discuss their variants.

\subsection{Reconstruction Objective}
\label{ssec:reconstruction_objective}
Let $\phi_\theta : \calX \to \calX, \bx \mapsto \phi_\theta(\bx)$ be a feature map from the data space $\calX$ onto itself that is composed of an \emph{encoding} function $\enc : \calX \to \calZ$ (the \emph{encoder}) and a \emph{decoding} function $\dec : \calZ \to \calX$ (the \emph{decoder}), that is, $\phi_\theta \equiv (\dec \circ \enc)_\theta$ where $\theta$ holds the parameters of both the encoder and decoder.
We call $\calZ$ the \emph{latent space} and $\enc(\bx) = \bz$ the \emph{latent representation} (or \emph{embedding} or \emph{code}) of $\bx$.
The reconstruction objective then is to learn $\phi_\theta$ such that $\phi_\theta(\bx) = \dec(\enc(\bx)) = \hat{\bx} \approx \bx$, that is, to find some encoding and decoding transformation so that $\bx$ is reconstructed with minimal error, usually measured in Euclidean distance.
Given unlabeled data $\bx_1, \ldots, \bx_n \in \calX$, the reconstruction objective is given by
\begin{equation}
\label{eqn:rec_objective}
    \min_\theta \quad \frac{1}{n} \sum_{i=1}^n \Vert \bx_i - (\dec \circ \enc)_\theta(\bx_i) \Vert^2 + \calR,
\end{equation}
where $\calR$ again denotes the different forms of regularization that various methods introduce, for example on the parameters $\theta$, the structure of the encoding and decoding transformations, or the geometry of latent space $\calZ$.
Without any restrictions, the reconstruction objective \eqref{eqn:rec_objective} would be optimally solved by the identity map $\phi_\theta \equiv \mathrm{id}$, but then of course nothing would be learned from the data.
In order to learn something useful, structural assumptions about the data-generating process are therefore necessary.
We here identify two principal assumptions: the manifold and the prototype assumptions.

\subsubsection{The Manifold Assumption}
The manifold assumption asserts that the data lives (approximately) on some lower-dimensional (possibly non-linear and non-convex) manifold $\calM$ that is embedded within the data space $\calX$\,---\,that is $\calM \subset \calX$ with $\dim(\calM) < \dim(\calX)$.
In this case $\calX$ is sometimes also called the \emph{ambient} or \emph{observation space}.
For natural images observed in pixel space, for instance, the manifold captures the structure of scenes as well as variation due to rotation and translation, changes in color, shape, size, texture, and so on. 
For human voices observed in audio signal space, the manifold captures variation due to the words being spoken as well as person-to-person variation in the anatomy and physiology of the vocal folds.
The (approximate) manifold assumption implies that there exists a lower-dimensional latent space $\calZ$ as well as functions $\enc: \calX \mapsto \calZ$ and $\dec: \calZ \mapsto \calX$ such that for all $x \in \calX$, $x \approx \dec(\enc(x))$.
Consequently, the generating distribution $\Pgen$ can be represented as the push-forward through $\dec$ of a latent distribution $\Pgen_Z$. Equivalently, the latent distribution $\Pgen_Z$ is the push-forward of $\Pgen$ through $\enc$. 

The goal of learning is therefore to learn the pair of functions $\enc$ and $\dec$ so that  $\dec(\enc(\calX)) \approx \calM \subset \calX$.
Methods that incorporate the manifold assumption usually restrict the latent space $\calZ \subseteq \bbR^d$ to have much lower dimensionality $d$ than the data space $\calX \subseteq \bbR^D$ (i.e., $d \ll D$).
The manifold assumption is also widespread in related unsupervised learning tasks such as manifold learning itself \cite{lee2007,pless2009}, dimensionality reduction \cite{kohonen1990self,scholkopf1998,kohonen2001,vanderMaaten2009}, disentanglement \cite{schmidhuber1992,locatello2019}, and representation learning in general \cite{bengio2013,tschannen2018}.

\subsubsection{The Prototype Assumption}
\label{sssec:prototype}
The prototype assumption asserts that there exists a finite number of prototypical elements in the data space $\calX$ that characterize the data well.
We can model this assumption in terms of a data-generating distribution that depends on a discrete latent categorical variable $Z \in \calZ = \{1, \ldots, K\}$ that captures some $K$ prototypes or modes of the data distribution.
This prototype assumption is also common in clustering and classification when we assume a collection of prototypical instances represent clusters or classes well.
With the reconstruction objective under the prototype assumption, we aim to learn an encoding function that for $\bx \in \calX$ identifies a $\enc(\bx) = k \in \{1, \ldots, K\}$ and a decoding function $k \mapsto \dec(k) = \bc_k $ that maps to some $k$-th prototype (or some prototypical distribution or mixture of prototypes more generally) such that the reconstruction error $\Vert \bx - \bc_k \Vert$ becomes minimal.
In contrast to the manifold assumption where we aim to describe the data by some continuous mapping, under the (most basic) prototype assumption we characterize the data by a discrete set of vectors $\{\bc_1, \ldots, \bc_K \} \subseteq \calX$.
The method of representing a data distribution by a set of prototype vectors is also known as Vector Quantization (VQ) \cite{linde1980algorithm,gersho1992}.

\subsubsection{The Reconstruction Anomaly Score}
A model that is trained on the reconstruction objective must extract salient features and characteristic patterns from the data in its encoding\,---\,subject to imposed model assumptions\,---\,so that its decoding from the compressed latent representation achieves low reconstruction error (e.g., feature correlations and dependencies, recurring patterns, cluster structure, statistical redundancy, etc.).
Assuming that the training data $\bx_1, \ldots, \bx_n \in \calX$ includes mostly normal points, we therefore expect a reconstruction-based model to produce a \emph{low} reconstruction error for normal instances and a \emph{high} reconstruction error for anomalies.
For this reason, the anomaly score is usually also directly defined by the reconstruction error:
\begin{equation}
\label{eqn:rec_score}
    s(\bx) = \Vert \bx - (\dec \circ \enc)_{\theta}(\bx) \Vert^2.
\end{equation}
For models that have learned some truthful manifold structure or prototypical representation, a high reconstruction error would then detect off-manifold or non-prototypical instances.

Most reconstruction methods do not follow any probabilistic motivation, and a point $\bx$ gets flagged anomalous simply because it does not conform to its `idealized' representation $\dec(\enc(\bx)) = \hat{\bx}$ under the encoding and decoding process.
However, some reconstruction methods also have probabilistic interpretations, such as PCA \cite{tipping1999}, or even are derived from probabilistic objectives such as Bayesian PCA \cite{bishop1999} or VAEs \cite{kingma2014}.
These methods are again related to density (level set) estimation (under specific assumptions about some latent structure), usually in the sense that a high reconstruction error indicates low density regions and vice versa.

\subsection{Principal Component Analysis}
A common way to formulate the Principal Component Analysis (PCA) objective is to seek an orthogonal basis $W$ in data space $\calX \subseteq \bbR^D$ that maximizes the empirical variance of the (centered) data $\bx_1, \ldots, \bx_n \in \calX$:
\begin{equation}
\label{eqn:pca_variance}
    \max_W \; \sum_{i=1}^n \Vert W \bx_i \Vert^2 \quad \text{s.t.} \; WW^\top = I.
\end{equation}
Solving this objective results in a well-known eigenvalue problem, since the optimal basis is given by the eigenvectors of the empirical covariance matrix where the respective eigenvalues correspond to the component-wise variances \cite{jolliffe2002}.
The $d \leq D$ components that explain most of the variance\,---\,the principal components\,---\,are then given by the $d$ eigenvectors that have the largest eigenvalues.

Several works have adapted PCA for anomaly detection \cite{hawkins1974,jackson1979,parra1996,shyu2003,huang2007,dutta2007,sharan2018}, which can be considered the default reconstruction baseline. 
From a reconstruction perspective, the objective to find an orthogonal projection $W^\top W$ to a $d$-dimensional linear subspace (which is the case for $W \in \bbR^{d{\times}D}$ with $WW^\top = I$) such that the mean squared reconstruction error is minimized,
\begin{equation}
\label{eqn:pca_project}
    \min_W \; \sum_{i=1}^n \Vert \bx_i - W^\top W \bx_i \Vert^2 \quad \text{s.t.} \; WW^\top = I,
\end{equation}
yields exactly the same PCA solution.
So PCA optimally solves the reconstruction objective \eqref{eqn:rec_objective} for a linear encoder $\enc(\bx) = W\bx = \bz$ and transposed linear decoder $\dec(\bz) = W^\top \bz$ with constraint $WW^\top = I$.
For linear PCA, we can also readily identify its probabilistic interpretation \cite{tipping1999}, namely that the data distribution follows from the linear transformation $X = W^{\top} Z + \varepsilon$ of a $d$-dimensional latent Gaussian distribution $Z \sim \calN(\bm{0}, I)$, possibly with added noise $\varepsilon \sim \calN(\bm{0}, \sigma^2 I)$, so that $\Pgen \equiv \calN(\bm{0}, W^{\top} W + \sigma^2 I)$.
Maximizing the likelihood of this Gaussian over the encoding and decoding parameter $W$ again yields PCA as the optimal solution \cite{tipping1999}.
Hence, PCA assumes the data lives on a $d$-dimensional ellipsoid embedded in data space $\calX \subseteq \bbR^D$.
Standard PCA therefore provides an illustrative example for the connections between density estimation and reconstruction.

Linear PCA, of course, is limited to data encodings that can only exploit linear feature correlations.
Kernel PCA \cite{scholkopf1998} introduced a non-linear generalization of component analysis by extending the PCA objective to non-linear kernel feature maps and taking advantage of the `kernel trick'.
For a PSD kernel $k(\bx, \tilde{\bx})$ with feature map $\phi_k : \calX \to \calF_k$, kernel PCA solves the reconstruction objective \eqref{eqn:pca_project} in feature space $\calF_k$,
\begin{equation}
\label{eqn:kernel_pca}
    \min_W \; \sum_{i=1}^n \Vert \phi_k(\bx_i) - W^\top W \phi_k(\bx_i) \Vert^2 \quad \text{s.t.} \; WW^\top = I,
\end{equation}
which results in an eigenvalue problem of the kernel matrix \cite{scholkopf1998}.
For kernel PCA, the reconstruction error can again serve as an anomaly score. It can be computed implicitly via the dual \cite{hoffmann2007}.
This reconstruction from linear principal components in feature space $\calF_k$ corresponds to a reconstruction from some non-linear subspace or manifold in input space $\calX$ \cite{ham2004}.
Replacing the reconstruction $W^\top W \phi_k(\bx)$ in \eqref{eqn:kernel_pca} with a prototype $\bc \in \calF_k$ yields a reconstruction model that considers the squared error to the kernel mean, since the prototype is optimally solved by $\bc = \tfrac{1}{n} \sum_{i=1}^n \phi(\bx_i)$ for the $L^2$-distance.
For RBF kernels, this prototype model is (up to a multiplicative constant) equivalent to kernel density estimation \cite{hoffmann2007}, which provides a link between kernel reconstruction and nonparametric density estimation methods.
Finally, Robust PCA variants have been introduced as well \cite{kwak2008,nguyen2009,candes2011,xiao2013}, which account for data contamination or noise (cf., \ref{sssec:unsupervised}).

%%%%%%%%%%%%%%%%%%%%%%%%%%%%%%%%%%%%%%%%%%%%%%%%%%%%%%%%%%%%%%%%%%%%%%%%%%%%%%%%
\begin{figure}[!t]
\centering
\setlength{\fboxsep}{0em}

\parbox{.32\linewidth}{\centering \footnotesize \sffamily PCA (AUC=66.8)}
\parbox{.32\linewidth}{\centering \footnotesize \sffamily kPCA (AUC=94.0)}
\parbox{.32\linewidth}{\centering \footnotesize \sffamily AE (AUC=97.9)}

\vspace{0.7mm}

\fbox{\includegraphics[width=0.32\linewidth]{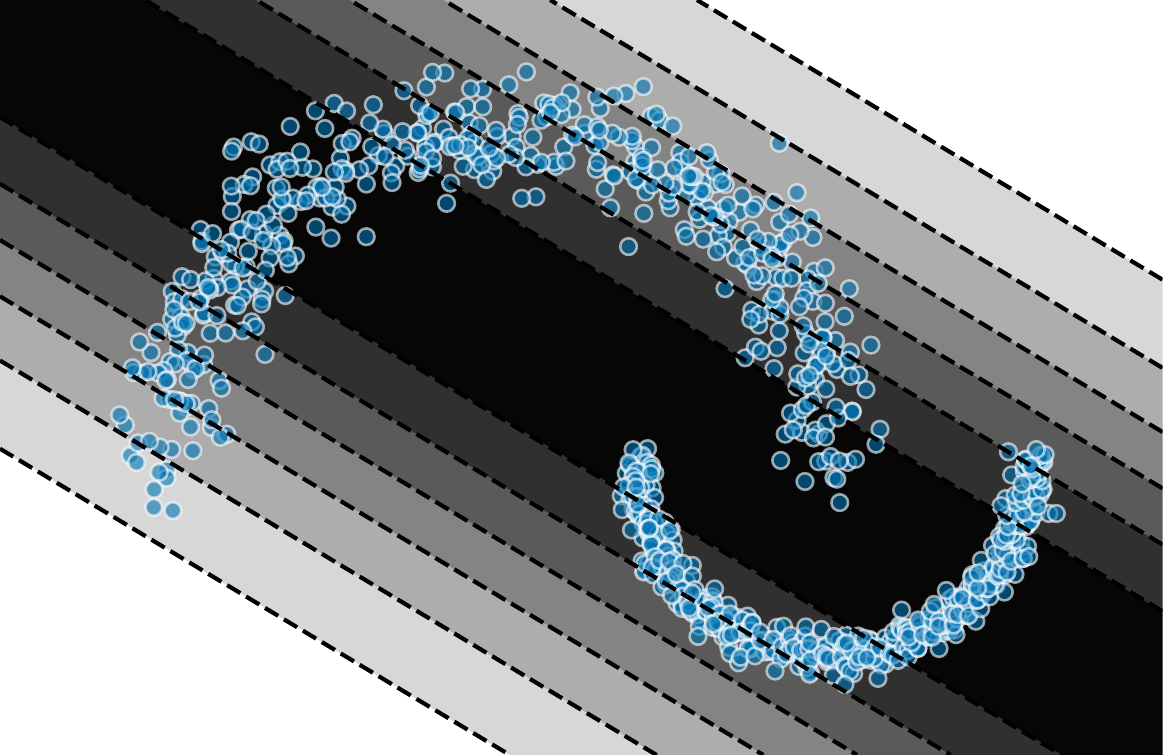}}
\fbox{\includegraphics[width=0.32\linewidth]{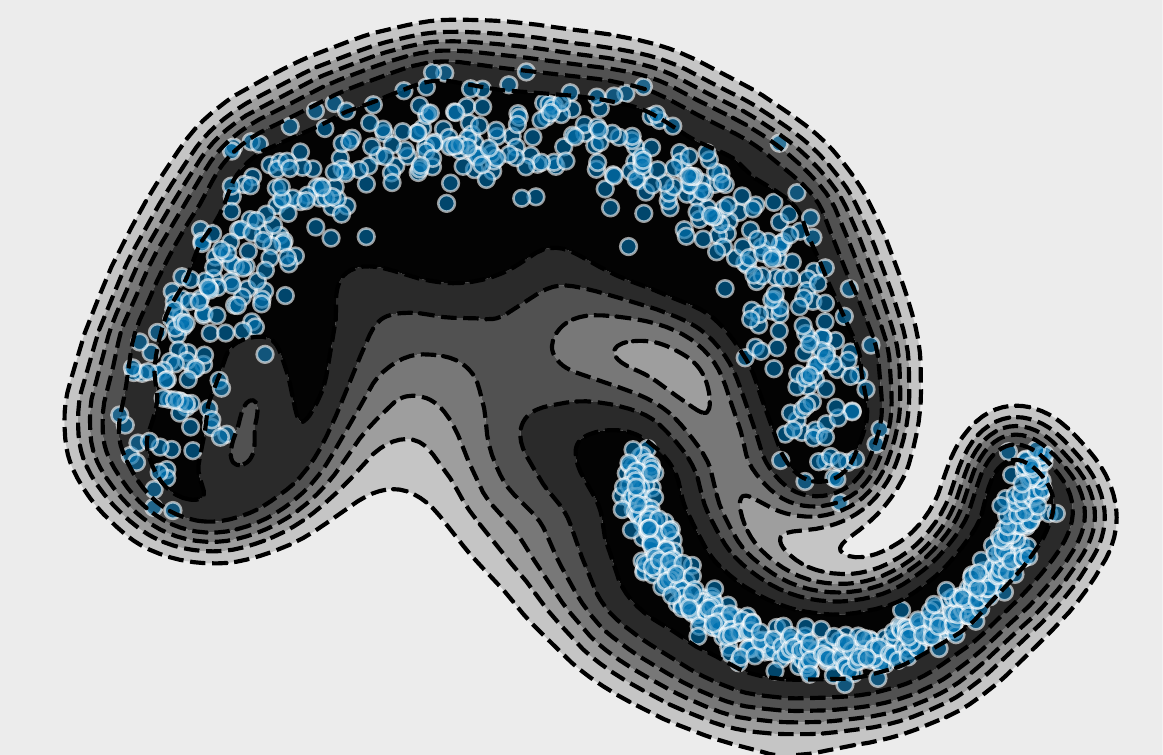}}
\fbox{\includegraphics[width=0.32\linewidth]{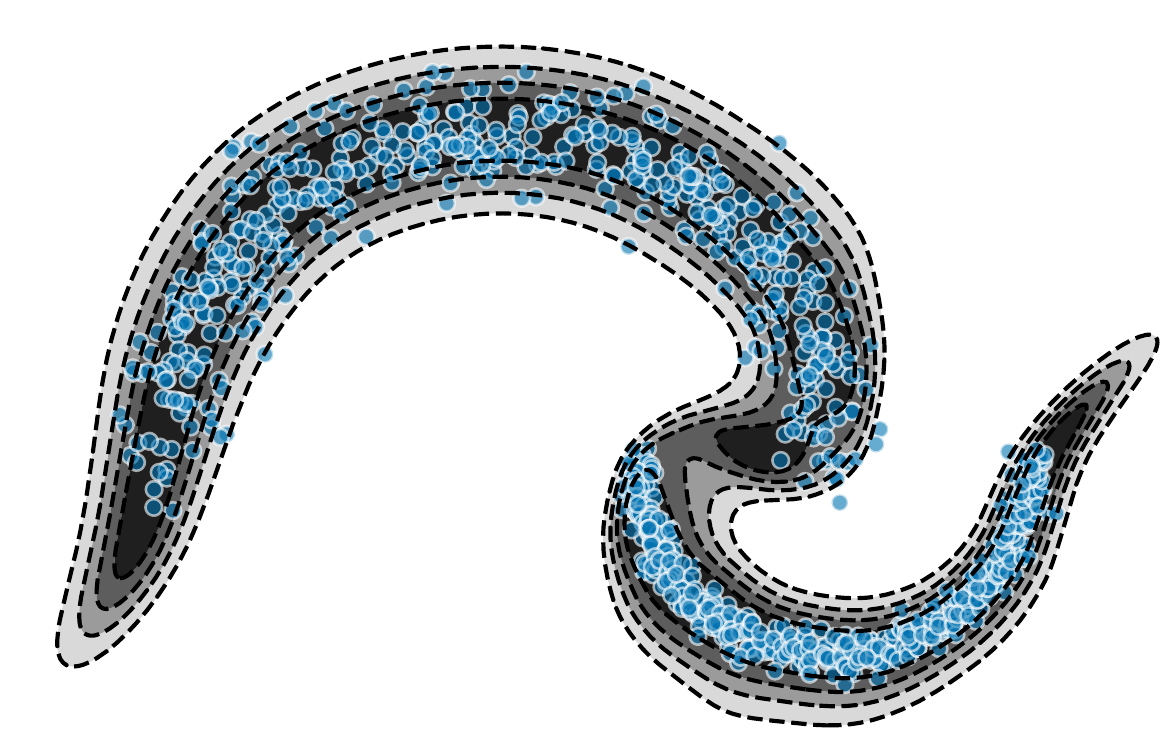}}

\caption{Reconstruction models on the \emph{Big Moon, Small Moon} toy example (see Fig.~\ref{fig:2D_toy_example}). PCA finds the linear subspace with the lowest reconstruction error under an orthogonal projection of the data. Kernel PCA (kPCA) solves (linear) component analysis in kernel feature space which enables an optimal reconstruction from (kernel-induced) non-linear components in input space. An autoencoder (AE) with one-dimensional latent code learns a one-dimensional, non-linear manifold in input space having minimal reconstruction error.}
\label{fig:2D_toy_reconstruction}
\end{figure}
%%%%%%%%%%%%%%%%%%%%%%%%%%%%%%%%%%%%%%%%%%%%%%%%%%%%%%%%%%%%%%%%%%%%%%%%%%%%%%%%

\subsection{Autoencoders}
\label{ssec:autoencoders}

Autoencoders are reconstruction models that use neural networks for the encoding and decoding of data.
They were originally introduced during the 80s \cite{oja1982,rumelhart1986,ballard1987,hinton1989} primarily as methods to perform non-linear dimensionality reduction \cite{kramer1991,hinton2006}, yet they have also been studied early on for anomaly detection \cite{japkowicz1995,hawkins2002}.
Today, deep autoencoders are among the most widely adopted methods for deep anomaly detection in the literature \cite{chalapathy2017,chen2017b,principi2017,zhou2017,zong2018,aytekin2018,chen2018,pawlowski2018,abati2019,huang2019,gong2019,oza2019b,nguyen2019,kim2020} likely owing to their long history and easy-to-use standard variants.
The standard autoencoder objective is given by
\begin{equation}
\label{eqn:autoencoder_objective}
    \min_\omega \quad \frac{1}{n} \sum_{i=1}^n \Vert \bx_i - (\dec \circ \enc)_\omega(\bx_i) \Vert^2 + \calR,
\end{equation}
which is a realization of the general reconstruction objective \eqref{eqn:rec_objective} with $\theta = \omega$, that is, the optimization is carried out over the weights $\omega$ of the neural network encoder and decoder.
A common way to regularize autoencoders is by mapping to a lower dimensional `bottleneck' representation $\enc(\bx) = \bz \in \calZ$ through the encoder network, which enforces data compression and effectively limits the dimensionality of the manifold or subspace to be learned.
If linear networks are used, such an autoencoder in fact recovers the same optimal subspace as spanned by the PCA eigenvectors \cite{baldi1989,oja1992principal}.
In Fig.~\ref{fig:2D_toy_reconstruction}, we show a comparison of three canonical reconstruction models (PCA, kPCA, and AE) trained on the \emph{Big Moon, Small Moon} toy data set, each using a different feature representation (raw input, kernel, and neural network), resulting in different manifolds.
Apart from a `bottleneck', a number of different ways to regularize autoencoders have been introduced in the literature.
Following ideas of sparse coding \cite{olshausen1996,olshausen1997,lewicki2000,lee2007b}, sparse autoencoders \cite{makhzani2014,zeng2018} regularize the (possibly higher-dimensional, over-complete) latent code towards sparsity, for example via $L^1$ Lasso penalization \cite{arpit2016}.
Denoising autoencoders (DAEs) \cite{vincent2008,vincent2010} explicitly feed noise-corrupted inputs $\tilde{\bx} = \bx + \varepsilon$ into the network which is then trained to reconstruct the original inputs $\bx$. 
DAEs thus provide a way to specify a noise model for $\varepsilon$ (cf., \ref{sssec:unsupervised}), which has been applied for noise-robust acoustic novelty detection \cite{marchi2015}, for instance.
In situations in which the training data is already corrupted with noise or unknown anomalies, robust deep autoencoders \cite{zhou2017}, which split the data into well-represented and corrupted parts similar to robust PCA \cite{candes2011}, have been proposed.
Contractive autoencoders (CAEs) \cite{rifai2011} propose to penalize the Frobenius norm of the Jacobian of the encoder activations with respect to the inputs to obtain a smoother and more robust latent representation.
Such ways of regularization influence the geometry and shape of the subspace or manifold that is learned by the autoencoder, for example by imposing some degree of smoothness or introducing invariances towards certain types of input corruptions or transformations \cite{huang2019}.
Hence, these regularization choices should again reflect the specific assumptions of a given anomaly detection task.

Besides the deterministic variants above, probabilistic autoencoders have also been proposed, which again establish a connection to density estimation.
The most explored class of probabilistic autoencoders are Variational Autoencoders (VAEs) \cite{kingma2014,rezende2014,kingma2019}, as introduced in section \ref{sssec:vae} through the lens of neural generative models, which approximately maximize the data likelihood (or evidence) by maximizing the ELBO.
From a reconstruction perspective, VAEs adopt a stochastic autoencoding process, which is realized by encoding and decoding the parameters of distributions (e.g., Gaussians) through the encoder and decoder networks, from which the latent code and reconstruction then can be sampled.
For a standard Gaussian VAE, for example, where $q(\bz | \bx) \sim \calN(\bmu_{\bx}, \diag(\bsigma_{\bx}^2))$, $p(\bz) \sim \calN(\bm{0}, I)$, and $p(\bx | \bz) \sim \calN(\bmu_{\bz}, I)$ with encoder $\phi_{e,\omega'}(\bx) = (\bmu_{\bx}, \bsigma_{\bx})$ and decoder $\phi_{d,\omega}(\bz) = \bmu_{\bz}$, the empirical ELBO objective \eqref{eqn:elbo} becomes
\begin{equation}\label{eqn:vae_objective}
\begin{split}
    \min_{\omega, \omega'} \; & \, \frac{1}{n} \sum_{i=1}^n \sum_{j=1}^M \; \Big[ \tfrac{1}{2} \Vert \bx_i - \bmu_{\bz_{ij}} \Vert^2\\ 
    &+ D_{\KL}\left(\calN(\bz_{ij}; \bmu_{\bx_i}, \diag(\bsigma_{\bx_i}^2)) \Vert \calN(\bz_{ij}; \bm{0}, I)\right) \Big],
\end{split}
\end{equation}
where $\bz_{i1}, \ldots, \bz_{iM}$ are $M$ Monte Carlo samples drawn from the encoding distribution $\bz \sim q(\bz | \bx_i)$ of $\bx_i$.
Hence, such a VAE is trained to minimize the mean reconstruction error over samples from an encoded latent Gaussian that is regularized to be close to a standard isotropic Gaussian.
VAEs have been used in various forms for anomaly detection \cite{an2015,xu2018,you2019}, for instance on multimodal sequential data with LSTMs in robot-assisted feeding \cite{park2018} and for new physics mining at the Large Hadron Collider \cite{cerri2019}.
Other probabilistic autoencoders that have been applied to anomaly detection are Adversarial Autoencoders (AAEs) \cite{makhzani2015,principi2017,chen2018}. 
By adopting an adversarial loss to regularize and match the latent encoding distribution, AAEs can employ any arbitrary prior $p(\bz)$, as long as sampling is feasible.

Finally, other autoencoder variants that have been applied to anomaly detection include RNN-based autoencoders \cite{malhotra2016,lu2017,kiran2018,kieu2019}, convolutional autoencoders \cite{pawlowski2018}, autoencoder ensembles \cite{chen2017b,kieu2019} and variants that constrain the gradients \cite{kwon2020} or actively control the latent code topology \cite{hofer2019} of an autoencoder.
Autoencoders also have been utilized in two-step approaches that use autoencoders for dimensionality reduction and apply traditional methods on the learned embeddings \cite{erfani2016,amarbayasgalan2018,sarafijanovic2019}.

\subsection{Prototypical Clustering}
\label{ssec:clustering}

Clustering methods that make the prototype assumption provide another approach to reconstruction-based anomaly detection.
As mentioned above, the reconstruction error here is usually given by the distance of a point to its nearest prototype, which ideally has been learned to represent a distinct mode of the normal data distribution.
Prototypical clustering methods \cite{jain2010} include the well-known Vector Quantization (VQ) algorithms $k$-means, $k$-medians, and $k$-medoids, which define a Voronoi partitioning \cite{voronoi1908a,voronoi1908b} over the metric space where they are applied\,---\,typically the input space $\calX$.
Kernel variants of $k$-means have also been studied \cite{dhillon2004} and considered for anomaly detection \cite{gornitz2017}.
GMMs with a finite number of $k$ mixtures (cf., section \ref{ssec:density_classic}) have been used for (soft) prototypical clustering as well. Here, the distance to each cluster (or mixture component) is given by the Mahalanobis distance that is defined by the covariance matrix of the respective Gaussian mixture component \cite{amruthnath2018}.

More recently, deep learning approaches to clustering have also been introduced \cite{xie2016,van2017,razavi2019,kampffmeyer2019}, some also based on $k$-means \cite{yang2017}, and adopted for anomaly detection \cite{aytekin2018,caron2018,amarbayasgalan2018}.
As in deep one-class classification (cf., section \ref{ssec:deep-one-class}), a persistent question in deep clustering is how to effectively regularize against a feature map collapse \cite{bojanowski2017b}. 
Note that whereas for deep clustering methods the reconstruction error is measured in latent space $\calZ$, for deep autoencoders it is measured in the input space $\calX$ after decoding.
Thus, a latent feature collapse (i.e., a constant encoder $\enc \equiv \bc \in \calZ$) would result in a constant decoding (the data mean at optimum) for an autoencoder, which generally is a suboptimal solution of \eqref{eqn:autoencoder_objective}.
For this reason, autoencoders seem less susceptible to a feature collapse, though they have also been observed to converge to bad local optima under SGD optimization, specifically if they employ bias terms \cite{ruff2018}.

%%%%%%%%%%%%%%%%%%%%%%%%%%%%%%%%%%%%%%%%%%%%%%%%%%%%%%%%%%%%%%%%%%%%%%%%%%%%%%%%
\begin{table*}[th]
    \caption{Anomaly detection methods identified with our unifying view (last column contains representative references).}
    \label{tab:taxonomy_methods}
    \centering
    \resizebox{\linewidth}{!}{\begin{tabular}{llllllllc}
    \toprule
    Method                  & Loss $\ell(s, y)$     & Model $f_\theta(\bx)$                                             & \multicolumn{2}{l}{Feature Map $\phi(\bx)$}   & Parameter $\theta$    & Regularization $\calR(f, \phi, \theta)$                                   & Bayes?        & References \\
    \midrule
    Parametric Density      & $-\log(s)$            & $p(\bx | \theta)$                                                 & $\bx$ & (input)                               & $\theta$              & choice of density class $\{p_{\theta} \, | \, \theta \in \Theta\}$        & \xmark & \cite{bishop2006,murphy2012} \\[1.5pt]
    Gaussian/Mahalanobis    & $-\log(s)$            & $\calN(\bx | \bmu, \Sigma)$                                       & $\bx$ & (input)                               & $(\bmu, \Sigma)$      & --                                                                        & \xmark & \cite{bishop2006,murphy2012} \\[1.5pt]
    GMM                     & $-\log(s)$            & $\sum_k \pi_k \, \calN(\bx | \bmu_k, \Sigma_k)$                   & $\bx$ & (input)                               & $(\pi, \bmu, \Sigma)$ & number of mixture components $K$                                          & latent & \cite{theodoridis2020} \\[1.5pt]
    KDE                     & $-\log(s)$            & $\exp(-\Vert \phi_k(\bx) - \bmu \Vert^2)$                         & $\phi_k(\bx)$ & (kernel)                      & $\bmu$                & kernel hyperparameters (e.g., bandwidth $h$)                              & \xmark & \cite{roberts1994b,bishop1994} \\[1.5pt]
    EBMs                    & $-\log(s)$            & $\tfrac{1}{Z(\theta)} \exp(-E(\phi(\bx), \bz; \theta))$           &  $\phi(\bx)$  & (various)                     & $\theta$              & latent prior $p(\bz)$ & latent & \cite{lecun2006,zhai2016} \\[1.5pt]
    Normalizing Flows       & $-\log(s)$            & $p_{\bz}(\phi_{\omega}^{-1}(\bx)) \, |\det J_{\phi_{\omega}^{-1}}(\bx)|$ & $\phi_\omega(\bx)$ & (neural)                 & $\omega$            & base distribution $p_{\bz}(\bz)$; diffeomorphism architecture & \xmark & \cite{papamakarios19,nachman20} \\[1.5pt]
    GAN ($D$-based)         & $-\log(s)$ & $\sigma(\langle \bw,\psi_\omega(\bx)\rangle)$ & $\psi_\omega(\bx)$ & (neural) & $(\bw,\omega)$ & adversarial training & \xmark & \cite{schlegl2019,sabokrou2020} \\[1.5pt]
    \midrule
    Min.\ Vol.\ Sphere      & $\max(0, s)$          & $\Vert \bx - \bc \Vert^2 - R^2$                                   & $\bx$ & (input)                               & $(\bc, R)$            & $\nu R^2$                                                     & \xmark & \cite{tax2001} \\[1.5pt]
    Min.\ Vol.\ Ellipsoid   & $\max(0, s)$          & $(\bx - \bc)^\top \Sigma^{-1} (\bx - \bc) - R^2$                  & $\bx$ & (input)                               & $(\bc, R, \Sigma)$    & $\nu (\tfrac{1}{2} \Vert \Sigma \Vert_{\Fr}^2 + R^2)$          & \xmark & \cite{rousseeuw1999} \\[1.5pt]
    SVDD                    & $\max(0, s)$          & $\Vert \phi_k(\bx) - \bc \Vert^2 - R^2$                           & $\phi_k(\bx)$ & (kernel)                      & $(\bc, R)$            & $\nu R^2$                                                     & \xmark & \cite{tax2004} \\[1.5pt]
    Semi-Sup.\ SVDD         & $\max(0, ys)$         & $\Vert \phi_k(\bx) - \bc \Vert^2 - R^2$                           & $\phi_k(\bx)$ & (kernel)                      & $(\bc, R)$            & $\nu R^2$                                                     & \xmark & \cite{tax2004,gornitz2013} \\[1.5pt]
    Soft Deep SVDD          & $\max(0, s)$          & $\Vert \phi_\omega(\bx) - \bc \Vert^2 - R^2$                           & $\phi_\omega(\bx)$ & (neural)                      & $(\bc, R, \omega)$         & $\nu R^2$; weight decay; collapse reg.\ (various)  & \xmark & \cite{ruff2018} \\[1.5pt]
    OC Deep SVDD            & $s$                   & $\Vert \phi_\omega(\bx) - \bc \Vert^2$                                 & $\phi_\omega(\bx)$ & (neural)                      & $(\bc, \omega)$            & weight decay; collapse reg.\ (various)               & \xmark & \cite{ruff2018} \\[1.5pt]
    Deep SAD                & $s^y$                 & $\Vert \phi_\omega(\bx) - \bc \Vert^2$                                 & $\phi_\omega(\bx)$ & (neural)                      & $(\bc, \omega)$            & weight decay                                  & \xmark & \cite{ruff2020} \\[1.5pt]
    OC-SVM                  & $\max(0, s)$          & $\rho - \langle \bw, \phi_k(\bx) \rangle$                         & $\phi_k(\bx)$ & (kernel)                      & $(\bw, \rho)$         & $\nu (\tfrac{1}{2} \Vert \bw \Vert^2 - \rho)$                  & \xmark & \cite{scholkopf2001} \\[1.5pt]
    OC-NN                   & $\max(0, s)$          & $\rho - \langle \bw, \phi_\omega(\bx) \rangle$                         & $\phi_\omega(\bx)$ & (neural)                      & $(\bw, \rho, \omega)$      & $\nu (\tfrac{1}{2} \Vert \bw \Vert^2 - \rho)$; weight decay & \xmark & \cite{chalapathy2018b} \\[1.5pt]
    Bayesian DD             & $\max(0, s)$          & $\Vert \phi_k(\bx) - \bc \Vert^2 - R^2$                           & $\phi_k(\bx)$ & (kernel)                      & $(\bc, R)$            & $\bc=\sum_i\alpha_i\phi_k(\bx_i)$ with prior $\alpha\sim\mathcal N(\boldsymbol\mu,\Sigma)$ & fully & \cite{ghasemi2012} \\[1.5pt]
    GT                      & $-\log(s)$            & $\prod_k \sigma_k(\langle \bw, \phi_\omega(T_k(\bx))\rangle)$   & $\phi_\omega(\bx)$ & (neural)                 & $(\bw,\omega)$        & transformations $\calT = \{T_1, \ldots, T_K\}$ for self-labeling & \xmark & \cite{golan2018,hendrycks2019d} \\[1.5pt]
    GOAD (CE)               & $-\log(s)$            & $\prod_k \sigma_k(-\Vert \phi_\omega(T_k(\bx)) - \bc_k \Vert^2)$  & $\phi_\omega(\bx)$ & (neural)                 & $(\bc_1, \ldots, \bc_K, \omega)$        & transformations $\calT = \{T_1, \ldots, T_K\}$ for self-labeling & \xmark & \cite{bergman2020b} \\[1.5pt]
    BCE (supervised)        & $-y \log(s){-}\tfrac{1{-}y}{2}\log(1{-}s)$ & $\sigma(\langle \bw, \phi_\omega(\bx)\rangle)$                & $\phi_\omega(\bx)$ & (neural)                      & $(\bw,\omega)$             & weight decay & \xmark & \cite{ruff2020b} \\[1.5pt]
    BNN (supervised)        & $-y \log(s){-}\tfrac{1{-}y}{2}\log(1{-}s)$ & $\sigma(\langle \bw, \phi_\omega(\bx)\rangle)$                & $\phi_\omega(\bx)$ & (neural)                      & $(\bw,\omega)$             & prior $p(\bw,\omega)$ & fully & \cite{mackay1992,blundell2015} \\
    \midrule
    PCA                     & $s$                   & $\Vert \bx - W^\top W \bx \Vert_2^2$                                      & $\bx$ & (input)                       & $W$                         & $W W^\top = I$                                                & \xmark & \cite{hawkins1974} \\[1.5pt]
    Robust PCA              & $s$                   & $\Vert \bx - W^\top W \bx \Vert_1$                                        & $\bx$ & (input)                       & $W$                         & $W W^\top = I$                                                & \xmark & \cite{kwak2008} \\[1.5pt]
    Probabilistic PCA       & $-\log(s)$            & $\calN(\bx | \bm{0}, W^\top W + \sigma^2 I)$                                   & $\bx$ & (input)                       & $(W, \sigma^2)$           & linear latent Gauss model $\bx = W^\top \bz + \varepsilon$    & latent & \cite{tipping1999} \\[1.5pt]
    Bayesian PCA            & $-\log(s)$            & $\calN(\bx | \bm{0}, W^\top W + \sigma^2 I) \, p(W | \bm{\alpha})$                & $\bx$ & (input)                       & $(W, \sigma^2)$           & linear latent Gauss model with prior $p(W | \bm{\alpha})$     & fully & \cite{bishop1999} \\[1.5pt]
    Kernel PCA              & $s$                   & $\Vert \phi_k(\bx) - W^\top W \phi_k(\bx) \Vert^2$                        & $\phi_k(\bx)$ & (kernel)              & $W$                         & $W W^\top = I$                                                & \xmark & \cite{scholkopf1998,hoffmann2007} \\[1.5pt]
    Autoencoder             & $s$                   & $\Vert \bx - \phi_\omega(\bx) \Vert_2^2$                                       & $\phi_\omega(\bx)$ & (neural)              & $\omega$                         & advers.\ (AAE), contract.\ (CAE), denois.\ (DAE), etc.\                       & \xmark & \cite{zhou2017,kim2020} \\[1.5pt]
    VAE                     & $-\log(s)$            & $p_{\phi_\omega}(\bx | \bz) $                                                  & $\phi_\omega(\bx)$ & (neural)              & $\omega$                         & latent prior $p(\bz)$                                         & latent & \cite{an2015,kingma2019} \\[1.5pt]
    GAN ($G$-based)         & $-\log(s)$            & $p_{\phi_\omega}(\bx | \bz) $                                                  & $\phi_\omega(\bx)$ & (neural)              & $\omega$                         & adversarial training and latent prior $p(\bz)$                       & latent & \cite{schlegl2017,deecke2018} \\[1.5pt]
    $k$-means               & $s$                   & $\Vert \bx - \argmin_{\bc_k} \Vert \bx - \bc_k \Vert_2 \Vert_2^2$         & $\bx$ & (input)                       & $(\bc_1, \ldots, \bc_K)$  & number of prototypes $K$                                      & \xmark & \cite{jain2010,theodoridis2020} \\[1.5pt]
    $k$-medians             & $s$                   & $\Vert \bx - \argmin_{\bc_k} \Vert \bx - \bc_k \Vert_1 \Vert_1$           & $\bx$ & (input)                       & $(\bc_1, \ldots, \bc_K)$  & number of prototypes $K$                                      & \xmark & \cite{jain2010} \\[1.5pt]
    VQ                      & $s$                   & $\Vert \bx - \dec(\argmin_{\bc_k} \Vert \enc(\bx) - \bc_k \Vert) \Vert$   & $\phi(\bx)$   & (various)                     & $(\bc_1, \ldots, \bc_K)$  & number of prototypes $K$                                      & \xmark & \cite{linde1980algorithm,gersho1992} \\
    \bottomrule
\end{tabular}}
\end{table*}
%%%%%%%%%%%%%%%%%%%%%%%%%%%%%%%%%%%%%%%%%%%%%%%%%%%%%%%%%%%%%%%%%%%%%%%%%%%%%%%%

\section{A Unifying View of Anomaly Detection}
\label{sec:taxonomy}

In this section, we present a unifying view of the anomaly detection problem.
We identify specific anomaly detection modeling components that allow us to characterize the many methods discussed above in a systematic way.
Importantly, this view reveals connections that enable the transfer of algorithmic ideas between existing anomaly detection methods. 
Thus it uncovers promising directions for future research such as transferring concepts and ideas from kernel-based anomaly detection to deep methods and vice versa.

\subsection{Modeling Dimensions of the Anomaly Detection Problem}

We identify the following five components or \emph{modeling dimensions} for anomaly detection:

\begin{mdframed}[backgroundcolor=aliceblue,outerlinecolor=darkblue,outerlinewidth=0.5,roundcorner=4pt,innerleftmargin=2pt,innertopmargin=5pt]
\begin{tabular}{ll}
D1 \bf Loss           & $\ell: \bbR \times \calY \to \bbR, (s, y) \mapsto \ell(s , y)$\\[1.5pt]
D2 \bf Model          & $f_\theta: \calX \to \bbR, \bx \mapsto f_\theta(\bx)$\\[1.5pt]
D3 \bf Feature Map    & $\bx \mapsto \phi(\bx)$\\[1.5pt]
D4 \bf Regularization & $\calR(f, \phi, \theta)$\\[1.5pt]
D5 \bf Inference Mode & Frequentist or Bayesian $\theta \sim p(\theta)$
\end{tabular}
\end{mdframed}

Dimension D1 \textbf{Loss} is the (scalar) loss function that is applied to the output of some model $f_\theta(\bx)$.
Semi-supervised or supervised methods use loss functions that incorporate labels, but for the many unsupervised anomaly detection methods we have $\ell(s , y) = \ell(s)$.
D2 \textbf{Model} defines the specific model $f_\theta$ that maps an input $\bx \in \calX$ to some scalar value that is evaluated by the loss. 
We have arranged our previous three sections along this modeling dimension where we covered certain groups of methods that formulate models based on common principles, namely probabilistic modeling, one-class classification, and reconstruction.
Due to the close link between anomaly detection and density estimation (cf., \ref{sssec:density_level_set}), many of the methods formulate a likelihood model $f_\theta(\bx) = p_\theta(\bx \, | \, \calD_n)$ with negative log-loss $\ell(s) = - \log(s)$, that is, they have a negative log-likelihood objective, where $\calD_n = \{\bx_1, \ldots, \bx_n\}$ denotes the training data.
Dimension D3 captures the \textbf{Feature Map} $\bx \mapsto \phi(\bx)$ that is used in a model $f_\theta$.
This could be an (implicit) feature map $\phi_k(\bx)$ defined by some given kernel $k$ in kernel methods, for example, or an (explicit) neural network feature map $\phi_\omega(\bx)$ that is learned and parameterized with network weights $\omega$ in deep learning methods.
With dimension D4 \textbf{Regularization}, we capture various forms of regularization $\calR(f, \phi, \theta)$ of the model $f_\theta$, the feature map $\phi$, and their parameters $\theta$ in a broader sense.
Note that $\theta$ here may include both model parameters as well as feature map parameters, that is, $\theta = (\theta_f, \theta_\phi)$ in general.
$\theta_f$ could be the distributional parameters of a parametric density model, for instance, and $\theta_\phi$ the weights of a neural network.
Our last modeling dimension D5 describes the \textbf{Inference Mode}, specifically whether a method performs Bayesian inference \cite{theodoridis2020}.

The identification of the above modeling dimensions enables us to formulate a general anomaly detection learning objective that encompasses a broad range of anomaly detection methods:

\begin{mdframed}[backgroundcolor=aliceblue,outerlinecolor=darkblue,outerlinewidth=0.5,roundcorner=4pt,innerleftmargin=2pt,innertopmargin=5pt]
\begin{equation}
\label{eqn:unified_AD_objective}
    \min_\theta \quad \frac{1}{n} \sum_{i=1}^n \ell(f_\theta(\bx_i), y_i) + \calR(f, \phi, \theta) . \tag{$\ast$}
\end{equation}
\end{mdframed}

\noindent Denoting the minimum of \eqref{eqn:unified_AD_objective} by $\theta^*$, the anomaly score of a test input $\tilde{\bx}$ is computed via the model $f_{\theta^*}(\tilde{\bx})$. 
In the Bayesian case, where the objective in \eqref{eqn:unified_AD_objective} is the negative log-likelihood of a posterior $p(\theta \, | \, \mathcal{D}_n)$ induced by a prior distribution $p(\theta)$, we can predict in a fully Bayesian fashion via the expected model $\mathbb{E}_{\theta\sim p(\theta \, | \, \mathcal{D}_n)}f_{\theta}(\bx)$.
In Table \ref{tab:taxonomy_methods}, we describe many well-known anomaly detection methods using our unifying view.

\subsection{Comparative Discussion}

Below we compare the various approaches in light of our unifying view and discuss how this view enables the transfer of concepts between existing anomaly detection methods.
Table \ref{tab:taxonomy_methods} shows that the probabilistic methods are largely based on the negative log-likelihood objective.
The resulting negative log-likelihood anomaly scores provide a (usually continuous) ranking that is generally more informative than a binary density level set detector (cf., section \ref{sssec:threshold_vs_score}).
Reconstruction methods provide such a ranking as well, with the anomaly score given by the difference of a data instance and its reconstruction under the model.
Besides ranking and detecting anomalies, such scores make it possible to also rank inliers, which can be used, for example, to judge cluster memberships or determine prototypes (cf., section \ref{ssec:clustering}).
Reconstruction is particularly well suited when the data follows some manifold or prototypical structure (cf., section \ref{ssec:reconstruction_objective}).
In comparison, standard one-class classification methods, which aim to estimate a discriminative level set boundary (cf., section \ref{sec:one-class}), usually do not rank inliers.
This is typically incorporated into the learning objective via a hinge loss as can be seen in Table \ref{tab:taxonomy_methods}.
One-class classification is generally more sample-efficient and more robust to non-representative sampling of the normal data (e.g., a sampling bias towards specific normal modes) \cite{tax2001}, but is consequentially also less informative.
However, an inlier ranking for one-class classification can still be obtained via the distance of a point to the decision boundary, but such an approximate ranking may not faithfully represent in-distribution modes etc.
In addition to the theoretical comparison and discussion of anomaly detection methods in regard of our unifying view, we will present an empirical evaluation that includes methods from all three groups (probabilistic, one-class classification, and reconstruction) and three types of feature maps (raw input, kernel, and neural network) in section \ref{ssec:comparison}, where find that the detection performance in different data scenarios is very heterogeneous among the methods (with an advantage for deep methods on the more complex, semantic detection tasks). 
This exemplifies the fact that there is no simple `silver bullet' solution to the anomaly detection problem.

In addition to providing a framework for comparing methods, our unifying view also allows to identify concepts that may be transferred between shallow and deep anomaly detection methods in a systematic manner.
We discuss a few explicit examples to illustrate this point here.
Table \ref{tab:taxonomy_methods} shows that both the (kernel) SVDD and Deep SVDD employ a hypersphere model.
This connection can be used to transfer adaptations of the hypersphere model from one world to another (from shallow to deep or vice versa).
The adoption of semi-supervised \cite{gornitz2013,ruff2019b,ruff2020} or multi-sphere \cite{gornitz2017,ghafoori2020,bergman2020b} model extensions give successful examples for such a transfer.
Next, note in Table \ref{tab:taxonomy_methods}, that deep autoencoders usually consider the reconstruction error in the original data space $\calX$ after a neural network encoding and decoding.
In comparison, kernel PCA defines the error in kernel feature space $\calF_k$.
One might ask if using the reconstruction error in some neural feature space may also be useful for autoencoders, for instance to shift detection towards higher-level feature spaces.
Recent work that includes the reconstruction error over the hidden layers of an autoencoder \cite{kim2020} indeed suggests that this concept can improve detection performance.
Another question one might ask when comparing the reconstruction models in Table \ref{tab:taxonomy_methods} is if including the prototype assumption (cf., section \ref{sssec:prototype}) could also be useful in deep autoencoding and how this can be achieved practically.
The VQ-VAE model, which introduces a discrete codebook between the neural encoder and decoder, presents a way to incorporate this concept that has shown to result in reconstructions with improved quality and coherence in some settings \cite{van2017,razavi2019}.
Besides these existing proof-of-concepts for transferring ideas, which we have motivated here from our unifying view, we outline further potential combinations to explore in future research in section \ref{ssec:unexplored_modeling_combinations}.

\subsection{Distance-based Anomaly Detection}

Our unifying view focuses on anomaly detection methods that formulate some loss-based learning objective.
Apart from these methods, there also exists a rich literature on purely `distance-based' anomaly detection methods and algorithms that have been studied extensively in the data mining community in particular.
Many of these algorithms follow a \emph{lazy learning} paradigm, in which there is no a priori training phase of learning a model, but instead new test points are evaluated with respect to the training instances only as they occur.
We here group these methods as `distance-based' without further granularity, but remark that various taxonomies for these types of methods have been proposed \cite{chandola2009,aggarwal2017}.
Examples of such methods include nearest-neighbor-based methods \cite{knorr2000,ramaswamy2000,harmeling2006,zhao2009,gu2019} such as LOF \cite{breunig2000} and partitioning tree-based methods \cite{juszczak2009} such as Isolation Forest \cite{liu2008,Guha2016}.
These methods usually also aim to capture the high-density regions of the data in some manner, for instance by scaling distances in relation to local neighborhoods \cite{breunig2000}, and thus are mostly consistent with the formal anomaly detection problem definition presented in section \ref{sec:introduction_to_AD}.
The majority of these algorithms have been studied and applied in the original input space $\calX$.
Few of them have been considered in the context of deep learning, but some hybrid anomaly detection approaches exist, which apply distance-based algorithms on top of deep neural feature maps from pre-trained networks (e.g., \cite{bergman2020}).

\section{Evaluation and Explanation}
\label{sec:evaluation_and_xai}

The theoretical considerations and unifying view above provide useful insights about the characteristics and underlying modeling assumptions of the different anomaly detection methods. What matters the most to the practitioner, however, is to evaluate how well an anomaly detection method performs on real data. In this section, we first present different aspects of evaluation, in particular, the problem of {\em building} a dataset that includes meaningful anomalies, and the problem of robustly {\em evaluating} an anomaly detection model on the collected data. In a second step, we will look at the limitations of classical evaluation techniques, specifically, their inability to directly inspect and verify the exact strategy employed by some model for detection, for instance, which input variables a model uses for prediction. We then present `Explainable AI' approaches for enabling such deeper inspection of a model.

\subsection{Building Anomaly Detection Benchmarks}
\label{ssec:benchmarks}

Unlike standard supervised datasets, there is an intrinsic difficulty in building anomaly detection benchmarks: Anomalies are rare and some of them may have never been observed before they manifest themselves in practice. Existing anomaly benchmarks typically rely on one of the following strategies:
\begin{enumerate}
    \item {\em $k$-classes-out:} Start from a binary or multi-class dataset and declare one or more classes to be normal and the rest to be anomalous. Due to the semantic homogeneity of the resulting `anomalies,' such a benchmark may not be a good simulacrum of real anomalies. For example, simple low-level anomalies (e.g., additive noise) may not be tested for.
    \item {\em Synthetic:} Start from an existing supervised or unsupervised dataset and generate synthetic anomalies (e.g., \cite{DBLP:conf/sp/GlasserL13,mnist-c,hendrycks2019b}). Having full control over anomalies is desirable from a statistical view point, to get robust error estimates. However, the characteristics of real anomalies may be unknown or difficult to generate.
    \item {\em Real-world:} Consider a dataset that contains anomalies and have them labeled by a human expert. This is the ideal case. In addition to the anomaly label, the human can augment a sample with an annotation of which exact features are responsible for the anomaly (e.g., a segmentation mask in the context of image data).
\end{enumerate}
We provide examples of anomaly detection benchmarks and datasets falling into these three categories in Table \ref{tab:benchmarks}.

Although all three approaches are capable of producing anomalous data, we note that real anomalies may exhibit much wider and finer variations compared to those in the dataset. In \emph{adversarial cases}, anomalies may be designed maliciously to avoid detection (e.g., in fraud and cybersecurity scenarios \cite{laskov2004intrusion,stokes2008,kloft2012security,emmott2013,Emmott2016,lavin2015}). 

%%%%%%%%%%%%%%%%%%%%%%%%%%%%%%%%%%%%%%%%%%%%%%%%%%%%%%%%%%%%%%%%%%%%%%%%%%%%%%%%
\begin{table}[!t]
    \caption{Existing anomaly detection benchmarks.}
    \label{tab:benchmarks}
    \centering
    \begin{tabular}{rl}
    \toprule
    {\bf $k$-classes-out} & \parbox{.7\linewidth}{(Fashion-)MNIST, CIFAR-10, STL-10, ImageNet}\\\midrule
    {\bf Synthetic} & \parbox{.7\linewidth}{MNIST-C \cite{mnist-c}, ImageNet-C \cite{hendrycks2019b}, ImageNet-P \cite{hendrycks2019b}, ImageNet-O \cite{hendrycks2019e}} \\\midrule
    {\bf Real-world} & \parbox{.7\linewidth}{{\em Industrial:} MVTec-AD \cite{bergmann2019}, PCB \cite{huang2019pcb}\\
    {\em Medical:} CAMELYON16 \cite{bejnordi2017,tuluptceva2020}, NIH Chest X-ray \cite{Wang_2017,tuluptceva2020}, MOOD \cite{mood}, HCP/BRATS \cite{chen2018}, Neuropathology \cite{faust2018,naud2020}\\
    {\em Security:} Credit-card-fraud \cite{pozzolo2018}, URL \cite{ma2009}, UNSW-NB15 \cite{moustafa2015}\\
    {\em Time series:} NAB \cite{ahmad2017}, Yahoo \cite{laptev2015}\\
    {\em Misc.:} Emmott \cite{Emmott2016}, ELKI \cite{campos2016}, ODDS \cite{rayana2016}, UCI \cite{domingues2018,Dua:2019}}\\
    \bottomrule
\end{tabular}
\end{table}
%%%%%%%%%%%%%%%%%%%%%%%%%%%%%%%%%%%%%%%%%%%%%%%%%%%%%%%%%%%%%%%%%%%%%%%%%%%%%%%%

\subsection{Evaluating Anomaly Detectors}
\label{ssec:evaluation}

Most applications come with different costs for false alarms (type I error) and missed anomalies (type II error). Hence, it is common to consider the decision function
\begin{equation}
\begin{split}
    \text{decide} \; \left\{ \begin{array}{lll} \text{anomaly} & \text{if} & s(\bx) \geq \tau\\ \text{inlier} & \text{if} & s(\bx) < \tau, \end{array} \right.
\end{split}
\end{equation}
where $s$ denotes the anomaly score, and adjust the decision threshold $\tau$ in a way that (i) minimizes the costs associated to the type I and type II errors on the collected validation data, or (ii) accommodates the hard constraints of the environment in which the anomaly detection system will be deployed.

\medskip

To illustrate this, consider an example in financial fraud detection: anomaly alarms are typically sent to a fraud analyst who must decide whether to open an investigation into the potentially fraudulent activity.
There is typically a fixed number of analysts. Suppose they can only handle $k$ alarms per day, that is, the $k$ examples with the highest predicted anomaly score. In this scenario, the measure to optimize is the \emph{precision@$k$}, since we want to maximize the number of anomalies contained in those $k$ alarms.

In contrast, consider a credit card company that places an automatic hold on a credit card when an anomaly alarm is reported. False alarms result in angry customers and reduced revenue, so the goal is to maximize the number of true alarms subject to a constraint on the percentage of false alarms. 
The corresponding measure is to maximize \emph{recall@$k$}\,---\,where $k$ is the number of false alarms.

However, it is often the case that application-related costs and constraints are not fully specified or vary over time. With such restrictions, it is desirable to have a measure that evaluates the performance of anomaly detection models under a broad range of possible application scenarios, or analogously, a broad range of decision thresholds $\tau$.
The Area Under the ROC Curve (AUROC or simply AUC) provides an evaluation measure that considers the full range of decision thresholds on a given test set \cite{bradley1997,fawcett2006}. 
The ROC curve plots all the (false alarm rate, recall)-pairs that result from iterating over all thresholds which cover every possible test set decision split, and the area under this curve is the AUC measure. 
A convenient property of the AUC is that the random guessing baseline always achieves an AUC of 0.5, regardless of whether there is an imbalance between anomalies and normal instances in the test set. 
This makes AUC easy to interpret and comparable over different application scenarios, which is one of the reasons why AUC is the most commonly used performance measure in anomaly detection \cite{ding2014,campos2016}. 
One caveat of AUC is that it can produce overly optimistic scores in case of highly imbalanced test sets \cite{davis2006,ahmed2020}.  % Few anomalies and plenty of normal instances can drive a small false alarm rate, although precision might be bad.
In such cases, the Area Under the Precision-Recall Curve (AUPRC) is more informative and appropriate to use \cite{davis2006,ahmed2020}. 
The PR curve plots all the (precision, recall)-pairs that result from iterating over all possible test set decision thresholds. 
AUPRC therefore is preferable to AUROC when precision is more relevant than the false alarm rate. 
A common robust way to compute AUPRC is via Average Precision (AP) \cite{boyd2013}. 
One downside of AUPRC (or AP) is that the random guessing baseline is given by the fraction of anomalies in the test set and thus varies between applications. 
This makes AUPRC (or AP) generally harder to interpret and less comparable over different application scenarios. 
In scenarios where there is no clear preference for precision or the false alarm rate, we recommend to ideally report both threshold-independent measures for a comprehensive evaluation.

\subsection{Comparison on MNIST-C and MVTec-AD}
\label{ssec:comparison}

In the following, we apply the AUC measure to compare a selection of anomaly detection methods from the three major approaches (probabilistic, one-class, reconstruction) and three types of feature representation (raw input, kernel, and neural network). We perform the comparison on the synthetic MNIST-C and real-world MVTec-AD datasets. MNIST-C is MNIST extended with a set of fifteen types of corruptions (e.g., blurring, added stripes, impulse noise, etc). 
MVTec-AD consists of fifteen image sets from industrial production, where anomalies correspond to manufacturing defects. 
These images sometimes take the form of textures (e.g., wood, grid) or objects (e.g., toothbrush, screw). 
For MNIST-C, models are trained on the standard MNIST training set and then tested on each corruption separately. 
We measure the AUC separating the corrupted from the uncorrupted test set. 
For MVTec-AD, we train distinct models on each of the fifteen image sets and measure the AUC on the corresponding test set. 
Results for each model are shown in Tables \ref{tab:mnist-c_auc} and \ref{tab:mvtec_auc}. We provide the training details of each model in Appendix \ref{appendix:training_details}.

%%%%%%%%%%%%%%%%%%%%%%%%%%%%%%%%%%%%%%%%%%%%%%%%%%%%%%%%%%%%%%%%%%%%%%%%%%%%%%%%
\begin{table}[th]
    \caption{AUC detection performance on MNIST-C.}
    \label{tab:mnist-c_auc}
    \centering
    \resizebox{\columnwidth}{!}{\begin{tabular}{rccccccccc}\toprule
& Gaussian & MVE & PCA & KDE & SVDD & kPCA & AGAN & DOCC & AE\\\midrule
          brightness & \bf 100.0 &      99.0 & \bf 100.0 & \bf 100.0 &     100.0 & \bf 100.0 & \bf 100.0 &      13.7 & \bf 100.0\\
         canny edges &      99.4 &      68.4 & \bf 100.0 &      78.9 &      96.3 &      99.9 &     100.0 &      97.9 &     100.0\\
         dotted line &      99.9 &      62.9 &      99.3 &      68.5 &      70.0 &      92.6 &      91.5 &      86.4 & \bf 100.0\\
                 fog &     100.0 &      89.6 &      98.1 &      62.1 &      92.3 &      91.3 & \bf 100.0 &      17.4 &     100.0\\
          glass blur &      79.5 &      34.7 &      70.7 &       8.0 &      49.1 &      27.1 & \bf 100.0 &      31.1 &      99.6\\
       impulse noise & \bf 100.0 &      69.0 & \bf 100.0 &      98.0 &      99.7 & \bf 100.0 & \bf 100.0 &      97.5 & \bf 100.0\\
         motion blur &      38.1 &      43.4 &      24.3 &       8.1 &      50.2 &      18.3 & \bf 100.0 &      70.7 &      95.1\\
              rotate &      31.3 &      54.7 &      24.9 &      37.1 &      57.7 &      38.7 & \bf  93.2 &      65.5 &      53.4\\
               scale &       7.5 &      20.7 &      14.5 &       5.0 &      36.5 &      19.6 &      68.1 & \bf  79.8 &      40.4\\
               shear &      63.7 &      58.1 &      55.5 &      49.9 &      58.2 &      54.1 & \bf  94.9 &      64.6 &      70.6\\
          shot noise &      94.9 &      43.2 &      97.1 &      41.6 &      63.4 &      81.5 &      96.7 &      51.5 & \bf  99.7\\
             spatter & \bf  99.8 &      52.6 &      85.0 &      44.5 &      57.3 &      64.5 &      99.0 &      68.2 &      97.4\\
              stripe & \bf 100.0 &      99.9 & \bf 100.0 & \bf 100.0 &     100.0 & \bf 100.0 & \bf 100.0 &     100.0 & \bf 100.0\\
           translate &      94.5 &      73.9 &      96.3 &      76.2 &      91.8 &      94.8 &      97.3 & \bf  98.8 &      92.2\\
              zigzag &      99.9 &      72.5 &     100.0 &      84.0 &      87.7 &      99.4 &      98.3 &      94.3 & \bf 100.0\\
\bottomrule\end{tabular}
}
\end{table}
%%%%%%%%%%%%%%%%%%%%%%%%%%%%%%%%%%%%%%%%%%%%%%%%%%%%%%%%%%%%%%%%%%%%%%%%%%%%%%%%

%%%%%%%%%%%%%%%%%%%%%%%%%%%%%%%%%%%%%%%%%%%%%%%%%%%%%%%%%%%%%%%%%%%%%%%%%%%%%%%%
\begin{table}[ht]
    \caption{AUC detection performance on MVTec-AD.}
    \label{tab:mvtec_auc}
    \centering
    \resizebox{\columnwidth}{!}{\begin{tabular}{rrccccccccc}\toprule
& & Gaussian & MVE & PCA & KDE & SVDD & kPCA & AGAN & DOCC & AE\\\midrule
\multirow{5}{*}{\rotatebox[origin=c]{90}{Textures}}
&      carpet &      48.8 &      63.5 &      45.6 &      34.8 &      48.7 &      41.9 &      83.1 & \bf  90.6 &      36.8\\
&        grid &      60.6 &      67.8 &      81.8 &      71.7 &      80.4 &      76.7 & \bf  91.7 &      52.4 &      74.6\\
&     leather &      39.6 &      49.5 &      60.3 &      41.5 &      57.3 &      61.1 &      58.6 & \bf  78.3 &      64.0\\
&        tile &      68.5 &      79.7 &      56.4 &      68.9 &      73.3 &      63.2 &      74.1 & \bf  96.5 &      51.8\\
&        wood &      54.0 &      80.1 &      90.4 & \bf  94.7 &      94.1 &      90.6 &      74.5 &      91.6 &      88.5\\\midrule
\multirow{10}{*}{\rotatebox[origin=c]{90}{Objects}}
&      bottle &      78.9 &      67.0 &      97.4 &      83.3 &      89.3 &      96.3 &      90.6 & \bf  99.6 &      95.0\\
&       cable &      56.5 &      71.9 &      77.6 &      66.9 &      73.1 &      75.6 &      69.7 & \bf  90.9 &      57.3\\
&     capsule &      71.6 &      65.1 &      75.7 &      56.2 &      61.3 &      71.5 &      60.7 & \bf  91.0 &      52.5\\
&    hazelnut &      67.6 &      80.4 &      89.1 &      69.9 &      74.3 &      83.8 & \bf  96.4 &      95.0 &      90.5\\
&   metal nut &      54.7 &      45.1 &      56.4 &      33.3 &      54.3 &      59.0 &      79.3 & \bf  85.2 &      45.5\\
&        pill &      65.5 &      71.5 & \bf  82.5 &      69.1 &      76.2 &      80.7 &      64.6 &      80.4 &      76.0\\
&       screw &      53.5 &      35.5 &      67.9 &      36.9 &       8.6 &      46.7 & \bf  99.6 &      86.9 &      77.9\\
&  toothbrush &      93.9 &      76.1 & \bf  98.3 &      93.3 &      96.1 & \bf  98.3 &      70.8 &      96.4 &      49.4\\
&  transistor &      70.2 &      64.8 &      81.8 &      72.4 &      74.8 &      80.0 &      78.8 & \bf  90.8 &      51.2\\
&      zipper &      50.1 &      65.2 &      82.8 &      61.4 &      68.6 &      81.0 &      69.7 & \bf  92.4 &      35.0\\
\bottomrule\end{tabular}
}
\end{table}
%%%%%%%%%%%%%%%%%%%%%%%%%%%%%%%%%%%%%%%%%%%%%%%%%%%%%%%%%%%%%%%%%%%%%%%%%%%%%%%%

A first striking observation is the heterogeneity in performance of the various methods on the different corruptions and defect classes. 
For example, AGAN performs generally well on MNIST-C but is systematically outperformed by the Deep One-Class Classification (DOCC) model on MVTec-AD. 
Also, the more powerful nonlinear models are not better on every class, and simple `shallow' models occasionally outperform their deeper counterparts. 
For instance, the simple Gaussian model reaches top performance on MNIST-C:spatter, linear PCA ranks highest on \mbox{MVTec-AD:toothbrush}, and KDE ranks highest on MVTec-AD:wood. 
The fact that some of the simplest models sometimes perform well highlights the strong differences in modeling structure of each anomaly detection model. 
Since the MNIST-C and MVTec-AD test sets are not highly imbalanced, we see the same trends when using Average Precision (AP) as an evaluation measure as to be expected \cite{davis2006}.
We provide the detection performance results in AP in Appendix \ref{appendix:avg_precision}.

However, what is still unclear is whether the measured model performance faithfully reflects the performance on a broader set of anomalies (i.e., the generalization performance) or whether some methods only benefit from the specific (possibly non-representative) types of anomalies that have been collected in the test set.
In other words, assuming that all models achieve 100\% test accuracy (e.g., MNIST-C:stripe), can we conclude that all models will perform well on a broad range of anomalies? This problem has been already highlighted in the context of supervised learning, and explanation methods can be applied to uncover such potential hidden weaknesses of models, also known as `Clever Hanses' \cite{lapuschkin-natcom19}.

\subsection{Explaining Anomalies}
\label{ssec:xai}

In the following, we consider techniques that augment anomaly predictions with explanations. These techniques enable us to better understand the generalization properties and detection strategies used by different anomaly models, and in turn to also address some of the limitations of classical validation procedures. 
Producing explanations of model predictions is already common in supervised learning, and this field is often referred to as Explainable AI (or XAI) \cite{DBLP:series/lncs/11700}. Popular XAI methods include LIME \cite{DBLP:conf/kdd/Ribeiro0G16}, (Guided) Grad-CAM \cite{DBLP:conf/iccv/SelvarajuCDVPB17}, integrated gradients \cite{sundararajan2017,qi2019visualizing}, and Layer-wise Relevance Propagation (LRP) \cite{bach-plos15}. Grad-CAM and LRP rely on the structure of the network to produce robust explanations.

Explainable AI has recently also been brought to unsupervised learning and, in particular, to anomaly detection \cite{micenkova2013,Siddiqui2019,kauffmann2020,kauffmann2020a,liznerski2020,Sipple2020}. Unlike supervised learning, which is largely dominated by neural networks \cite{krizhevsky2012,lecun2015,litjens17}, state-of-the-art methods for unsupervised learning are much more heterogeneous, including neural networks but also kernel-based, centroid-based, or probability-based models. In such a heterogeneous setting, it is difficult to build explanation methods that allow for a consistent comparison of detection strategies of the multiple anomaly detection models. Two directions to achieve such consistent explanations are particularly promising:
\begin{enumerate}
    \item Model-agnostic explanation techniques (e.g., sampling-based) that apply transparently to any model, whether it is a neural network or something different (e.g., \cite{micenkova2013}).
    \item A conversion of non-neural network models into functionally equivalent neural networks, or \emph{neuralization}, so that existing approaches for explaining neural networks (e.g., LRP \cite{bach-plos15}) can be applied \cite{kauffmann2020,kauffmann2020a}.
\end{enumerate}

\begin{figure}[ht]
    \centering
    \includegraphics[width=\linewidth]{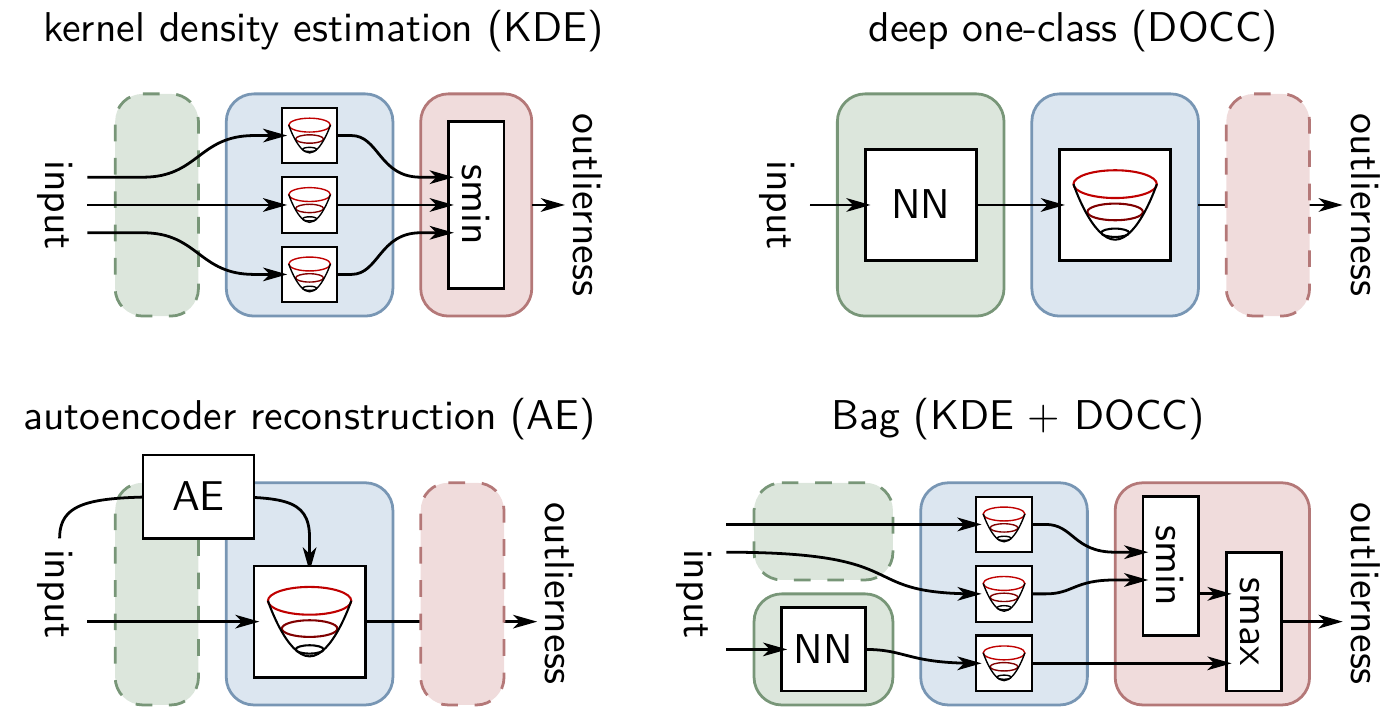}
    \caption{\setlength{\fboxsep}{1pt} An illustration of the \emph{neuralization} concept that reformulates models as strictly equivalent neural networks. 
    Here, kernel density estimation (KDE), deep one-class classification (DOCC), and autoencoder (AE) are expressed as a three-layer architecture \cite{kauffmann2020a}: (i) \colorbox{neural_green}{feature extraction} $\rightarrow$ (ii) \colorbox{neural_blue}{distance computation} $\rightarrow$ (iii) \colorbox{neural_red}{pooling}. The `neuralized' formulation enables to apply LRP \cite{bach-plos15} for explaining anomalies.
    A bag of models (here KDE and DOCC) can also be expressed in this way.}
    \label{fig:neuralization}
\end{figure}

In the following, we demonstrate a neuralization approach. It has been shown that numerous anomaly detection models, in particular kernel-based models such as KDE or one-class SVMs, can be rewritten as strictly equivalent neural networks \cite{kauffmann2020,kauffmann2020a}. The neuralized equivalents of a model may not be unique, and explanations obtained with LRP consequently depend on the chosen network structure \cite{DBLP:journals/corr/abs-1906-07633}. Here, we aim to find a single structure that fits many models. We show examples of neuralized models in Fig.~\ref{fig:neuralization}. They typically organize into a three-layer architecture, from left to right: feature extraction, distance computation, and pooling.

%%%%%%%%%%%%%%%%%%%%%%%%%%%%%%%%%%%%%%%%%%%%%%%%%%%%%%%%%%%%%%%%%%%%%%%%%%%%%%%%
%%%%%%%%%%%%%%%%%%%%%%%%%%%%%%%%%%%%%%%%%%%%%%%%%%%%%%%%%%%%%%%%%%%%%%%%%%%%%%%%

For example, the KDE model, usually expressed as
$
\textstyle f(\bx) = \frac1n \sum_{i=1}^n \exp(-\gamma\,\|\bx-\bx_i\|^2),
$
can have its negative log-likelihood $s(\bx) = -\log f(\bx)$ rewritten as a two-layer network:
\begin{align*}
h_j &= \gamma\,\|\bx-\bx_j\|^2 + \log n & \text{(layer 1)}\\
s(\bx) &= \text{smin}_j \{ h_j \} & \text{(layer 2)}
\end{align*}
where \textit{smin} is a soft min-pooling of the type log-sum-exp (see \cite{kauffmann2020} for further details).

Once the model has been converted into a neural network, we can apply explanation techniques such as LRP \cite{bach-plos15} to produce an explanation of the anomaly prediction. In this case, the LRP algorithm will take the score at the output of the model, propagate to `winners' in the pool, then assign the score to directions in the input or feature space that contribute the most to the distance, and if necessary propagate the signal further down the feature hierarchy (see the Supplement of \cite{kauffmann2020a} for how this is done exactly).

\begin{figure}[ht]
    \centering
    \includegraphics[width=\linewidth]{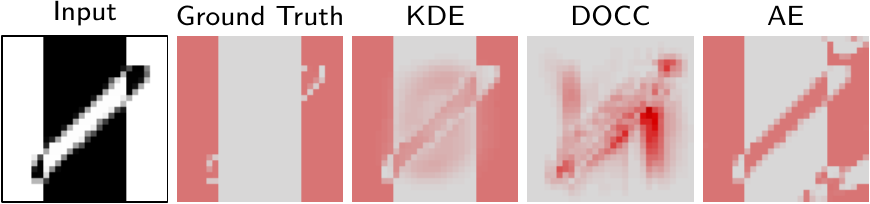}
    \caption{An example of anomaly explanations.
    The input is an anomalous digit 1 from MNIST-C:stripe that has been corrupted by inverting the pixels in the left and right vertical stripes. 
    The ground truth explanation highlights the anomalous pixels in red. 
    The kernel density estimator (KDE), deep one-class classification (DOCC), and autoencoder (AE) detect the stripe anomalies accurately, but the LRP explanations show that the strategies are very different:
    KDE highlights the anomaly, but also some regions of the digit itself. 
    DOCC strongly emphasizes vertical edges. 
    The AE produces a result similar to KDE but with decision artifacts in the corners of the image and on the digit itself.}
    \label{fig:anomaly_localization}
\end{figure}

Fig.~\ref{fig:anomaly_localization} shows from left to right an anomaly from the MNIST-C dataset, the ground-truth explanation (the squared difference between the digit before and after corruption) as well as LRP explanations for three anomaly detection models (KDE, DOCC, and AE).

%%%%%%%%%%%%%%%%%%%%%%%%%%%%%%%%%%%%%%%%%%%%%%%%%%%%%%%%%%%%%%%%%%%%%%%%%%%%%%%%
%%%%%%%%%%%%%%%%%%%%%%%%%%%%%%%%%%%%%%%%%%%%%%%%%%%%%%%%%%%%%%%%%%%%%%%%%%%%%%%%

From these observations, it is clear that each model, although predicting with 100\% accuracy on the current data, will have different generalization properties and vulnerabilities when encountering subsequent anomalies. We will work through an example in section \ref{ssec:example_mvtec} to show how explanations can help to diagnose and improve a detection model.

\medskip

To conclude, we emphasize that a standard quantitative evaluation can be imprecise or even misleading when the available data is not fully representative, and in that case, explanations can be produced to more comprehensively assess the quality of an anomaly detection model.

\section{Worked-Through Examples}
\label{sec:guidelines_and_examples}

In this section, we work through two specific, real-world examples to exemplify the modeling and evaluation process and provide some best practices.

\subsection{Example 1: Thyroid Disease Detection}
\label{ssec:example_uci}

In the first example our goal is to learn a model to detect thyroid gland dysfunctions such as hyperthyroidism.
The Thyroid dataset\footnote{Available from the ODDS Library \cite{rayana2016} at \href{http://odds.cs.stonybrook.edu/}{http://odds.cs.stonybrook.edu/}} includes $n$ = 3772 data instances and has $D$ = 6 real-valued features.
It contains a total of 93 ($\sim$2.5\%) anomalies.
For a quantitative evaluation, we consider a dataset split of 60:10:30 corresponding to the training, validation, and test sets respectively, while preserving the ratio of $\sim$2.5\% anomalies in each of the sets.

We choose the OC-SVM \cite{scholkopf2001} with standard RBF kernel $k(\bx, \tilde{\bx}) = \exp(-\gamma \Vert \bx - \tilde{\bx} \Vert^2 )$ as a method for this task since the data is real-valued, low-dimensional, and the OC-SVM scales sufficiently well for this comparatively small dataset.
In addition, the $\nu$-parameter formulation (see Eq.\ \eqref{eqn:ocsvm}) enables us to use our prior knowledge and thus approximately control the false alarm rate $\alpha$ and with it implicitly also the miss rate, which leads to our first recommendation:

\begin{center}
\fbox{\textbf{Assess the risks of false alarms and missed anomalies}}
\end{center}

Calibrating the false alarm rate and miss rate of a detection model can make the difference between life or death in a medical context such as disease detection.
Though the consequences must not always be as dramatic as in a medical setting, it is important to carefully consider the risks and costs involved with type I and type II errors in advance.
In our example, a false alarm would suggest a thyroid dysfunction although the patient is healthy. 
On the other hand, a missed alarm would occur if the model recognizes a patient with a dysfunction as healthy.
Such asymmetric risks, with a greater expected loss for anomalies that go undetected, are very common in medical diagnosis \cite{huynh1998,petticrew2000,pepe2003,zhou2011}.
Given only $D$ = 6 measurements per data record, we therefore seek to learn a detector with a miss rate ideally close to zero, at the cost of an increased false alarm rate. 
Patients falsely ascribed with a dysfunction by such a detector could then undergo further, more elaborate clinical testing to verify the disease.
Assuming our data is representative and $\sim$12\%\footnote{\href{https://www.thyroid.org/}{https://www.thyroid.org/}} of the population is at risk of thyroid dysfunction, we choose a slightly higher $\nu = 0.15$ to further increase the robustness against potential data contamination (here the training set contains $\sim$2.5\% contamination in the form of unlabeled anomalies).
We then train the model and choose the kernel scale $\gamma$ according to the best AUC we observe on the small, labeled validation set which includes 9 labeled anomalies. We select $\gamma$ from $\gamma \in \{ (2^i D)^{-1} \, | \, i = -5, \ldots, 5 \}$, that is, from a $\log_2$ span that accounts for the dimensionality $D$.

Following the above, we observe a rather poor best validation set AUC of 83.9\% at $\gamma = (2^{-5} D)^{-1}$, which is the largest value from the hyperparameter range.
This is an indication that we forgot an important preprocessing step, namely:

\begin{center}
\fbox{\textbf{Apply feature scaling to normalize value ranges}}
\end{center}

Any method that relies on computing distances, including kernel methods, requires the features to be scaled to similar ranges to prevent features with wider value ranges from dominating the distances. If this is not done, it can cause anomalies that deviate on smaller scale features to be undetected. Similar reasoning also holds for clustering and classification (e.g., see discussions in \cite{dudaHart1973} or \cite{theodoridis2009}).
Min-max normalization or standardization are common choices, but since we assume there might be some contamination, we apply a robust feature scaling via the median and interquartile range. Remember that scaling parameters should be computed using only information from the training data and then applied to all of the data.
After we have scaled the features, we observe a much improved best validation set AUC of 98.6\% at $\gamma = (2^{2} D)^{-1}$.
The so-trained and selected model finally achieves a test set AUC of 99.2\%, a false alarm rate of 14.8\% (i.e., close to our a priori specified $\nu=0.15$), and a miss rate of zero.

\subsection{Example 2: MVTec Industrial Inspection}
\label{ssec:example_mvtec}

In our second example, we consider the task of detecting anomalies in wood images from the MVTec-AD dataset. Unlike the first worked-through example, the MVTec data is high-dimensional and corresponds to arrays of pixel values. Hence, all input features are already on a similar scale (between $-1$ and $+1$) and thus we do not need to apply feature rescaling.

Following the standard model training\,/\,validation procedure, we train a set of models on the training data, select their hyperparameters on hold out data (e.g., a few inliers and anomalies extracted from the test set), and then evaluate their performance on the remainder of the test set. Table \ref{tab:wood-auc} shows the AUC performance of the nine models in our benchmark.

\begin{table}[ht]
    \caption{AUC detection performance on the MVTec-AD `wood' class.}
    \label{tab:wood-auc}
    \centering
\resizebox{\columnwidth}{!}{\begin{tabular}{ccccccccc}\toprule
Gaussian & MVE & PCA & KDE & SVDD & kPCA & AGAN & DOCC & AE\\\midrule
 54.0 &      80.1 &      90.4 & \bf  94.7 &      94.1 &      90.6 &      74.5 &      91.6 &      88.5\\\bottomrule
    \end{tabular}}
\end{table}

We observe that the best performing model is the kernel density estimation (KDE). This is particularly surprising, because this model does not compute the kinds of higher-level image features that deep models, such as DOCC, learn and apply. An examination of the dataset shows that the anomalies involve properties such as small perforations and stains that do not require high-level semantic information to be detected. But is that the only reason why the performance of KDE is so high? In order to get insight into the strategy used by KDE to arrive at its prediction, we employ the neuralization/LRP approach presented in section \ref{ssec:xai}.

\begin{center}
\fbox{\textbf{Apply XAI to analyze model predictions}}
\end{center}

\begin{figure}[ht]
    \centering
    \includegraphics[page=1]{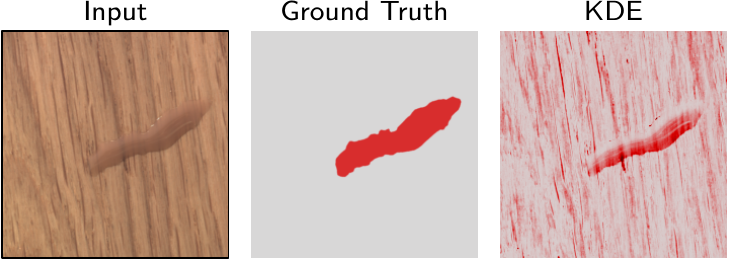}
    \caption{The input image, ground-truth source of anomaly (here, a stain of liquid), and the explanation of the KDE anomaly prediction. The KDE model assigns high relevance to the wood grain instead of the liquid stain. This discrepancy between the ground truth and model explanation reveals a `Clever Hans' strategy used by the KDE model.}
    \label{fig:cleverhans}
\end{figure}

Fig.~\ref{fig:cleverhans} shows an example of an image along with its ground-truth pixel-level anomaly as well as the computed pixel-wise explanation for KDE.

%%%%%%%%%%%%%%%%%%%%%%%%%%%%%%%%%%%%%%%%%%%%%%%%%%%%%%%%%%%%%%%%%%%%%%%%%%%%%%%%
%%%%%%%%%%%%%%%%%%%%%%%%%%%%%%%%%%%%%%%%%%%%%%%%%%%%%%%%%%%%%%%%%%%%%%%%%%%%%%%%

Ideally, we would like the model to make its decision based on the actual anomaly (here, the liquid stain), and therefore, we would expect the ground-truth annotation and the KDE explanation to coincide. 
However, it is clear from inspection of the explanation that KDE is {\em not} looking at the true cause of the anomaly and is looking instead at the vertical stripes present everywhere in the input image. This discrepancy between the explanation and the ground truth can be observed on other images of the `wood' class. 
The high AUC score of KDE thus must be due to a spurious correlation in the test set between the reaction of the model to these stripes and the presence of anomalies.
We call this a `Clever Hans' effect \cite{lapuschkin-natcom19}, because just like the horse Hans, who could correctly answer arithmetic problems by reading unintended reactions of his master\footnote{\url{https://en.wikipedia.org/wiki/Clever_Hans}}, the model appears to work because of a spurious correlation.
Obviously the KDE model is unlikely to generalize well when the anomalies and the stripes become decoupled (e.g., as we observe more data or under some adversarial manipulation).
This illustrates the importance of generating explanations to identify these kinds of failures. Once we have identified the problem, how can we change our anomaly detection strategy so that it is more robust and generalizes better?

\begin{center}
\fbox{\textbf{Improve the model based on explanations}}
\end{center}

In practice, there are various approaches to improve the model based on explanation feedback:

\begin{enumerate}
    \item {\em Data extension:} We can extend the data with missing training cases, for instance anomalous wood examples that lack stripes or normal wood examples that have stripes to break to spurious correlation between stripes and anomalies. When further data collection is not possible, synthetic data extension schemes such as blurring or sharpening can also be considered.
    \item {\em Model extension:} If the first approach is not sufficient, or if the model is simply not capable of implementing the necessary prediction structure, the model itself can be changed (e.g., using a more flexible deep model). In other cases, the model may have enough representation power but is statistically inefficient (e.g., subject to the curse of dimensionality). In that case, adding structure (e.g., convolutions) or regularization can also help to learn a model with an appropriate prediction strategy.
    \item {\em Ensembles:} If all considered models have their own strengths and weaknesses, ensemble approaches can be considered. Ensembles have a conceptual justification in the context of anomaly detection \cite{kauffmann2020a}, and they have been shown to work well empirically \cite{DBLP:conf/kdd/LazarevicK05,DBLP:conf/dasfaa/VuAG10}.
\end{enumerate}

Once the model has been improved using these strategies, explanations can be recomputed and examined to verify that the decision strategy has been corrected. If that is not the case, the process can be iterated until we reach a satisfactory model.

In our wood example, we have observed that KDE reacts strongly to the vertical strains. Based on this observation, we replace the Gaussian kernel with a Mahalanobis kernel that effectively applies a horizontal Gaussian blur to the images before computing the distance. This has the effect of reducing the strain patterns, but keeping the anomalies intact. This increases the explanation accuracy of the model from an average cosine similarity of 0.34 to 0.38 on the ground truth explanations. Fig.~\ref{fig:cleverhans_fixed} shows the explanation of the improved model. Implementation details can be found in Appendix \ref{appendix:xai}. The AUC drops to 87\%, which corresponds to a more realistic estimate of the generalization abilities of the KDE model, which previously was biased by the spurious correlation.

%%%%%%%%%%%%%%%%%%%%%%%%%%%%%%%%%%%%%%%%%%%%%%%%%%%%%%%%%%%%%%%%%%%%%%%%%%%%%%%%
\begin{figure}[ht]
    \centering
    \includegraphics[page=2]{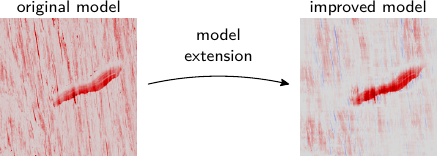}
    \caption{Explanations of the original (left) and improved (right) KDE model. The Gaussian kernel strongly reacts to the vertical wood stripes (left). After replacing it with a Mahalanobis kernel (right) that is less sensitive to high horizontal frequencies, the model focuses on the true source of anomaly considerably better.}
    \label{fig:cleverhans_fixed}
\end{figure}
%%%%%%%%%%%%%%%%%%%%%%%%%%%%%%%%%%%%%%%%%%%%%%%%%%%%%%%%%%%%%%%%%%%%%%%%%%%%%%%%

\section{Conclusion and Outlook}
\label{sec:future_research}
Anomaly detection is a blossoming field of broad theoretical and practical interest across the disciplines. 
In this work, we have given a review of the past and present state of anomaly detection research, established a systematic unifying view, and discussed many practical aspects. 
While we have included some of our own contributions, we hope that we have fulfilled our aim of providing a balanced and comprehensive snapshot of this exciting research field.
Focus was given to a solid theoretical basis, which then allowed us put today's two main lines of development into perspective: the more classical kernel world and the more recent world of deep learning and representation learning for anomaly detection.

We will conclude our review by turning to what lies ahead.
Below, we highlight some critical open challenges\,---\,of which there are many\,---\,and identify a number of potential avenues for future research that we hope will provide useful guidance.

\subsection{Unexplored Combinations of Modeling Dimensions}
\label{ssec:unexplored_modeling_combinations}
As can be seen in Fig.~\ref{fig:taxonomy_intro} and Table \ref{tab:taxonomy_methods}, there is a zoo of different anomaly detection algorithms that have historically been explored along various dimensions. 
This review has shown conceptual similarities between anomaly detection members from kernel methods and deep learning. 
Note, however, that the exploration of novel algorithms has been substantially different in both domains, which offers unique possibilities to explore new methodology: steps that have been pursued in kernel learning but not in deep anomaly detection could be transferred (or vice versa) and powerful new directions could emerge. 
In other words, ideas could be readily transferred from kernels to deep learning and back, and novel combinations in our unifying view would emerge. 

Let us now discuss some specific opportunities to clarify this point. Consider the problem of robustness to noise and contamination or signal to noise ratio. For shallow methods, the problem is well studied, and we have many effective methods \cite{kwak2008,JMLR:v9:braun08a,huber2009,candes2011,kim2012,xiao2013,liu2014}.
In deep anomaly detection, very little work has addressed this problem.
A second example is the application of Bayesian methods. Bayesian inference has been mostly considered for shallow methods \cite{bishop1999,ghasemi2012}, owing to the prohibitive cost or intractability of exact Bayesian inference in deep neural networks. 
Recent progress in approximate Bayesian inference and Bayesian neural networks \cite{blundell2015,gal2016,lakshminarayanan2017,kendall2017,ovadia2019} raise the possibility of developing methods that complement anomaly scores with uncertainty estimates or uncertainty estimates of their respective explanations \cite{bykov2020much}.
In the area of semi-supervised anomaly detection, ideas have already been successfully transferred from kernel learning \cite{tax2001,gornitz2013} to deep methods \cite{ruff2020} for one-class classification.
But probabilistic and reconstruction methods that can make use of labeled anomalies are less explored.
For time-series anomaly detection \cite{fox1972,tsay1988,tsay2000,gupta2014,lavin2015,samek2017robust}, where forecasting (i.e., conditional density estimation) models are practical and widely deployed, semi-supervised extensions of such methods could lead to significant improvements in applications in which some labeled examples are available (e.g., learning from failure cases in monitoring tasks).
Concepts from density ratio estimation \cite{hido2011}, noise contrastive estimation \cite{gutmann2010}, or coding theory \cite{vandermeulen2020} could lead to novel semi-supervised methods in principled ways.
Finally, active learning strategies for anomaly detection \cite{pelleg2005,abe2006,stokes2008,gornitz2009}, which identify informative instances for labeling, have primarily only been explored for shallow detectors and could be extended to deep learning approaches.

This is a partial list of opportunities that we have noticed. Further analysis of our framework will likely expose additional directions for innovation.

\subsection{Bridging Related Lines of Research on Robustness}
\label{ssec:related_lines}

Other recent lines of research on robust deep learning are closely related to anomaly detection or may even be interpreted as special instances of the problem.
These include out-of-distribution detection, model calibration, uncertainty estimation, and adversarial examples or attacks.
Bridging these lines of research by working out the nuances of the specific problem formulations can be insightful for connecting concepts and transferring ideas to jointly advance research.

A basic approach to creating robust classifiers is to endow them with the ability to reject input objects that are likely to be misclassified. This is known as the problem of \textit{classification with a reject option}, and it has been studied extensively \cite{chow1957,chow1970,bartlett2008,tax2008,grandvalet2009,cortes2016,geifman2017}. However, this work focuses on objects that fall near the decision boundary where the classifier is uncertain. 

One approach to making the rejection decision is to calibrate the classification probabilities and then reject objects for which no class is predicted to have high probability following Chow's optimal rejection rule \cite{chow1970}. Consequently, many researchers have developed techniques for calibrating the probabilities of classifiers \cite{platt1999,guo2017,lakshminarayanan2017,devries2018,lee2018,ni2019,meinke2020} or for Bayesian uncertainty quantification \cite{mackay1992,mackay1998,blundell2015,gal2016,kendall2017,ovadia2019}.

Recent work has begun to address other reasons for rejecting an input object. \emph{Out-of-distribution (OOD) detection} considers cases where the object is drawn from a distribution different from the training distribution $\Pnorm$ \cite{hendrycks2017,liang2018,lee2018,lee2018b,choi2020,meinke2020}. From a formal standpoint, it is impossible to determine whether an input $x$ is drawn from one of two distributions $\bbP_1$ and $\bbP_2$ if both distributions have support at $x$. Consequently, the OOD problem reduces to determining whether $x$ lies outside regions of high density in $\Pnorm$, which is exactly the anomaly detection problem we have described in this review. 

A second reason to reject an input object is because it belongs to a class that was not part of the training data. This is the problem of \emph{open set recognition}. Such objects can also be regarded as being generated by a distribution $\Pgen^-$, so this problem also fits within our framework and can be addressed with the algorithms described here. Nonetheless, researchers have developed a separate set of methods for open set recognition \cite{scheirer2012,scheirer2014,bendale2016,shu2017,liu2018a,zhang2020}, and an important goal for future research is to evaluate these methods from the anomaly detection perspective and to evaluate anomaly detection algorithms from the open set perspective.  

In rejection, out-of-distribution, and open set recognition problems, there is an additional source of information that is not available in standard anomaly detection problems: the class labels of the objects. Hence, the learning task combines classification with anomaly detection. Formally, the goal is to train a classifier on labeled data $(\bx_1, y_1), \ldots, (\bx_n, y_n)$ with class labels $y \in \{1, \ldots, k\}$ while also developing some measure to decide whether an unlabeled test point $\tilde{\bx}$ should be rejected (for any of the reasons listed above). The class label information tells us about the structure of $\Pnorm$ and allows us to model it as a joint distribution $\Pnorm \equiv \Pgen_{X,Y}$. Methods for rejection, out-of-distribution, and open set recognition all take advantage of this additional structure. Note that the labels $y$ are different from the labels that mark normal or anomalous points in supervised or semi-supervised anomaly detection (cf., section \ref{ssec:data_settings}).

Research on the unresolved and fundamental issue of adversarial examples and attacks \cite{szegedy2014,goodfellow2015,carlini2017,tramer2017,biggio2018,athalye2018,madry2018,carlini2019,zhang2019b,ilyas2019} is related to anomaly detection as well.
We may interpret adversarial attacks as extremely hard-to-detect out-of-distribution samples \cite{lakshminarayanan2017}, as they are specifically crafted to target the decision boundary and confidence of a learned classifier.
Standard adversarial attacks find a small perturbation $\delta$ for an input $\bx$ so that $\tilde{\bx} = \bx + \delta$ yields some class prediction desired by the attacker. For instance, a perturbed image of a dog may be indistinguishable from the original to the human's eye, yet the predicted label changes from `dog' to `cat'.
Note that such an adversarial example $\tilde{\bx}$ still likely is (and probably should) be normal under the data marginal $\Pgen_X$ (an imperceptibly perturbed image of a dog shows a dog after all!) but the pair $(\tilde{\bx}, \text{`cat'})$ should be anomalous under the joint $\Pgen_{X,Y}$ \cite{che2019}.
Methods for OOD detection have been found to also increase adversarial robustness \cite{lakshminarayanan2017,shalev2018,hendrycks2019c,hendrycks2019d,choi2020}, some of which model the class conditional distributions for detection \cite{lee2018b,che2019}, for the reason just described.

The above highlights the connection of these lines of research towards the general goal of robust deep models.
Thus, we believe that connecting ideas and concepts in these lines (e.g., the use of spherical models in both anomaly detection \cite{ruff2018,bergman2020b} and OOD \cite{shalev2018,dhamija2018}) may help them to advance together.
Finally, the assessment of the robustness of neural networks and their fail-safe design and integration are topics of high practical relevance that have recently found their way in international standardization initiatives (cf., section \ref{ssec:AD_motivation}). Beyond doubt, understanding the brittleness of deep networks (also in context of their explanations \cite{dombrowski2019explanations}) will be critical for their adoption in anomaly detection applications that involve malicious attackers such as fraudsters or network intruders.

\subsection{Interpretability and Trustworthiness}

Much of anomaly detection research has been devoted to developing new methods that improve detection accuracy.
In most applications, however, accuracy alone is not sufficient \cite{caruana2015,kauffmann2020a} and further criteria such as interpretability \cite{montavon2018methods,samek2020} and trustworthiness \cite{lee2004,jiang2018,ovadia2019} are equally critical as demonstrated in sections \ref{sec:evaluation_and_xai} and \ref{sec:guidelines_and_examples}.
For researchers and practitioners alike \cite{lipton2017} it is vital to understand the underlying reasons for how a specific anomaly detection model reaches a particular prediction.
Interpretable, explanatory feedback enhances model transparency, which is indispensable for accountable decision-making \cite{goodman2017}, uncovering model failures such as Clever Hans behavior \cite{lapuschkin-natcom19,kauffmann2020a}, and understanding model vulnerabilities that can be insightful for improving a model or system. This is especially relevant in safety-critical environments \cite{amodei2016,richter2017}.
Existing work on interpretable anomaly detection has considered finding subspaces of anomaly-discriminative features \cite{micenkova2013,dang2013,dang2014,duan2015,vinh2016,macha2018}, deducing sequential feature explanations \cite{Siddiqui2019}, using feature-wise reconstruction errors \cite{schlegl2019,bergmann2019}, employing fully convolutional architectures \cite{liznerski2020}, and explaining anomalies via integrated gradients \cite{Sipple2020} or LRP \cite{kauffmann2020,kauffmann2020a}.
In relation to the vast body of literature though, research on interpretability and trustworthiness in anomaly detection has seen comparatively little attention.
The fact that anomalies may not share similar patterns (i.e., their heterogeneity) poses a challenge for their explanation, which also distinguishes this setting from interpreting supervised classification models.
Furthermore, anomalies might arise due to the presence of abnormal patterns, but conversely also due to a lack of normal patterns.
While for the former case an explanation that highlights the abnormal features is satisfactory, how should an explanation for missing features be conceptualized? 
For example given the MNIST dataset of digits, what should an explanation of an anomalous all-black image be?
The matters of interpretability and trustworthiness get more pressing as the task and data become more complex.
Effective solutions of complex tasks will necessarily require more powerful methods, for which explanations become generally harder to interpret.
We thus believe that future research in this direction will be imperative.

\subsection{The Need for Challenging and Open Datasets}

Challenging problems with clearly defined evaluation criteria on publicly available benchmark datasets are invaluable for measuring progress and moving a field forward.
The significance of the ImageNet database \cite{deng2009}, together with corresponding competitions and challenges \cite{russakovsky2015}, for progressing computer vision and supervised deep learning in the last decade give a prime example of this.
Currently, the standard evaluation practices in deep anomaly detection \cite{ruff2018,akcay2018,golan2018,hendrycks2019a,abati2019,perera2019a,wang2019b,hendrycks2019d,wang2019c,ruff2020,bergman2020b,kim2020}, out-of-distribution detection \cite{hendrycks2017,liang2018,lee2018,lee2018b,nalisnick2019,ren2019,choi2020,serra2020}, and open set recognition \cite{scheirer2012,scheirer2014,bendale2016,shu2017,liu2018a} still extensively repurpose classification datasets by deeming some dataset classes to be anomalous or considering in-distribution vs.\ out-of-distribution dataset combinations (e.g., training a model on Fashion-MNIST clothing items and regarding MNIST digits to be anomalous).
Although these synthetic protocols have some value, it has been questioned how well they reflect real progress on challenging anomaly detection tasks \cite{ruff2020b,ahmed2020}.
Moreover, we think the tendency that only few methods seem to dominate most of the benchmark datasets in the work cited above is alarming, since it suggests a bias towards evaluating only the upsides of newly proposed methods, yet often critically leaving out an analysis of their downsides and limitations.
This situation suggests a lack of diversity in the current evaluation practices and the benchmarks being used.
In the spirit of \emph{all models are wrong} \cite{box1976}, we stress that more research effort should go into studying when and how certain models are wrong and behave like Clever Hanses. We need to understand the trade-offs that different methods make. For example, some methods are likely making a trade-off between detecting low-level vs.\ high-level semantic anomalies (cf., section \ref{sssec:anomaly_types} and \cite{ahmed2020}). 
The availability of more diverse and challenging datasets would be of great benefit in this regard.
Recent datasets such as MVTec-AD \cite{bergmann2019} and competitions such as the Medical Out-of-Distribution Analysis Challenge \cite{mood} provide excellent examples, but the field needs many more challenging open datasets to foster progress.

\subsection{Weak Supervision and Self-Supervised Learning}
\label{ssec:weak_supervision}

The bulk of anomaly detection research has been studying the problem in absence of any kind of supervision, that is, in an unsupervised setting (cf., section \ref{sssec:unsupervised}).
Recent work suggests, however, that significant performance improvements on complex detection tasks seem achievable through various forms of weak supervision and self-supervised learning.

\emph{Weak supervision} or \emph{weakly supervised learning} describes learning from imperfectly or scarcely labeled data \cite{ratner2017,zhou2018,roh2019}.
Labels might be inaccurate (e.g., due to labeling errors or uncertainty) or incomplete (e.g., covering only few normal modes or specific anomalies).
Current work on semi-supervised anomaly detection indicates that including even only few labeled anomalies can already yield remarkable performance improvements on complex data \cite{pang2019,daniel2019,ruff2020,tuluptceva2020,ruff2020b,liznerski2020}.
A key challenge here is to formulate and optimize such methods so that they generalize well to novel anomalies.
Combining these semi-supervised methods with active learning techniques helps identifying informative candidates for labeling \cite{pelleg2005,abe2006,stokes2008,gornitz2009}. It is an effective strategy for designing anomaly detection systems that continuously improve via expert feedback loops \cite{Siddiqui2019,das2020}. This approach has not yet been explored for deep detectors, though.
Outlier exposure \cite{hendrycks2019a}, that is, using massive amounts of data that is publicly available in some domains (e.g., stock photos for computer vision or the English Wikipedia for NLP) as auxiliary negative samples (cf., section \ref{ssec:negative_samples}), can also be viewed as a form of weak supervision (imperfectly labeled anomalies).
Though such negative samples may not coincide with ground-truth anomalies, we believe such contrasting can be beneficial for learning characteristic representations of normal concepts in many domains (e.g., using auxiliary log data to well characterize the normal logs of a specific computer system \cite{nedelkoski2020}).
So far, this has been little explored in applications.
Transfer learning approaches to anomaly detection also follow the idea of distilling more domain knowledge into a model, for example, through using and possibly fine-tuning pre-trained (supervised) models \cite{oza2019,perera2019c,ouardini2019,bergman2020,kauffmann2020a}.
Overall, weak forms of supervision or domain priors may be essential for achieving effective solutions in semantic anomaly detection tasks that involve high-dimensional data, as has also been found in other unsupervised learning tasks such as disentanglement \cite{locatello2019,shu2020,locatello2020}.
Hence, we think that developing effective methods for weakly supervised anomaly detection will contribute to advancing the state of the art.

\emph{Self-supervised learning} describes the learning of representations through solving auxiliary tasks, for example, next sentence and masked words prediction \cite{devlin2019}, future frame prediction in videos \cite{mathieu2016}, or the prediction of transformations applied to images \cite{chen2020} such as colorization \cite{zhang2016}, cropping \cite{doersch2015,noroozi2016}, or rotation \cite{gidaris2018}.
These auxiliary prediction tasks do not require (ground-truth) labels for learning and can thus be applied to unlabeled data, which makes self-supervised learning particularly appealing for anomaly detection.
Self-supervised methods that have been introduced for visual anomaly detection train multi-class classification models based on pseudo labels that correspond to various geometric transformations (e.g., flips, translations, rotations, etc.) \cite{golan2018,hendrycks2019d,wang2019c}.
An anomaly score can then be derived from the softmax activation statistics of a so-trained classifier, assuming that a high prediction uncertainty (close to a uniform distribution) indicates anomalies.
These methods have shown significant performance improvements on the common $k$-classes-out image benchmarks (see Table \ref{tab:benchmarks}).
Bergman and Hoshen \cite{bergman2020b} have recently proposed a generalization of this idea to non-image data, called GOAD, which is based on random affine transformations.
We can identify GOAD and self-supervised methods based on geometric transformations (GT) as classification-based approaches within our unifying view (see Table \ref{tab:taxonomy_methods}).
Other recent and promising self-supervised approaches are based on contrastive learning \cite{chen2020,winkens2020,tack2020}.
In a broader context, the interesting question will be to what extent self-supervision can facilitate the learning of semantic representations.
There is some evidence that self-supervised learning helps to improve the detection of semantic anomalies and thus exhibits inductive biases towards semantic representations \cite{ahmed2020}.
On the other hand, there also exists evidence showing that self-supervision mainly improves learning of effective feature representations for low-level statistics \cite{asano2020}.
Hence, this research question remains to be answered, but bears great potential for many domains where large amounts of unlabeled data are available.

\subsection{Foundation and Theory}

The recent progress in anomaly detection research has also raised more fundamental questions.
These include open questions about the out-of-distribution generalization properties of various methods presented in this review, the definition of anomalies in high-dimensional spaces, and information-theoretic interpretations of the problem.

Nalisnick et al.\ \cite{nalisnick2019} have recently observed that deep generative models (DGMs) (cf., section \ref{sec:probabilistic}) such as normalizing flows, VAEs, or autoregressive models can often assign higher likelihood to anomalies than to in-distribution samples.
For example, models trained on Fashion-MNIST clothing items can systematically assign higher likelihood to MNIST digits \cite{nalisnick2019}.
This counter-intuitive finding, which has been replicated in subsequent work \cite{choi2018,hendrycks2019a,ren2019,nalisnick2019b,grathwohl20,serra2020}, revealed that there is a critical lack of theoretical understanding of these models.
Solidifying evidence \cite{ren2019,serra2020,schirrmeister2020,kirichenko2020} indicates that one reason seems to be that the likelihood in current DGMs is still largely biased towards low-level background statistics. Consequently, simpler data points attain higher likelihood (e.g., MNIST digits under models trained on Fashion-MNIST, but not vice versa).
Another critical remark in this context is that for (truly) high-dimensional data, the region with highest density must not necessarily coincide with the region of highest probability mass (called the \emph{typical set}), that is, the region where data points most likely occur \cite{nalisnick2019b}.
For instance, while the highest density of a $D$-dimensional standard Gaussian distribution is given at the origin, points sampled from the distribution concentrate around an annulus with radius $\sqrt{D}$ for large $D$ \cite{vershynin2018}.
Therefore, points close to the origin have high density, but are unlikely to occur.
This mismatch questions the standard theoretical density (level set) problem formulation (cf., section \ref{ssec:problem_definition}) and use of likelihood-based anomaly detectors for some settings.
Hence, theoretical research aimed at understanding the above phenomenon and DGMs themselves presents an exciting research opportunity.

Similar observations suggest that reconstruction-based models can systematically well reconstruct simpler out-of-distribution points that sit within the convex hull of the data. For example, an anomalous all-black image can be well reconstructed by an autoencoder trained on MNIST digits \cite{tong2020}.
An even simpler example is the perfect reconstruction of points that lie within the linear subspace spanned by the principal components of a PCA model, even in regions far away from the normal training data (e.g., along the principal component in Fig.~\ref{fig:2D_toy_reconstruction}).
While such out-of-distribution generalization properties might be desirable for general representation learning \cite{krueger2020}, such behavior critically can be undesirable for anomaly detection.
Therefore, we stress that more theoretical research on understanding such out-of-distribution generalization properties or biases, especially for more complex models, will be necessary.

Finally, the push towards deep learning also presents new opportunities to interpret and analyze the anomaly detection problem from different theoretical angles. Autoencoders, for example, can be understood from an information theory perspective \cite{shannon1948} as adhering to the \emph{Infomax principle} \cite{linsker1988,bell1995,hjelm2019} by implicitly maximizing the mutual information between the input and latent code\,---\,subject to structural constraints or regularization of the code (e.g., `bottleneck', latent prior, sparsity, etc.)\,---\,via the reconstruction objective \cite{vincent2008}.
Similarly, information-theoretic perspectives of VAEs have been formulated showing that these models can be viewed as making a rate-distortion trade-off \cite{berger2003} when balancing the latent compression (negative rate) and reconstruction accuracy (distortion) \cite{higgins2017,alemi2018}.
This view has recently been used to draw a connection between VAEs and Deep SVDD, where the latter can be seen as a special case that only seeks to minimize the rate (maximize compression) \cite{park2020}.
Overall, anomaly detection has been studied comparatively less from an information-theoretic perspective \cite{lee2001,host2019}, yet we think this could be fertile ground for building a better theoretical understanding of representation learning for anomaly detection.

\smallskip
Concluding, we firmly believe that anomaly detection in all its exciting variants will also in the future remain an indispensable practical tool in the quest to obtain robust learning models that perform well on complex data.

\appendices
\section{Notation and Abbreviations}

For reference, we provide the notation and abbreviations used in this work in Tables \ref{tab:notation} and \ref{tab:abbrev} respectively.

%%%%%%%%%%%%%%%%%%%%%%%%%%%%%%%%%%%%%%%%%%%%%%%%%%%%%%%%%%%%%%%%%%%%%%%%%%%%%%%%
\begin{table}[ht]
    \caption{Notation Conventions}
    \label{tab:notation}
    \centering
    \begin{tabular}{ll}
    \toprule
    Symbol                  & Description \\
    \midrule
    $\bbN$                  & The natural numbers \\
    $\bbR$                  & The real numbers \\
    $D$                     & The input data dimensionality $D \in \bbN$ \\
    $\calX$                 & The input data space $\calX \subseteq \bbR^D$ \\
    $\calY$                 & The labels $\calY = \{\pm 1\}$ (${+1}$ : normal; ${-1}$ : anomaly) \\
    $\bx$                   & A vector, e.g.\ a data point $\bx \in \calX$ \\
    $\calD_n$               & An unlabeled dataset $\calD_n = \{ \bx_1, \ldots, \bx_n \}$ of size $n$ \\
    $\Pgen, \pgen$          & The data-generating distribution and pdf \\
    $\Pnorm, \pnorm$        & The normal data distribution and pdf \\
    $\Pout, \pout$          & The anomaly distribution and pdf \\
    $\phat$                 & An estimated pdf \\
    $\varepsilon$           & An error or noise distribution\\
    $\supp(p)$              & \makecell[tl]{The support of a data distribution $\mathbb{P}$ with density $p$,\\ i.e.\ $\{ \bx \in \mathcal{X} \, | \, p(\bx) > 0 \}$} \\
    $\calA$                 & The set of anomalies \\
    $C_\alpha$              & An $\alpha$-density level set \\
    $\hat{C}_\alpha$        & An $\alpha$-density level set estimator \\
    $\tau_\alpha$           & The threshold $\tau_\alpha \geq 0$ corresponding to $C_\alpha$ \\
    $c_\alpha(\bx)$         & The threshold anomaly detector corresponding to $C_\alpha$ \\
    $s(\bx)$                & An anomaly score function $s: \calX \to \bbR$ \\
    $\bbInd_{A}(\bx)$       & The indicator function for some set $A$ \\
    $\ell(s,y)$             & A loss function $\ell : \bbR \times \{\pm 1 \} \to \bbR$ \\
    $f_\theta(\bx)$         & A model $f_\theta : \calX \to \bbR$ with parameters $\theta$ \\
    $k(\bx, \tilde{\bx})$   & A kernel $k: \calX \times \calX \to \bbR$ \\
    $\calF_k$               & The RKHS or feature space of kernel $k$ \\
    $\phi_k(\bx)$           & The feature map $\phi_k : \calX \to \calF_k$ of kernel $k$ \\
    $\phi_\omega(\bx)$      & A neural network $\bx \mapsto \phi_\omega(\bx)$ with weights $\omega$ \\
    \bottomrule
\end{tabular}

\end{table}
%%%%%%%%%%%%%%%%%%%%%%%%%%%%%%%%%%%%%%%%%%%%%%%%%%%%%%%%%%%%%%%%%%%%%%%%%%%%%%%%

%%%%%%%%%%%%%%%%%%%%%%%%%%%%%%%%%%%%%%%%%%%%%%%%%%%%%%%%%%%%%%%%%%%%%%%%%%%%%%%%
\begin{table}[ht]
    \caption{List of Abbreviations}
    \label{tab:abbrev}
    \centering
    \begin{tabular}{ll}
    \toprule
    Abbreviation        & Description \\
    \midrule
    AD      & Anomaly Detection \\
    AE		& Autoencoder \\
    AP		& Average Precision \\
    AAE		& Adversarial Autoencoder \\
    AUPRC   & Area Under the Precision-Recall Curve \\
    AUROC	& Area Under the ROC curve \\
    CAE		& Contrastive Autoencoder \\
    DAE		& Denoising Autoencoder \\
    DGM		& Deep Generative Model \\
    DSVDD	& Deep Support Vector Data Description \\
    DSAD	& Deep Semi-supervised Anomaly Detection \\
    EBM		& Energy Based Model \\
    ELBO	& Evidence Lower Bound \\
    GAN		& Generative Adversarial Network \\
    GMM		& Gaussian Mixture Model \\
    GT		& Geometric Transformations \\
    iForest & Isolation Forest \\
    KDE		& Kernel Density Estimation \\
    $k$-NN	& $k$-Nearest Neighbors \\
    kPCA	& Kernel Principal Component Analysis \\
    LOF		& Local Outlier Factor \\
    LPUE	& Learning from Positive and Unlabeled Examples \\
    LSTM	& Long short-term memory \\
    MCMC	& Markov chain Monte Carlo \\
    MCD		& Minimum Covariance Determinant \\
    MVE		& Minimum Volume Ellipsoid \\
    OOD		& Out-of-distribution \\
    OE		& Outlier Exposure \\
    OC-NN	& One-Class Neural Network \\
    OC-SVM	& One-Class Support Vector Machine \\
    pPCA	& Probabilistic Principal Component Analysis \\
    PCA		& Principal Component Analysis \\
    pdf		& Probability density function \\
    PSD		& Positive semidefinite \\
    RBF		& Radial basis function \\
    RKHS	& Reproducing Kernel Hilbert Space \\
    rPCA	& Robust Principal Component Analysis \\
    SGD		& Stochastic Gradient Descent \\
    SGLD	& Stochastic Gradient Langevin Dynamics \\
    SSAD	& Semi-Supervised Anomaly Detection \\
    SVDD	& Support Vector Data Description \\
    VAE		& Variational Autoencoder \\
    VQ		& Vector Quantization \\
    XAI		& Explainable AI \\
    \bottomrule
\end{tabular}

\end{table}
%%%%%%%%%%%%%%%%%%%%%%%%%%%%%%%%%%%%%%%%%%%%%%%%%%%%%%%%%%%%%%%%%%%%%%%%%%%%%%%%

\section{Additional Details on Experimental Evaluation}
\label{appendix:exp_details}

\subsection{Average Precision on MNIST-C and MVTec-AD}
\label{appendix:avg_precision}

We provide the detection performance measured in Average Precision (AP) of the experimental evaluation on MNIST-C and MVTec-AD from section \ref{ssec:comparison} in Tables \ref{tab:mnist-c_ap} and \ref{tab:mvtec_ap} respectively.
As can be seen (and as to be expected \cite{davis2006}), the performance in AP here shows the same trends as AUC (cf., Tables \ref{tab:mnist-c_auc} and \ref{tab:mvtec_auc} in section \ref{ssec:comparison}), since the MNIST-C and MVTec-AD test sets are not highly imbalanced.

%%%%%%%%%%%%%%%%%%%%%%%%%%%%%%%%%%%%%%%%%%%%%%%%%%%%%%%%%%%%%%%%%%%%%%%%%%%%%%%%
\begin{table}[th]
    \caption{AP detection performance on MNIST-C.}
    \label{tab:mnist-c_ap}
    \centering
    \resizebox{\columnwidth}{!}{\begin{tabular}{rccccccccc}\toprule
& Gaussian & MVE & PCA & KDE & SVDD & kPCA & AGAN & DOCC & AE\\\midrule
          brightness & \bf 100.0 &      98.0 & \bf 100.0 & \bf 100.0 &     100.0 & \bf 100.0 & \bf 100.0 &      32.9 & \bf 100.0\\
         canny edges &      99.1 &      58.8 & \bf 100.0 &      71.8 &      96.6 &      99.9 &     100.0 &      97.7 &     100.0\\
         dotted line &      99.9 &      56.8 &      99.0 &      63.4 &      67.9 &      90.9 &      88.8 &      81.5 & \bf  99.9\\
                 fog &     100.0 &      88.3 &      98.7 &      75.5 &      94.2 &      94.2 & \bf 100.0 &      34.8 &     100.0\\
          glass blur &      78.6 &      42.0 &      65.5 &      31.5 &      45.9 &      36.2 & \bf 100.0 &      37.6 &      99.6\\
       impulse noise & \bf 100.0 &      59.8 & \bf 100.0 &      97.1 &      99.6 & \bf 100.0 & \bf 100.0 &      96.2 & \bf 100.0\\
         motion blur &      52.6 &      44.3 &      37.3 &      31.5 &      47.1 &      33.9 & \bf 100.0 &      66.5 &      93.8\\
              rotate &      44.1 &      52.2 &      38.3 &      42.3 &      56.3 &      43.5 & \bf  93.6 &      66.0 &      53.1\\
               scale &      31.9 &      34.5 &      33.0 &      31.2 &      39.4 &      34.4 &      61.9 & \bf  70.2 &      42.5\\
               shear &      72.7 &      62.0 &      64.2 &      52.5 &      59.0 &      60.0 & \bf  95.5 &      66.5 &      70.4\\
          shot noise &      93.6 &      44.8 &      97.3 &      42.7 &      60.4 &      81.7 &      96.8 &      49.0 & \bf  99.7\\
             spatter & \bf  99.8 &      50.5 &      82.6 &      45.8 &      54.8 &      61.2 &      99.2 &      63.2 &      97.1\\
              stripe & \bf 100.0 &      99.9 & \bf 100.0 & \bf 100.0 &     100.0 & \bf 100.0 & \bf 100.0 &     100.0 & \bf 100.0\\
           translate &      95.5 &      64.8 &      97.0 &      73.7 &      92.2 &      95.7 &      97.2 & \bf  98.6 &      93.7\\
              zigzag &      99.8 &      64.6 &     100.0 &      79.4 &      86.5 &      99.3 &      98.0 &      94.8 & \bf 100.0\\
\bottomrule\end{tabular}
}
\end{table}
%%%%%%%%%%%%%%%%%%%%%%%%%%%%%%%%%%%%%%%%%%%%%%%%%%%%%%%%%%%%%%%%%%%%%%%%%%%%%%%%

%%%%%%%%%%%%%%%%%%%%%%%%%%%%%%%%%%%%%%%%%%%%%%%%%%%%%%%%%%%%%%%%%%%%%%%%%%%%%%%%
\begin{table}[ht]
    \caption{AP detection performance on MVTec-AD.}
    \label{tab:mvtec_ap}
    \centering
    \resizebox{\columnwidth}{!}{\begin{tabular}{rrccccccccc}\toprule
& & Gaussian & MVE & PCA & KDE & SVDD & kPCA & AGAN & DOCC & AE\\\midrule
\multirow{5}{*}{\rotatebox[origin=c]{90}{Textures}}
&      carpet &      77.3 &      86.9 &      71.0 &      70.2 &      77.4 &      69.8 &      94.3 & \bf  97.2 &      70.9\\
&        grid &      79.9 &      80.8 &      91.7 &      85.5 &      89.2 &      88.7 & \bf  97.4 &      75.4 &      84.8\\
&     leather &      72.9 &      81.1 &      85.8 &      75.3 &      83.6 &      86.3 &      82.1 & \bf  92.3 &      87.7\\
&        tile &      84.4 &      91.6 &      80.5 &      85.1 &      86.9 &      83.9 &      88.8 & \bf  98.6 &      78.1\\
&        wood &      82.0 &      93.8 &      97.0 & \bf  98.5 &      98.3 &      97.1 &      92.0 &      97.6 &      96.8\\\midrule
\multirow{10}{*}{\rotatebox[origin=c]{90}{Objects}}
&      bottle &      92.3 &      86.2 &      99.2 &      94.2 &      96.7 &      98.9 &      97.2 & \bf  99.9 &      98.5\\
&       cable &      73.2 &      76.6 &      85.9 &      78.5 &      82.9 &      84.2 &      81.2 & \bf  94.1 &      71.3\\
&     capsule &      92.3 &      89.3 &      93.0 &      85.9 &      88.7 &      92.0 &      84.3 & \bf  97.9 &      82.8\\
&    hazelnut &      81.9 &      89.3 &      94.2 &      83.2 &      85.7 &      90.9 & \bf  98.1 &      97.5 &      95.0\\
&   metal nut &      86.3 &      82.6 &      86.5 &      75.0 &      86.0 &      87.4 &      92.7 & \bf  96.3 &      77.0\\
&        pill &      91.8 &      93.8 & \bf  96.5 &      91.7 &      95.0 &      96.1 &      90.6 &      95.6 &      94.5\\
&       screw &      78.0 &      71.4 &      86.6 &      69.1 &      55.4 &      77.0 & \bf  99.8 &      95.1 &      90.3\\
&  toothbrush &      97.6 &      87.6 & \bf  99.4 &      97.4 &      98.5 & \bf  99.4 &      86.9 &      98.7 &      73.9\\
&  transistor &      70.5 &      54.7 &      80.7 &      70.1 &      74.1 &      79.7 &      71.2 & \bf  90.0 &      51.4\\
&      zipper &      81.0 &      84.2 &      91.8 &      82.8 &      87.9 &      91.5 &      85.7 & \bf  97.8 &      79.3\\
\bottomrule\end{tabular}
}
\end{table}
%%%%%%%%%%%%%%%%%%%%%%%%%%%%%%%%%%%%%%%%%%%%%%%%%%%%%%%%%%%%%%%%%%%%%%%%%%%%%%%%

\subsection{Training Details}
\label{appendix:training_details}

For PCA, we compute the reconstruction error whilst maintaining 90\% of variance of the training data. 
We do the same for kPCA, and additionally choose the kernel width such that 50\% neighbors capture 50\% of total similarity scores.
For MVE, we use the fast minimum covariance determinant estimator \cite{rousseeuw1999} with a default support fraction of 0.9 and a contamination rate parameter of 0.01.
To facilitate MVE computation on MVTec-AD, we first reduce the dimensionality via PCA retaining 90\% of variance.
For KDE, we choose the bandwidth parameter to maximize the likelihood of a small hold-out set from the training data.
For SVDD, we consider $\nu \in \{0.01, 0.05, 0.1, 0.2\}$ and select the kernel scale using a small labeled hold-out set.
The deep one-class classifier applies a whitening transform on the representations after the first fully-connected layer of a pre-trained VGG16 model (on MVTec-AD) or a CNN classifier trained on the EMNIST letter subset (on MNIST-C).
For the AE on MNIST-C, we use a LeNet-type encoder that has two convolutional layers with max-pooling followed by two fully connected layers that map to an encoding of 64 dimensions, and construct the decoder symmetrically.
On MVTec-AD, we use an encoder-decoder architecture as presented in \cite{huang2019} which maps to a bottleneck of 512 dimensions. Both, the encoder and decoder here consist of four blocks having two 3${\times}$3 convolutional layers followed by max-pooling or upsampling respectively. 
We train the AE such that the reconstruction error of a small training hold-out set is minimized.
For AGAN, we use the AE encoder and decoder architecture for the discriminator and generator networks respectively, where we train the GAN until convergence to a stable equilibrium.

\subsection{Explaining KDE}
\label{appendix:xai}
The model can be neuralized as described in Section \ref{ssec:xai}, replacing the squared Euclidean distance in the first layer with a squared Mahalanobis distance. The heatmaps of both models (KDE and Mahalanobis KDE) are computed as
\[
R = \tfrac12\sum_{j=1}^n (\bx_j - \bx) \odot \nabla_{\bx_j}s(\bx),
\]
where $\odot$ denotes element-wise multiplication.
This implements a Taylor-type decomposition as described in \cite{kauffmann2020}.

\subsection{Open Source Software, Tutorials, and Demos}
For the implementation of the shallow MVE and SVDD models, we have used the \texttt{scikit-learn} library \cite{scikit-learn} available at \url{https://scikit-learn.org/}.
For the implementation of the shallow Gauss, PCA, KDE, and kPCA models as well as the deep AGAN, DOCC, RealNVP, and AE models, we have used the \texttt{PyTorch} library \cite{pytorch} available at \url{https://pytorch.org/}.
Implementations of the Deep SVDD and Deep SAD methods are available at \url{https://github.com/lukasruff/}.
Tutorials, demos, and code for explainable AI techniques, in particular LRP, can be found at \url{http://www.heatmapping.org/}.
In the spirit of \emph{the need for open source software in machine learning} \cite{sonnenburg2007}, a similar collection of tutorials, demos, and code on anomaly detection methods are in the making and will be made available at \url{http://www.pyano.org/}.

\section*{Acknowledgments}
We kindly thank the reviewers and the editor for their constructive feedback which helped to improve this work.
LR acknowledges support by the German Federal Ministry of Education and Research (BMBF) in the project ALICE III (01IS18049B).
RV, KRM, GM, and WS acknowledge support by the Berlin Institute for the Foundations of Learning and Data (BIFOLD) sponsored by the BMBF.
MK acknowledges support by the German Research Foundation (DFG) award KL 2698/2-1 and by the BMBF awards 01IS18051A and 031B0770E.
KRM, GM, and WS acknowledge financial support by the BMBF for the Berlin Center for Machine Learning (01IS18037A-I) and under the Grants 01IS14013A-E, 031L0207A-D. KRM also acknowledges financial support by BMBF Grants 01GQ1115 and 01GQ0850; DFG under Grant Math+, EXC 2046/1, Project ID 390685689 and partial support by the Institute of Information \& Communications Technology Planning \& Evaluation (IITP) grants funded by the Korea Government (No. 2017-0-00451, Development of BCI based Brain and Cognitive Computing Technology for Recognizing User’s Intentions using Deep Learning) and funded by the Korea Government (No. 2019-0-00079,  Artificial Intelligence Graduate School Program, Korea University).
TD acknowledges financial support from the US Defense Advanced Research Projects Agency (DARPA) under Contract Nos. HR001119C0112, FA8750-19-C-0092, and HR001120C0022.
Correspondence to MK, TD and KRM.

\bibliographystyle{IEEEtran}
\bibliography{IEEEabrv,references}

% Generated by IEEEtran.bst, version: 1.14 (2015/08/26)
\begin{thebibliography}{100}
\providecommand{\url}[1]{#1}
\csname url@samestyle\endcsname
\providecommand{\newblock}{\relax}
\providecommand{\bibinfo}[2]{#2}
\providecommand{\BIBentrySTDinterwordspacing}{\spaceskip=0pt\relax}
\providecommand{\BIBentryALTinterwordstretchfactor}{4}
\providecommand{\BIBentryALTinterwordspacing}{\spaceskip=\fontdimen2\font plus
\BIBentryALTinterwordstretchfactor\fontdimen3\font minus
  \fontdimen4\font\relax}
\providecommand{\BIBforeignlanguage}[2]{{%
\expandafter\ifx\csname l@#1\endcsname\relax
\typeout{** WARNING: IEEEtran.bst: No hyphenation pattern has been}%
\typeout{** loaded for the language `#1'. Using the pattern for}%
\typeout{** the default language instead.}%
\else
\language=\csname l@#1\endcsname
\fi
#2}}
\providecommand{\BIBdecl}{\relax}
\BIBdecl

\bibitem{pearson1901}
K.~Pearson, ``On lines and planes of closest fit to systems of points in
  space,'' \emph{The London, Edinburgh, and Dublin Philosophical Magazine and
  Journal of Science}, vol.~2, no.~11, pp. 559--572, 1901.

\bibitem{hotelling1933}
H.~Hotelling, ``Analysis of a complex of statistical variables into principal
  components.'' \emph{Journal of Educational Psychology}, vol.~24, no.~6, pp.
  417--441, 1933.

\bibitem{scholkopf1998}
B.~Sch{\"o}lkopf, A.~Smola, and K.-R. M{\"u}ller, ``Nonlinear component
  analysis as a kernel eigenvalue problem,'' \emph{Neural Computation},
  vol.~10, no.~5, pp. 1299--1319, 1998.

\bibitem{hoffmann2007}
H.~Hoffmann, ``Kernel {PCA} for novelty detection,'' \emph{Pattern
  Recognition}, vol.~40, no.~3, pp. 863--874, 2007.

\bibitem{huber2009}
P.~J. Huber and E.~M. Ronchetti, \emph{Robust Statistics}, 2nd~ed.\hskip 1em
  plus 0.5em minus 0.4em\relax John Wiley \& Sons, 2009.

\bibitem{scholkopf2001}
B.~Sch{\"o}lkopf, J.~C. Platt, J.~Shawe-Taylor, A.~J. Smola, and R.~C.
  Williamson, ``Estimating the support of a high-dimensional distribution,''
  \emph{Neural Computation}, vol.~13, no.~7, pp. 1443--1471, 2001.

\bibitem{tax2004}
D.~M.~J. Tax and R.~P.~W. Duin, ``Support vector data description,''
  \emph{Machine Learning}, vol.~54, no.~1, pp. 45--66, 2004.

\bibitem{knorr2000}
E.~M. Knorr, R.~T. Ng, and V.~Tucakov, ``Distance-based outliers: algorithms
  and applications,'' \emph{The VLDB Journal}, vol.~8, no.~3, pp. 237--253,
  2000.

\bibitem{ramaswamy2000}
S.~Ramaswamy, R.~Rastogi, and K.~Shim, ``Efficient algorithms for mining
  outliers from large data sets,'' in \emph{Proceedings of the {ACM} {SIGMOD}
  International Conference on Management of Data}, 2000, pp. 427--438.

\bibitem{breunig2000}
M.~M. Breunig, H.-P. Kriegel, R.~T. Ng, and J.~Sander, ``{LOF}: {I}dentifying
  density-based local outliers,'' in \emph{Proceedings of the {ACM} {SIGMOD}
  International Conference on Management of Data}, 2000, pp. 93--104.

\bibitem{rosenblatt1956}
M.~Rosenblatt, ``Remarks on some nonparametric estimates of a density
  function,'' \emph{The Annals of Mathematical Statistics}, vol.~27, no.~3, pp.
  832--837, 1956.

\bibitem{parzen1962}
E.~Parzen, ``On estimation of a probability density function and mode,''
  \emph{The Annals of Mathematical Statistics}, vol.~33, no.~3, pp. 1065--1076,
  1962.

\bibitem{edgeworth1887}
F.~Y. Edgeworth, ``On discordant observations,'' \emph{The London, Edinburgh,
  and Dublin Philosophical Magazine and Journal of Science}, vol.~23, no.~5,
  pp. 364--375, 1887.

\bibitem{kuhn-1970}
T.~S. Kuhn, \emph{The Structure of Scientific Revolutions}.\hskip 1em plus
  0.5em minus 0.4em\relax Chicago: University of Chicago Press, 1970.

\bibitem{patcha2007}
A.~Patcha and J.-M. Park, ``An overview of anomaly detection techniques:
  Existing solutions and latest technological trends,'' \emph{Computer
  Networks}, vol.~51, no.~12, pp. 3448--3470, 2007.

\bibitem{liao2013}
H.-J. Liao, C.-H.~R. Lin, Y.-C. Lin, and K.-Y. Tung, ``Intrusion detection
  system: A comprehensive review,'' \emph{Journal of Network and Computer
  Applications}, vol.~36, no.~1, pp. 16--24, 2013.

\bibitem{ahmed2016}
M.~Ahmed, A.~N. Mahmood, and J.~Hu, ``A survey of network anomaly detection
  techniques,'' \emph{Journal of Network and Computer Applications}, vol.~60,
  pp. 19--31, 2016.

\bibitem{kwon2017}
D.~Kwon, H.~Kim, J.~Kim, S.~C. Suh, I.~Kim, and K.~J. Kim, ``A survey of deep
  learning-based network anomaly detection,'' \emph{Cluster Computing},
  vol.~10, pp. 1--13, 2017.

\bibitem{xin2018}
Y.~Xin, L.~Kong, Z.~Liu, Y.~Chen, Y.~Li, H.~Zhu, M.~Gao, H.~Hou, and C.~Wang,
  ``Machine learning and deep learning methods for cybersecurity,'' \emph{IEEE
  Access}, vol.~6, pp. 35\,365--35\,381, 2018.

\bibitem{malaiya2018}
R.~K. Malaiya, D.~Kwon, J.~Kim, S.~C. Suh, H.~Kim, and I.~Kim, ``An empirical
  evaluation of deep learning for network anomaly detection,'' in
  \emph{International Conference on Computing, Networking and Communications},
  2018, pp. 893--898.

\bibitem{bolton2002}
R.~J. Bolton and D.~J. Hand, ``Statistical fraud detection: A review,''
  \emph{Statistical Science}, vol.~17, no.~3, pp. 235--255, 2002.

\bibitem{bhattacharyya2011}
S.~Bhattacharyya, S.~Jha, K.~Tharakunnel, and J.~C. Westland, ``Data mining for
  credit card fraud: A comparative study,'' \emph{Decision Support Systems},
  vol.~50, no.~3, pp. 602--613, 2011.

\bibitem{joudaki2015}
H.~Joudaki, A.~Rashidian, B.~Minaei-Bidgoli, M.~Mahmoodi, B.~Geraili,
  M.~Nasiri, and M.~Arab, ``Using data mining to detect health care fraud and
  abuse: A review of literature,'' \emph{Global Journal of Health Science},
  vol.~7, no.~1, pp. 194--202, 2015.

\bibitem{ahmed2016b}
M.~Ahmed, A.~N. Mahmood, and M.~R. Islam, ``A survey of anomaly detection
  techniques in financial domain,'' \emph{Future Generation Computer Systems},
  vol.~55, pp. 278--288, 2016.

\bibitem{abdallah2016}
A.~Abdallah, M.~A. Maarof, and A.~Zainal, ``Fraud detection system: A survey,''
  \emph{Journal of Network and Computer Applications}, vol.~68, pp. 90--113,
  2016.

\bibitem{vanCapelleveen2016}
G.~van Capelleveen, M.~Poel, R.~M. Mueller, D.~Thornton, and J.~van
  Hillegersberg, ``Outlier detection in healthcare fraud: A case study in the
  medicaid dental domain,'' \emph{International Journal of Accounting
  Information Aystems}, vol.~21, pp. 18--31, 2016.

\bibitem{zheng2018}
Y.-J. Zheng, X.-H. Zhou, W.-G. Sheng, Y.~Xue, and S.-Y. Chen, ``Generative
  adversarial network based telecom fraud detection at the receiving bank,''
  \emph{Neural Networks}, vol. 102, pp. 78--86, 2018.

\bibitem{rabatel2011}
J.~Rabatel, S.~Bringay, and P.~Poncelet, ``Anomaly detection in monitoring
  sensor data for preventive maintenance,'' \emph{Expert Systems with
  Applications}, vol.~38, no.~6, pp. 7003--7015, 2011.

\bibitem{marzat2012}
J.~Marzat, H.~Piet-Lahanier, F.~Damongeot, and E.~Walter, ``Model-based fault
  diagnosis for aerospace systems: a survey,'' \emph{Proceedings of the
  Institution of Mechanical Engineers, Part G: Journal of Aerospace
  Engineering}, vol. 226, no.~10, pp. 1329--1360, 2012.

\bibitem{marti2015}
L.~Mart{\'\i}, N.~Sanchez-Pi, J.~M. Molina, and A.~C.~B. Garcia, ``Anomaly
  detection based on sensor data in petroleum industry applications,''
  \emph{Sensors}, vol.~15, no.~2, pp. 2774--2797, 2015.

\bibitem{yan2015}
W.~Yan and L.~Yu, ``On accurate and reliable anomaly detection for gas turbine
  combustors: A deep learning approach,'' in \emph{Annual Conference of the
  Prognostics and Health Management Society}, vol.~6, 2015.

\bibitem{lopez2017}
F.~Lopez, M.~Saez, Y.~Shao, E.~C. Balta, J.~Moyne, Z.~M. Mao, K.~Barton, and
  D.~Tilbury, ``Categorization of anomalies in smart manufacturing systems to
  support the selection of detection mechanisms,'' \emph{Robotics and
  Automation Letters}, vol.~2, no.~4, pp. 1885--1892, 2017.

\bibitem{hundman2018}
K.~Hundman, V.~Constantinou, C.~Laporte, I.~Colwell, and T.~Soderstrom,
  ``Detecting spacecraft anomalies using {LSTM}s and nonparametric dynamic
  thresholding,'' in \emph{International Conference on Knowledge Discovery \&
  Data Mining}, 2018, pp. 387--395.

\bibitem{atha2018}
D.~J. Atha and M.~R. Jahanshahi, ``Evaluation of deep learning approaches based
  on convolutional neural networks for corrosion detection,'' \emph{Structural
  Health Monitoring}, vol.~17, no.~5, pp. 1110--1128, 2018.

\bibitem{ramotsoela2018}
D.~Ramotsoela, A.~Abu-Mahfouz, and G.~Hancke, ``A survey of anomaly detection
  in industrial wireless sensor networks with critical water system
  infrastructure as a case study,'' \emph{Sensors}, vol.~18, no.~8, p. 2491,
  2018.

\bibitem{zhao2019}
R.~Zhao, R.~Yan, Z.~Chen, K.~Mao, P.~Wang, and R.~X. Gao, ``Deep learning and
  its applications to machine health monitoring,'' \emph{Mechanical Systems and
  Signal Processing}, vol. 115, pp. 213--237, 2019.

\bibitem{borghesi2019}
A.~Borghesi, A.~Bartolini, M.~Lombardi, M.~Milano, and L.~Benini, ``Anomaly
  detection using autoencoders in high performance computing systems,'' in
  \emph{{AAAI} Conference on Artificial Intelligence}, vol.~33, 2019, pp.
  9428--9433.

\bibitem{Sipple2020}
J.~Sipple, ``Interpretable , multidimensional , multimodal anomaly detection
  with negative sampling for detection of device failure,'' in
  \emph{International Conference on Machine Learning}, 2020, pp. 4368--4377.

\bibitem{golmohammadi2015}
K.~Golmohammadi and O.~R. Zaiane, ``Time series contextual anomaly detection
  for detecting market manipulation in stock market,'' in \emph{{IEEE}
  International Conference on Data Science and Advanced Analytics}, 2015, pp.
  1--10.

\bibitem{golmohammadi2017}
------, ``Sentiment analysis on twitter to improve time series contextual
  anomaly detection for detecting stock market manipulation,'' in
  \emph{International Conference on Big Data Analytics and Knowledge
  Discovery}, 2017, pp. 327--342.

\bibitem{rabaoui2008}
A.~Rabaoui, M.~Davy, S.~Rossignol, and N.~Ellouze, ``Using one-class svms and
  wavelets for audio surveillance,'' \emph{IEEE Transactions on Information
  Forensics and Security}, vol.~3, no.~4, pp. 763--775, 2008.

\bibitem{marchi2015}
E.~Marchi, F.~Vesperini, F.~Eyben, S.~Squartini, and B.~Schuller, ``A novel
  approach for automatic acoustic novelty detection using a denoising
  autoencoder with bidirectional lstm neural networks,'' in \emph{International
  Conference on Acoustics, Speech, and Signal Processing}, 2015, pp.
  1996--2000.

\bibitem{lim2017}
H.~Lim, J.~Park, and Y.~Han, ``Rare sound event detection using 1d
  convolutional recurrent neural networks,'' in \emph{Workshop on Detection and
  Classification of Acoustic Scenes and Events}, 2017, pp. 80--84.

\bibitem{principi2017}
E.~Principi, F.~Vesperini, S.~Squartini, and F.~Piazza, ``Acoustic novelty
  detection with adversarial autoencoders,'' in \emph{International Joint
  Conference on Neural Networks}, 2017, pp. 3324--3330.

\bibitem{koizumi2018}
Y.~Koizumi, S.~Saito, H.~Uematsu, Y.~Kawachi, and N.~Harada, ``Unsupervised
  detection of anomalous sound based on deep learning and the neyman--pearson
  lemma,'' \emph{{IEEE} Transactions on Audio, Speech, and Language
  Processing}, vol.~27, no.~1, pp. 212--224, 2018.

\bibitem{tarassenko1995}
L.~Tarassenko, P.~Hayton, N.~Cerneaz, and M.~Brady, ``Novelty detection for the
  identification of masses in mammograms,'' in \emph{International Conference
  on Artificial Neural Networks}, 1995, pp. 442--447.

\bibitem{chauhan2015}
S.~Chauhan and L.~Vig, ``Anomaly detection in {ECG} time signals via deep long
  short-term memory networks,'' in \emph{{IEEE} International Conference on
  Data Science and Advanced Analytics}, 2015, pp. 1--7.

\bibitem{leibig2017}
C.~Leibig, V.~Allken, M.~S. Ayhan, P.~Berens, and S.~Wahl, ``Leveraging
  uncertainty information from deep neural networks for disease detection,''
  \emph{Scientific Reports}, vol.~7, no.~1, pp. 1--14, 2017.

\bibitem{litjens2017}
G.~Litjens, T.~Kooi, B.~E. Bejnordi, A.~A.~A. Setio, F.~Ciompi, M.~Ghafoorian,
  J.~A. Van Der~Laak, B.~Van~Ginneken, and C.~I. S{\'a}nchez, ``A survey on
  deep learning in medical image analysis,'' \emph{Medical Image Analysis},
  vol.~42, pp. 60--88, 2017.

\bibitem{schlegl2017}
T.~Schlegl, P.~Seeb{\"o}ck, S.~M. Waldstein, U.~Schmidt-Erfurth, and G.~Langs,
  ``Unsupervised anomaly detection with generative adversarial networks to
  guide marker discovery,'' in \emph{International Conference on Information
  Processing in Medical Imaging}, 2017, pp. 146--157.

\bibitem{chen2018}
X.~Chen and E.~Konukoglu, ``Unsupervised detection of lesions in brain {MRI}
  using constrained adversarial auto-encoders,'' in \emph{Medical Imaging with
  Deep Learning}, 2018.

\bibitem{iakovidis2018}
D.~K. Iakovidis, S.~V. Georgakopoulos, M.~Vasilakakis, A.~Koulaouzidis, and
  V.~P. Plagianakos, ``Detecting and locating gastrointestinal anomalies using
  deep learning and iterative cluster unification,'' \emph{IEEE Transactions on
  Medical Imaging}, vol.~37, no.~10, pp. 2196--2210, 2018.

\bibitem{latif2018}
S.~Latif, M.~Usman, R.~Rana, and J.~Qadir, ``Phonocardiographic sensing using
  deep learning for abnormal heartbeat detection,'' \emph{IEEE Sensors
  Journal}, vol.~18, no.~22, pp. 9393--9400, 2018.

\bibitem{pawlowski2018}
N.~Pawlowski, M.~C. Lee, M.~Rajchl, S.~McDonagh, E.~Ferrante, K.~Kamnitsas,
  S.~Cooke, S.~Stevenson, A.~Khetani, T.~Newman, F.~Zeiler, R.~Digby, J.~P.
  Coles, D.~Rueckert, D.~K. Menon, V.~F.~J. Newcombe, and B.~Glocker,
  ``Unsupervised lesion detection in brain {CT} using bayesian convolutional
  autoencoders,'' in \emph{Medical Imaging with Deep Learning}, 2018.

\bibitem{baur2019}
C.~Baur, B.~Wiestler, S.~Albarqouni, and N.~Navab, ``Fusing unsupervised and
  supervised deep learning for white matter lesion segmentation,'' in
  \emph{Medical Imaging with Deep Learning}, 2019, pp. 63--72.

\bibitem{schlegl2019}
T.~Schlegl, P.~Seeb{\"o}ck, S.~M. Waldstein, G.~Langs, and U.~Schmidt-Erfurth,
  ``f-{AnoGAN}: Fast unsupervised anomaly detection with generative adversarial
  networks,'' \emph{Medical Image Analysis}, vol.~54, pp. 30--44, 2019.

\bibitem{seebock2019}
P.~Seeb{\"o}ck, J.~I. Orlando, T.~Schlegl, S.~M. Waldstein, H.~Bogunovi{\'c},
  S.~Klimscha, G.~Langs, and U.~Schmidt-Erfurth, ``Exploiting epistemic
  uncertainty of anatomy segmentation for anomaly detection in retinal {OCT},''
  \emph{IEEE Transactions on Medical Imaging}, vol.~39, no.~1, pp. 87--98,
  2019.

\bibitem{guo2020}
P.~Guo, Z.~Xue, Z.~Mtema, K.~Yeates, O.~Ginsburg, M.~Demarco, L.~R. Long,
  M.~Schiffman, and S.~Antani, ``Ensemble deep learning for cervix image
  selection toward improving reliability in automated cervical precancer
  screening,'' \emph{Diagnostics}, vol.~10, no.~7, p. 451, 2020.

\bibitem{naud2020}
L.~Naud and A.~Lavin, ``Manifolds for unsupervised visual anomaly detection,''
  \emph{arXiv preprint arXiv:2006.11364}, 2020.

\bibitem{tuluptceva2020}
N.~Tuluptceva, B.~Bakker, I.~Fedulova, H.~Schulz, and D.~V. Dylov, ``Anomaly
  detection with deep perceptual autoencoders,'' \emph{arXiv preprint
  arXiv:2006.13265}, 2020.

\bibitem{wong2003}
W.-K. Wong, A.~W. Moore, G.~F. Cooper, and M.~M. Wagner, ``Bayesian network
  anomaly pattern detection for disease outbreaks,'' in \emph{International
  Conference on Machine Learning}, 2003, pp. 808--815.

\bibitem{wong2005}
W.-K. Wong, A.~Moore, G.~Cooper, and M.~Wagner, ``What’s strange about recent
  events (wsare): An algorithm for the early detection of disease outbreaks,''
  \emph{Journal of Machine Learning Research}, vol.~6, no. Dec, pp. 1961--1998,
  2005.

\bibitem{blender1997}
R.~Blender, K.~Fraedrich, and F.~Lunkeit, ``Identification of cyclone-track
  regimes in the north atlantic,'' \emph{Quarterly Journal of the Royal
  Meteorological Society}, vol. 123, no. 539, pp. 727--741, 1997.

\bibitem{verbesselt2012}
J.~Verbesselt, A.~Zeileis, and M.~Herold, ``Near real-time disturbance
  detection using satellite image time series,'' \emph{Remote Sensing of
  Environment}, vol. 123, pp. 98--108, 2012.

\bibitem{fisher2017}
W.~D. Fisher, T.~K. Camp, and V.~V. Krzhizhanovskaya, ``Anomaly detection in
  earth dam and levee passive seismic data using support vector machines and
  automatic feature selection,'' \emph{Journal of Computational Science},
  vol.~20, pp. 143--153, 2017.

\bibitem{flach2017}
M.~Flach, F.~Gans, A.~Brenning, J.~Denzler, M.~Reichstein, E.~Rodner,
  S.~Bathiany, P.~Bodesheim, Y.~Guanche, S.~Sippel \emph{et~al.},
  ``Multivariate anomaly detection for earth observations: a comparison of
  algorithms and feature extraction techniques,'' \emph{Earth System Dynamics},
  vol.~8, no.~3, pp. 677--696, 2017.

\bibitem{wu2018}
Y.~Wu, Y.~Lin, Z.~Zhou, D.~C. Bolton, J.~Liu, and P.~Johnson, ``Deepdetect: A
  cascaded region-based densely connected network for seismic event
  detection,'' \emph{IEEE Transactions on Geoscience and Remote Sensing},
  vol.~57, no.~1, pp. 62--75, 2018.

\bibitem{jiang2020}
T.~Jiang, Y.~Li, W.~Xie, and Q.~Du, ``Discriminative reconstruction constrained
  generative adversarial network for hyperspectral anomaly detection,''
  \emph{IEEE Transactions on Geoscience and Remote Sensing}, 2020.

\bibitem{oprea2002}
T.~I. Oprea, ``Chemical space navigation in lead discovery,'' \emph{Current
  Opinion in Chemical Biology}, vol.~6, no.~3, pp. 384--389, 2002.

\bibitem{gromski2019}
P.~S. Gromski, A.~B. Henson, J.~M. Granda, and L.~Cronin, ``How to explore
  chemical space using algorithms and automation,'' \emph{Nature Reviews
  Chemistry}, vol.~3, no.~2, pp. 119--128, 2019.

\bibitem{min2017}
S.~Min, B.~Lee, and S.~Yoon, ``Deep learning in bioinformatics,''
  \emph{Briefings in Bioinformatics}, vol.~18, no.~5, pp. 851--869, 2017.

\bibitem{tomlins2005}
S.~A. Tomlins, D.~R. Rhodes, S.~Perner, S.~M. Dhanasekaran, R.~Mehra, X.-W.
  Sun, S.~Varambally, X.~Cao, J.~Tchinda, R.~Kuefer \emph{et~al.}, ``Recurrent
  fusion of {TMPRSS2} and {ETS} transcription factor genes in prostate
  cancer,'' \emph{Science}, vol. 310, no. 5748, pp. 644--648, 2005.

\bibitem{tibshirani2007}
R.~Tibshirani and T.~Hastie, ``Outlier sums for differential gene expression
  analysis,'' \emph{Biostatistics}, vol.~8, no.~1, pp. 2--8, 2007.

\bibitem{cerri2019}
O.~Cerri, T.~Q. Nguyen, M.~Pierini, M.~Spiropulu, and J.-R. Vlimant,
  ``Variational autoencoders for new physics mining at the large hadron
  collider,'' \emph{Journal of High Energy Physics}, vol. 2019, no.~5, p.~36,
  2019.

\bibitem{kharkov2020}
Y.~A. Kharkov, V.~Sotskov, A.~Karazeev, E.~Kiktenko, and A.~Fedorov,
  ``Revealing quantum chaos with machine learning,'' \emph{Physical Review B},
  vol. 101, no.~6, p. 064406, 2020.

\bibitem{protopapas2006}
P.~Protopapas, J.~Giammarco, L.~Faccioli, M.~Struble, R.~Dave, and C.~Alcock,
  ``Finding outlier light curves in catalogues of periodic variable stars,''
  \emph{Monthly Notices of the Royal Astronomical Society}, vol. 369, no.~2,
  pp. 677--696, 2006.

\bibitem{dutta2007}
H.~Dutta, C.~Giannella, K.~Borne, and H.~Kargupta, ``Distributed top-k outlier
  detection from astronomy catalogs using the {DEMAC} system,'' in \emph{{SIAM}
  International Conference on Data Mining}, 2007, pp. 473--478.

\bibitem{henrion2013}
M.~Henrion, D.~J. Mortlock, D.~J. Hand, and A.~Gandy, ``Classification and
  anomaly detection for astronomical survey data,'' in \emph{Astrostatistical
  Challenges for the New Astronomy}.\hskip 1em plus 0.5em minus 0.4em\relax
  Springer New York, 2013, pp. 149--184.

\bibitem{reyes2020}
E.~Reyes and P.~A. Est{\'e}vez, ``Transformation based deep anomaly detection
  in astronomical images,'' \emph{arXiv preprint arXiv:2005.07779}, 2020.

\bibitem{bengio2013}
Y.~Bengio, A.~Courville, and P.~Vincent, ``Representation learning: A review
  and new perspectives,'' \emph{{IEEE} Transactions on Pattern Analysis and
  Machine Intelligence}, vol.~35, no.~8, pp. 1798--1828, 2013.

\bibitem{lecun2015}
Y.~LeCun, Y.~Bengio, and G.~Hinton, ``Deep learning,'' \emph{Nature}, vol. 521,
  no. 7553, pp. 436--444, 2015.

\bibitem{schmidhuber2015}
J.~Schmidhuber, ``Deep learning in neural networks: An overview,'' \emph{Neural
  Networks}, vol.~61, pp. 85--117, 2015.

\bibitem{goodfellow2016}
I.~Goodfellow, Y.~Bengio, and A.~Courville, \emph{Deep learning}.\hskip 1em
  plus 0.5em minus 0.4em\relax MIT press, 2016.

\bibitem{krizhevsky2012}
A.~Krizhevsky, I.~Sutskever, and G.~E. Hinton, ``Imagenet classification with
  deep convolutional neural networks,'' in \emph{Advances in Neural Information
  Processing Systems}, 2012, pp. 1097--1105.

\bibitem{simonyan2015}
K.~Simonyan and A.~Zisserman, ``Very deep convolutional networks for
  large-scale image recognition,'' in \emph{International Conference on
  Learning Representations}, 2015.

\bibitem{szegedy2015}
C.~Szegedy, W.~Liu, Y.~Jia, P.~Sermanet, S.~Reed, D.~Anguelov, D.~Erhan,
  V.~Vanhoucke, and A.~Rabinovich, ``Going deeper with convolutions,'' in
  \emph{{IEEE} Conference on Computer Vision and Pattern Recognition}, 2015,
  pp. 1--9.

\bibitem{long2015}
J.~Long, E.~Shelhamer, and T.~Darrell, ``Fully convolutional networks for
  semantic segmentation,'' in \emph{{IEEE} Conference on Computer Vision and
  Pattern Recognition}, 2015, pp. 3431--3440.

\bibitem{ren2015}
S.~Ren, K.~He, R.~Girshick, and J.~Sun, ``Faster {R-CNN}: Towards real-time
  object detection with region proposal networks,'' in \emph{Advances in Neural
  Information Processing Systems}, 2015, pp. 91--99.

\bibitem{gatys2016}
L.~A. Gatys, A.~S. Ecker, and M.~Bethge, ``Image style transfer using
  convolutional neural networks,'' in \emph{{IEEE} Conference on Computer
  Vision and Pattern Recognition}, 2016, pp. 2414--2423.

\bibitem{he2016}
K.~He, X.~Zhang, S.~Ren, and J.~Sun, ``Deep residual learning for image
  recognition,'' in \emph{{IEEE} Conference on Computer Vision and Pattern
  Recognition}, 2016, pp. 770--778.

\bibitem{redmon2016}
J.~Redmon, S.~Divvala, R.~Girshick, and A.~Farhadi, ``You only look once:
  Unified, real-time object detection,'' in \emph{{IEEE} Conference on Computer
  Vision and Pattern Recognition}, 2016, pp. 779--788.

\bibitem{karras2019}
T.~Karras, S.~Laine, and T.~Aila, ``A style-based generator architecture for
  generative adversarial networks,'' in \emph{{IEEE} Conference on Computer
  Vision and Pattern Recognition}, 2019, pp. 4401--4410.

\bibitem{xie2020}
Q.~Xie, M.-T. Luong, E.~Hovy, and Q.~V. Le, ``Self-training with noisy student
  improves imagenet classification,'' in \emph{{IEEE} Conference on Computer
  Vision and Pattern Recognition}, 2020, pp. 10\,687--10\,698.

\bibitem{lee2009b}
H.~Lee, P.~Pham, Y.~Largman, and A.~Y. Ng, ``Unsupervised feature learning for
  audio classification using convolutional deep belief networks,'' in
  \emph{Advances in Neural Information Processing Systems}, 2009, pp.
  1096--1104.

\bibitem{dahl2011}
G.~E. Dahl, D.~Yu, L.~Deng, and A.~Acero, ``Context-dependent pre-trained deep
  neural networks for large-vocabulary speech recognition,'' \emph{{IEEE}
  Transactions on Audio, Speech, and Language Processing}, vol.~20, no.~1, pp.
  30--42, 2011.

\bibitem{mohamed2011}
A.-r. Mohamed, G.~E. Dahl, and G.~Hinton, ``Acoustic modeling using deep belief
  networks,'' \emph{{IEEE} Transactions on Audio, Speech, and Language
  Processing}, vol.~20, no.~1, pp. 14--22, 2011.

\bibitem{hinton2012}
G.~Hinton, L.~Deng, D.~Yu, G.~E. Dahl, A.-r. Mohamed, N.~Jaitly, A.~Senior,
  V.~Vanhoucke, P.~Nguyen, T.~N. Sainath, and B.~Kingsbury, ``Deep neural
  networks for acoustic modeling in speech recognition: The shared views of
  four research groups,'' \emph{IEEE Signal Processing Magazine}, vol.~29,
  no.~6, pp. 82--97, 2012.

\bibitem{graves2013}
A.~Graves, A.-r. Mohamed, and G.~Hinton, ``Speech recognition with deep
  recurrent neural networks,'' in \emph{International Conference on Acoustics,
  Speech, and Signal Processing}, 2013, pp. 6645--6649.

\bibitem{hannun2014}
A.~Hannun, C.~Case, J.~Casper, B.~Catanzaro, G.~Diamos, E.~Elsen, R.~Prenger,
  S.~Satheesh, S.~Sengupta, A.~Coates \emph{et~al.}, ``Deep speech: Scaling up
  end-to-end speech recognition,'' \emph{arXiv preprint arXiv:1412.5567}, 2014.

\bibitem{amodei2016b}
D.~Amodei, S.~Ananthanarayanan, R.~Anubhai, J.~Bai, E.~Battenberg, C.~Case,
  J.~Casper, B.~Catanzaro, Q.~Cheng, G.~Chen, J.~Chen, J.~Chen, Z.~Chen,
  M.~Chrzanowski, A.~Coates, G.~Diamos, K.~Ding, N.~Du, E.~Elsen, J.~Engel,
  W.~Fang, L.~Fan, C.~Fougner, L.~Gao, C.~Gong, A.~Hannun, T.~Han, L.~Johannes,
  B.~Jiang, C.~Ju, B.~Jun, P.~LeGresley, L.~Lin, J.~Liu, Y.~Liu, W.~Li, X.~Li,
  D.~Ma, S.~Narang, A.~Ng, S.~Ozair, Y.~Peng, R.~Prenger, S.~Qian, Z.~Quan,
  J.~Raiman, V.~Rao, S.~Satheesh, D.~Seetapun, S.~Sengupta, K.~Srinet,
  A.~Sriram, H.~Tang, L.~Tang, C.~Wang, J.~Wang, K.~Wang, Y.~Wang, Z.~Wang,
  Z.~Wang, S.~Wu, L.~Wei, B.~Xiao, W.~Xie, Y.~Xie, D.~Yogatama, B.~Yuan,
  J.~Zhan, and Z.~Zhu, ``Deep speech 2: End-to-end speech recognition in
  english and mandarin,'' in \emph{International Conference on Machine
  Learning}, vol.~48, 2016, pp. 173--182.

\bibitem{chan2016}
W.~Chan, N.~Jaitly, Q.~Le, and O.~Vinyals, ``Listen, attend and spell: A neural
  network for large vocabulary conversational speech recognition,'' in
  \emph{International Conference on Acoustics, Speech, and Signal Processing},
  2016, pp. 4960--4964.

\bibitem{chorowski2019}
J.~Chorowski, R.~J. Weiss, S.~Bengio, and A.~van~den Oord, ``Unsupervised
  speech representation learning using {WaveNet} autoencoders,'' \emph{{IEEE}
  Transactions on Audio, Speech, and Language Processing}, vol.~27, no.~12, pp.
  2041--2053, 2019.

\bibitem{schneider2019}
S.~Schneider, A.~Baevski, R.~Collobert, and M.~Auli, ``wav2vec: Unsupervised
  pre-training for speech recognition,'' in \emph{Interspeech}, 2019, pp.
  3465--3469.

\bibitem{bengio2003}
Y.~Bengio, R.~Ducharme, P.~Vincent, and C.~Jauvin, ``A neural probabilistic
  language model,'' \emph{Journal of Machine Learning Research}, vol.~3, no.
  Feb, pp. 1137--1155, 2003.

\bibitem{mikolov2013}
T.~Mikolov, I.~Sutskever, K.~Chen, G.~S. Corrado, and J.~Dean, ``Distributed
  representations of words and phrases and their compositionality,'' in
  \emph{Advances in Neural Information Processing Systems}, 2013, pp.
  3111--3119.

\bibitem{pennington2014}
J.~Pennington, R.~Socher, and C.~D. Manning, ``{GloVe}: Global vectors for word
  representation,'' in \emph{Conference on Empirical Methods in Natural
  Language Processing}, 2014, pp. 1532--1543.

\bibitem{cho2014}
K.~Cho, B.~van Merri{\"e}nboer, C.~Gulcehre, D.~Bahdanau, F.~Bougares,
  H.~Schwenk, and Y.~Bengio, ``Learning phrase representations using {RNN}
  encoder{--}decoder for statistical machine translation,'' in \emph{Conference
  on Empirical Methods in Natural Language Processing}, 2014, pp. 1724--1734.

\bibitem{bojanowski2017}
P.~Bojanowski, E.~Grave, A.~Joulin, and T.~Mikolov, ``Enriching word vectors
  with subword information,'' \emph{Transactions of the Association for
  Computational Linguistics}, vol.~5, pp. 135--146, 2017.

\bibitem{joulin2017}
A.~Joulin, {\'E}.~Grave, P.~Bojanowski, and T.~Mikolov, ``Bag of tricks for
  efficient text classification,'' in \emph{Conference of the European Chapter
  of the Association for Computational Linguistics}, 2017, pp. 427--431.

\bibitem{peters2018}
M.~Peters, M.~Neumann, M.~Iyyer, M.~Gardner, C.~Clark, K.~Lee, and
  L.~Zettlemoyer, ``Deep contextualized word representations,'' in \emph{North
  American Chapter of the Association for Computational Linguistics}, 2018, pp.
  2227--2237.

\bibitem{devlin2019}
J.~Devlin, M.-W. Chang, K.~Lee, and K.~Toutanova, ``{BERT}: Pre-training of
  deep bidirectional transformers for language understanding,'' in \emph{North
  American Chapter of the Association for Computational Linguistics}, 2019, pp.
  4171--4186.

\bibitem{wu2019}
C.-S. Wu, A.~Madotto, E.~Hosseini-Asl, C.~Xiong, R.~Socher, and P.~Fung,
  ``Transferable multi-domain state generator for task-oriented dialogue
  systems,'' in \emph{Annual Meeting of the Association for Computational
  Linguistics}, 2019, pp. 808--819.

\bibitem{brown2020}
T.~B. Brown, B.~Mann, N.~Ryder, M.~Subbiah, J.~Kaplan, P.~Dhariwal,
  A.~Neelakantan, P.~Shyam, G.~Sastry, A.~Askell, S.~Agarwal, A.~Herbert-Voss,
  G.~Krueger, T.~Henighan, R.~Child, A.~Ramesh, D.~M. Ziegler, J.~Wu,
  C.~Winter, C.~Hesse, M.~Chen, E.~Sigler, M.~Litwin, S.~Gray, B.~Chess,
  J.~Clark, C.~Berner, S.~McCandlish, A.~Radford, I.~Sutskever, and D.~Amodei,
  ``Language models are few-shot learners,'' in \emph{Advances in Neural
  Information Processing Systems}, 2020.

\bibitem{lengauer2007bioinformatics}
T.~Lengauer, O.~Sander, S.~Sierra, A.~Thielen, and R.~Kaiser, ``Bioinformatics
  prediction of {HIV} coreceptor usage,'' \emph{Nature Biotechnology}, vol.~25,
  no.~12, p. 1407, 2007.

\bibitem{baldi2014searching}
P.~Baldi, P.~Sadowski, and D.~Whiteson, ``Searching for exotic particles in
  high-energy physics with deep learning,'' \emph{Nature Communications},
  vol.~5, p. 4308, 2014.

\bibitem{schutt2017quantum}
K.~T. Sch{\"u}tt, F.~Arbabzadah, S.~Chmiela, K.-R. M{\"u}ller, and
  A.~Tkatchenko, ``Quantum-chemical insights from deep tensor neural
  networks,'' \emph{Nature Communications}, vol.~8, no.~1, pp. 1--8, 2017.

\bibitem{carleo2017solving}
G.~Carleo and M.~Troyer, ``Solving the quantum many-body problem with
  artificial neural networks,'' \emph{Science}, vol. 355, no. 6325, pp.
  602--606, 2017.

\bibitem{schnorb}
K.~Sch{\"u}tt, M.~Gastegger, A.~Tkatchenko, K.-R. M{\"u}ller, and R.~Maurer,
  ``Unifying machine learning and quantum chemistry with a deep neural network
  for molecular wavefunctions,'' \emph{Nature Communications}, vol.~10, p.
  5024, 2019.

\bibitem{jurmeister2019machine}
P.~Jurmeister, M.~Bockmayr, P.~Seegerer, T.~Bockmayr, D.~Treue, G.~Montavon,
  C.~Vollbrecht, A.~Arnold, D.~Teichmann, K.~Bressem \emph{et~al.}, ``Machine
  learning analysis of {DNA} methylation profiles distinguishes primary lung
  squamous cell carcinomas from head and neck metastases,'' \emph{Science
  Translational Medicine}, vol.~11, no. 509, 2019.

\bibitem{klauschen2018scoring}
F.~Klauschen, K.-R. M{\"u}ller, A.~Binder, M.~Bockmayr, M.~H{\"a}gele,
  P.~Seegerer, S.~Wienert, G.~Pruneri, S.~de~Maria, S.~Badve \emph{et~al.},
  ``Scoring of tumor-infiltrating lymphocytes: From visual estimation to
  machine learning,'' \emph{Seminars in Cancer Biology}, vol.~52, no.~2, p.
  151, 2018.

\bibitem{arcadu2019deep}
F.~Arcadu, F.~Benmansour, A.~Maunz, J.~Willis, Z.~Haskova, and M.~Prunotto,
  ``Deep learning algorithm predicts diabetic retinopathy progression in
  individual patients,'' \emph{npj Digital Medicine}, vol.~2, no.~1, pp. 1--9,
  2019.

\bibitem{ardila2019end}
D.~Ardila, A.~P. Kiraly, S.~Bharadwaj, B.~Choi, J.~J. Reicher, L.~Peng, D.~Tse,
  M.~Etemadi, W.~Ye, G.~Corrado \emph{et~al.}, ``End-to-end lung cancer
  screening with three-dimensional deep learning on low-dose chest computed
  tomography,'' \emph{Nature Medicine}, vol.~25, no.~6, pp. 954--961, 2019.

\bibitem{esteva2019guide}
A.~Esteva, A.~Robicquet, B.~Ramsundar, V.~Kuleshov, M.~DePristo, K.~Chou,
  C.~Cui, G.~Corrado, S.~Thrun, and J.~Dean, ``A guide to deep learning in
  healthcare,'' \emph{Nature Medicine}, vol.~25, no.~1, pp. 24--29, 2019.

\bibitem{faust2018}
K.~Faust, Q.~Xie, D.~Han, K.~Goyle, Z.~Volynskaya, U.~Djuric, and P.~Diamandis,
  ``Visualizing histopathologic deep learning classification and anomaly
  detection using nonlinear feature space dimensionality reduction,''
  \emph{{BMC} Bioinformatics}, vol.~19, no.~1, p. 173, 2018.

\bibitem{chalapathy2017}
R.~Chalapathy, A.~K. Menon, and S.~Chawla, ``Robust, deep and inductive anomaly
  detection,'' in \emph{Joint European Conference on Machine Learning and
  Knowledge Discovery in Databases}, 2017, pp. 36--51.

\bibitem{chen2017b}
J.~Chen, S.~Sathe, C.~C. Aggarwal, and D.~S. Turaga, ``Outlier {D}etection with
  {A}utoencoder {E}nsembles,'' in \emph{{SIAM} International Conference on Data
  Mining}, 2017, pp. 90--98.

\bibitem{zhou2017}
C.~Zhou and R.~C. Paffenroth, ``Anomaly detection with robust deep
  autoencoders,'' in \emph{International Conference on Knowledge Discovery \&
  Data Mining}, 2017, pp. 665--674.

\bibitem{zong2018}
B.~Zong, Q.~Song, M.~R. Min, W.~Cheng, C.~Lumezanu, D.~Cho, and H.~Chen, ``Deep
  autoencoding {G}aussian mixture model for unsupervised anomaly detection,''
  in \emph{International Conference on Learning Representations}, 2018.

\bibitem{aytekin2018}
C.~Aytekin, X.~Ni, F.~Cricri, and E.~Aksu, ``Clustering and unsupervised
  anomaly detection with $l_2$ normalized deep auto-encoder representations,''
  in \emph{International Joint Conference on Neural Networks}, 2018, pp. 1--6.

\bibitem{abati2019}
D.~Abati, A.~Porrello, S.~Calderara, and R.~Cucchiara, ``Latent space
  autoregression for novelty detection,'' in \emph{{IEEE} Conference on
  Computer Vision and Pattern Recognition}, 2019, pp. 481--490.

\bibitem{huang2019}
C.~Huang, J.~Cao, F.~Ye, M.~Li, Y.~Zhang, and C.~Lu, ``Inverse-transform
  autoencoder for anomaly detection,'' \emph{arXiv preprint arXiv:1911.10676},
  2019.

\bibitem{gong2019}
D.~Gong, L.~Liu, V.~Le, B.~Saha, M.~R. Mansour, S.~Venkatesh, and A.~v.~d.
  Hengel, ``Memorizing normality to detect anomaly: Memory-augmented deep
  autoencoder for unsupervised anomaly detection,'' in \emph{International
  Conference on Computer Vision}, 2019, pp. 1705--1714.

\bibitem{oza2019b}
P.~Oza and V.~M. Patel, ``{C2AE}: Class conditioned auto-encoder for open-set
  recognition,'' in \emph{{IEEE} Conference on Computer Vision and Pattern
  Recognition}, 2019, pp. 2307--2316.

\bibitem{nguyen2019}
D.~T. Nguyen, Z.~Lou, M.~Klar, and T.~Brox, ``Anomaly detection with
  multiple-hypotheses predictions,'' in \emph{International Conference on
  Machine Learning}, vol.~97, 2019, pp. 4800--4809.

\bibitem{kim2020}
K.~H. Kim, S.~Shim, Y.~Lim, J.~Jeon, J.~Choi, B.~Kim, and A.~S. Yoon, ``{RaPP}:
  Novelty detection with reconstruction along projection pathway,'' in
  \emph{International Conference on Learning Representations}, 2020.

\bibitem{erfani2016}
S.~M. Erfani, S.~Rajasegarar, S.~Karunasekera, and C.~Leckie,
  ``High-dimensional and large-scale anomaly detection using a linear one-class
  {SVM} with deep learning,'' \emph{Pattern Recognition}, vol.~58, pp.
  121--134, 2016.

\bibitem{ruff2018}
L.~Ruff, R.~A. Vandermeulen, N.~G{\"o}rnitz, L.~Deecke, S.~A. Siddiqui,
  A.~Binder, E.~M{\"u}ller, and M.~Kloft, ``Deep one-class classification,'' in
  \emph{International Conference on Machine Learning}, vol.~80, 2018, pp.
  4390--4399.

\bibitem{sabokrou2018}
M.~Sabokrou, M.~Khalooei, M.~Fathy, and E.~Adeli, ``Adversarially learned
  one-class classifier for novelty detection,'' in \emph{{IEEE} Conference on
  Computer Vision and Pattern Recognition}, 2018, pp. 3379--3388.

\bibitem{oza2019}
P.~Oza and V.~M. Patel, ``One-class convolutional neural network,''
  \emph{{IEEE} Signal Processing Letters}, vol.~26, no.~2, pp. 277--281, 2019.

\bibitem{ruff2019}
L.~Ruff, Y.~Zemlyanskiy, R.~Vandermeulen, T.~Schnake, and M.~Kloft,
  ``Self-attentive, multi-context one-class classification for unsupervised
  anomaly detection on text,'' in \emph{Annual Meeting of the Association for
  Computational Linguistics}, 2019, pp. 4061--4071.

\bibitem{perera2019a}
P.~Perera, R.~Nallapati, and B.~Xiang, ``{OCGAN}: One-class novelty detection
  using {GAN}s with constrained latent representations,'' in \emph{{IEEE}
  Conference on Computer Vision and Pattern Recognition}, 2019, pp. 2898--2906.

\bibitem{perera2019c}
P.~Perera and V.~M. Patel, ``Learning deep features for one-class
  classification,'' \emph{IEEE Transactions on Image Processing}, vol.~28,
  no.~11, pp. 5450--5463, 2019.

\bibitem{wang2019d}
J.~Wang and A.~Cherian, ``{GODS}: Generalized one-class discriminative
  subspaces for anomaly detection,'' in \emph{International Conference on
  Computer Vision}, 2019, pp. 8201--8211.

\bibitem{ruff2020}
L.~Ruff, R.~A. Vandermeulen, N.~G{\"o}rnitz, A.~Binder, E.~M{\"u}ller, K.-R.
  M{\"u}ller, and M.~Kloft, ``Deep semi-supervised anomaly detection,'' in
  \emph{International Conference on Learning Representations}, 2020.

\bibitem{ghafoori2020}
Z.~Ghafoori and C.~Leckie, ``Deep multi-sphere support vector data
  description,'' in \emph{{SIAM} International Conference on Data Mining},
  2020, pp. 109--117.

\bibitem{chalapathy2018}
R.~Chalapathy, E.~Toth, and S.~Chawla, ``Group anomaly detection using deep
  generative models,'' in \emph{European Conference on Machine Learning and
  Principles and Practice of Knowledge Discovery in Databases}, 2018, pp.
  173--189.

\bibitem{deecke2018}
L.~Deecke, R.~A. Vandermeulen, L.~Ruff, S.~Mandt, and M.~Kloft, ``Image anomaly
  detection with generative adversarial networks,'' in \emph{European
  Conference on Machine Learning and Principles and Practice of Knowledge
  Discovery in Databases}, 2018, pp. 3--17.

\bibitem{akcay2018}
S.~Akcay, A.~Atapour-Abarghouei, and T.~P. Breckon, ``Ganomaly: Semi-supervised
  anomaly detection via adversarial training,'' in \emph{Computer Vision --
  ACCV 2018}, C.~V. Jawahar, H.~Li, G.~Mori, and K.~Schindler, Eds.\hskip 1em
  plus 0.5em minus 0.4em\relax Cham: Springer International Publishing, 2019,
  pp. 622--637.

\bibitem{choi2018}
H.~Choi, E.~Jang, and A.~A. Alemi, ``{WAIC}, but why? generative ensembles for
  robust anomaly detection,'' \emph{arXiv preprint arXiv:1810.01392}, 2018.

\bibitem{pidhorskyi2018}
S.~Pidhorskyi, R.~Almohsen, and G.~Doretto, ``Generative probabilistic novelty
  detection with adversarial autoencoders,'' in \emph{Advances in Neural
  Information Processing Systems}, 2018, pp. 6822--6833.

\bibitem{zenati2018a}
H.~Zenati, M.~Romain, C.-S. Foo, B.~Lecouat, and V.~Chandrasekhar,
  ``Adversarially learned anomaly detection,'' in \emph{{IEEE} International
  Conference on Data Mining}, 2018, pp. 727--736.

\bibitem{golan2018}
I.~Golan and R.~El-Yaniv, ``Deep anomaly detection using geometric
  transformations,'' in \emph{Advances in Neural Information Processing
  Systems}, 2018, pp. 9758--9769.

\bibitem{hendrycks2019d}
D.~Hendrycks, M.~Mazeika, S.~Kadavath, and D.~Song, ``Using self-supervised
  learning can improve model robustness and uncertainty,'' in \emph{Advances in
  Neural Information Processing Systems}, 2019, pp. 15\,637--15\,648.

\bibitem{wang2019c}
S.~Wang, Y.~Zeng, X.~Liu, E.~Zhu, J.~Yin, C.~Xu, and M.~Kloft, ``Effective
  end-to-end unsupervised outlier detection via inlier priority of
  discriminative network,'' in \emph{Advances in Neural Information Processing
  Systems}, 2019, pp. 5960--5973.

\bibitem{bergman2020b}
L.~Bergman and Y.~Hoshen, ``Classification-based anomaly detection for general
  data,'' in \emph{International Conference on Learning Representations}, 2020.

\bibitem{tack2020}
J.~Tack, S.~Mo, J.~Jeong, and J.~Shin, ``{CSI}: Novelty detection via
  contrastive learning on distributionally shifted instances,'' \emph{Advances
  in Neural Information Processing Systems}, 2020.

\bibitem{markou2003a}
M.~Markou and S.~Singh, ``Novelty detection: a review—part 1: statistical
  approaches,'' \emph{Signal Processing}, vol.~83, no.~12, pp. 2481--2497, Dec.
  2003.

\bibitem{markou2003b}
------, ``Novelty detection: a review—part 2: neural network based
  approaches,'' \emph{Signal Processing}, vol.~83, no.~12, pp. 2499--2521, Dec.
  2003.

\bibitem{hodge2004}
V.~Hodge and J.~Austin, ``A survey of outlier detection methodologies,''
  \emph{Artificial Intelligence Review}, vol.~22, no.~2, pp. 85--126, Oct.
  2004.

\bibitem{walfish2006}
S.~Walfish, ``A review of statistical outlier methods,'' \emph{Pharmaceutical
  Technology}, vol.~30, no.~11, pp. 1--5, 2006.

\bibitem{chandola2009}
V.~Chandola, A.~Banerjee, and V.~Kumar, ``Anomaly detection: A survey,''
  \emph{ACM Computing Surveys}, vol.~41, no.~3, pp. 1--58, 2009.

\bibitem{hadi2009}
A.~S. Hadi, R.~Imon, and M.~Werner, ``Detection of outliers,'' \emph{Wiley
  Interdisciplinary Reviews: Computational Statistics}, vol.~1, no.~1, pp.
  57--70, 2009.

\bibitem{gogoi2011}
P.~Gogoi, D.~K. Bhattacharyya, B.~Borah, and J.~K. Kalita, ``A survey of
  outlier detection methods in network anomaly identification,'' \emph{Computer
  Journal}, vol.~54, no.~4, pp. 570--588, 2011.

\bibitem{singh2012}
K.~Singh and S.~Upadhyaya, ``Outlier detection: applications and techniques,''
  \emph{International Journal of Computer Science Issues}, vol.~9, no.~1, p.
  307, 2012.

\bibitem{zimek2012}
A.~Zimek, E.~Schubert, and H.-P. Kriegel, ``A survey on unsupervised outlier
  detection in high-dimensional numerical data,'' \emph{Statistical Analysis
  and Data Mining: The ASA Data Science Journal}, vol.~5, no.~5, pp. 363--387,
  2012.

\bibitem{aguinis2013}
H.~Aguinis, R.~K. Gottfredson, and H.~Joo, ``Best-practice recommendations for
  defining, identifying, and handling outliers,'' \emph{Organizational Research
  Methods}, vol.~16, no.~2, pp. 270--301, 2013.

\bibitem{zhang2013}
J.~Zhang, ``Advancements of outlier detection: A survey,'' \emph{ICST
  Transactions on Scalable Information Systems}, vol.~13, no.~1, pp. 1--26,
  2013.

\bibitem{pimentel2014}
M.~A. Pimentel, D.~A. Clifton, L.~Clifton, and L.~Tarassenko, ``A review of
  novelty detection,'' \emph{Signal Processing}, vol.~99, pp. 215--249, 2014.

\bibitem{gupta2014}
M.~Gupta, J.~Gao, C.~C. Aggarwal, and J.~Han, ``Outlier detection for temporal
  data: A survey,'' \emph{Transactions on Knowledge and Data Engineering},
  vol.~26, no.~9, pp. 2250--2267, 2014.

\bibitem{agrawal2015}
S.~Agrawal and J.~Agrawal, ``Survey on anomaly detection using data mining
  techniques,'' \emph{Procedia Computer Science}, vol.~60, pp. 708--713, 2015.

\bibitem{akoglu2015}
L.~Akoglu, H.~Tong, and D.~Koutra, ``Graph based anomaly detection and
  description: A survey,'' \emph{Data Mining and Knowledge Discovery}, vol.~29,
  no.~3, pp. 626--688, 2015.

\bibitem{ranshous2015}
S.~Ranshous, S.~Shen, D.~Koutra, S.~Harenberg, C.~Faloutsos, and N.~F.
  Samatova, ``Anomaly detection in dynamic networks: A survey,'' \emph{Wiley
  Interdisciplinary Reviews: Computational Statistics}, vol.~7, no.~3, pp.
  223--247, 2015.

\bibitem{tamboli2016}
J.~Tamboli and M.~Shukla, ``A survey of outlier detection algorithms for data
  streams,'' in \emph{Proceedings of the 3rd International Conference on
  Computing for Sustainable Global Development}, 2016, pp. 3535--3540.

\bibitem{goldstein2016}
M.~Goldstein and S.~Uchida, ``A comparative evaluation of unsupervised anomaly
  detection algorithms for multivariate data,'' \emph{PLoS ONE}, vol.~11,
  no.~4, p. e0152173, 2016.

\bibitem{xu2019}
X.~Xu, H.~Liu, and M.~Yao, ``Recent progress of anomaly detection,''
  \emph{Complexity}, vol. 2019, pp. 1--11, 2019.

\bibitem{wang2019}
H.~Wang, M.~J. Bah, and M.~Hammad, ``Progress in outlier detection techniques:
  A survey,'' \emph{{IEEE} Access}, vol.~7, pp. 107\,964--108\,000, 2019.

\bibitem{barnett1994}
V.~Barnett and T.~Lewis, \emph{Outliers in Statistical Data}, 3rd~ed.\hskip 1em
  plus 0.5em minus 0.4em\relax Wiley, 1994.

\bibitem{rousseeuw2005}
P.~J. Rousseeuw and A.~M. Leroy, \emph{Robust Regression and Outlier
  Detection}.\hskip 1em plus 0.5em minus 0.4em\relax John Wiley \& Sons, 2005.

\bibitem{aggarwal2017}
C.~C. Aggarwal, \emph{Outlier Analysis}, 2nd~ed.\hskip 1em plus 0.5em minus
  0.4em\relax Springer International Publishing, 2017.

\bibitem{chalapathy2019}
R.~Chalapathy and S.~Chawla, ``Deep learning for anomaly detection: A survey,''
  \emph{arXiv preprint arXiv:1901.03407}, 2019.

\bibitem{diMattia2019}
F.~D. Mattia, P.~Galeone, M.~D. Simoni, and E.~Ghelfi, ``A survey on {GAN}s for
  anomaly detection,'' \emph{arXiv preprint arXiv:1906.11632}, 2019.

\bibitem{pang2020}
G.~Pang, C.~Shen, L.~Cao, and A.~van~den Hengel, ``Deep learning for anomaly
  detection: A review,'' \emph{arXiv preprint arXiv:2007.02500}, 2020.

\bibitem{muller2001introduction}
K.-R. M{\"u}ller, S.~Mika, G.~R{\"a}tsch, K.~Tsuda, and B.~Sch{\"o}lkopf, ``An
  introduction to kernel-based learning algorithms,'' \emph{IEEE Transactions
  on Neural Networks}, vol.~12, no.~2, pp. 181--201, 2001.

\bibitem{guyon2003}
I.~Guyon and A.~Elisseeff, ``An introduction to variable and feature
  selection,'' \emph{Journal of Machine Learning Research}, vol.~3, no. Mar,
  pp. 1157--1182, 2003.

\bibitem{garcia2015}
S.~Garc{\'\i}a, J.~Luengo, and F.~Herrera, \emph{Data Preprocessing in Data
  Mining}, 1st~ed.\hskip 1em plus 0.5em minus 0.4em\relax Springer
  International Publishing Switzerland, 2015.

\bibitem{rumsfeld2011}
D.~Rumsfeld, \emph{Known and Unknown: A Memoir}.\hskip 1em plus 0.5em minus
  0.4em\relax Penguin Group USA, 2011.

\bibitem{anscombe1960}
F.~J. Anscombe, ``Rejection of outliers,'' \emph{Technometrics}, vol.~2, no.~2,
  pp. 123--146, May 1960.

\bibitem{grubbs1969}
F.~E. Grubbs, ``Procedures for detecting outlying observations in samples,''
  \emph{Technometrics}, vol.~11, no.~1, pp. 1--21, 1969.

\bibitem{hawkins1980}
D.~M. Hawkins, \emph{Identification of Outliers}.\hskip 1em plus 0.5em minus
  0.4em\relax Springer Netherlands, 1980, vol.~11.

\bibitem{bergmann2019}
P.~Bergmann, M.~Fauser, D.~Sattlegger, and C.~Steger, ``{MVTec AD}--a
  comprehensive real-world dataset for unsupervised anomaly detection,'' in
  \emph{{IEEE} Conference on Computer Vision and Pattern Recognition}, 2019,
  pp. 9592--9600.

\bibitem{song2007}
X.~Song, M.~Wu, C.~Jermaine, and S.~Ranka, ``Conditional anomaly detection,''
  \emph{IEEE Transactions on Knowledge and Data Engineering}, vol.~19, no.~5,
  pp. 631--645, 2007.

\bibitem{smets2009}
K.~Smets, B.~Verdonk, and E.~M. Jordaan, ``Discovering novelty in
  spatio/temporal data using one-class support vector machines,'' in
  \emph{International Joint Conference on Neural Networks}, 2009, pp.
  2956--2963.

\bibitem{chandola2010}
V.~Chandola, A.~Banerjee, and V.~Kumar, ``Anomaly detection for discrete
  sequences: A survey,'' \emph{IEEE Transactions on Knowledge and Data
  Engineering}, vol.~24, no.~5, pp. 823--839, 2010.

\bibitem{lu2017}
W.~Lu, Y.~Cheng, C.~Xiao, S.~Chang, S.~Huang, B.~Liang, and T.~Huang,
  ``Unsupervised sequential outlier detection with deep architectures,''
  \emph{IEEE Transactions on Image Processing}, vol.~26, no.~9, pp. 4321--4330,
  2017.

\bibitem{samek2017robust}
W.~Samek, S.~Nakajima, M.~Kawanabe, and K.-R. M{\"u}ller, ``On robust parameter
  estimation in brain--computer interfacing,'' \emph{Journal of Neural
  Engineering}, vol.~14, no.~6, p. 061001, 2017.

\bibitem{xiong2011}
L.~Xiong, B.~P{\'o}czos, and J.~G. Schneider, ``Group anomaly detection using
  flexible genre models,'' in \emph{Advances in Neural Information Processing
  Systems}, 2011, pp. 1071--1079.

\bibitem{muandet2013}
K.~Muandet and B.~Sch{\"o}lkopf, ``One-class support measure machines for group
  anomaly detection,'' in \emph{Conference on Uncertainty in Artificial
  Intelligence}, 2013, pp. 449--458.

\bibitem{yu2015}
R.~Yu, X.~He, and Y.~Liu, ``{GLAD}: Group anomaly detection in social media
  analysis,'' \emph{ACM Transactions on Knowledge Discovery from Data},
  vol.~10, no.~2, pp. 1--22, 2015.

\bibitem{bontemps2016}
L.~Bontemps, J.~McDermott, N.-A. Le-Khac \emph{et~al.}, ``Collective anomaly
  detection based on long short-term memory recurrent neural networks,'' in
  \emph{International Conference on Future Data and Security
  Engineering}.\hskip 1em plus 0.5em minus 0.4em\relax Springer International
  Publishing, 2016, pp. 141--152.

\bibitem{ahmed2020}
F.~Ahmed and A.~Courville, ``Detecting semantic anomalies,'' in \emph{{AAAI}
  Conference on Artificial Intelligence}, 2020, pp. 3154--3162.

\bibitem{fox1972}
A.~J. Fox, ``Outliers in time series,'' \emph{Journal of the Royal Statistical
  Society: Series B (Methodological)}, vol.~34, no.~3, pp. 350--363, 1972.

\bibitem{tsay1988}
R.~S. Tsay, ``Outliers, level shifts, and variance changes in time series,''
  \emph{Journal of Forecasting}, vol.~7, no.~1, pp. 1--20, 1988.

\bibitem{tsay2000}
R.~S. Tsay, D.~Pe{\~n}a, and A.~E. Pankratz, ``Outliers in multivariate time
  series,'' \emph{Biometrika}, vol.~87, no.~4, pp. 789--804, Dec. 2000.

\bibitem{lavin2015}
A.~Lavin and S.~Ahmad, ``Evaluating real-time anomaly detection algorithms--the
  numenta anomaly benchmark,'' in \emph{International Conference on Machine
  Learning and Applications}.\hskip 1em plus 0.5em minus 0.4em\relax IEEE,
  2015, pp. 38--44.

\bibitem{chawla2006}
S.~Chawla and P.~Sun, ``{SLOM}: a new measure for local spatial outliers,''
  \emph{Knowledge and Information Systems}, vol.~9, no.~4, pp. 412--429, 2006.

\bibitem{schubert2014}
E.~Schubert, A.~Zimek, and H.-P. Kriegel, ``Local outlier detection
  reconsidered: a generalized view on locality with applications to spatial,
  video, and network outlier detection,'' \emph{Data Mining and Knowledge
  Discovery}, vol.~28, no.~1, pp. 190--237, 2014.

\bibitem{noble2003}
C.~C. Noble and D.~J. Cook, ``Graph-based anomaly detection,'' in
  \emph{International Conference on Knowledge Discovery \& Data Mining}, 2003,
  pp. 631--636.

\bibitem{honer2017minimizing}
J.~H{\"o}ner, S.~Nakajima, A.~Bauer, K.-R. M{\"u}ller, and N.~G{\"o}rnitz,
  ``Minimizing trust leaks for robust sybil detection,'' in \emph{International
  Conference on Machine Learning}, vol.~70, 2017, pp. 1520--1528.

\bibitem{ahmed2018}
M.~Ahmed, ``Collective anomaly detection techniques for network traffic
  analysis,'' \emph{Annals of Data Science}, vol.~5, no.~4, pp. 497--512, 2018.

\bibitem{locatello2019}
F.~Locatello, S.~Bauer, M.~Lucic, G.~R\"{a}tsch, S.~Gelly, B.~Sch{\"o}lkopf,
  and O.~Bachem, ``Challenging common assumptions in the unsupervised learning
  of disentangled representations,'' in \emph{International Conference on
  Machine Learning}, vol.~97, 2019, pp. 4114--4124.

\bibitem{scholkopf2002}
B.~Sch{\"o}lkopf and A.~J. Smola, \emph{Learning with Kernels}.\hskip 1em plus
  0.5em minus 0.4em\relax MIT press, 2002.

\bibitem{steinwart2005}
I.~Steinwart, D.~Hush, and C.~Scovel, ``A classification framework for anomaly
  detection,'' \emph{Journal of Machine Learning Research}, vol.~6, no. Feb,
  pp. 211--232, 2005.

\bibitem{chapelle2006}
O.~Chapelle, B.~Sch{\"o}lkopf, and A.~Zien, \emph{Semi-Supervised
  Learning}.\hskip 1em plus 0.5em minus 0.4em\relax The {MIT} Press, Cambridge,
  Massachusetts, 2006.

\bibitem{polonik1995}
W.~Polonik, ``Measuring mass concentrations and estimating density contour
  clusters-an excess mass approach,'' \emph{The Annals of Statistics}, vol.~23,
  no.~3, pp. 855--881, 1995.

\bibitem{tsybakov1997}
A.~B. Tsybakov, ``On nonparametric estimation of density level sets,''
  \emph{The Annals of Statistics}, vol.~25, no.~3, pp. 948--969, 1997.

\bibitem{ben1997}
S.~Ben-David and M.~Lindenbaum, ``Learning distributions by their density
  levels: A paradigm for learning without a teacher,'' \emph{Journal of
  Computer and System Sciences}, vol.~55, no.~1, pp. 171--182, 1997.

\bibitem{rigollet2009}
P.~Rigollet, R.~Vert \emph{et~al.}, ``Optimal rates for plug-in estimators of
  density level sets,'' \emph{Bernoulli}, vol.~15, no.~4, pp. 1154--1178, 2009.

\bibitem{polonik1997}
W.~Polonik, ``Minimum volume sets and generalized quantile processes,''
  \emph{Stochastic Processes and Their Applications}, vol.~69, no.~1, pp.
  1--24, 1997.

\bibitem{garcia2003}
J.~N. Garcia, Z.~Kutalik, K.-H. Cho, and O.~Wolkenhauer, ``Level sets and
  minimum volume sets of probability density functions,'' \emph{International
  Journal of Approximate Reasoning}, vol.~34, no.~1, pp. 25--47, 2003.

\bibitem{scott2006}
C.~D. Scott and R.~D. Nowak, ``Learning minimum volume sets,'' \emph{Journal of
  Machine Learning Research}, vol.~7, no. Apr, pp. 665--704, 2006.

\bibitem{ghaoui2003}
L.~E. Ghaoui, M.~I. Jordan, and G.~R. Lanckriet, ``Robust novelty detection
  with single-class {MPM},'' in \emph{Advances in Neural Information Processing
  Systems}, 2003, pp. 929--936.

\bibitem{menon2018}
A.~K. Menon and R.~C. Williamson, ``A loss framework for calibrated anomaly
  detection,'' in \emph{Advances in Neural Information Processing Systems},
  2018, pp. 1494--1504.

\bibitem{tax1999}
D.~M.~J. Tax and R.~P.~W. Duin, ``Support vector domain description,''
  \emph{Pattern Recognition Letters}, vol.~20, no.~11, pp. 1191--1199, 1999.

\bibitem{tax2001}
D.~M.~J. Tax, ``One-class classification,'' Ph.D. dissertation, Delft
  University of Technology, 2001.

\bibitem{clemencon2013}
S.~Cl{\'e}men{\c{c}}on and J.~Jakubowicz, ``Scoring anomalies: a m-estimation
  formulation,'' in \emph{International Conference on Artificial Intelligence
  and Statistics}, 2013, pp. 659--667.

\bibitem{goix2015}
N.~Goix, A.~Sabourin, and S.~Cl{\'e}men{\c{c}}on, ``On anomaly ranking and
  excess-mass curves,'' in \emph{International Conference on Artificial
  Intelligence and Statistics}, 2015, pp. 287--295.

\bibitem{hampel2005}
F.~R. Hampel, E.~M. Ronchetti, P.~J. Rousseeuw, and W.~A. Stahel, \emph{Robust
  Statistics: The Approach Based on Influence Functions}.\hskip 1em plus 0.5em
  minus 0.4em\relax John Wiley \& Sons, 2005.

\bibitem{liu2006}
Y.~Liu and Y.~F. Zheng, ``Minimum enclosing and maximum excluding machine for
  pattern description and discrimination,'' in \emph{International Conference
  on Pattern Recognition}, 2006, pp. 129--132.

\bibitem{gornitz2013}
N.~G{\"o}rnitz, M.~Kloft, K.~Rieck, and U.~Brefeld, ``Toward supervised anomaly
  detection,'' \emph{Journal of Artificial Intelligence Research}, vol.~46, pp.
  235--262, 2013.

\bibitem{min2018}
E.~Min, J.~Long, Q.~Liu, J.~Cui, Z.~Cai, and J.~Ma, ``{SU-IDS}: A
  semi-supervised and unsupervised framework for network intrusion detection,''
  in \emph{International Conference on Cloud Computing and Security}, 2018, pp.
  322--334.

\bibitem{kiran2018}
B.~R. Kiran, D.~M. Thomas, and R.~Parakkal, ``An overview of deep learning
  based methods for unsupervised and semi-supervised anomaly detection in
  videos,'' \emph{Journal of Imaging}, vol.~4, no.~2, p.~36, 2018.

\bibitem{Siddiqui2018}
A.~Siddiqui, A.~Fern, T.~G. Dietterich, R.~Wright, A.~Theriault, and D.~W.
  Archer, ``Feedback-guided anomaly discovery via online optimization,'' in
  \emph{International Conference on Knowledge Discovery \& Data Mining}, 2018,
  pp. 2200--2209.

\bibitem{hendrycks2019a}
D.~Hendrycks, M.~Mazeika, and T.~G. Dietterich, ``Deep anomaly detection with
  outlier exposure,'' in \emph{International Conference on Learning
  Representations}, 2019.

\bibitem{denis1998}
F.~Denis, ``{PAC} learning from positive statistical queries,'' in
  \emph{International Conference on Algorithmic Learning Theory}, 1998, pp.
  112--126.

\bibitem{zhang2008}
B.~Zhang and W.~Zuo, ``Learning from positive and unlabeled examples: A
  survey,'' in \emph{Proceedings of the {IEEE} International Symposium on
  Information Processing}, 2008, pp. 650--654.

\bibitem{duPlessis2014}
M.~C. Du~Plessis, G.~Niu, and M.~Sugiyama, ``Analysis of learning from positive
  and unlabeled data,'' in \emph{Advances in Neural Information Processing
  Systems}, 2014, pp. 703--711.

\bibitem{munoz2010}
J.~Mu{\~n}oz-Mar{\'\i}, F.~Bovolo, L.~G{\'o}mez-Chova, L.~Bruzzone, and
  G.~Camp-Valls, ``Semi-{S}upervised {O}ne-{C}lass {S}upport {V}ector
  {M}achines for {C}lassification of {R}emote {S}ensing {S}ata,'' \emph{IEEE
  Transactions on Geoscience and Remote Sensing}, vol.~48, no.~8, pp.
  3188--3197, 2010.

\bibitem{blanchard2010}
G.~Blanchard, G.~Lee, and C.~Scott, ``Semi-supervised novelty detection,''
  \emph{Journal of Machine Learning Research}, vol.~11, no. Nov, pp.
  2973--3009, 2010.

\bibitem{song2017}
H.~Song, Z.~Jiang, A.~Men, and B.~Yang, ``A hybrid semi-supervised anomaly
  detection model for high-dimensional data,'' \emph{Computational Intelligence
  and Neuroscience}, 2017.

\bibitem{liu2018a}
S.~Liu, R.~Garrepalli, T.~Dietterich, A.~Fern, and D.~Hendrycks, ``Open
  category detection with {PAC} guarantees,'' in \emph{International Conference
  on Machine Learning}, vol.~80, 2018, pp. 3169--3178.

\bibitem{hendrycks2017}
D.~Hendrycks and K.~Gimpel, ``A baseline for detecting misclassified and
  out-of-distribution examples in neural networks,'' in \emph{International
  Conference on Learning Representations}, 2017.

\bibitem{che2019}
T.~Che, X.~Liu, S.~Li, Y.~Ge, R.~Zhang, C.~Xiong, and Y.~Bengio, ``Deep
  verifier networks: Verification of deep discriminative models with deep
  generative models,'' \emph{arXiv preprint arXiv:1911.07421}, 2019.

\bibitem{schirrmeister2020}
R.~T. Schirrmeister, Y.~Zhou, T.~Ball, and D.~Zhang, ``Understanding anomaly
  detection with deep invertible networks through hierarchies of distributions
  and features,'' in \emph{Advances in Neural Information Processing Systems},
  2020.

\bibitem{boracchi2018}
G.~Boracchi, D.~Carrera, C.~Cervellera, and D.~Maccio, ``{QuantTree}:
  Histograms for change detection in multivariate data streams,'' in
  \emph{International Conference on Machine Learning}, vol.~80, 2018, pp.
  639--648.

\bibitem{sugiyama2007covariate}
M.~Sugiyama, M.~Krauledat, and K.-R. M{\"u}ller, ``Covariate shift adaptation
  by importance weighted cross validation,'' \emph{Journal of Machine Learning
  Research}, vol.~8, no. May, pp. 985--1005, 2007.

\bibitem{quionero2009dataset}
J.~Quionero-Candela, M.~Sugiyama, A.~Schwaighofer, and N.~D. Lawrence,
  \emph{Dataset Shift in Machine Learning}.\hskip 1em plus 0.5em minus
  0.4em\relax The MIT Press, 2009.

\bibitem{sugiyama2012machine}
M.~Sugiyama and M.~Kawanabe, \emph{Machine Learning in Non-Stationary
  Environments: Introduction to Covariate Shift Adaptation}.\hskip 1em plus
  0.5em minus 0.4em\relax MIT press, 2012.

\bibitem{baehrens2010explain}
D.~Baehrens, T.~Schroeter, S.~Harmeling, M.~Kawanabe, K.~Hansen, and K.-R.
  M{\"u}ller, ``How to explain individual classification decisions,''
  \emph{Journal of Machine Learning Research}, vol.~11, no. Jun, pp.
  1803--1831, 2010.

\bibitem{montavon2018methods}
G.~Montavon, W.~Samek, and K.-R. M{\"u}ller, ``Methods for interpreting and
  understanding deep neural networks,'' \emph{Digital Signal Processing},
  vol.~73, pp. 1--15, 2018.

\bibitem{lapuschkin-natcom19}
S.~Lapuschkin, S.~W{\"{a}}ldchen, A.~Binder, G.~Montavon, W.~Samek, and K.-R.
  M{\"{u}}ller, ``Unmasking {C}lever {H}ans predictors and assessing what
  machines really learn,'' \emph{Nature Communications}, vol.~10, p. 1096,
  2019.

\bibitem{DBLP:series/lncs/11700}
W.~Samek, G.~Montavon, A.~Vedaldi, L.~K. Hansen, and K.-R. M{\"{u}}ller, Eds.,
  \emph{Explainable {AI:} Interpreting, Explaining and Visualizing Deep
  Learning}, ser. Lecture Notes in Computer Science.\hskip 1em plus 0.5em minus
  0.4em\relax Springer International Publishing, 2019, vol. 11700.

\bibitem{hardle1990applied}
W.~H{\"a}rdle, \emph{Applied Nonparametric Regression}, ser. Econometric
  Society Monographs.\hskip 1em plus 0.5em minus 0.4em\relax Cambridge
  university press, 1990.

\bibitem{laurikkala00}
J.~Laurikkala, M.~Juhola, E.~Kentala, N.~Lavrac, S.~Miksch, and B.~Kavsek,
  ``Informal identification of outliers in medical data,'' in \emph{5th
  International Workshop on Intelligent Data Analysis in Medicine and
  Pharmacology}, vol.~1, 2000, pp. 20--24.

\bibitem{jain88}
A.~K. Jain and R.~C. Dubes, \emph{aa}.\hskip 1em plus 0.5em minus 0.4em\relax
  USA: Prentice-Hall, Inc., 1988.

\bibitem{roberts1994b}
S.~Roberts and L.~Tarassenko, ``A probabilistic resource allocating network for
  novelty detection,'' \emph{Neural Computation}, vol.~6, no.~2, pp. 270--284,
  Mar. 1994.

\bibitem{bishop1994}
C.~M. Bishop, ``Novelty detection and neural network validation,'' \emph{{IEE}
  Proceedings - Vision, Image and Signal Processing}, vol. 141, no.~4, pp.
  217--222, 1994.

\bibitem{devroye85}
L.~Devroye and L.~Gy\"orfi, \emph{Nonparametric Density Estimation: The L1
  View}.\hskip 1em plus 0.5em minus 0.4em\relax New York; Chichester: John
  Wiley \& Sons, 1985.

\bibitem{fruhwirth06}
S.~Fr{\"u}hwirth-Schnatter, \emph{Finite Mixture and Markov Switching
  Models}.\hskip 1em plus 0.5em minus 0.4em\relax Springer New York, 2006.

\bibitem{kim2012}
J.~Kim and C.~D. Scott, ``Robust kernel density estimation,'' \emph{Journal of
  Machine Learning Research}, vol.~13, no.~82, pp. 2529--2565, 2012.

\bibitem{vandermeulen13}
R.~Vandermeulen and C.~Scott, ``Consistency of robust kernel density
  estimators,'' in \emph{Conference on Learning Theory}, 2013, pp. 568--591.

\bibitem{amruthnath2018}
N.~Amruthnath and T.~Gupta, ``A research study on unsupervised machine learning
  algorithms for early fault detection in predictive maintenance,'' in
  \emph{International Conference on Industrial Engineering and Applications
  (ICIEA)}.\hskip 1em plus 0.5em minus 0.4em\relax IEEE, 2018, pp. 355--361.

\bibitem{fahlman83}
S.~E. Fahlman, G.~E. Hinton, and T.~J. Sejnowski, ``Massively parallel
  architectures for {AI}: {NETL}, {T}histle, and {B}oltzmann machines,'' in
  \emph{{AAAI} Conference on Artificial Intelligence}, 1983, pp. 109--113.

\bibitem{hopfield1982}
J.~J. Hopfield, ``Neural networks and physical systems with emergent collective
  computational abilities,'' \emph{Proceedings of the National Academy of
  Sciences of the United States of America}, vol.~79, no.~8, pp. 2554--2558,
  1982.

\bibitem{lecun2006}
Y.~LeCun, S.~Chopra, R.~Hadsell, M.~Ranzato, and F.~Huang, ``A tutorial on
  energy-based learning,'' in \emph{Predicting structured data}, G.~Bakir,
  T.~Hofman, B.~Scholkopt, A.~Smola, and B.~Taskar, Eds.\hskip 1em plus 0.5em
  minus 0.4em\relax MIT Press, 2006.

\bibitem{hinton2002}
G.~E. Hinton, ``Training products of experts by minimizing contrastive
  divergence,'' \emph{Neural Computation}, vol.~14, no.~8, pp. 1771--1800, Aug.
  2002.

\bibitem{welling2011}
M.~Welling and Y.~W. Teh, ``Bayesian learning via stochastic gradient langevin
  dynamics,'' in \emph{International Conference on Machine Learning}, 2011, pp.
  681--688.

\bibitem{grathwohl20}
W.~Grathwohl, K.-C. Wang, J.-H. Jacobsen, D.~Duvenaud, M.~Norouzi, and
  K.~Swersky, ``Your classifier is secretly an energy based model and you
  should treat it like one,'' in \emph{International Conference on Learning
  Representations}, 2020.

\bibitem{hinton06dbn}
G.~E. Hinton, S.~Osindero, and Y.-W. Teh, ``A fast learning algorithm for deep
  belief nets,'' \emph{Neural Computation}, vol.~18, no.~7, pp. 1527--1554,
  Jul. 2006.

\bibitem{salakhutdinov09}
R.~Salakhutdinov and G.~Hinton, ``Deep {B}oltzmann machines,'' in
  \emph{International Conference on Artificial Intelligence and Statistics},
  2009, pp. 448--455.

\bibitem{ngiam11}
J.~Ngiam, Z.~Chen, P.~W. Koh, and A.~Ng, ``Learning deep energy models,'' in
  \emph{International Conference on Machine Learning}, 2011, pp. 1105--1112.

\bibitem{zhai2016}
S.~Zhai, Y.~Cheng, W.~Lu, and Z.~Zhang, ``Deep structured energy based models
  for anomaly detection,'' in \emph{International Conference on Machine
  Learning}, vol.~48, 2016, pp. 1100--1109.

\bibitem{kingma2014}
D.~P. Kingma and M.~Welling, ``Auto-encoding variational bayes,'' in
  \emph{International Conference on Learning Representations}, 2014.

\bibitem{rezende2014}
D.~J. Rezende, S.~Mohamed, and D.~Wierstra, ``Stochastic backpropagation and
  approximate inference in deep generative models,'' in \emph{International
  Conference on Machine Learning}, vol.~32, 2014, pp. 1278--1286.

\bibitem{kingma2019}
D.~P. Kingma and M.~Welling, ``An introduction to variational autoencoders,''
  \emph{Foundations and Trends{\textregistered} in Machine Learning}, vol.~12,
  no.~4, pp. 307--392, 2019.

\bibitem{goodfellow2014}
I.~Goodfellow, J.~Pouget-Abadie, M.~Mirza, B.~Xu, D.~Warde-Farley, S.~Ozair,
  A.~Courville, and Y.~Bengio, ``Generative adversarial nets,'' in
  \emph{Advances in Neural Information Processing Systems}, 2014, pp.
  2672--2680.

\bibitem{xu2018}
H.~Xu, W.~Chen, N.~Zhao, Z.~Li, J.~Bu, Z.~Li, Y.~Liu, Y.~Zhao, D.~Pei, Y.~Feng
  \emph{et~al.}, ``Unsupervised anomaly detection via variational auto-encoder
  for seasonal {KPI}s in web applications,'' in \emph{World Wide Web
  Conference}, 2018, pp. 187--196.

\bibitem{nalisnick2019}
E.~Nalisnick, A.~Matsukawa, Y.~W. Teh, D.~Gorur, and B.~Lakshminarayanan, ``Do
  deep generative models know what they don't know?'' in \emph{International
  Conference on Learning Representations}, 2019.

\bibitem{an2015}
J.~An and S.~Cho, ``Variational autoencoder based anomaly detection using
  reconstruction probability,'' \emph{Special Lecture on IE}, vol.~2, pp.
  1--18, 2015.

\bibitem{salimans16}
T.~Salimans, I.~Goodfellow, W.~Zaremba, V.~Cheung, A.~Radford, X.~Chen, and
  X.~Chen, ``Improved techniques for training {GAN}s,'' in \emph{Advances in
  Neural Information Processing Systems}, 2016, pp. 2234--2242.

\bibitem{arjovsky17}
M.~Arjovsky, S.~Chintala, and L.~Bottou, ``{W}asserstein generative adversarial
  networks,'' in \emph{International Conference on Machine Learning}, vol.~70,
  2017, pp. 214--223.

\bibitem{gulrajani17}
I.~Gulrajani, F.~Ahmed, M.~Arjovsky, V.~Dumoulin, and A.~C. Courville,
  ``Improved training of wasserstein {GAN}s,'' in \emph{Advances in Neural
  Information Processing Systems}, 2017, pp. 5767--5777.

\bibitem{dinh2014nice}
L.~Dinh, D.~Krueger, and Y.~Bengio, ``{NICE:} non-linear independent components
  estimation,'' in \emph{International Conference on Learning Representations},
  2015.

\bibitem{papamakarios19}
G.~Papamakarios, E.~Nalisnick, D.~J. Rezende, S.~Mohamed, and
  B.~Lakshminarayanan, ``Normalizing flows for probabilistic modeling and
  inference,'' \emph{arXiv preprint arXiv:1912.02762}, 2019.

\bibitem{kobyzev2020}
I.~Kobyzev, S.~Prince, and M.~Brubaker, ``Normalizing flows: An introduction
  and review of current methods,'' \emph{{IEEE} Transactions on Pattern
  Analysis and Machine Intelligence}, 2020.

\bibitem{dinh17}
L.~Dinh, J.~Sohl{-}Dickstein, and S.~Bengio, ``Density estimation using real
  {NVP},'' in \emph{International Conference on Learning Representations},
  2017.

\bibitem{huang18}
C.-W. Huang, D.~Krueger, A.~Lacoste, and A.~Courville, ``Neural autoregressive
  flows,'' in \emph{International Conference on Machine Learning}, vol.~80,
  2018, pp. 2078--2087.

\bibitem{noe2019boltzmann}
F.~No{\'e}, S.~Olsson, J.~K{\"o}hler, and H.~Wu, ``Boltzmann generators:
  Sampling equilibrium states of many-body systems with deep learning,''
  \emph{Science}, vol. 365, no. 6457, p. eaaw1147, 2019.

\bibitem{nachman20}
B.~Nachman and D.~Shih, ``Anomaly detection with density estimation,''
  \emph{Physical Review D}, vol. 101, p. 075042, Apr 2020.

\bibitem{wellhausen20}
L.~{Wellhausen}, R.~{Ranftl}, and M.~{Hutter}, ``Safe robot navigation via
  multi-modal anomaly detection,'' \emph{Robotics and Automation Letters},
  vol.~5, no.~2, pp. 1326--1333, 2020.

\bibitem{kirichenko2020}
P.~Kirichenko, P.~Izmailov, and A.~G. Wilson, ``Why normalizing flows fail to
  detect out-of-distribution data,'' in \emph{Advances in Neural Information
  Processing Systems}, 2020.

\bibitem{mirza14}
M.~Mirza and S.~Osindero, ``Conditional generative adversarial nets,''
  \emph{arXiv preprint arXiv:1411.1784}, 2014.

\bibitem{suh16}
S.~{Suh}, D.~H. {Chae}, H.~{Kang}, and S.~{Choi}, ``Echo-state conditional
  variational autoencoder for anomaly detection,'' in \emph{International Joint
  Conference on Neural Networks}, 2016, pp. 1015--1022.

\bibitem{abdelhamed19}
A.~Abdelhamed, M.~A. Brubaker, and M.~S. Brown, ``Noise flow: Noise modeling
  with conditional normalizing flows,'' in \emph{International Conference on
  Computer Vision}, 2019, pp. 3165--3173.

\bibitem{li2019}
D.~Li, D.~Chen, B.~Jin, L.~Shi, J.~Goh, and S.-K. Ng, ``{MAD-GAN}: Multivariate
  anomaly detection for time series data with generative adversarial
  networks,'' in \emph{International Conference on Artificial Neural Networks},
  2019, pp. 703--716.

\bibitem{bowman2016}
S.~R. Bowman, L.~Vilnis, O.~Vinyals, A.~Dai, R.~Jozefowicz, and S.~Bengio,
  ``Generating sentences from a continuous space,'' in \emph{The {SIGNLL}
  Conference on Computational Natural Language Learning}, Aug. 2016, pp.
  10--21.

\bibitem{chen18}
L.~Chen, S.~Dai, C.~Tao, H.~Zhang, Z.~Gan, D.~Shen, Y.~Zhang, G.~Wang,
  R.~Zhang, and L.~Carin, ``Adversarial text generation via feature-mover's
  distance,'' in \emph{Advances in Neural Information Processing Systems},
  2018, pp. 4666--4677.

\bibitem{jin18}
W.~Jin, R.~Barzilay, and T.~Jaakkola, ``Junction tree variational autoencoder
  for molecular graph generation,'' in \emph{International Conference on
  Machine Learning}, vol.~80, 2018, pp. 2323--2332.

\bibitem{bojchevski18}
A.~Bojchevski, O.~Shchur, D.~Z{\"u}gner, and S.~G{\"u}nnemann, ``{N}et{GAN}:
  Generating graphs via random walks,'' in \emph{International Conference on
  Machine Learning}, vol.~80, 2018, pp. 610--619.

\bibitem{liao19}
R.~Liao, Y.~Li, Y.~Song, S.~Wang, W.~Hamilton, D.~K. Duvenaud, R.~Urtasun, and
  R.~Zemel, ``Efficient graph generation with graph recurrent attention
  networks,'' in \emph{Advances in Neural Information Processing Systems},
  2019, pp. 4255--4265.

\bibitem{vapnik1998}
V.~Vapnik, \emph{Statistical Learning Theory}.\hskip 1em plus 0.5em minus
  0.4em\relax Wiley, 1998.

\bibitem{moya1993}
M.~M. Moya, M.~W. Koch, and L.~D. Hostetler, ``One-class classifier networks
  for target recognition applications,'' in \emph{Proceedings World Congress on
  Neural Networks}, 1993, pp. 797--801.

\bibitem{moya1996}
M.~M. Moya and D.~R. Hush, ``Network constraints and multi-objective
  optimization for one-class classification,'' \emph{Neural Networks}, vol.~9,
  no.~3, pp. 463--474, 1996.

\bibitem{khan2014}
S.~S. Khan and M.~G. Madden, ``One-class classification: taxonomy of study and
  review of techniques,'' \emph{The Knowledge Engineering Review}, vol.~29,
  no.~3, pp. 345--374, 2014.

\bibitem{minter1975}
T.~Minter, ``Single-class classification,'' in \emph{LARS Symposia}, 1975,
  p.~54.

\bibitem{elYaniv2007}
R.~El-Yaniv and M.~Nisenson, ``Optimal single-class classification
  strategies,'' in \emph{Advances in Neural Information Processing Systems},
  2007, pp. 377--384.

\bibitem{rousseeuw1985}
P.~J. Rousseeuw, ``Multivariate estimation with high breakdown point,''
  \emph{Mathematical Statistics and Applications}, vol.~8, pp. 283--297, 1985.

\bibitem{rousseeuw1999}
P.~J. Rousseeuw and K.~V. Driessen, ``A fast algorithm for the minimum
  covariance determinant estimator,'' \emph{Technometrics}, vol.~41, no.~3, pp.
  212--223, 1999.

\bibitem{munoz2006}
A.~{Mu{\~n}oz} and J.~M. {Moguerza}, ``Estimation of high-density regions using
  one-class neighbor machines,'' \emph{{IEEE} Transactions on Pattern Analysis
  and Machine Intelligence}, vol.~28, no.~3, pp. 476--480, 2006.

\bibitem{scholkopf1999input}
B.~Sch\"olkopf, S.~Mika, C.~J. Burges, P.~Knirsch, K.-R. M\"uller, G.~R\"atsch,
  and A.~J. Smola, ``Input space versus feature space in kernel-based
  methods,'' \emph{IEEE Transactions on Neural Networks}, vol.~10, no.~5, pp.
  1000--1017, 1999.

\bibitem{manevitz2001}
L.~M. Manevitz and M.~Yousef, ``One-class {SVM}s for document classification,''
  \emph{Journal of Machine Learning Research}, vol.~2, no. Dec, pp. 139--154,
  2001.

\bibitem{vert2006}
R.~Vert and J.-P. Vert, ``Consistency and convergence rates of one-class {SVM}s
  and related algorithms,'' \emph{Journal of Machine Learning Research},
  vol.~7, no. May, pp. 817--854, 2006.

\bibitem{lee2007a}
G.~Lee and C.~D. Scott, ``The one class support vector machine solution path,''
  in \emph{International Conference on Acoustics, Speech, and Signal
  Processing}, vol.~2, 2007, pp. 521--524.

\bibitem{sjostrand2006}
K.~Sj{\"o}strand and R.~Larsen, ``The entire regularization path for the
  support vector domain description,'' in \emph{International Conference on
  Medical Image Computing and Computer-Assisted Intervention}, 2006, pp.
  241--248.

\bibitem{lee2009}
G.~Lee and C.~Scott, ``Nested support vector machines,'' \emph{IEEE
  Transactions on Signal Processing}, vol.~58, no.~3, pp. 1648--1660, 2009.

\bibitem{glazer2013}
A.~Glazer, M.~Lindenbaum, and S.~Markovitch, ``q-{OCSVM}: A q-quantile
  estimator for high-dimensional distributions,'' in \emph{Advances in Neural
  Information Processing Systems}, 2013, pp. 503--511.

\bibitem{gornitz2017}
N.~G{\"o}rnitz, L.~A. Lima, K.-R. M{\"u}ller, M.~Kloft, and S.~Nakajima,
  ``Support vector data descriptions and $k$-means clustering: One class?''
  \emph{IEEE Transactions on Neural Networks and Learning Systems}, vol.~29,
  no.~9, pp. 3994--4006, 2017.

\bibitem{das2010}
S.~Das, B.~L. Matthews, A.~N. Srivastava, and N.~C. Oza, ``Multiple kernel
  learning for heterogeneous anomaly detection: Algorithm and aviation safety
  case study,'' in \emph{International Conference on Knowledge Discovery \&
  Data Mining}, 2010, pp. 47--56.

\bibitem{gautam2019}
C.~Gautam, R.~Balaji, K.~Sudharsan, A.~Tiwari, and K.~Ahuja, ``Localized
  multiple kernel learning for anomaly detection: One-class classification,''
  \emph{Knowledge-Based Systems}, vol. 165, pp. 241--252, 2019.

\bibitem{ratsch2002}
G.~R\"{a}tsch, S.~Mika, B.~Sch\"{o}lkopf, and K.-R. M{\"{u}}ller,
  ``Constructing boosting algorithms from {SVM}s: An application to one-class
  classification,'' \emph{{IEEE} Transactions on Pattern Analysis and Machine
  Intelligence}, vol.~24, no.~9, pp. 1184--1199, Sep. 2002.

\bibitem{roth2005}
V.~Roth, ``Outlier detection with one-class kernel fisher discriminants,'' in
  \emph{Advances in Neural Information Processing Systems}, 2005, pp.
  1169--1176.

\bibitem{roth2006}
------, ``Kernel fisher discriminants for outlier detection,'' \emph{Neural
  Computation}, vol.~18, no.~4, pp. 942--960, 2006.

\bibitem{dufrenois2014}
F.~Dufrenois, ``A one-class kernel fisher criterion for outlier detection,''
  \emph{IEEE Transactions on Neural Networks and Learning Systems}, vol.~26,
  no.~5, pp. 982--994, 2014.

\bibitem{ghasemi2012}
A.~Ghasemi, H.~R. Rabiee, M.~T. Manzuri, and M.~H. Rohban, ``A bayesian
  approach to the data description problem,'' in \emph{{AAAI} Conference on
  Artificial Intelligence}, 2012, pp. 907--913.

\bibitem{stolpe2013}
M.~Stolpe, K.~Bhaduri, K.~Das, and K.~Morik, ``Anomaly detection in vertically
  partitioned data by distributed {C}ore {V}ector {M}achines,'' in
  \emph{European Conference on Machine Learning and Principles and Practice of
  Knowledge Discovery in Databases}, 2013, pp. 321--336.

\bibitem{jiang2019}
H.~Jiang, H.~Wang, W.~Hu, D.~Kakde, and A.~Chaudhuri, ``Fast incremental {SVDD}
  learning algorithm with the {G}aussian kernel,'' in \emph{{AAAI} Conference
  on Artificial Intelligence}, vol.~33, 2019, pp. 3991--3998.

\bibitem{liu2014}
W.~Liu, G.~Hua, and J.~R. Smith, ``Unsupervised one-class learning for
  automatic outlier removal,'' in \emph{{IEEE} Conference on Computer Vision
  and Pattern Recognition}, 2014, pp. 3826--3833.

\bibitem{wu2020}
P.~Wu, J.~Liu, and F.~Shen, ``A deep one-class neural network for anomalous
  event detection in complex scenes,'' \emph{IEEE Transactions on Neural
  Networks and Learning Systems}, vol.~31, no.~7, pp. 2609--2622, 2020.

\bibitem{chalapathy2018b}
R.~Chalapathy, A.~K. Menon, and S.~Chawla, ``Anomaly detection using one-class
  neural networks,'' \emph{arXiv preprint arXiv:1802.06360}, 2018.

\bibitem{lecun2012}
Y.~LeCun, L.~Bottou, G.~B. Orr, and K.-R. M{\"u}ller, ``Efficient {BackProp},''
  in \emph{Neural Networks: Tricks of the Trade - Second Edition}, ser. Lecture
  Notes in Computer Science, G.~Montavon, G.~B. Orr, and K.-R. M{\"u}ller,
  Eds.\hskip 1em plus 0.5em minus 0.4em\relax Springer, Berlin, Heidelberg,
  2012, vol. 7700, pp. 9--48.

\bibitem{kingma2015}
D.~P. Kingma and J.~Ba, ``Adam: {A} method for stochastic optimization,'' in
  \emph{International Conference on Learning Representations}, 2015.

\bibitem{goh2017}
\BIBentryALTinterwordspacing
G.~Goh, ``Why momentum really works,'' \emph{Distill}, 2017. [Online].
  Available: \url{http://distill.pub/2017/momentum}
\BIBentrySTDinterwordspacing

\bibitem{ruff2020b}
L.~Ruff, R.~A. Vandermeulen, B.~J. Franks, K.-R. M{\"u}ller, and M.~Kloft,
  ``Rethinking assumptions in deep anomaly detection,'' \emph{arXiv preprint
  arXiv:2006.00339}, 2020.

\bibitem{goyal2020}
S.~Goyal, A.~Raghunathan, M.~Jain, H.~V. Simhadri, and P.~Jain, ``{DROCC}: Deep
  robust one-class classification,'' in \emph{International Conference on
  Machine Learning}, 2020, pp. 11\,335--11\,345.

\bibitem{kauffmann2020a}
J.~Kauffmann, L.~Ruff, G.~Montavon, and K.-R. M{\"{u}}ller, ``The {C}lever
  {H}ans effect in anomaly detection,'' \emph{arXiv preprint arXiv:2006.10609},
  2020.

\bibitem{chong2020}
P.~Chong, L.~Ruff, M.~Kloft, and A.~Binder, ``Simple and effective prevention
  of mode collapse in deep one-class classification,'' in \emph{International
  Joint Conference on Neural Networks}, 2020, pp. 1--9.

\bibitem{pang2019}
G.~Pang, C.~Shen, and A.~van~den Hengel, ``Deep anomaly detection with
  deviation networks,'' in \emph{International Conference on Knowledge
  Discovery \& Data Mining}, 2019, pp. 353--362.

\bibitem{liznerski2020}
P.~Liznerski, L.~Ruff, R.~A. Vandermeulen, B.~J. Franks, M.~Kloft, and K.-R.
  M{\"u}ller, ``Explainable deep one-class classification,'' in
  \emph{International Conference on Learning Representations}, 2021.

\bibitem{shen2020}
L.~Shen, Z.~Li, and J.~Kwok, ``Timeseries anomaly detection using temporal
  hierarchical one-class network,'' in \emph{Advances in Neural Information
  Processing Systems}, 2020.

\bibitem{sabokrou2020}
M.~Sabokrou, M.~Fathy, G.~Zhao, and E.~Adeli, ``Deep end-to-end one-class
  classifier,'' \emph{IEEE Transactions on Neural Networks and Learning
  Systems}, pp. 1--10, 2020.

\bibitem{hastie2009}
T.~Hastie, R.~Tibshirani, and J.~Friedman, \emph{The Elements of Statistical
  Learning: Data Mining, Inference, and Prediction}, 2nd~ed.\hskip 1em plus
  0.5em minus 0.4em\relax Springer New York, 2009.

\bibitem{steinbuss2020}
G.~Steinbuss and K.~B{\"o}hm, ``Generating artificial outliers in the absence
  of genuine ones -- a survey,'' \emph{arXiv preprint arXiv:2006.03646}, 2020.

\bibitem{theiler2003}
J.~P. Theiler and D.~M. Cai, ``Resampling approach for anomaly detection in
  multispectral images,'' in \emph{Algorithms and Technologies for
  Multispectral, Hyperspectral, and Ultraspectral Imagery IX}, vol. 5093.\hskip
  1em plus 0.5em minus 0.4em\relax International Society for Optics and
  Photonics, 2003, pp. 230--240.

\bibitem{davenport2006}
M.~A. {Davenport}, R.~G. {Baraniuk}, and C.~D. {Scott}, ``Learning minimum
  volume sets with support vector machines,'' in \emph{IEEE Signal Processing
  Society Workshop on Machine Learning for Signal Processing}, 2006, pp.
  301--306.

\bibitem{fan2004}
W.~Fan, M.~Miller, S.~Stolfo, W.~Lee, and P.~Chan, ``Using artificial anomalies
  to detect unknown and known network intrusions,'' \emph{Knowledge and
  Information Systems}, vol.~6, no.~5, pp. 507--527, 2004.

\bibitem{cheema2016}
P.~Cheema, N.~L.~D. Khoa, M.~Makki~Alamdari, W.~Liu, Y.~Wang, F.~Chen, and
  P.~Runcie, ``On structural health monitoring using tensor analysis and
  support vector machine with artificial negative data,'' in \emph{Proceedings
  of the 25th {ACM} International on Conference on Information and Knowledge
  Management}, 2016, pp. 1813--1822.

\bibitem{abe2006}
N.~Abe, B.~Zadrozny, and J.~Langford, ``Outlier detection by active learning,''
  in \emph{International Conference on Knowledge Discovery \& Data Mining},
  2006, pp. 504--509.

\bibitem{stokes2008}
J.~W. Stokes, J.~C. Platt, J.~Kravis, and M.~Shilman, ``{ALADIN}: Active
  learning of anomalies to detect intrusions,'' Microsoft Research, Technical
  Report MSR-TR-2008-24, 2008.

\bibitem{gornitz2009}
N.~G{\"o}rnitz, M.~Kloft, and U.~Brefeld, ``Active and semi-supervised data
  domain description,'' in \emph{European Conference on Machine Learning and
  Principles and Practice of Knowledge Discovery in Databases}, 2009, pp.
  407--422.

\bibitem{pelleg2005}
D.~Pelleg and A.~W. Moore, ``Active learning for anomaly and rare-category
  detection,'' in \emph{Advances in Neural Information Processing Systems},
  2005, pp. 1073--1080.

\bibitem{du2019}
M.~Du, Z.~Chen, C.~Liu, R.~Oak, and D.~Song, ``Lifelong anomaly detection
  through unlearning,'' in \emph{Proceedings of the {ACM} {SIGSAC} Conference
  on Computer and Communications Security}, 2019, pp. 1283--1297.

\bibitem{japkowicz1995}
N.~Japkowicz, C.~Myers, and M.~Gluck, ``A novelty detection approach to
  classification,'' in \emph{International Joint Conferences on Artificial
  Intelligence}, vol.~1, 1995, pp. 518--523.

\bibitem{hawkins2002}
S.~Hawkins, H.~He, G.~Williams, and R.~Baxter, ``Outlier detection using
  replicator neural networks,'' in \emph{International Conference on Data
  Warehousing and Knowledge Discovery}, vol. 2454, 2002, pp. 170--180.

\bibitem{lee2007}
J.~A. Lee and M.~Verleysen, \emph{Nonlinear Dimensionality Reduction}.\hskip
  1em plus 0.5em minus 0.4em\relax Springer New York, 2007.

\bibitem{pless2009}
R.~Pless and R.~Souvenir, ``A survey of manifold learning for images,''
  \emph{IPSJ Transactions on Computer Vision and Applications}, vol.~1, pp.
  83--94, 2009.

\bibitem{kohonen1990self}
T.~Kohonen, ``The self-organizing map,'' \emph{Proceedings of the {IEEE}},
  vol.~78, no.~9, pp. 1464--1480, Sep. 1990.

\bibitem{kohonen2001}
------, \emph{Self-Organizing Maps}, 3rd~ed.\hskip 1em plus 0.5em minus
  0.4em\relax Springer, Berlin, Heidelberg, 2001.

\bibitem{vanderMaaten2009}
L.~van~der Maaten, E.~Postma, and J.~van~den Herik, ``Dimensionality reduction:
  A comparative review,'' Tilburg centre for Creative Computing (TiCC), Tilburg
  University, Technical Report TiCC-TR 2009-005, 2009.

\bibitem{schmidhuber1992}
J.~Schmidhuber, ``Learning factorial codes by predictability minimization,''
  \emph{Neural Computation}, vol.~4, no.~6, pp. 863--879, 1992.

\bibitem{tschannen2018}
M.~Tschannen, O.~Bachem, and M.~Lucic, ``Recent advances in autoencoder-based
  representation learning,'' in \emph{3rd Workshop on Bayesian Deep Learning
  (NeurIPS 2018)}, 2018.

\bibitem{linde1980algorithm}
Y.~Linde, A.~Buzo, and R.~Gray, ``An algorithm for vector quantizer design,''
  \emph{IEEE Transactions on Communications}, vol.~28, no.~1, pp. 84--95, Jan.
  1980.

\bibitem{gersho1992}
A.~Gersho and R.~M. Gray, \emph{Vector Quantization and Signal Compression},
  ser. The Springer International Series in Engineering and Computer
  Science.\hskip 1em plus 0.5em minus 0.4em\relax Springer, Boston, MA, 1992,
  vol. 159.

\bibitem{tipping1999}
M.~E. Tipping and C.~M. Bishop, ``Probabilistic principal component analysis,''
  \emph{Journal of the Royal Statistical Society. Series B (Statistical
  Methodology)}, vol.~61, no.~3, pp. 611--622, 1999.

\bibitem{bishop1999}
C.~M. Bishop, ``Bayesian {PCA},'' in \emph{Advances in Neural Information
  Processing Systems}, 1999, pp. 382--388.

\bibitem{jolliffe2002}
I.~T. Jolliffe, \emph{Principal Component Analysis}, 2nd~ed., ser. Springer
  Series in Statistics.\hskip 1em plus 0.5em minus 0.4em\relax New York, NY:
  Springer, 2002.

\bibitem{hawkins1974}
D.~M. Hawkins, ``The detection of errors in multivariate data using principal
  components,'' \emph{Journal of the American Statistical Association},
  vol.~69, no. 346, pp. 340--344, 1974.

\bibitem{jackson1979}
J.~E. Jackson and G.~S. Mudholkar, ``Control procedures for residuals
  associated with principal component analysis,'' \emph{Technometrics},
  vol.~21, no.~3, pp. 341--349, Aug. 1979.

\bibitem{parra1996}
L.~Parra, G.~Deco, and S.~Miesbach, ``Statistical independence and novelty
  detection with information preserving nonlinear maps,'' \emph{Neural
  Computation}, vol.~8, no.~2, pp. 260--269, 1996.

\bibitem{shyu2003}
M.-L. Shyu, S.-C. Chen, K.~Sarinnapakorn, and L.~Chang, ``A novel anomaly
  detection scheme based on principal component classifier,'' in \emph{{IEEE}
  International Conference on Data Mining}, 2003, pp. 353--365.

\bibitem{huang2007}
L.~Huang, X.~Nguyen, M.~Garofalakis, M.~I. Jordan, A.~Joseph, and N.~Taft,
  ``In-network {PCA} and anomaly detection,'' in \emph{Advances in Neural
  Information Processing Systems}, 2007, pp. 617--624.

\bibitem{sharan2018}
V.~Sharan, P.~Gopalan, and U.~Wieder, ``Efficient anomaly detection via matrix
  sketching,'' in \emph{Advances in Neural Information Processing Systems},
  2018, pp. 8069--8080.

\bibitem{ham2004}
J.~Ham, D.~D. Lee, S.~Mika, and B.~Sch{\"o}lkopf, ``A kernel view of the
  dimensionality reduction of manifolds,'' in \emph{International Conference on
  Machine Learning}, 2004, p.~47.

\bibitem{kwak2008}
N.~Kwak, ``Principal component analysis based on {L1}-norm maximization,''
  \emph{{IEEE} Transactions on Pattern Analysis and Machine Intelligence},
  vol.~30, no.~9, pp. 1672--1680, 2008.

\bibitem{nguyen2009}
M.~H. Nguyen and F.~Torre, ``Robust kernel principal component analysis,'' in
  \emph{Advances in Neural Information Processing Systems}, 2009, pp.
  1185--1192.

\bibitem{candes2011}
E.~J. Cand\`{e}s, X.~Li, Y.~Ma, and J.~Wright, ``Robust principal component
  analysis?'' \emph{Journal of the ACM}, vol.~58, no.~3, pp. 1--37, 2011.

\bibitem{xiao2013}
Y.~Xiao, H.~Wang, W.~Xu, and J.~Zhou, ``{L1} norm based {KPCA} for novelty
  detection,'' \emph{Pattern Recognition}, vol.~46, no.~1, pp. 389--396, 2013.

\bibitem{oja1982}
E.~Oja, ``A simplified neuron model as a principal component analyzer,''
  \emph{Journal of Mathematical Biology}, vol.~15, no.~3, pp. 267--273, 1982.

\bibitem{rumelhart1986}
D.~E. Rumelhart, G.~E. Hinton, and R.~J. Williams, ``Learning internal
  representations by error propagation,'' in \emph{Parallel Distributed
  Processing -- Explorations in the Microstructure of Cognition}.\hskip 1em
  plus 0.5em minus 0.4em\relax MIT Press, 1986, ch.~8, pp. 318--362.

\bibitem{ballard1987}
D.~H. Ballard, ``Modular learning in neural networks,'' in \emph{{AAAI}
  Conference on Artificial Intelligence}, 1987, pp. 279--284.

\bibitem{hinton1989}
G.~E. Hinton, ``Connectionist learning procedures,'' \emph{Artificial
  Intelligence}, vol.~40, no.~1, pp. 185--234, 1989.

\bibitem{kramer1991}
M.~A. Kramer, ``Nonlinear principal component analysis using autoassociative
  neural networks,'' \emph{AIChE journal}, vol.~37, no.~2, pp. 233--243, 1991.

\bibitem{hinton2006}
G.~E. Hinton and R.~R. Salakhutdinov, ``Reducing the dimensionality of data
  with neural networks,'' \emph{Science}, vol. 313, no. 5786, pp. 504--507,
  Jul. 2006.

\bibitem{baldi1989}
P.~Baldi and K.~Hornik, ``Neural networks and principal component analysis:
  Learning from examples without local minima,'' \emph{Neural Networks},
  vol.~2, no.~1, pp. 53--58, 1989.

\bibitem{oja1992principal}
E.~Oja, ``Principal components, minor components, and linear neural networks,''
  \emph{Neural Networks}, vol.~5, no.~6, pp. 927--935, 1992.

\bibitem{olshausen1996}
B.~A. Olshausen and D.~J. Field, ``Emergence of simple-cell receptive field
  properties by learning a sparse code for natural images,'' \emph{Nature},
  vol. 381, no. 6583, pp. 607--609, 1996.

\bibitem{olshausen1997}
------, ``Sparse coding with an overcomplete basis set: A strategy employed by
  {V1}?'' \emph{Vision Research}, vol.~37, no.~23, pp. 3311--3325, 1997.

\bibitem{lewicki2000}
M.~S. Lewicki and T.~J. Sejnowski, ``Learning overcomplete representations,''
  \emph{Neural Computation}, vol.~12, no.~2, pp. 337--365, 2000.

\bibitem{lee2007b}
H.~Lee, A.~Battle, R.~Raina, and A.~Y. Ng, ``Efficient sparse coding
  algorithms,'' in \emph{Advances in Neural Information Processing Systems},
  2007, pp. 801--808.

\bibitem{makhzani2014}
A.~Makhzani and B.~Frey, ``$k$-sparse autoencoders,'' in \emph{International
  Conference on Learning Representations}, 2014.

\bibitem{zeng2018}
N.~Zeng, H.~Zhang, B.~Song, W.~Liu, Y.~Li, and A.~M. Dobaie, ``Facial
  expression recognition via learning deep sparse autoencoders,''
  \emph{Neurocomputing}, vol. 273, pp. 643--649, 2018.

\bibitem{arpit2016}
D.~Arpit, Y.~Zhou, H.~Ngo, and V.~Govindaraju, ``Why regularized auto-encoders
  learn sparse representation?'' in \emph{International Conference on Machine
  Learning}, vol.~48, 2016, pp. 136--144.

\bibitem{vincent2008}
P.~Vincent, H.~Larochelle, Y.~Bengio, and P.-A. Manzagol, ``Extracting and
  composing robust features with denoising autoencoders,'' in
  \emph{International Conference on Machine Learning}, 2008, pp. 1096--1103.

\bibitem{vincent2010}
P.~Vincent, H.~Larochelle, I.~Lajoie, Y.~Bengio, P.-A. Manzagol, and L.~Bottou,
  ``Stacked denoising autoencoders: Learning useful representations in a deep
  network with a local denoising criterion,'' \emph{Journal of Machine Learning
  Research}, vol.~11, no. Dec, pp. 3371--3408, 2010.

\bibitem{rifai2011}
S.~Rifai, P.~Vincent, X.~Muller, X.~Glorot, and Y.~Bengio, ``Contractive
  auto-encoders: Explicit invariance during feature extraction,'' in
  \emph{International Conference on Machine Learning}, 2011, pp. 833--840.

\bibitem{you2019}
S.~You, K.~C. Tezcan, X.~Chen, and E.~Konukoglu, ``Unsupervised lesion
  detection via image restoration with a normative prior,'' in
  \emph{International Conference on Medical Imaging with Deep Learning}, 2019,
  pp. 540--556.

\bibitem{park2018}
D.~Park, Y.~Hoshi, and C.~C. Kemp, ``A multimodal anomaly detector for
  robot-assisted feeding using an {LSTM}-based variational autoencoder,''
  \emph{Robotics and Automation Letters}, vol.~3, no.~3, pp. 1544--1551, 2018.

\bibitem{makhzani2015}
A.~Makhzani, J.~Shlens, N.~Jaitly, I.~Goodfellow, and B.~Frey, ``Adversarial
  autoencoders,'' \emph{arXiv preprint arXiv:1511.05644}, 2015.

\bibitem{malhotra2016}
P.~Malhotra, A.~Ramakrishnan, G.~Anand, L.~Vig, P.~Agarwal, and G.~Shroff,
  ``{LSTM}-based encoder-decoder for multi-sensor anomaly detection,''
  \emph{arXiv preprint arXiv:1607.00148}, 2016.

\bibitem{kieu2019}
T.~Kieu, B.~Yang, C.~Guo, and C.~S. Jensen, ``Outlier detection for time series
  with recurrent autoencoder ensembles,'' in \emph{International Joint
  Conferences on Artificial Intelligence}, 2019, pp. 2725--2732.

\bibitem{kwon2020}
G.~Kwon, M.~Prabhushankar, D.~Temel, and G.~AlRegib, ``Backpropagated gradient
  representations for anomaly detection,'' in \emph{European Conference on
  Computer Vision}, 2020, pp. 206--226.

\bibitem{hofer2019}
C.~D. Hofer, R.~Kwitt, M.~Dixit, and M.~Niethammer, ``Connectivity-optimized
  representation learning via persistent homology,'' in \emph{International
  Conference on Machine Learning}, vol.~97, 2019, pp. 2751--2760.

\bibitem{amarbayasgalan2018}
T.~Amarbayasgalan, B.~Jargalsaikhan, and K.~H. Ryu, ``Unsupervised novelty
  detection using deep autoencoders with density based clustering,''
  \emph{Applied Sciences}, vol.~8, no.~9, p. 1468, 2018.

\bibitem{sarafijanovic2019}
N.~Sarafijanovic-Djukic and J.~Davis, ``Fast distance-based anomaly detection
  in images using an inception-like autoencoder,'' in \emph{International
  Conference on Discovery Science}.\hskip 1em plus 0.5em minus 0.4em\relax
  Springer, Cham, 2019, pp. 493--508.

\bibitem{jain2010}
A.~K. Jain, ``Data clustering: 50 years beyond k-means,'' \emph{Pattern
  Recognition Letters}, vol.~31, no.~8, pp. 651--666, 2010.

\bibitem{voronoi1908a}
G.~Voronoi, ``Nouvelles applications des param{\`e}tres continus {\`a} la
  th{\'e}orie des formes quadratiques. {P}remier m{\'e}moire. {S}ur quelques
  propri{\'e}t{\'e}s des formes quadratiques positives parfaites.''
  \emph{Journal f{\"u}r die Reine und Angewandte Mathematik}, vol. 1908, no.
  133, pp. 97--178, 1908.

\bibitem{voronoi1908b}
------, ``Nouvelles applications des param{\`e}tres continus {\`a} la
  th{\'e}orie des formes quadratiques. {D}euxi{\`e}me m{\'e}moire. {R}echerches
  sur les parall{\'e}llo{\`e}dres primitifs.'' \emph{Journal f{\"u}r die Reine
  und Angewandte Mathematik}, vol. 1908, no. 134, pp. 198--287, 1908.

\bibitem{dhillon2004}
I.~S. Dhillon, Y.~Guan, and B.~Kulis, ``Kernel k-means, spectral clustering and
  normalized cuts,'' in \emph{International Conference on Knowledge Discovery
  \& Data Mining}, 2004, pp. 551--556.

\bibitem{xie2016}
J.~Xie, R.~Girshick, and A.~Farhadi, ``Unsupervised deep embedding for
  clustering analysis,'' in \emph{International Conference on Machine
  Learning}, vol.~48, 2016, pp. 478--487.

\bibitem{van2017}
A.~Van Den~Oord, O.~Vinyals \emph{et~al.}, ``Neural discrete representation
  learning,'' in \emph{Advances in Neural Information Processing Systems},
  2017, pp. 6306--6315.

\bibitem{razavi2019}
A.~Razavi, A.~van~den Oord, and O.~Vinyals, ``Generating diverse high-fidelity
  images with {VQ-VAE}-2,'' in \emph{Advances in Neural Information Processing
  Systems}, 2019, pp. 14\,866--14\,876.

\bibitem{kampffmeyer2019}
M.~Kampffmeyer, S.~L{\o}kse, F.~M. Bianchi, L.~Livi, A.-B. Salberg, and
  R.~Jenssen, ``Deep divergence-based approach to clustering,'' \emph{Neural
  Networks}, vol. 113, pp. 91--101, 2019.

\bibitem{yang2017}
B.~Yang, X.~Fu, N.~D. Sidiropoulos, and M.~Hong, ``Towards k-means-friendly
  spaces: Simultaneous deep learning and clustering,'' in \emph{International
  Conference on Machine Learning}, vol.~70, 2017, pp. 3861--3870.

\bibitem{caron2018}
M.~Caron, P.~Bojanowski, A.~Joulin, and M.~Douze, ``Deep clustering for
  unsupervised learning of visual features,'' in \emph{European Conference on
  Computer Vision}, 2018, pp. 132--149.

\bibitem{bojanowski2017b}
P.~Bojanowski and A.~Joulin, ``Unsupervised learning by predicting noise,'' in
  \emph{International Conference on Machine Learning}, vol.~70, 2017, pp.
  517--526.

\bibitem{bishop2006}
C.~M. Bishop, \emph{Pattern Recognition and Machine Learning}.\hskip 1em plus
  0.5em minus 0.4em\relax Springer New York, 2006.

\bibitem{murphy2012}
K.~P. Murphy, \emph{Machine Learning: A Probabilistic Perspective}.\hskip 1em
  plus 0.5em minus 0.4em\relax MIT press, 2012.

\bibitem{theodoridis2020}
S.~Theodoridis, \emph{Machine Learning: A Bayesian and Optimization
  Perspective}, 2nd~ed.\hskip 1em plus 0.5em minus 0.4em\relax Academic Press,
  2020.

\bibitem{mackay1992}
D.~J.~C. MacKay, ``A practical bayesian framework for backpropagation
  networks,'' \emph{Neural Computation}, vol.~4, no.~3, pp. 448--472, 1992.

\bibitem{blundell2015}
C.~Blundell, J.~Cornebise, K.~Kavukcuoglu, and D.~Wierstra, ``Weight
  uncertainty in neural networks,'' in \emph{International Conference on
  Machine Learning}, vol.~37, 2015, pp. 1613--1622.

\bibitem{ruff2019b}
L.~Ruff, R.~A. Vandermeulen, N.~G{\"o}rnitz, A.~Binder, E.~M{\"u}ller, and
  M.~Kloft, ``Deep support vector data description for unsupervised and
  semi-supervised anomaly detection,'' in \emph{ICML 2019 Workshop on
  Uncertainty \& Robustness in Deep Learning}, 2019.

\bibitem{harmeling2006}
S.~Harmeling, G.~Dornhege, D.~M.~J. Tax, F.~Meinecke, and K.-R. M{\"u}ller,
  ``From outliers to prototypes: ordering data,'' \emph{Neurocomputing},
  vol.~69, no. 13-15, pp. 1608--1618, 2006.

\bibitem{zhao2009}
M.~Zhao and V.~Saligrama, ``Anomaly detection with score functions based on
  nearest neighbor graphs,'' in \emph{Advances in Neural Information Processing
  Systems}, 2009, pp. 2250--2258.

\bibitem{gu2019}
X.~Gu, L.~Akoglu, and A.~Rinaldo, ``Statistical analysis of nearest neighbor
  methods for anomaly detection,'' in \emph{Advances in Neural Information
  Processing Systems}, 2019, pp. 10\,923--10\,933.

\bibitem{juszczak2009}
P.~Juszczak, D.~M.~J. Tax, E.~Pe, R.~P.~W. Duin \emph{et~al.}, ``Minimum
  spanning tree based one-class classifier,'' \emph{Neurocomputing}, vol.~72,
  no.~7, pp. 1859--1869, 2009.

\bibitem{liu2008}
F.~T. Liu, K.~M. Ting, and Z.-H. Zhou, ``Isolation forest,'' in \emph{{IEEE}
  International Conference on Data Mining}, 2008, pp. 413--422.

\bibitem{Guha2016}
S.~Guha, N.~Mishra, G.~Roy, and O.~Schrijvers, ``Robust random cut forest based
  anomaly detection on streams,'' in \emph{International Conference on Machine
  Learning}, vol.~48, 2016, pp. 2712--2721.

\bibitem{bergman2020}
L.~Bergman, N.~Cohen, and Y.~Hoshen, ``Deep nearest neighbor anomaly
  detection,'' \emph{arXiv preprint arXiv:2002.10445}, 2020.

\bibitem{DBLP:conf/sp/GlasserL13}
J.~Glasser and B.~Lindauer, ``Bridging the gap: {A} pragmatic approach to
  generating insider threat data,'' in \emph{{IEEE} Symposium on Security and
  Privacy Workshops}, 2013, pp. 98--104.

\bibitem{mnist-c}
N.~Mu and J.~Gilmer, ``{MNIST-C:} {A} robustness benchmark for computer
  vision,'' \emph{arXiv preprint arXiv:1906.02337}, 2019.

\bibitem{hendrycks2019b}
D.~Hendrycks and T.~Dietterich, ``Benchmarking neural network robustness to
  common corruptions and perturbations,'' in \emph{International Conference on
  Learning Representations}, 2019.

\bibitem{laskov2004intrusion}
P.~Laskov, C.~Sch{\"a}fer, I.~Kotenko, and K.-R. M{\"u}ller, ``Intrusion
  detection in unlabeled data with quarter-sphere support vector machines,''
  \emph{{PIK} - Praxis der Informationsverarbeitung und Kommunikation},
  vol.~27, no.~4, pp. 228--236, 2004.

\bibitem{kloft2012security}
M.~Kloft and P.~Laskov, ``Security analysis of online centroid anomaly
  detection,'' \emph{Journal of Machine Learning Research}, vol.~13, no. 118,
  pp. 3681--3724, 2012.

\bibitem{emmott2013}
A.~F. Emmott, S.~Das, T.~Dietterich, A.~Fern, and W.-K. Wong, ``Systematic
  construction of anomaly detection benchmarks from real data,'' in \emph{KDD
  2013 Workshop on Outlier Detection and Description}, 2013, pp. 16--21.

\bibitem{Emmott2016}
A.~Emmott, S.~Das, T.~Dietterich, A.~Fern, and W.-K. Wong, ``A meta-analysis of
  the anomaly detection problem,'' \emph{arXiv preprint arXiv:1503.01158},
  2016.

\bibitem{hendrycks2019e}
D.~Hendrycks, K.~Zhao, S.~Basart, J.~Steinhardt, and D.~Song, ``Natural
  adversarial examples,'' \emph{arXiv preprint arXiv:1907.07174}, 2019.

\bibitem{huang2019pcb}
W.~Huang and P.~Wei, ``A {PCB} dataset for defects detection and
  classification,'' \emph{arXiv preprint arXiv:1901.08204}, 2019.

\bibitem{bejnordi2017}
B.~E. Bejnordi, M.~Veta, P.~J. Van~Diest, B.~Van~Ginneken, N.~Karssemeijer,
  G.~Litjens, J.~A. Van Der~Laak, M.~Hermsen, Q.~F. Manson, M.~Balkenhol
  \emph{et~al.}, ``Diagnostic assessment of deep learning algorithms for
  detection of lymph node metastases in women with breast cancer,''
  \emph{Journal of the American Medical Association}, vol. 318, no.~22, pp.
  2199--2210, 2017.

\bibitem{Wang_2017}
X.~Wang, Y.~Peng, L.~Lu, Z.~Lu, M.~Bagheri, and R.~M. Summers, ``{ChestX-ray8}:
  Hospital-scale chest x-ray database and benchmarks on weakly-supervised
  classification and localization of common thorax diseases,'' in \emph{{IEEE}
  Conference on Computer Vision and Pattern Recognition}, 2017, pp. 2097--2106.

\bibitem{mood}
\BIBentryALTinterwordspacing
D.~Zimmerer, J.~Petersen, G.~Köhler, P.~Jäger, P.~Full, T.~Roß, T.~Adler,
  A.~Reinke, L.~Maier-Hein, and K.~Maier-Hein, ``Medical out-of-distribution
  analysis challenge,'' Mar. 2020. [Online]. Available:
  \url{https://doi.org/10.5281/zenodo.3784230}
\BIBentrySTDinterwordspacing

\bibitem{pozzolo2018}
A.~{Dal Pozzolo}, G.~{Boracchi}, O.~{Caelen}, C.~{Alippi}, and G.~{Bontempi},
  ``Credit card fraud detection: A realistic modeling and a novel learning
  strategy,'' \emph{IEEE Transactions on Neural Networks and Learning Systems},
  vol.~29, no.~8, pp. 3784--3797, 2018.

\bibitem{ma2009}
J.~Ma, L.~K. Saul, S.~Savage, and G.~M. Voelker, ``Identifying suspicious
  {URL}s: An application of large-scale online learning,'' in
  \emph{International Conference on Machine Learning}, 2009, pp. 681--688.

\bibitem{moustafa2015}
N.~{Moustafa} and J.~{Slay}, ``{UNSW-NB15}: a comprehensive data set for
  network intrusion detection systems,'' in \emph{Military Communications and
  Information Systems Conference}, 2015, pp. 1--6.

\bibitem{ahmad2017}
S.~Ahmad, A.~Lavin, S.~Purdy, and Z.~Agha, ``Unsupervised real-time anomaly
  detection for streaming data,'' \emph{Neurocomputing}, vol. 262, pp.
  134--147, 2017.

\bibitem{laptev2015}
N.~Laptev, S.~Amizadeh, and I.~Flint, ``Generic and scalable framework for
  automated time-series anomaly detection,'' in \emph{International Conference
  on Knowledge Discovery \& Data Mining}, 2015, pp. 1939--1947.

\bibitem{campos2016}
G.~O. Campos, A.~Zimek, J.~Sander, R.~J. Campello, B.~Micenkov{\'a},
  E.~Schubert, I.~Assent, and M.~E. Houle, ``On the evaluation of unsupervised
  outlier detection: measures, datasets, and an empirical study,'' \emph{Data
  Mining and Knowledge Discovery}, vol.~30, no.~4, pp. 891--927, 2016.

\bibitem{rayana2016}
\BIBentryALTinterwordspacing
S.~Rayana, ``{ODDS} library,'' 2016. [Online]. Available:
  \url{http://odds.cs.stonybrook.edu}
\BIBentrySTDinterwordspacing

\bibitem{domingues2018}
R.~Domingues, M.~Filippone, P.~Michiardi, and J.~Zouaoui, ``A comparative
  evaluation of outlier detection algorithms: Experiments and analyses,''
  \emph{Pattern Recognition}, vol.~74, pp. 406--421, 2018.

\bibitem{Dua:2019}
\BIBentryALTinterwordspacing
D.~Dua and C.~Graff, ``{UCI} machine learning repository,'' 2017. [Online].
  Available: \url{http://archive.ics.uci.edu/ml}
\BIBentrySTDinterwordspacing

\bibitem{bradley1997}
A.~P. Bradley, ``The use of the area under the {ROC} curve in the evaluation of
  machine learning algorithms,'' \emph{Pattern Recognition}, vol.~30, no.~7,
  pp. 1145--1159, 1997.

\bibitem{fawcett2006}
T.~Fawcett, ``An introduction to {ROC} analysis,'' \emph{Pattern Recognition
  Letters}, vol.~27, no.~8, pp. 861--874, 2006.

\bibitem{ding2014}
X.~Ding, Y.~Li, A.~Belatreche, and L.~P. Maguire, ``An experimental evaluation
  of novelty detection methods,'' \emph{Neurocomputing}, vol. 135, pp.
  313--327, 2014.

\bibitem{davis2006}
J.~Davis and M.~Goadrich, ``The relationship between precision-recall and {ROC}
  curves,'' in \emph{International Conference on Machine Learning}, 2006, pp.
  233--240.

\bibitem{boyd2013}
K.~Boyd, K.~H. Eng, and C.~D. Page, ``Area under the precision-recall curve:
  Point estimates and confidence intervals,'' in \emph{European Conference on
  Machine Learning and Principles and Practice of Knowledge Discovery in
  Databases}, 2013, pp. 451--466.

\bibitem{DBLP:conf/kdd/Ribeiro0G16}
M.~T. Ribeiro, S.~Singh, and C.~Guestrin, ``"{W}hy should {I} trust you?":
  {E}xplaining the predictions of any classifier,'' in \emph{International
  Conference on Knowledge Discovery \& Data Mining}, 2016, pp. 1135--1144.

\bibitem{DBLP:conf/iccv/SelvarajuCDVPB17}
R.~R. Selvaraju, M.~Cogswell, A.~Das, R.~Vedantam, D.~Parikh, and D.~Batra,
  ``Grad-{CAM}: {V}isual explanations from deep networks via gradient-based
  localization,'' in \emph{{International Conference on Computer
  Vision}}.\hskip 1em plus 0.5em minus 0.4em\relax {IEEE} Computer Society,
  2017, pp. 618--626.

\bibitem{sundararajan2017}
M.~Sundararajan, A.~Taly, and Q.~Yan, ``Axiomatic attribution for deep
  networks,'' in \emph{International Conference on Machine Learning}, vol.~70,
  2017, pp. 3319--3328.

\bibitem{qi2019visualizing}
Z.~Qi, S.~Khorram, and F.~Li, ``Visualizing deep networks by optimizing with
  integrated gradients,'' in \emph{CVPR 2019 Workshops}, vol.~2, 2019.

\bibitem{bach-plos15}
S.~Bach, A.~Binder, G.~Montavon, F.~Klauschen, K.-R. M{\"u}ller, and W.~Samek,
  ``On pixel-wise explanations for non-linear classifier decisions by
  layer-wise relevance propagation,'' \emph{PLoS ONE}, vol.~10, no.~7, p.
  e0130140, 07 2015.

\bibitem{micenkova2013}
B.~Micenkov{\'a}, R.~T. Ng, X.-H. Dang, and I.~Assent, ``Explaining outliers by
  subspace separability,'' in \emph{{IEEE} International Conference on Data
  Mining}, 2013, pp. 518--527.

\bibitem{Siddiqui2019}
M.~D. Siddiqui, A.~Fern, T.~G. Dietterich, and W.~K. Wong, ``{Sequential
  feature explanations for anomaly detection},'' \emph{ACM Transactions on
  Knowledge Discovery from Data}, vol.~13, no.~1, pp. 1--22, 2019.

\bibitem{kauffmann2020}
J.~Kauffmann, K.-R. M{\"u}ller, and G.~Montavon, ``Towards explaining
  anomalies: A deep {T}aylor decomposition of one-class models,'' \emph{Pattern
  Recognition}, vol. 101, p. 107198, 2020.

\bibitem{litjens17}
G.~Litjens, T.~Kooi, B.~E. Bejnordi, A.~A.~A. Setio, F.~Ciompi, M.~Ghafoorian,
  J.~A. W.~M. van~der Laak, B.~van Ginneken, and C.~I. S{\'{a}}nchez, ``A
  survey on deep learning in medical image analysis,'' \emph{Medical Image
  Analysis}, vol.~42, pp. 60--88, 2017.

\bibitem{DBLP:journals/corr/abs-1906-07633}
J.~Kauffmann, M.~Esders, G.~Montavon, W.~Samek, and K.-R. M{\"{u}}ller, ``From
  clustering to cluster explanations via neural networks,'' \emph{arXiv
  preprint arXiv:1906.07633}, 2019.

\bibitem{huynh1998}
P.~T. Huynh, A.~M. Jarolimek, and S.~Daye, ``The false-negative mammogram.''
  \emph{Radiographics}, vol.~18, no.~5, pp. 1137--1154, 1998.

\bibitem{petticrew2000}
M.~Petticrew, A.~Sowden, D.~Lister-Sharp, and K.~Wright, ``False-negative
  results in screening programmes: systematic review of impact and
  implications,'' \emph{Health Technology Assess.}, vol.~4, no.~5, pp. 1--120,
  2000.

\bibitem{pepe2003}
M.~S. Pepe, \emph{The Statistical Evaluation of Medical Tests for
  Classification and Prediction}.\hskip 1em plus 0.5em minus 0.4em\relax Oxford
  University Press, 2003.

\bibitem{zhou2011}
X.-H. Zhou, N.~A. Obuchowski, and D.~K. McClish, \emph{Statistical Methods in
  Diagnostic Medicine}, 2nd~ed.\hskip 1em plus 0.5em minus 0.4em\relax John
  Wiley \& Sons, 2011.

\bibitem{dudaHart1973}
R.~O. Duda and P.~E. Hart, \emph{Pattern Classification and Scene
  Analysis}.\hskip 1em plus 0.5em minus 0.4em\relax John Willey \& Sons, 1973.

\bibitem{theodoridis2009}
S.~Theodoridis and K.~Koutroumbas, \emph{Pattern Recognition}, 4th~ed.\hskip
  1em plus 0.5em minus 0.4em\relax Academic Press, 2009.

\bibitem{DBLP:conf/kdd/LazarevicK05}
A.~Lazarevic and V.~Kumar, ``Feature bagging for outlier detection,'' in
  \emph{International Conference on Knowledge Discovery \& Data Mining}, 2005,
  pp. 157--166.

\bibitem{DBLP:conf/dasfaa/VuAG10}
H.~V. Nguyen, H.~H. Ang, and V.~Gopalkrishnan, ``Mining outliers with ensemble
  of heterogeneous detectors on random subspaces,'' in \emph{{DASFAA} {(1)}},
  ser. Lecture Notes in Computer Science, vol. 5981.\hskip 1em plus 0.5em minus
  0.4em\relax Springer, Berlin, Heidelberg, 2010, pp. 368--383.

\bibitem{JMLR:v9:braun08a}
M.~L. Braun, J.~M. Buhmann, and K.-R. M{{\"u}}ller, ``On relevant dimensions in
  kernel feature spaces,'' \emph{Journal of Machine Learning Research}, vol.~9,
  no. Aug, pp. 1875--1908, 2008.

\bibitem{gal2016}
Y.~Gal and Z.~Ghahramani, ``Dropout as a bayesian approximation: Representing
  model uncertainty in deep learning,'' in \emph{International Conference on
  Machine Learning}, vol.~48, 2016, pp. 1050--1059.

\bibitem{lakshminarayanan2017}
B.~Lakshminarayanan, A.~Pritzel, and C.~Blundell, ``Simple and scalable
  predictive uncertainty estimation using deep ensembles,'' in \emph{Advances
  in Neural Information Processing Systems}, 2017, pp. 6402--6413.

\bibitem{kendall2017}
A.~Kendall and Y.~Gal, ``What uncertainties do we need in bayesian deep
  learning for computer vision?'' in \emph{Advances in Neural Information
  Processing Systems}, 2017, pp. 5574--5584.

\bibitem{ovadia2019}
Y.~Ovadia, E.~Fertig, J.~Ren, Z.~Nado, D.~Sculley, S.~Nowozin, J.~Dillon,
  B.~Lakshminarayanan, and J.~Snoek, ``Can you trust your model's uncertainty?
  evaluating predictive uncertainty under dataset shift,'' in \emph{Advances in
  Neural Information Processing Systems}, 2019, pp. 13\,991--14\,002.

\bibitem{bykov2020much}
K.~Bykov, M.~M.-C. H{\"o}hne, K.-R. M{\"u}ller, S.~Nakajima, and M.~Kloft,
  ``How much can i trust you?--quantifying uncertainties in explaining neural
  networks,'' \emph{arXiv preprint arXiv:2006.09000}, 2020.

\bibitem{hido2011}
S.~Hido, Y.~Tsuboi, H.~Kashima, M.~Sugiyama, and T.~Kanamori, ``Statistical
  outlier detection using direct density ratio estimation,'' \emph{Knowledge
  and Information Systems}, vol.~26, no.~2, pp. 309--336, 2011.

\bibitem{gutmann2010}
M.~Gutmann and A.~Hyv{\"a}rinen, ``Noise-contrastive estimation: A new
  estimation principle for unnormalized statistical models,'' in
  \emph{International Conference on Artificial Intelligence and Statistics},
  2010, pp. 297--304.

\bibitem{vandermeulen2020}
R.~A. Vandermeulen, R.~Saitenmacher, and A.~Ritchie, ``A proposal for
  supervised density estimation,'' in \emph{NeurIPS 2020 Pre-registration
  Workshop}, 2020.

\bibitem{chow1957}
C.~K. Chow, ``An optimum character recognition system using decision
  functions,'' \emph{IRE Transactions on Electronic Computers}, vol. EC-6,
  no.~4, pp. 247--254, Dec. 1957.

\bibitem{chow1970}
------, ``On optimum recognition error and reject tradeoff,'' \emph{IEEE
  Transactions on Information Theory}, vol.~16, no.~1, pp. 41--46, 1970.

\bibitem{bartlett2008}
P.~L. Bartlett and M.~H. Wegkamp, ``Classification with a reject option using a
  hinge loss,'' \emph{Journal of Machine Learning Research}, vol.~9, no. Aug,
  pp. 1823--1840, 2008.

\bibitem{tax2008}
D.~M.~J. Tax and R.~P.~W. Duin, ``Growing a multi-class classifier with a
  reject option,'' \emph{Pattern Recognition Letters}, vol.~29, no.~10, pp.
  1565--1570, 2008.

\bibitem{grandvalet2009}
Y.~Grandvalet, A.~Rakotomamonjy, J.~Keshet, and S.~Canu, ``Support vector
  machines with a reject option,'' in \emph{Advances in Neural Information
  Processing Systems}, 2009, pp. 537--544.

\bibitem{cortes2016}
C.~Cortes, G.~DeSalvo, and M.~Mohri, ``Learning with rejection,'' in
  \emph{International Conference on Algorithmic Learning Theory}, 2016, pp.
  67--82.

\bibitem{geifman2017}
Y.~Geifman and R.~El-Yaniv, ``Selective classification for deep neural
  networks,'' in \emph{Advances in Neural Information Processing Systems},
  2017, pp. 4878--4887.

\bibitem{platt1999}
J.~C. Platt, ``Probabilistic outputs for support vector machines and
  comparisons to regularized likelihood methods,'' \emph{Advances in Large
  Margin Classifiers}, vol.~10, no.~3, pp. 61--74, 1999.

\bibitem{guo2017}
C.~Guo, G.~Pleiss, Y.~Sun, and K.~Q. Weinberger, ``On calibration of modern
  neural networks,'' in \emph{International Conference on Machine Learning},
  vol.~70, 2017, pp. 1321--1330.

\bibitem{devries2018}
T.~DeVries and G.~W. Taylor, ``Learning confidence for out-of-distribution
  detection in neural networks,'' \emph{arXiv preprint arXiv:1802.04865}, 2018.

\bibitem{lee2018}
K.~Lee, H.~Lee, K.~Lee, and J.~Shin, ``Training confidence-calibrated
  classifiers for detecting out-of-distribution samples,'' in
  \emph{International Conference on Learning Representations}, 2018.

\bibitem{ni2019}
C.~Ni, N.~Charoenphakdee, J.~Honda, and M.~Sugiyama, ``On the calibration of
  multiclass classification with rejection,'' in \emph{Advances in Neural
  Information Processing Systems}, 2019, pp. 2586--2596.

\bibitem{meinke2020}
A.~Meinke and M.~Hein, ``Towards neural networks that provably know when they
  don't know,'' in \emph{International Conference on Learning Representations},
  2020.

\bibitem{mackay1998}
D.~J.~C. {MacKay} and M.~N. Gibbs, ``Density networks,'' in \emph{Statistics
  and Neural Networks: Advances at the Interface}.\hskip 1em plus 0.5em minus
  0.4em\relax USA: Oxford University Press, 1998, pp. 129--146.

\bibitem{liang2018}
S.~Liang, Y.~Li, and R.~Srikant, ``Enhancing the reliability of
  out-of-distribution image detection in neural networks,'' in
  \emph{International Conference on Learning Representations}, 2018.

\bibitem{lee2018b}
K.~Lee, K.~Lee, H.~Lee, and J.~Shin, ``A simple unified framework for detecting
  out-of-distribution samples and adversarial attacks,'' in \emph{Advances in
  Neural Information Processing Systems}, 2018, pp. 7167--7177.

\bibitem{choi2020}
S.~Choi and S.-Y. Chung, ``Novelty detection via blurring,'' in
  \emph{International Conference on Learning Representations}, 2020.

\bibitem{scheirer2012}
W.~J. Scheirer, A.~de~Rezende~Rocha, A.~Sapkota, and T.~E. Boult, ``Toward open
  set recognition,'' \emph{{IEEE} Transactions on Pattern Analysis and Machine
  Intelligence}, vol.~35, no.~7, pp. 1757--1772, 2012.

\bibitem{scheirer2014}
W.~J. Scheirer, L.~P. Jain, and T.~E. Boult, ``Probability models for open set
  recognition,'' \emph{{IEEE} Transactions on Pattern Analysis and Machine
  Intelligence}, vol.~36, no.~11, pp. 2317--2324, 2014.

\bibitem{bendale2016}
A.~Bendale and T.~E. Boult, ``Towards open set deep networks,'' in \emph{{IEEE}
  Conference on Computer Vision and Pattern Recognition}, 2016, pp. 1563--1572.

\bibitem{shu2017}
L.~Shu, H.~Xu, and B.~Liu, ``{DOC}: Deep open classification of text
  documents,'' in \emph{Conference on Empirical Methods in Natural Language
  Processing}, 2017, pp. 2911--2916.

\bibitem{zhang2020}
H.~Zhang, A.~Li, J.~Guo, and Y.~Guo, ``Hybrid models for open set
  recognition,'' \emph{arXiv preprint arXiv:2003.12506}, 2020.

\bibitem{szegedy2014}
C.~Szegedy, W.~Zaremba, I.~Sutskever, J.~Bruna, D.~Erhan, I.~Goodfellow, and
  R.~Fergus, ``Intriguing properties of neural networks,'' in
  \emph{International Conference on Learning Representations}, 2014.

\bibitem{goodfellow2015}
I.~J. Goodfellow, J.~Shlens, and C.~Szegedy, ``Explaining and harnessing
  adversarial examples,'' in \emph{International Conference on Learning
  Representations}, 2015.

\bibitem{carlini2017}
N.~Carlini and D.~Wagner, ``Towards evaluating the robustness of neural
  networks,'' in \emph{{IEEE} Symposium on Security and Privacy}.\hskip 1em
  plus 0.5em minus 0.4em\relax IEEE, 2017, pp. 39--57.

\bibitem{tramer2017}
F.~Tram{\`e}r, A.~Kurakin, N.~Papernot, I.~Goodfellow, D.~Boneh, and
  P.~McDaniel, ``Ensemble adversarial training: Attacks and defenses,'' in
  \emph{International Conference on Learning Representations}, 2018.

\bibitem{biggio2018}
B.~Biggio and F.~Roli, ``Wild patterns: Ten years after the rise of adversarial
  machine learning,'' \emph{Pattern Recognition}, vol.~84, pp. 317--331, 2018.

\bibitem{athalye2018}
A.~Athalye, N.~Carlini, and D.~Wagner, ``Obfuscated gradients give a false
  sense of security: Circumventing defenses to adversarial examples,'' in
  \emph{International Conference on Machine Learning}, vol.~80, 2018, pp.
  274--283.

\bibitem{madry2018}
A.~Madry, A.~Makelov, L.~Schmidt, D.~Tsipras, and A.~Vladu, ``Towards deep
  learning models resistant to adversarial attacks,'' in \emph{International
  Conference on Learning Representations}, 2018.

\bibitem{carlini2019}
N.~Carlini, A.~Athalye, N.~Papernot, W.~Brendel, J.~Rauber, D.~Tsipras,
  I.~Goodfellow, A.~Madry, and A.~Kurakin, ``On evaluating adversarial
  robustness,'' \emph{arXiv preprint arXiv:1902.06705}, 2019.

\bibitem{zhang2019b}
H.~Zhang, Y.~Yu, J.~Jiao, E.~Xing, L.~El~Ghaoui, and M.~Jordan, ``Theoretically
  principled trade-off between robustness and accuracy,'' in
  \emph{International Conference on Machine Learning}, vol.~97, 2019, pp.
  7472--7482.

\bibitem{ilyas2019}
A.~Ilyas, S.~Santurkar, D.~Tsipras, L.~Engstrom, B.~Tran, and A.~Madry,
  ``Adversarial examples are not bugs, they are features,'' in \emph{Advances
  in Neural Information Processing Systems}, 2019, pp. 125--136.

\bibitem{shalev2018}
G.~Shalev, Y.~Adi, and J.~Keshet, ``Out-of-distribution detection using
  multiple semantic label representations,'' in \emph{Advances in Neural
  Information Processing Systems}, 2018, pp. 7375--7385.

\bibitem{hendrycks2019c}
D.~Hendrycks, K.~Lee, and M.~Mazeika, ``Using pre-training can improve model
  robustness and uncertainty,'' in \emph{International Conference on Machine
  Learning}, vol.~97, 2019, pp. 2712--2721.

\bibitem{dhamija2018}
A.~R. Dhamija, M.~G{\"u}nther, and T.~Boult, ``Reducing network
  agnostophobia,'' in \emph{Advances in Neural Information Processing Systems},
  2018, pp. 9157--9168.

\bibitem{dombrowski2019explanations}
A.-K. Dombrowski, M.~Alber, C.~Anders, M.~Ackermann, K.-R. M{\"u}ller, and
  P.~Kessel, ``Explanations can be manipulated and geometry is to blame,'' in
  \emph{Advances in Neural Information Processing Systems}, 2019, pp.
  13\,589--13\,600.

\bibitem{caruana2015}
R.~Caruana, Y.~Lou, J.~Gehrke, P.~Koch, M.~Sturm, and N.~Elhadad,
  ``Intelligible models for healthcare: Predicting pneumonia risk and hospital
  30-day readmission,'' in \emph{International Conference on Knowledge
  Discovery \& Data Mining}, 2015, pp. 1721--1730.

\bibitem{samek2020}
W.~Samek, G.~Montavon, S.~Lapuschkin, C.~J. Anders, and K.-R. M{\"u}ller,
  ``Toward interpretable machine learning: Transparent deep neural networks and
  beyond,'' \emph{arXiv preprint arXiv:2003.07631}, 2020.

\bibitem{lee2004}
J.~D. Lee and K.~A. See, ``Trust in automation: Designing for appropriate
  reliance,'' \emph{Human Factors}, vol.~46, no.~1, pp. 50--80, 2004.

\bibitem{jiang2018}
H.~Jiang, B.~Kim, M.~Guan, and M.~Gupta, ``To trust or not to trust a
  classifier,'' in \emph{Advances in Neural Information Processing Systems},
  2018, pp. 5541--5552.

\bibitem{lipton2017}
Z.~C. Lipton, ``The doctor just won't accept that!'' in \emph{NIPS 2017
  Interpretable ML Symposium}, 2017.

\bibitem{goodman2017}
B.~Goodman and S.~Flaxman, ``European union regulations on algorithmic
  decision-making and a “right to explanation”,'' \emph{AI Magazine},
  vol.~38, no.~3, pp. 50--57, 2017.

\bibitem{amodei2016}
D.~Amodei, C.~Olah, J.~Steinhardt, P.~Christiano, J.~Schulman, and D.~Man{\'e},
  ``Concrete problems in ai safety,'' \emph{arXiv preprint arXiv:1606.06565},
  2016.

\bibitem{richter2017}
C.~Richter and N.~Roy, ``Safe visual navigation via deep learning and novelty
  detection,'' in \emph{Robotics: Science and Systems XIII}, 2017.

\bibitem{dang2013}
X.~H. Dang, B.~Micenkov{\'a}, I.~Assent, and R.~T. Ng, ``Local outlier
  detection with interpretation,'' in \emph{European Conference on Machine
  Learning and Principles and Practice of Knowledge Discovery in Databases},
  2013, pp. 304--320.

\bibitem{dang2014}
X.~H. Dang, I.~Assent, R.~T. Ng, A.~Zimek, and E.~Schubert, ``Discriminative
  features for identifying and interpreting outliers,'' in \emph{International
  Conference on Data Engineering}.\hskip 1em plus 0.5em minus 0.4em\relax IEEE,
  2014, pp. 88--99.

\bibitem{duan2015}
L.~Duan, G.~Tang, J.~Pei, J.~Bailey, A.~Campbell, and C.~Tang, ``Mining
  outlying aspects on numeric data,'' \emph{Data Mining and Knowledge
  Discovery}, vol.~29, no.~5, pp. 1116--1151, 2015.

\bibitem{vinh2016}
N.~X. Vinh, J.~Chan, S.~Romano, J.~Bailey, C.~Leckie, K.~Ramamohanarao, and
  J.~Pei, ``Discovering outlying aspects in large datasets,'' \emph{Data Mining
  and Knowledge Discovery}, vol.~30, no.~6, pp. 1520--1555, 2016.

\bibitem{macha2018}
M.~Macha and L.~Akoglu, ``Explaining anomalies in groups with characterizing
  subspace rules,'' \emph{Data Mining and Knowledge Discovery}, vol.~32, no.~5,
  pp. 1444--1480, 2018.

\bibitem{deng2009}
J.~Deng, W.~Dong, R.~Socher, L.-J. Li, K.~Li, and L.~Fei-Fei, ``{ImageNet}: A
  large-scale hierarchical image database,'' in \emph{{IEEE} Conference on
  Computer Vision and Pattern Recognition}, 2009, pp. 248--255.

\bibitem{russakovsky2015}
O.~Russakovsky, J.~Deng, H.~Su, J.~Krause, S.~Satheesh, S.~Ma, Z.~Huang,
  A.~Karpathy, A.~Khosla, M.~Bernstein, A.~C. Berg, and L.~Fei-Fei,
  ``{ImageNet} large scale visual recognition challenge,'' \emph{International
  Journal of Computer Vision}, vol. 115, no.~3, pp. 211--252, 2015.

\bibitem{wang2019b}
J.~Wang, S.~Sun, and Y.~Yu, ``Multivariate triangular quantile maps for novelty
  detection,'' in \emph{Advances in Neural Information Processing Systems},
  2019, pp. 5061--5072.

\bibitem{ren2019}
J.~Ren, P.~J. Liu, E.~Fertig, J.~Snoek, R.~Poplin, M.~Depristo, J.~Dillon, and
  B.~Lakshminarayanan, ``Likelihood ratios for out-of-distribution detection,''
  in \emph{Advances in Neural Information Processing Systems}, 2019, pp.
  14\,680--14\,691.

\bibitem{serra2020}
J.~Serr{\`a}, D.~{\'A}lvarez, V.~G{\'o}mez, O.~Slizovskaia, J.~F.
  N{\'u}{\~n}ez, and J.~Luque, ``Input complexity and out-of-distribution
  detection with likelihood-based generative models,'' in \emph{International
  Conference on Learning Representations}, 2020.

\bibitem{box1976}
G.~E. Box, ``Science and statistics,'' \emph{Journal of the American
  Statistical Association}, vol.~71, no. 356, pp. 791--799, 1976.

\bibitem{ratner2017}
A.~Ratner, S.~H. Bach, H.~Ehrenberg, J.~Fries, S.~Wu, and C.~R{\'e}, ``Snorkel:
  Rapid training data creation with weak supervision,'' \emph{Proceedings of
  the {VLDB} Endowment}, vol.~11, no.~3, pp. 269--282, 2017.

\bibitem{zhou2018}
Z.-H. Zhou, ``A brief introduction to weakly supervised learning,''
  \emph{National Science Review}, vol.~5, no.~1, pp. 44--53, 2018.

\bibitem{roh2019}
Y.~Roh, G.~Heo, and S.~E. Whang, ``A survey on data collection for machine
  learning: A big data - ai integration perspective,'' \emph{IEEE Transactions
  on Knowledge and Data Engineering}, 2019.

\bibitem{daniel2019}
T.~Daniel, T.~Kurutach, and A.~Tamar, ``Deep variational semi-supervised
  novelty detection,'' \emph{arXiv preprint arXiv:1911.04971}, 2019.

\bibitem{das2020}
S.~Das, W.-K. Wong, T.~Dietterich, A.~Fern, and A.~Emmott, ``Discovering
  anomalies by incorporating feedback from an expert,'' \emph{Transactions on
  Knowledge Discovery from Data}, vol.~14, no.~4, pp. 1--32, 2020.

\bibitem{nedelkoski2020}
S.~Nedelkoski, J.~Bogatinovski, A.~Acker, J.~Cardoso, and O.~Kao,
  ``Self-attentive classification-based anomaly detection in unstructured
  logs,'' \emph{arXiv preprint arXiv:2008.09340}, 2020.

\bibitem{ouardini2019}
K.~Ouardini, H.~Yang, B.~Unnikrishnan, M.~Romain, C.~Garcin, H.~Zenati, J.~P.
  Campbell, M.~F. Chiang, J.~Kalpathy-Cramer, V.~Chandrasekhar \emph{et~al.},
  ``Towards practical unsupervised anomaly detection on retinal images,'' in
  \emph{Domain Adaptation and Representation Transfer and Medical Image
  Learning with Less Labels and Imperfect Data}.\hskip 1em plus 0.5em minus
  0.4em\relax Springer, Cham, 2019, pp. 225--234.

\bibitem{shu2020}
R.~Shu, Y.~Chen, A.~Kumar, S.~Ermon, and B.~Poole, ``Weakly supervised
  disentanglement with guarantees,'' in \emph{International Conference on
  Learning Representations}, 2020.

\bibitem{locatello2020}
F.~Locatello, B.~Poole, G.~R{\"a}tsch, B.~Sch{\"o}lkopf, O.~Bachem, and
  M.~Tschannen, ``Weakly-supervised disentanglement without compromises,'' in
  \emph{International Conference on Machine Learning}, 2020, pp. 7753--7764.

\bibitem{mathieu2016}
M.~Mathieu, C.~Couprie, and Y.~LeCun, ``Deep multi-scale video prediction
  beyond mean square error,'' in \emph{International Conference on Learning
  Representations}, 2016.

\bibitem{chen2020}
T.~Chen, S.~Kornblith, M.~Norouzi, and G.~Hinton, ``A simple framework for
  contrastive learning of visual representations,'' in \emph{International
  Conference on Machine Learning}, 2020, pp. 10\,709--10\,719.

\bibitem{zhang2016}
R.~Zhang, P.~Isola, and A.~A. Efros, ``Colorful image colorization,'' in
  \emph{European Conference on Computer Vision}, 2016, pp. 649--666.

\bibitem{doersch2015}
C.~Doersch, A.~Gupta, and A.~A. Efros, ``Unsupervised visual representation
  learning by context prediction,'' in \emph{International Conference on
  Computer Vision}, 2015, pp. 1422--1430.

\bibitem{noroozi2016}
M.~Noroozi and P.~Favaro, ``Unsupervised learning of visual representations by
  solving jigsaw puzzles,'' in \emph{European Conference on Computer Vision},
  2016, pp. 69--84.

\bibitem{gidaris2018}
S.~Gidaris, P.~Singh, and N.~Komodakis, ``Unsupervised representation learning
  by predicting image rotations,'' in \emph{International Conference on
  Learning Representations}, 2018.

\bibitem{winkens2020}
J.~Winkens, R.~Bunel, A.~G. Roy, R.~Stanforth, V.~Natarajan, J.~R. Ledsam,
  P.~MacWilliams, P.~Kohli, A.~Karthikesalingam, S.~Kohl \emph{et~al.},
  ``Contrastive training for improved out-of-distribution detection,''
  \emph{arXiv preprint arXiv:2007.05566}, 2020.

\bibitem{asano2020}
Y.~M. Asano, C.~Rupprecht, and A.~Vedaldi, ``A critical analysis of
  self-supervision, or what we can learn from a single image,'' in
  \emph{International Conference on Learning Representations}, 2020.

\bibitem{nalisnick2019b}
E.~Nalisnick, A.~Matsukawa, Y.~W. Teh, and B.~Lakshminarayanan, ``Detecting
  out-of-distribution inputs to deep generative models using typicality,'' in
  \emph{4th Workshop on Bayesian Deep Learning (NeurIPS 2019)}, 2019.

\bibitem{vershynin2018}
R.~Vershynin, \emph{High-Dimensional Probability: An Introduction with
  Applications in Data Science}.\hskip 1em plus 0.5em minus 0.4em\relax
  Cambridge University Press, 2018, vol.~47.

\bibitem{tong2020}
A.~Tong, G.~Wolf, and S.~Krishnaswamyt, ``Fixing bias in reconstruction-based
  anomaly detection with lipschitz discriminators,'' in \emph{{IEEE}
  International Workshop on Machine Learning for Signal Processing}.\hskip 1em
  plus 0.5em minus 0.4em\relax IEEE, 2020, pp. 1--6.

\bibitem{krueger2020}
D.~Krueger, E.~Caballero, J.-H. Jacobsen, A.~Zhang, J.~Binas, R.~L. Priol, and
  A.~Courville, ``Out-of-distribution generalization via risk extrapolation,''
  \emph{arXiv preprint arXiv:2003.00688}, 2020.

\bibitem{shannon1948}
C.~E. Shannon, ``A mathematical theory of communication,'' \emph{The Bell
  System Technical Journal}, vol.~27, no.~3, pp. 379--423, July, October 1948.

\bibitem{linsker1988}
R.~Linsker, ``Self-organization in a perceptual network,'' \emph{Computer},
  vol.~21, no.~3, pp. 105--117, 1988.

\bibitem{bell1995}
A.~J. Bell and T.~J. Sejnowski, ``An information-maximization approach to blind
  separation and blind deconvolution,'' \emph{Neural Computation}, vol.~7,
  no.~6, pp. 1129--1159, 1995.

\bibitem{hjelm2019}
R.~D. Hjelm, A.~Fedorov, S.~Lavoie-Marchildon, K.~Grewal, P.~Bachman,
  A.~Trischler, and Y.~Bengio, ``Learning deep representations by mutual
  information estimation and maximization,'' in \emph{International Conference
  on Learning Representations}, 2019.

\bibitem{berger2003}
T.~Berger, ``Rate-distortion theory,'' \emph{Wiley Encyclopedia of
  Telecommunications}, 2003.

\bibitem{higgins2017}
I.~Higgins, L.~Matthey, A.~Pal, C.~Burgess, X.~Glorot, M.~Botvinick,
  S.~Mohamed, and A.~Lerchner, ``$\beta$-{VAE}: Learning basic visual concepts
  with a constrained variational framework,'' in \emph{International Conference
  on Learning Representations}, 2017.

\bibitem{alemi2018}
A.~Alemi, B.~Poole, I.~Fischer, J.~Dillon, R.~A. Saurous, and K.~Murphy,
  ``Fixing a broken {ELBO},'' in \emph{International Conference on Machine
  Learning}, vol.~80, 2018, pp. 159--168.

\bibitem{park2020}
S.~Park, G.~Adosoglou, and P.~M. Pardalos, ``Interpreting rate-distortion of
  variational autoencoder and using model uncertainty for anomaly detection,''
  \emph{arXiv preprint arXiv:2005.01889}, 2020.

\bibitem{lee2001}
W.~Lee and D.~Xiang, ``Information-theoretic measures for anomaly detection,''
  in \emph{IEEE Symposium on Security and Privacy}.\hskip 1em plus 0.5em minus
  0.4em\relax IEEE, 2001, pp. 130--143.

\bibitem{host2019}
A.~H{\o}st-Madsen, E.~Sabeti, and C.~Walton, ``Data discovery and anomaly
  detection using atypicality: Theory,'' \emph{Transactions on Information
  Theory}, vol.~65, no.~9, pp. 5302--5322, 2019.

\bibitem{scikit-learn}
F.~Pedregosa, G.~Varoquaux, A.~Gramfort, V.~Michel, B.~Thirion, O.~Grisel,
  M.~Blondel, P.~Prettenhofer, R.~Weiss, V.~Dubourg, J.~Vanderplas, A.~Passos,
  D.~Cournapeau, M.~Brucher, M.~Perrot, and E.~Duchesnay, ``Scikit-learn:
  Machine learning in {P}ython,'' \emph{Journal of Machine Learning Research},
  vol.~12, pp. 2825--2830, 2011.

\bibitem{pytorch}
A.~Paszke, S.~Gross, F.~Massa, A.~Lerer, J.~Bradbury, G.~Chanan, T.~Killeen,
  Z.~Lin, N.~Gimelshein, L.~Antiga, A.~Desmaison, A.~Kopf, E.~Yang, Z.~DeVito,
  M.~Raison, A.~Tejani, S.~Chilamkurthy, B.~Steiner, L.~Fang, J.~Bai, and
  S.~Chintala, ``{PyTorch}: An imperative style, high-performance deep learning
  library,'' in \emph{Advances in Neural Information Processing Systems}, 2019,
  pp. 8026--8037.

\bibitem{sonnenburg2007}
S.~Sonnenburg, M.~L. Braun, C.~S. Ong, S.~Bengio, L.~Bottou, G.~Holmes,
  Y.~LeCun, K.-R. M{\"u}ller, F.~Pereira, C.~E. Rasmussen, G.~R\"{a}tsch,
  B.~Sch{\"o}lkopf, A.~Smola, P.~Vincent, J.~Weston, and R.~Williamson, ``The
  need for open source software in machine learning,'' \emph{Journal of Machine
  Learning Research}, vol.~8, no. Oct, pp. 2443--2466, 2007.

\end{thebibliography}

\end{document}